# SOFT NEUTROSOPHIC ALGEBRAIC STRUCTURES AND THEIR GENERALIZATION

**Florentin Smarandache**
e-mail: **smarand@unm.edu**

**Mumtaz Ali**
e-mail: **mumtazali770@yahoo.com**

e-mail: **bloomy_boy2006@yahoo.com**

**Muhammad Shabir**

e-mail: **mshbirbhatti@yahoo.co.uk**

# CONTENTS

**Chapter Two**

# SOFT NEUTROSOPHIC GROUPS AND SOFT NEUTROSOPHIC N-GROUPS

**Chapter Three**

# SOFT NEUTROSOPHIC SEMIGROUPS AND SOFT NEUTROSOPHIC N-SEMIGROUPS

**Chapter Four**

# SOFT NEUTROSOPHIC LOOPS AND SOFT NEUTROSOPHIC N-LOOPS

**Chapter Five**

**SOFT NEUTROSOPHIC LA-SEMIGROUPS AND SOFT NEUTROSOPHIC N-LA-SEMIGROUPS**

**Chapter Six**

# SUGGESTED PROBLEMS

# DEDICATION | MUMTAZ ALI

This book is dedicated to my dearest Farzana Ahmad. Her kind support and love always courage me to work on this work. Whenever I am upset and alone, her guidance provide me a better way for doing something better. I m so much thankful to her with whole heartedly and dedicate this book to her. This is my humble way of paying homage to her support, kindness and love.

# PREFACE

In this book the authors introduced the notions of soft neutrosophic algebraic structures. These soft neutrosophic algebraic structures are basically defined over the neutrosophic algebraic structures which means a parameterized collection of subsets of the neutrosophic algebraic structure. For instance , the existence of a soft neutrosophic group over a neutrosophic group or a soft neutrosophic semigroup over a neutrosophic semigroup, or a soft neutrosophic field over a neutrosophic field, or a soft neutrosophic LA-semigroup over a neutrosophic LA-semigroup, or a soft neutosophic loop over a neutrosophic loop. It is interesting to note that these notions are defined over finite and infinite neutrosophic algebraic structures. These structures are even bigger than the classical algebraic structures.

This book contains five chapters. Chapter one is about the introductory concepts. In chapter two the notions of soft neutrosophic group, soft neutrosophic bigroup and soft neutrosophic N-group are introduced and many fantastic properties are given with illustrative examples. Soft neutrosophic semigroup, soft neutrosophic bisemigroup and soft neutrosophic N-semigroup are placed in the second chapter with related properties. The chapter three is about soft neutrosophic loop with soft neutrosophic biloop and soft neutrosophic N-loop. Chapter four contained on soft neutrosophic LA-semigroup, soft neutrosophic bi-LA-semigroup and soft neutrosophic N-LA-semigroup. In the final chapter, there are number of suggestive problems.

# Chapter One

# INTRODUCTION

In this chapter we give certain basic concepts and notions. This chapter has 5 sections. The first section contains the notions of neutrosophic groups, neutrosophic bigroups abd neutrosophic N-groups. In section 2, we mentioned neutrosophic semigroups, neutrosophic bisemigroups and neutrosophic N-semigroups. Section 3 is about neutrosophic loops and their generalizations. In section 4 some introductory description are given about neutrosophic left almost semigroup abbrivated as neutrosophic LA-semigroup, neutrosophic bi-LA-semigroup and neutrosophic N-LA-semigroup and their related properties. In the last section, we gave some basic literature about soft set theory and their related properties which will help to understand the theory of soft sets.

## 1.1 Neutrosophic groups, Neutrosophic Bigroups, Neutrosophic N-groups and their Properties

Groups are the only perfect and rich algebraic structures than other algebraic strutures with one binary operation. Neutrosophic groups areeven bigger algebraic structures than groups with some kind of indeterminacy factor. Now we proceed to define a neutrosophic group.

**Definition 1.1.1:** Let $(G,*)$ be any group and let $\langle G \cup I \rangle = \{a + bI : a, b \in G\}$. Then the neutrosophic group is generated by $I$ and $G$ under $*$ denoted by $N(G) = \{\langle G \cup I \rangle, *\}$. $I$ is called the neutrosophic element with the property $I^2 = I$. For an integer $n$, $n + I$ and $nI$ are neutrosophic elements and $0.I = 0$.
$I^{-1}$, the inverse of $I$ is not defined and hence does not exist.

**Theorem 1.1.1:** Let $N(G)$ be a neutrosophic group. Then

1) $N(G)$ in general is not a group.
2) $N(G)$ always contains a group.

**Example 1.1.1:** $(N(Z), +)$, $(N(Q), +)$, $(N(R), +)$ and $(N(C), +)$ are neutrosophic groups of integer, rational, real and complex numbers respectively.

**Example 1.1.2:** Let $Z_7 = \{0, 1, 2, ..., 6\}$ be a group under addition modulo 7. $N(G) = \{\langle Z_7 \cup I \rangle, '+' \bmod ulo 7\}$ is a neutrosophic group which is in fact a group. For $N(G) = \{a + bI : a, b \in Z_7\}$ is a group under ` + ' modulo 7.

**Example 1.1.3:** Let $N(G) = \langle Z_4 \cup I \rangle$ is a neutrosophic group under addition modul 4, where

$$N(Z_4) = \begin{Bmatrix} 0,1,2,3,I,2I,3I,1+I,1+2I,1+3I, \\ 2+I,2+2I,2+3I,3+I,3+2I,3+3I \end{Bmatrix}$$

**Definition 1.1.2:** A pseudo neutrosophic group is defined as a neutrosophic group, which does not contain a proper subset which is a group.

**Example 1.1.4:** Let
$$N(Z_2) = \{0,1,I,1+I\}$$
be a neutrosophic group under addition modulo 2. Then $N(Z_2)$ is a pseudo neutrosophic group.

**Definition 1.1.3:** Let $N(G)$ be a neutrosophic group. Then,

1) A proper subset $N(H)$ of $N(G)$ is said to be a neutrosophic subgroup of $N(G)$ if $N(H)$ is a neutrosophic group, that is, $N(H)$ contains a proper subset which is a group.
2) $N(H)$ is said to be a pseudo neutrosophic subgroup if it does not contain a proper subset which is a group.

**Definition 1.1.4:** Let $N(G)$ be a finite neutrosophic group. Let $P$ be a proper subset of $N(G)$ which under the operations of $N(G)$ is a neutrosophic group. If $o(P) / o(N(G))$ then we call $P$ to be a Lagrange neutrosophic subgroup.

**Definition 1.1.5:** $N(G)$ is called weakly Lagrange neutrosophic group if $N(G)$ has at least one Lagrange neutrosophic subgroup.

**Definition 1.1.6:** $N(G)$ is called Lagrange free neutrosophic group if $N(G)$ has no Lagrange neutrosophic subgroup.

**Definition 1.1.7 :** Let $N(G)$ be a finite neutrosophic group. Suppose $L$ is a pseudo neutrosophic subgroup of $N(G)$ and if $o(L) / o(N(G))$ then we call $L$ to be a pseudo Lagrange neutrosophic subgroup.

**Definition 1.1.8:** If $N(G)$ has at least one pseudo Lagrange neutrosophic subgroup then we call $N(G)$ to be a weakly pseudo Lagrange neutrosophic group.

**Definition 1.1.9:** If $N(G)$ has no pseudo Lagrange neutrosophic subgroup then we call $N(G)$ to be pseudo Lagrange free neutrosophic group.

**Definition 1.1.10:** Let $N(G)$ be a neutrosophic group. We say a neutrosophic subgroup $H$ of $N(G)$ is normal if we can find $x$ and $y$ in $N(G)$ such that $H = xHy$ for all $x, y \in N(G)$ (Note $x = y$ or $y = x^{-1}$ can also occur).

**Definition 1.1.11:** A neutrosophic group $N(G)$ which has no nontrivial neutrosophic normal subgroup is called a simple neutrosophic group.

**Definition 1.1.12:** Let $N(G)$ be a neutrosophic group. A proper pseudo neutrosophic subgroup $P$ of $N(G)$ is said to be normal if we have $P = xPy$ for all $x, y \in N(G)$. A neutrosophic group is said to be pseudo simple neutrosophic group if $N(G)$ has no nontrivial pseudo normal subgroups.

**Definition 1.1.13:** Let $B_N(G) = \{B(G_1) \cup B(G_2), *_1, *_2\}$ be a non-empty subset with two binary operations on $B_N(G)$ satisfying the following conditions:

1. $B_N(G) = \{B(G_1) \cup B(G_2)\}$ where $B(G_1)$ and $B(G_2)$ are proper subsets of $B_N(G)$.
2. $(B(G_1), *_1)$ is a neutrosophic group.
3. $(B(G_2), *_2)$ is a group.

Then we define $(B_N(G), *_1, *_2)$ to be a neutrosophic bigroup. If both $B(G_1)$ and $B(G_2)$ are neutrosophic groups. We say $B_N(G)$ is a strong neutrosophic bigroup. If both the groups are not neutrosophic group, we say $B_N(G)$ is just a bigroup.

**Example 1.1.5:** Let $B_N(G) = \{B(G_1) \cup B(G_2)\}$ where $B(G_1) = \{g \,/\, g^9 = 1\}$ be a cyclic group of order 9 and $B(G_2) = \{1, 2, I, 2I\}$ neutrosophic group under multiplication modulo 3. We call $B_N(G)$ a neutrosophic bigroup.

**Example 1.1.6:** Let $B_N(G) = \{B(G_1) \cup B(G_2)\}$, where $B(G_1) = \{1, 2, 3, 4, I, 2I, 3I, 4I\}$ a neutrosophic group under multiplication modulo 5. $B(G_2) = \{0, 1, 2, I, 2I, 1+I, 2+I, 1+2I, 2+2I\}$ is a neutrosophic group under multiplication modulo 3. Clearly $B_N(G)$ is a strong neutrosophic bigroup.

**Definition 1.1.14:** Let $B_N(G) = \{B(G_1) \cup B(G_2), *_1, *_2\}$ be a neutrosophic bigroup. A proper subset $P = \{P_1 \cup P_2, *_1, *_2\}$ is a neutrosophic subbigroup of $B_N(G)$ if the following conditions are satisfied $P = \{P_1 \cup P_2, *_1, *_2\}$ is a neutrosophic bigroup under the operations $*_1, *_2$ i.e. $(P_1, *_1)$ is a neutrosophic subgroup of $(B_1, *_1)$ and $(P_2, *_2)$ is a subgroup of $(B_2, *_2)$. $P_1 = P \cap B_1$ and $P_2 = P \cap B_2$ are subgroups of $B_1$ and $B_2$ respectively. If both of $P_1$ and $P_2$ are not neutrosophic then we call $P = P_1 \cup P_2$ to be just a bigroup.

**Definition 1.1.15:** Let $B_N(G) = \{B(G_1) \cup B(G_2), *_1, *_2\}$ be a neutrosophic bigroup. If both $B(G_1)$ and $B(G_2)$ are commutative groups, then we call $B_N(G)$ to be a commutative bigroup.

**Definition 1.1.16:** Let $B_N(G) = \{B(G_1) \cup B(G_2), *_1, *_2\}$ be a neutrosophic bigroup. If both $B(G_1)$ and $B(G_2)$ are cyclic, we call $B_N(G)$ a cyclic bigroup.

**Definition 1.1.17:** Let $B_N(G) = \{B(G_1) \cup B(G_2), *_1, *_2\}$ be a neutrosophic bigroup. $P(G) = \{P(G_1) \cup P(G_2), *_1, *_2\}$ be a neutrosophic bigroup. $P(G) = \{P(G_1) \cup P(G_2), *_1, *_2\}$ is said to be a neutrosophic normal subbigroup of $B_N(G)$ if $P(G)$ is a neutrosophic subbigroup and both $P(G_1)$ and $P(G_2)$ are normal subgroups of $B(G_1)$ and $B(G_2)$ respectively.

**Definition 1.1.18:** Let $B_N(G) = \{B(G_1) \cup B(G_2), *_1, *_2\}$ be a neutrosophic bigroup of finite order. Let $P(G) = \{P(G_1) \cup P(G_2), *_1, *_2\}$ be a neutrosophic subbigroup of $B_N(G)$. If $o(P(G))/o(B_N(G))$ then we call $P(G)$ a Lagrange neutrosophic subbigroup, if every neutrosophic subbigroup $P$ is such that $o(P)/o(B_N(G))$ then we call $B_N(G)$ to be a Lagrange neutrosophic bigroup.

**Definition 1.1.19:** If $B_N(G)$ has atleast one Lagrange neutrosophic subbigroup then we call $B_N(G)$ to be a weak Lagrange neutrosophic bigroup.

**Definition 1.1.20:** If $B_N(G)$ has no Lagrange neutrosophic subbigroup then $B_N(G)$ is called Lagrange free neutrosophic bigroup.

**Definition 1.1.21:** Let $B_N(G) = \{B(G_1) \cup B(G_2), *_1, *_2\}$ be a neutrosophic bigroup. Suppose $P = \{P(G_1) \cup P(G_2), *_1, *_2\}$ and $K = \{K(G_1) \cup K(G_2), *_1, *_2\}$ be any two neutrosophic subbigroups. we say $P$ and $K$ are conjugate if each $P(G_i)$ is conjugate with $K(G_i), i = 1, 2$, then we say $P$ and $K$ are neutrosophic conjugate subbigroups of $B_N(G)$.

**Definition 1.1.22:** A set $(\langle G \cup I \rangle, +, \circ)$ with two binary operations ` + ' and ` $\circ$ ' is called a strong neutrosophic bigroup if

1. $\langle G \cup I \rangle = \langle G_1 \cup I \rangle \cup \langle G_2 \cup I \rangle$,
2. $(\langle G_1 \cup I \rangle, +)$ is a neutrosophic group and
3. $(\langle G_2 \cup I \rangle, \circ)$ is a neutrosophic group.

**Example 1.1.7:** Let $\{\langle G \cup I \rangle, *_1, *_2\}$ be a strong neutrosophic bigroup where $\langle G \cup I \rangle = \langle Z \cup I \rangle \cup \{0,1,2,3,4,I,2I,3I,4I\}$. $\langle Z \cup I \rangle$ under ` + ' is a neutrosophic group and $\{0,1,2,3,4,I,2I,3I,4I\}$ under multiplication modulo 5 is a neutrosophic group.

**Definition 1.1.23:** A subset $H \neq \phi$ of a strong neutrosophic bigroup $(\langle G \cup I \rangle, *, \circ)$ is called a strong neutrosophic subbigroup if $H$ itself is a strong neutrosophic bigroup under ` * ' and ` $\circ$ ' operations defined on $\langle G \cup I \rangle$.

**Definition 1.1.24:** Let $(\langle G \cup I \rangle, *, \circ)$ be a strong neutrosophic bigroup of finite order. Let $H \neq \phi$ be a strong neutrosophic subbigroup of $(\langle G \cup I \rangle, *, \circ)$. If $o(H)/o(\langle G \cup I \rangle)$ then we call H, a Lagrange strong neutrosophic subbigroup of $\langle G \cup I \rangle$. If every strong neutrosophic subbigroup of $\langle G \cup I \rangle$ is a Lagrange strong neutrosophic subbigroup then we call $\langle G \cup I \rangle$ a Lagrange strong neutrosophic bigroup.

**Definition 1.1.25:** If the strong neutrosophic bigroup has at least one Lagrange strong neutrosophic subbigroup then we call $\langle G \cup I \rangle$ a weakly Lagrange strong neutrosophic bigroup.

**Definition 1.1.26:** If $\langle G \cup I \rangle$ has no Lagrange strong neutrosophic subbigroup then we call $\langle G \cup I \rangle$ a Lagrange free strong neutrosophic bigroup.

**Definition 1.1.27:** Let $(\langle G \cup I \rangle, +, \circ)$ be a strong neutrosophic bigroup with $\langle G \cup I \rangle = \langle G_1 \cup I \rangle \cup \langle G_2 \cup I \rangle$. Let $(H, +, \circ)$ be a neutrosophic subbigroup where $H = H_1 \cup H_2$. We say $H$ is a neutrosophic normal subbigroup of $G$ if both $H_1$ and $H_2$ are neutrosophic normal subgroups of $\langle G_1 \cup I \rangle$ and $\langle G_2 \cup I \rangle$ respectively.

**Definition 1.1.28:** Let $G = \langle G_1 \cup G_2, *, \otimes \rangle$, be a neutrosophic bigroup. We say two neutrosophic strong subbigroups $H = H_1 \cup H_2$ and $K = K_1 \cup K_2$ are conjugate neutrosophic subbigroups of $\langle G \cup I \rangle = \langle G_1 \cup I \rangle \cup \langle G_2 \cup I \rangle$ if $H_1$ is conjugate to $K_1$ and $H_2$ is conjugate to $K_2$ as neutrosophic subgroups of $\langle G_1 \cup I \rangle$ and $\langle G_1 \cup I \rangle$ respectively.

**Definition 1.1.29:** Let $(\langle G \cup I \rangle, *_1, ..., *_N)$ be a nonempty set with $N$-binary operations defined on it. We say $\langle G \cup I \rangle$ is a strong neutrosophic $N$-group if the following conditions are true.

1) $\langle G \cup I \rangle = \langle G_1 \cup I \rangle \cup \langle G_2 \cup I \rangle \cup ... \cup \langle G_N \cup I \rangle$ where $\langle G_i \cup I \rangle$ are proper subsets of $\langle G \cup I \rangle$.
2) $(\langle G_i \cup I \rangle, *_i)$ is a neutrosophic group, $i = 1, 2, ..., N$.

3) If in the above definition we have
   a. $\langle G \cup I \rangle = G_1 \cup \langle G_2 \cup I \rangle \cup ... \cup \langle G_k \cup I \rangle \cup \langle G_{k+1} \cup I \rangle \cup ... \cup G_N$
   b. $(G_i, *_i)$ is a group for some i or
4) $(\langle G_j \cup I \rangle, *_j)$ is a neutrosophic group for some $j$. Then we call $\langle G \cup I \rangle$ to be a neutrosophic $N$-group.

**Example 1.1.8:** Let $\langle G \cup I \rangle = (\langle G_1 \cup I \rangle \cup \langle G_2 \cup I \rangle \cup \langle G_3 \cup I \rangle \cup \langle G_4 \cup I \rangle, *_1, *_2, *_3, *_4)$ be a neutrosophic 4-group where $\langle G_1 \cup I \rangle = \{1, 2, 3, 4, I, 2I, 3I, 4I\}$ neutrosophic group under multiplication modulo 5.
$\langle G_2 \cup I \rangle = \{0, 1, 2, I, 2I, 1+I, 2+I, 1+2I, 2+2I\}$ a neutrosophic group under multiplication modulo 3, $\langle G_3 \cup I \rangle = \langle Z \cup I \rangle$, a neutrosophic group under addition and $\langle G_4 \cup I \rangle = \{(a,b) : a, b \in \{1, I, 4, 4I\}\}$, component-wise multiplication modulo 5.
Hence $\langle G \cup I \rangle$ is a strong neutrosophic 4-group.

**Example 1.1.9:** Let $(\langle G \cup I \rangle = \langle G_1 \cup I \rangle \cup \langle G_2 \cup I \rangle \cup G_3 \cup G_4, *_1, *_2, *_3, *_4)$ be a neutrosophic 4-group, where $\langle G_1 \cup I \rangle = \{1, 2, 3, 4, I, 2I, 3I, 4I\}$ a neutrosophic group under multiplication modulo 5. $\langle G_2 \cup I \rangle = \{0, 1, I, 1+I\}$, a neutrosophic group under multiplication modulo 2. $G_3 = S_3$ and $G_4 = A_5$, the alternating group. $\langle G \cup I \rangle$ is a neutrosophic 4-group.

**Definition 1.1.30:** Let $(\langle G \cup I \rangle = \langle G_1 \cup I \rangle \cup \langle G_2 \cup I \rangle \cup ... \cup \langle G_N \cup I \rangle, *_1, ..., *_N)$ be a neutrosophic $N$-group. A proper subset $(P, *_1, ..., *_N)$ is said to be a neutrosophic sub $N$-group of $\langle G \cup I \rangle$ if $P = (P_1 \cup ... \cup P_N)$ and each $(P_i, *_i)$ is a neutrosophic subgroup (subgroup) of $(G_i, *_i), 1 \leq i \leq N$.
It is important to note $(P, *_i)$ for no $i$ is a neutrosophic group.
Thus we see a strong neutrosophic $N$-group can have 3 types of subgroups viz.

1. Strong neutrosophic sub $N$-groups.
2. Neutrosophic sub $N$-groups.
3. Sub $N$-groups.

Also a neutrosophic $N$-group can have two types of sub $N$-groups.

1. Neutrosophic sub $N$-groups.
2. Sub $N$-groups.

**Definition 1.1.31:** If $\langle G \cup I \rangle$ is a neutrosophic $N$-group and if $\langle G \cup I \rangle$ has a proper subset $T$ such that $T$ is a neutrosophic sub $N$-group and not a strong neutrosophic sub $N$-group and $o(T)/o(\langle G \cup I \rangle)$ then we call $T$ a Lagrange sub $N$-group. If every sub $N$-group of $\langle G \cup I \rangle$ is a Lagrange sub $N$-group then we call $\langle G \cup I \rangle$ a Lagrange $N$-group.

**Definition 1.1.32:** If $\langle G \cup I \rangle$ has atleast one Lagrange sub $N$-group then we call $\langle G \cup I \rangle$ a weakly Lagrange neutrosophic N-group.

**Definition 1.1.33:** If $\langle G \cup I \rangle$ has no Lagrange sub $N$-group then we call $\langle G \cup I \rangle$ to be a Lagrange free $N$-group.

**Definition 1.1.34:** Let $(\langle G \cup I \rangle = \langle G_1 \cup I \rangle \cup \langle G_2 \cup I \rangle \cup ... \cup \langle G_N \cup I \rangle, *_1,...,*_N)$ be a neutrosophic $N$-group. Suppose $H = \{H_1 \cup H_2 \cup ... \cup H_N, *_1,...,*_N\}$ and $K = \{K_1 \cup K_2 \cup ... \cup K_N, *_1,...,*_N\}$ are two sub $N$-groups of $\langle G \cup I \rangle$, we say $K$ is a conjugate to $H$ or $H$ is conjugate to $K$ if each $H_i$ is conjugate to $K_i$ $(i = 1, 2,..., N)$ as subgroups of $G_i$.

## 1.2 Neutrosophic Semigroup, Neutrosophic Bisemigroup, Neutrosophic N-semigroup and their Properties

In this section, we just recall the notions of neutrosophic semigroups, neutrosophic bisemigroups and neutrosophic N-semigroups. We also give briefly some of the important properties of it.

**Definition 1.2.1**: Let $S$ be a semigroup. The semigroup generated by $S$ and $I$ i.e. $S \cup I$ is denoted by $\langle S \cup I \rangle$ is defined to be a neutrosophic semigroup where $I$ is indeterminacy element and termed as neutrosophic element.
It is interesting to note that all neutrosophic semigroups contain a proper subset which is a semigroup.

**Example 1.2.1:** Let $Z = \{$the set of positive and negative integers with zero$\}$, $Z$ is only a semigroup under multiplication. Let $N(S) = \langle Z \cup I \rangle$ be the neutrosophic semigroup under multiplication. Clearly $Z \subset N(S)$ is a semigroup.

**Definition 1.2.2:** Let $N(S)$ be a neutrosophic semigroup. A proper subset $P$ of $N(S)$ is said to be a neutrosophic subsemigroup, if $P$ is a neutrosophic semigroup under the operation of $N(S)$. A neutrosophic semigroup $N(S)$ is said to have a subsemigroup if $N(S)$ has a proper subset which is a semigroup under the operation of $N(S)$.

**Theorem 1.2.1:** Let $N(S)$ be a neutrosophic semigroup. Suppose $P_1$ and $P_2$ be any two neutrosophic subsemigroups of $N(S)$. Then $P_1 \cup P_2$, the union of two neutrosophic subsemigroups in general need not be a neutrosophic subsemigroup.

**Definition 1.2.3:** A neutrosophic semigroup $N(S)$ which has an element $e$ in $N(S)$ such that $e*s = s*e = s$ for all $s \in N(S)$, is called as a neutrosophic monoid.

**Definition 1.2.4:** Let $N(S)$ be a neutrosophic monoid under the binary operation $*$. Suppose $e$ is the identity in $N(S)$, that is $s*e = e*s = s$ for all $s \in N(S)$. We call a proper subset $P$ of $N(S)$ to be a neutrosophic submonoid if
1. $P$ is a neutrosophic semigroup under '$*$'.
2. $e \in P$, i.e. $P$ is a monoid under '$*$'.

**Definition 1.2.5:** Let $N(S)$ be a neutrosophic semigroup under a binary operation $*$. $P$ be a proper subset of $N(S)$. $P$ is said to be a neutrosophic ideal of $N(S)$ if the following conditions are satisfied.

1. $P$ is a neutrosophic semigroup.
2. For all $p \in P$ and for all $s \in N(S)$ we have $p*s$ and $s*p$ are in $P$.

**Definition 1.2.6:** Let $N(S)$ be a neutrosophic semigroup. $P$ be a neutrosophic ideal of $N(S)$, $P$ is said to be a neutrosophic cyclic ideal or neutrosophic principal ideal if $P$ can be generated by a single element.

**Definition 1.2.7:** Let $(BN(S),*,\circ)$ be a non-empty set with two binary operations $*$ and $\circ$. $(BN(S),*,\circ)$ is said to be a neutrosophic bisemigroup if $BN(S) = P_1 \cup P_2$ where atleast one of $(P_1,*)$ or $(P_2,\circ)$ is a neutrosophic semigroup and other is just a semigroup. $P_1$ and $P_2$ are proper subsets of $BN(S)$.

If both $(P_1,*)$ and $(P_2,\circ)$ in the above definition are neutrosophic semigroups then we call $(BN(S),*,\circ)$ a neutrosophic strong bisemigroup. All neutrosophic strong bisemigroups are trivially neutrosophic bisemigroups.

**Example 1.2.2:** Let $(BN(S),*,\circ) = \{0,1,2,3,I,2I,3I,S(3),*,\circ\} = (P_1,*) \cup (P_2,\circ)$ where $(P_1,*) = \{0,1,2,3,,2I,3I\}$ and $(P_2,\circ) = (S(3),\circ)$. Clearly $(P_1,*)$ is a neutrosophic semigroup under multiplication modulo $4$. $(P_2,\circ)$ is just a semigroup. Thus $(BN(S),*,\circ)$ is a neutrosophic bisemigroup.

**Definition 1.2.8:** Let $(BN(S) = P_1 \cup P_2;: *,\circ)$ be a neutrosophic bisemigroup. A proper subset $(T,\circ,*)$ is said to be a neutrosophic subbisemigroup of $BN(S)$ if

1. $T = T_1 \cup T_2$ where $T_1 = P_1 \cap T$ and $T_2 = P_2 \cap T$ and
2. At least one of $(T_1,\circ)$ or $(T_2,*)$ is a neutrosophic semigroup.

**Definition 1.2.9:** Let $(BN(S) = P_1 \cup P, *, \circ)$ be a neutrosophic strong bisemigroup. A proper subset $T$ of $BN(S)$ is called the neutrosophic strong subbisemigroup if $T = T_1 \cup T_2$ with $T_1 = P_1 \cap T$ and $T_2 = P_2 \cap T$ and if both $(T_1, *)$ and $(T_2, \circ)$ are neutrosophic subsemigroups of $(P_1, *)$ and $(P_2, \circ)$ respectively. We call $T = T_1 \cup T_2$ to be a neutrosophic strong subbisemigroup, if atleast one of $(T_1, *)$ or $(T_2, \circ)$ is a semigroup then $T = T_1 \cup T_2$ is only a neutrosophic subsemigroup.

**Definition 1.2.10:** Let $(BN(S) = P_1 \cup P, *, \circ)$ be any neutrosophic bisemigroup. Let $J$ be a proper subset of $BN(S)$ such that $J_1 = J \cap P_1$ and $J_2 = J \cap P_2$ are ideals of $P_1$ and $P_2$ respectively. Then $J$ is called the neutrosophic biideal of $BN(S)$.

**Definition 1.2.11:** Let $(BN(S), *, \circ)$ be a neutrosophic strong bisemigroup where $BN(S) = P_1 \cup P_2$ with $(P_1, *)$ and $(P_2, \circ)$ be any two neutrosophic semigroups. Let $J$ be a proper subset of $BN(S)$ such that $J_1 = J \cap P_1$ and $J_2 = J \cap P_2$ are ideals of $P_1$ and $P_2$ respectively. Then $J$ is called the neutrosophic strong biideal of $BN(S)$.

**Note:** Union of any two neutrosophic biideals in general is not a neutrosophic biideal. This is true of neutrosophic strong biideals.

**Definition 1.2.12:** Let $\{S(N), *_1, ..., *_2\}$ be a non-empty set with $N$-binary operations defined on it. We call $S(N)$ a neutrosophic $N$-semigroup ($N$ a positive integer) if the following conditions are satisfied.

1) $S(N) = S_1 \cup ... S_N$ where each $S_i$ is a proper subset of $S(N)$ i.e. $S_i \subset S_j$ or $S_j \subset S_i$ if $i \neq j$.
2) $(S_i, *_i)$ is either a neutrosophic semigroup or a semigroup for $i = 1, 2, 3, ..., N$.

If all the $N$-semigroups $(S_i, *_i)$ are neutrosophic semigroups (i.e. for $i = 1, 2, 3, ..., N$) then we call $S(N)$ to be a neutrosophic strong $N$-semigroup.

**Example 1.2.3:** Let $S(N) = \{S_1 \cup S_2 \cup S_3 \cup S_4, *_1, *_2, *_3, *_4\}$ be a neutrosophic 4-semigroup where
$S_1 = \{Z_{12}$, semigroup under multiplication modulo $12\}$,
$S_2 = \{1, 2, 3, I, 2I, 3I$, semigroup under multiplication modulo $4\}$, a neutrosophic semigroup.
$S_3 = \left\{ \begin{bmatrix} a & b \\ c & d \end{bmatrix} : a, b, c, d \in \langle R \cup I \rangle \right\}$, neutrosophic semigroup under matrix multiplication and
$S_4 = \langle Z \cup I \rangle$, neutrosophic semigroup under multiplication.

**Definition 1.2.13:** Let $S(N) = \{S_1 \cup S_2 \cup .... S_N, *_1, *_2, ..., *_N\}$ be a neutrosophic $N$-semigroup. A proper subset $P = \{P_1 \cup P_2 \cup .... P_N, *_1, *_2, ..., *_N\}$ of $S(N)$ is said to be a neutrosophic $N$-subsemigroup if $P_i = P \cap S_i, i = 1, 2, ..., N$ are subsemigroups of $S_i$ in which atleast some of the subsemigroups are neutrosophic subsemigroups.

**Definition 1.2.14:** Let $S(N) = \{S_1 \cup S_2 \cup .... S_N, *_1, *_2, ..., *_N\}$ be a neutrosophic strong $N$-semigroup. A proper subset $T = \{T_1 \cup T_2 \cup .... \cup T_N, *_1, *_2, ..., *_N\}$ of $S(N)$ is said to be a neutrosophic strong sub $N$-semigroup if each $(T_i, *_i)$ is a neutrosophic subsemigroup of $(S_i, *_i)$ for $i = 1, 2, ..., N$ where $T_i = S_i \cap T$.

If only a few of the $(T_i, *_i)$ in $T$ are just subsemigroups of $(S_i, *_i)$, (i.e. $(T_i, *_i)$ are not neutrosophic subsemigroups then we call $T$ to be a sub $N$-semigroup of $S(N)$.

**Definition 1.2.15.** Let $S(N) = \{S_1 \cup S_2 \cup .... S_N, *_1, *_2, ..., *_N\}$ be a neutrosophic $N$-semigroup. A proper subset $P = \{P_1 \cup P_2 \cup .... \cup P_N, *_1, *_2, ..., *_N\}$ of $S(N)$ is said to be a neutrosophic $N$-subsemigroup, if the following conditions are true.

1. $P$ is a neutrosophic sub $N$-semigroup of $S(N)$.
2. Each $P_i = S \cap P_i, i = 1, 2, ..., N$ is an ideal of $S_i$.

Then $P$ is called or defined as the neutrosophic $N$-ideal of the neutrosophic $N$-semigroup $S(N)$.

**Definition 1.2.16:** Let $S(N) = \{S_1 \cup S_2 \cup .... S_N, *_1, *_2, ..., *_N\}$ be a neutrosophic strong $N$-semigroup. A proper subset $J = \{J_1 \cup J_2 \cup .... J_N, *_1, *_2, ..., *_N\}$ where $J_t = J \cap S_t$ for $t = 1, 2, ..., N$ is said to be a neutrosophic strong $N$-ideal of $S(N)$ if the following conditions are satisfied.

1. Each it is a neutrosophic subsemigroup of $S_t, t = 1, 2, ..., N$ i.e. It is a neutrosophic strong N-subsemigroup of $S(N)$.
2. Each it is a two sided ideal of $S_t$ for $t = 1, 2, ..., N$.

Similarly one can define neutrosophic strong $N$-left ideal or neutrosophic strong right ideal of $S(N)$.

A neutrosophic strong $N$-ideal is one which is both a neutrosophic strong $N$-left ideal and $N$-right ideal of $S(N)$.

## 1.3 Neutrosophic Loops, Neutrosophic Biloops, Neutrosophic N-loops and their Properties

This section is about the introductory concepts of neutrosophic loops, neutrosophic biloops and neutrosophic N-loops. Their relavent properties are also given in the present section.

**Definition 1.3.1:** A neutrosophic loop is generated by a loop $L$ and $I$ denoted by $\langle L \cup I \rangle$. A neutrosophic loop in general need not be a loop for $I^2 = I$ and $I$ may not have an inverse but every element in a loop has an inverse.

**Example 1.3.1:** Let $\langle L \cup I \rangle = \langle L_7(4) \cup I \rangle$ be a neutrosophic loop which is generated by the loop $L_7(4)$ and $I$.

**Example 1.3.2:** Let $\langle L_{15}(2) \cup I \rangle = \{e, 1, 2, 3, 4, ..., 15, eI, 1I, 2I, ..., 14I, 15I\}$ be another neutrosophic loop of order $32$.

**Definition 1.3.2:** Let $\langle L \cup I \rangle$ be a neutrosophic loop. A proper subset $\langle P \cup I \rangle$ of $\langle L \cup I \rangle$ is called the neutrosophic subloop, if $\langle P \cup I \rangle$ is itself a neutrosophic loop under the operations of $\langle L \cup I \rangle$.

**Definition 1.3.3:** Let $(\langle L \cup I \rangle, \circ)$ be a neutrosophic loop of finite order. A proper subset $P$ of $\langle L \cup I \rangle$ is said to be a Lagrange neutrosophic subloop, if $P$ is a neutrosophic subloop under the operation $\circ$ and $o(P)/o\langle L \cup I \rangle$.

**Definition 1.3.4:** If every neutrosophic subloop of $\langle L \cup I \rangle$ is Lagrange then we call $\langle L \cup I \rangle$ to be a Lagrange neutrosophic loop.

**Definition 1.3.5:** If $\langle L \cup I \rangle$ has no Lagrange neutrosophic subloop then we call $\langle L \cup I \rangle$ to be a Lagrange free neutrosophic loop.

**Definition 1.3.6:** If $\langle L \cup I \rangle$ has atleast one Lagrange neutrosophic subloop then we call $\langle L \cup I \rangle$ to be a weakly Lagrange neutrosophic loop.

**Definition 1.3.7:** Let $(\langle B \cup I \rangle, *_1, *_2)$ be a non-empty set with two binary operations $*_1, *_2$, $\langle B \cup I \rangle$ is a neutrosophic biloop if the following conditions are satisfied.

1. $\langle B \cup I \rangle = P_1 \cup P_2$ where $P_1$ and $P_2$ are proper subsets of $\langle B \cup I \rangle$.
2. $(P_1, *_1)$ is a neutrosophic loop.
3. $(P_2, *_2)$ is a group or a loop.

**Example 1.3.3:** Let $B = (B_1 \cup B_2, *_1, *_2)$ be a neutrosophic biloop of order 20, where $B_1 = \langle L_5(3) \cup I \rangle$ and $B_2 = \{g : g^8 = e\}$.

**Definition 1.3.8:** Let $(\langle B \cup I \rangle, *_1, *_2)$ be a neutrosophic biloop. A proper subset $P$ of $\langle B \cup I \rangle$ is said to be a neutrosophic subbiloop of $\langle B \cup I \rangle$ if $P = (P_1 \cup P_2, *_1, *_2)$ is itself a neutrosophic biloop under the operations of $\langle B \cup I \rangle$.

**Definition 1.3.9:** Let $B = (B_1 \cup B_2, *_1, *_2)$ be a finite neutrosophic biloop. Let $P = (P_1 \cup P_2, *_1, *_2)$ be a neutrosophic biloop. If $o(P)/o(B)$ then we call $P$ to be a Lagrange neutrosophic subbiloop of $B$.

**Definition 1.3.10:** If every neutrosophic subbiloop of $B$ is Lagrange then we call $B$ to be a Lagrange neutrosophic biloop.

**Definition 1.3.11:** If $B$ has atleast one Lagrange neutrosophic subbiloop then we call $B$ to be a weakly Lagrange neutrosophic biloop.

**Definition 1.3.12:** If $B$ has no Lagrange neutrosophic subbiloops then we call $B$ to be a Lagrange free neutrosophic biloop.

**Definition 1.3.13:** Let $S(B) = \{S(B_1) \cup S(B_2) \cup ... \cup S(B_n), *_1, *_2, ..., *_N\}$ be a non-empty neutrosophic set with $N$-binary operations. $S(B)$ is a neutrosophic $N$-loop if $S(B) = S(B_1) \cup S(B_2) \cup ... \cup S(B_n)$, $S(B_i)$ are proper subsets of $S(B)$ for $1 \leq i \leq N$ and some of $S(B_i)$ are neutrosophic loops and some of the $S(B_i)$ are groups.

**Example 1.3.4:** Let $S(B) = \{S(B_1) \cup S(B_2) \cup S(B_3), *_1, *_2, *_3\}$ be a neutrosophic 3-loop, where $S(B_1) = \langle L_5(3) \cup I \rangle$, $S(B_2) = \{g : g^{12} = e\}$ and $S(B_3) = S_3$.

**Definition 1.3.14:** Let $S(B) = \{S(B_1) \cup S(B_2) \cup ... \cup S(B_n), *_1, *_2, ..., *_N\}$ be a neutrosophic $N$-loop. A proper subset $(P, *_1, *_2, ..., *_N)$ of $S(B)$ is said to be a neutrosophic sub $N$-loop of $S(B)$ if $P$ itself is a neutrosophic $N$-loop under the operations of $S(B)$.

**Definition 1.3.15:** Let $(L = L_1 \cup L_2 \cup ... \cup L_N, *_1, *_2, ..., *_N)$ be a neutrosophic $N$-loop of finite order. Suppose $P$ is a proper subset of $L$, which is a neutrosophic sub $N$-loop. If $o(P)/o(L)$ then we call $P$ a Lagrange neutrosophic sub $N$-loop.

**Definition 1.3.16:** If every neutrosophic sub $N$-loop is Lagrange then we call $L$ to be a Lagrange neutrosophic $N$-loop.

**Definition 1.3.17:** If $L$ has atleast one Lagrange neutrosophic sub $N$-loop then we call $L$ to be a weakly Lagrange neutrosophic $N$-loop.

**Definition 1.3.18:** If $L$ has no Lagrange neutrosophic sub $N$-loop then we call $L$ to be a Lagrange free neutrosophic $N$-loop.

## 1.4 Neutrosophic LA-semigroup, Neutrosophic Bi-LA-semigroup, Neutrosophic N-LA-semigroup and their Properties

In this section we mention some of the basic and fundamental concepts of neutrosophic LA-semigroup, neutrosophic bi-LA-semigroup and neutrosophic N-LA-semigroup.

**Definition 1.4.1:** Let $(S,*)$ be an LA-semigroup and let $\langle S \cup I \rangle = \{a+bI : a,b \in S\}$. The neutrosophic LA-semigroup is generated by $S$ and $I$ under $*$ denoted as $N(S) = \{\langle S \cup I \rangle, *\}$, where $I$ is called the neutrosophic element with property $I^2 = I$. For an integer $n$, $n+I$ and $nI$ are neutrosophic elements and $0.I = 0$.

$I^{-1}$, the inverse of $I$ is not defined and hence does not exist.

Similarly we can define neutrosophic RA-semigroup on the same lines.

**Example 1.4.1:** Let $N(S) = \{1,2,3,4,1I,2I,3I,4I\}$ be a neutrosophic LA-semigroup with the following table.

| *  | 1  | 2  | 3  | 4  | 1I | 2I | 3I | 4I |
|----|----|----|----|----|----|----|----|----|
| 1  | 1  | 4  | 2  | 3  | 1I | 4I | 2I | 3I |
| 2  | 3  | 2  | 4  | 1  | 3I | 2I | 4I | 1I |
| 3  | 4  | 1  | 3  | 2  | 4I | 1I | 3I | 2I |
| 4  | 2  | 3  | 1  | 4  | 2I | 3I | 1I | 4I |
| 1I | 1I | 4I | 2I | 3I | 1I | 4I | 2I | 3I |
| 2I | 3I | 2I | 4I | 1I | 3I | 2I | 4I | 1I |
| 3I | 4I | 1I | 3I | 2I | 4I | 1I | 3I | 2I |
| 4I | 2I | 3I | 1I | 4I | 2I | 3I | 1I | 4I |

**Definition 1.4.2:** Let $N(S)$ be a neutrosophic LA-semigroup. An element $e \in N(S)$ is said to be left identity if $e*s = s$ for all $s \in N(S)$. Similarly $e$ is called right identity if $s*e = s$ for all $s \in N(S)$. $e$ is called two sided identity or simply identity if $e$ is left as well as right identity.

**Definition 1.4.3:** Let $N(S)$ be a neutrosophic LA-semigroup and $N(H)$ be a proper subset of $N(S)$. Then $N(H)$ is called a neutrosophic sub LA-semigroup if $N(H)$ itself is a neutrosophic LA-semigroup under the operation of $N(S)$.

**Definition 1.4.4:** A neutrosophic sub LA-semigroup $N(H)$ is called strong neutrosophic sub LA-semigroup or pure neutrosophic sub LA-semigroup if all the elements of $N(H)$ are neutrosophic elements.

**Definition 1.4.5:** Let $N(S)$ be a neutrosophic LA-semigroup and $N(K)$ be a subset of $N(S)$. Then $N(K)$ is called Left (right) neutrosophic ideal of $N(S)$ if $N(S)N(K) \subseteq N(K), \{N(K)N(S) \subseteq N(K)\}$.

If $N(K)$ is both left and right neutrosophic ideal, then $N(K)$ is called a two sided neutrosophic ideal or simply a neutrosophic ideal.

**Definition 1.4.6:** A neutrosophic ideal $N(K)$ is called strong neutrosophic ideal or pure neutrosophic ideal if all of its elements are neutrosophic elements.

**Definition 1.4.7:** Let $(BN(S), *, \circ)$ be a non-empty set with two binary operations $*$ and $\circ$. $(BN(S), *, \circ)$ is said to be a neutrosophic bi-LA-semigroup if $BN(S) = P_1 \cup P_2$ where atleast one of $(P_1, *)$ or $(P_2, \circ)$ is a neutrosophic LA-semigroup and other is just an LA- semigroup. $P_1$ and $P_2$ are proper subsets of $BN(S)$.

If both $(P_1, *)$ and $(P_2, \circ)$ in the above definition are neutrosophic LA-semigroups then we call $(BN(S), *, \circ)$ a neutrosophic strong bi-LA-semigroup.

**Example 1.4.2:** Let $BN(S) = \{\langle S_1 \cup I \rangle \cup \langle S_2 \cup I \rangle\}$ be a neutrosophic bi-LA-semigroup, where $\langle S_1 \cup I \rangle = \{1, 2, 3, 1I, 2I, 3I\}$ is a neutrosophic bi-LA-semigroup and $\langle S_2 \cup I \rangle = \{1, 2, 3, I, 2I, 3I\}$ is another neutrosophic LA-semigroup.

**Definition 1.4.8:** Let $(BN(S) = P_1 \cup P_2; *, \circ)$ be a neutrosophic bi-LA-semigroup. A proper subset $(T, \circ, *)$ is said to be a neutrosophic sub bi-LA-semigroup of $BN(S)$ if

1. $T = T_1 \cup T_2$ where $T_1 = P_1 \cap T$ and $T_2 = P_2 \cap T$ and
2. At least one of $(T_1, \circ)$ or $(T_2, *)$ is a neutrosophic LA-semigroup.

**Definition 1.4.9:** Let $(BN(S) = P_1 \cup P_2, *, \circ)$ be a neutrosophic bi-LA-semigroup. A proper subset $(T, \circ, *)$ is said to be a neutrosophic strong sub bi-LA-semigroup of $BN(S)$ if

1. $T = T_1 \cup T_2$ where $T_1 = P_1 \cap T$ and $T_2 = P_2 \cap T$ and
2. $(T_1, \circ)$ and $(T_2, *)$ are neutrosophic strong LA-semigroups.

**Definition 1.4.10:** Let $(BN(S) = P_1 \cup P_2, *, \circ)$ be any neutrosophic bi-LA-semigroup. Let $J$ be a proper subset of $BN(S)$ such that $J_1 = J \cap P_1$ and $J_2 = J \cap P_2$ are ideals of $P_1$ and $P_2$ respectively. Then $J$ is called the neutrosophic biideal of $BN(S)$.

**Definition 1.4.11:** Let $(BN(S), *, \circ)$ be a strong neutrosophic bi-LA-semigroup where $BN(S) = P_1 \cup P_2$ with $(P_1, *)$ and $(P_2, \circ)$ be any two neutrosophic LA-semigroups. Let $J$ be a proper subset of $BN(S)$ where $J = J_1 \cup J_2$ with $J_1 = J \cap P_1$ and $J_2 = J \cap P_2$ are neutrosophic ideals of the neutrosophic LA-semigroups $P_1$ and $P_2$ respectively. Then $J$ is called or defined as the strong neutrosophic biideal of $BN(S)$.

**Definition 1.4.12:** Let $\{S(N), *_1, ..., *_2\}$ be a non-empty set with $N$-binary operations defined on it. We call $S(N)$ a neutrosophic $N$-LA-semigroup ($N$ a positive integer) if the following conditions are satisfied.

1. $S(N) = S_1 \cup ... S_N$ where each $S_i$ is a proper subset of $S(N)$.
2. $(S_i, *_i)$ is either a neutrosophic LA-semigroup or an LA-semigroup for $i = 1, 2, 3, ..., N$.

If all the $N$-LA-semigroups $(S_i, *_i)$ are neutrosophic LA-semigroups (i.e. for $i = 1, 2, 3, ..., N$) then we call $S(N)$ to be a neutrosophic strong $N$-LA-semigroup.

**Example 1.4.3:** Let $S(N) = \{S_1 \cup S_2 \cup S_3, *_1, *_2, *_3\}$ be a neutrosophic 3-LA-semigroup where $S_1 = \{1,2,3,4,1I,2I,3I,4I\}$ is a neutrosophic LA-semigroup, $S_2 = \{1,2,3,1I,2I,3I\}$ be another neutrosophic bi-LA-semigroup with the following table and $S_3 = \{1,2,3,I,2I,3I\}$ is a neutrosophic LA-semigroup.

**Definition 1.4.13:** Let $S(N) = \{S_1 \cup S_2 \cup ....S_N, *_1, *_2,...,*_N\}$ be a neutrosophic $N$-LA-semigroup. A proper subset $P = \{P_1 \cup P_2 \cup ....P_N, *_1, *_2,...,*_N\}$ of $S(N)$ is said to be a neutrosophic sub $N$-LA-semigroup if $P_i = P \cap S_i, i = 1,2,...,N$ are sub LA-semigroups of $S_i$ in which atleast some of the sub LA-semigroups are neutrosophic sub LA-semigroups.

**Definition 1.4.14:** Let $S(N) = \{S_1 \cup S_2 \cup ....S_N, *_1, *_2,...,*_N\}$ be a neutrosophic strong $N$-LA-semigroup. A proper subset $T = \{T_1 \cup T_2 \cup .... \cup T_N, *_1, *_2,...,*_N\}$ of $S(N)$ is said to be a neutrosophic strong sub $N$-LA-semigroup if each $(T_i, *_i)$ is a neutrosophic sub LA-semigroup of $(S_i, *_i)$ for $i = 1,2,...,N$ where $T_i = S_i \cap T$.

**Definition 1.4.15:** Let $S(N) = \{S_1 \cup S_2 \cup ....S_N, *_1, *_2,...,*_N\}$ be a neutrosophic $N$-LA-semigroup. A proper subset $P = \{P_1 \cup P_2 \cup .... \cup P_N, *_1, *_2,...,*_N\}$ of $S(N)$ is said to be a neutrosophic $N$-ideal, if the following conditions are true.

1. $P$ is a neutrosophic sub $N$-LA-semigroup of $S(N)$.
2. Each $P_i = S \cap P_i, i = 1,2,...,N$ is an ideal of $S_i$.

**Definition 1.4.16:** Let $S(N) = \{S_1 \cup S_2 \cup .... S_N, *_1, *_2, ..., *_N\}$ be a neutrosophic strong $N$-LA-semigroup. A proper subset $J = \{J_1 \cup J_2 \cup .... J_N, *_1, *_2, ..., *_N\}$ where $J_t = J \cap S_t$ for $t = 1, 2, ..., N$ is said to be a neutrosophic strong $N$-ideal of $S(N)$ if the following conditions are satisfied.

1) Each it is a neutrosophic sub LA-semigroup of $S_t, t = 1, 2, ..., N$ i.e. It is a neutrosophic strong N-sub LA-semigroup of $S(N)$.
2) Each it is a two sided ideal of $S_t$ for $t = 1, 2, ..., N$.

Similarly one can define neutrosophic strong $N$-left ideal or neutrosophic strong right ideal of $S(N)$.

A neutrosophic strong $N$-ideal is one which is both a neutrosophic strong $N$-left ideal and $N$-right ideal of $S(N)$.

## 1.5 Soft Sets and their Properties

In this section, we give the the definition of soft set and their related properties are also mention in the present section.

Throughout this section $U$ refers to an initial universe, $E$ is a set of parameters, $P(U)$ is the power set of $U$, and $A, B \subset E$.

**Definition 1.5.1:** A pair $(F, A)$ is called a soft set over $U$ where $F$ is a mapping given by $F : A \to P(U)$.
In other words, a soft set over $U$ is a parameterized family of subsets of the universe $U$. For $a \in A$, $F(a)$ may be considered as the set of $a$-elements of the soft set $(F, A)$, or as the set of $a$-approximate elements of the soft set.

**Example 1.5.1:** Suppose that $U$ is the set of shops. $E$ is the set of parameters and each parameter is a word or sentence. Let

$$E = \left\{ \begin{array}{l} \text{high rent,normal rent,} \\ \text{in good condition,in bad condition} \end{array} \right\}.$$

Let us consider a soft set $(F, A)$ which describes the attractiveness of shops that Mr. $Z$ is taking on rent. Suppose that there are five houses in the universe $U = \{s_1, s_2, s_3, s_4, s_5\}$ under consideration, and that $A = \{a_1, a_2, a_3\}$ be the set of parameters where
$a_1$ stands for the parameter 'high rent,
$a_2$ stands for the parameter 'normal rent,
$a_3$ stands for the parameter 'in good condition.
Suppose that
$$F(a_1) = \{s_1, s_4\},$$
$$F(a_2) = \{s_2, s_5\},$$
$$F(a_3) = \{s_3\}.$$

The soft set $(F, A)$ is an approximated family $\{F(a_i), i = 1, 2, 3\}$ of subsets of the set $U$ which gives us a collection of approximate description of an object. Then $(F, A)$ is a soft set as a collection of approximations over $U$, where
$$F(a_1) = high\ rent = \{s_1, s_2\},$$
$$F(a_2) = normal\ rent = \{s_2, s_5\},$$
$$F(a_3) = in\ good\ condition = \{s_3\}.$$

**Definition 1.5.2:** For two soft sets $(F, A)$ and $(H, B)$ over $U$, $(F, A)$ is called a soft subset of $(H, B)$ if
1. $A \subseteq B$ and
2. $F(a) \subseteq H(a)$, for all $x \in A$.

This relationship is denoted by $(F, A) \subset (H, B)$. Similarly $(F, A)$ is called a soft superset of $(H, B)$ if $(H, B)$ is a soft subset of $(F, A)$ which is denoted by $(F, A) \supset (H, B)$.

**Definition 1.5.3:** Two soft sets $(F, A)$ and $(H, B)$ over $U$ are called soft equal if $(F, A)$ is a soft subset of $(H, B)$ and $(H, B)$ is a soft subset of $(F, A)$.

**Definition 1.5.4:** Let $(F, A)$ and $(K, B)$ be two soft sets over a common universe $U$ such that $A \cap B \neq \phi$. Then their restricted intersection is denoted by $(F, A) \cap_R (K, B) = (H, C)$ where $(H, C)$ is defined as $H(c) = F(c) \cap K(c)$ for all $c \in C = A \cap B$.

**Definition 1.5.5:** The extended intersection of two soft sets $(F, A)$ and $(K, B)$ over a common universe $U$ is the soft set $(H, C)$, where $C = A \cup B$, and for all $c \in C$, $H(c)$ is defined as
$$H(c) = \begin{cases} F(c) & \text{if } c \in A - B, \\ G(c) & \text{if } c \in B - A, \\ F(c) \cap G(c) & \text{if } c \in A \cap B. \end{cases}$$
We write $(F, A) \cap_\varepsilon (K, B) = (H, C)$.

**Definition 1.5.6:** The restricted union of two soft sets $(F, A)$ and $(K, B)$ over a common universe $U$ is the soft set $(H, C)$, where $C = A \cup B$, and for all $c \in C$, $H(c)$ is defined as $H(c) = F(c) \cup G(c)$ for all $c \in C$. We write it as $(F, A) \cup_R (K, B) = (H, C)$.

**Definition 1.5.7:** The extended union of two soft sets $(F, A)$ and $(K, B)$ over a common universe $U$ is the soft set $(H, C)$, where $C = A \cup B$, and for all $c \in C$, $H(c)$ is defined as

$$H(c) = \begin{cases} F(c) & \text{if } c \in A - B, \\ G(c) & \text{if } c \in B - A, \\ F(c) \cup G(c) & \text{if } c \in A \cap B. \end{cases}$$

We write $(F, A) \cup_{\varepsilon} (K, B) = (H, C)$.

# Chapter Two

# SOFT NEUTROSOPHIC GROUPS AND THEIR GENERALIZATION

This chapter has three sections. In first section, we introduce the important notions of soft neutrosophic groups over neutrosophic groups which is infact a collection of parameterized family of neutrosophic subgroups and we also give some of their characterization with sufficient amount of examples in this section. The second section deal with soft neutrosophic bigroups which are basically defined over neutrosophic bigroups. We also establish some basic and fundamental results of soft neutrosophic bigroups. The third section is about the generalization of soft neutrosophic groups. We defined soft neutrosophic N-groups over neutrosophic N-groups in this section and some of their basic properties are also investigated.

## 2.1 Soft Neutrosophic Group

In this section we introduced soft neutrosophic group which is a parameterized family of neutrosophic subgroups of the neutrosophic group. We also dfined a new notion called soft pseudo neutrosophic group. We study some of its interesting properties and explained this notion with many examples.

**Definition 2.1.1:** Let $N(G)$ be a neutrosophic group and $(F,A)$ be soft set ove r $N(G)$. Then $(F,A)$ is called soft neutrosophic group over $N(G)$ if and only if $F(a) \prec N(G)$, for all $a \in A$.

This situation is explained with the help of following examples.

**Example 2.1.1:** Let

$$N(Z_4) = \begin{Bmatrix} 0,1,2,3,I,2I,3I,1+I,1+2I,1+3I, \\ 2+I,2+2I,2+3I,3+I,3+2I,3+3I \end{Bmatrix}$$

be a neutrosophic group under addition modulo 4. and let $A = \{a_1, a_2, a_3, a_4\}$ be a set of parameters. Then $(F, A)$ is soft neutrosophic group over $N(Z_4)$, where

$$F(a_1) = \{0,1,2,3\}, F(a_2) = \{0,I,2I,3I\},$$
$$F(a_3) = \{0,2,2I,2+2I\},$$
$$F(a_4) = \{0,I,2I,3I,2,2+2I,2+I,2+3I\}.$$

**Example 2.1.2:** Let
$$N(G) = \{e,a,b,c,I,aI,bI,cI\}$$
be a neutrosophic group under multiplication where
$$a^2 = b^2 = c^2 = e, bc = cb = a, ac = ca = b, ab = ba = c.$$
Then $(F,A)$ is a soft neutrosophic group over $N(G)$, where

$$F(a_1) = \{e,a,I,aI\},$$
$$F(a_2) = \{e,b,I,bI\},$$
$$F(a_3) = \{e,c,I,cI\}.$$

**Example 2.1.3:** Let $(F,A)$ be a soft neutrosophic groups over $N(Z_2)$ under addition modulo $2$, where

$$F(a_1) = \{0,1\}, F(a_2) = \{0,I\}.$$

**Example 2.1.4:** *Let* $N(G) = \begin{Bmatrix} 0,1,2,3,4,5,I,2I,3I,4I,5I, \\ 1+I,2+I,3+I,...,5+5I \end{Bmatrix}$ *be a neutrosophic group under addition modulo* $6$. *Then* $(F,A)$ *is soft neutrosophic group over* $N(G)$, *where*
$$F(a_1) = \{0,3,3I,3+3I\},$$
$$F(a_2) = \{0,2,4,2+2I,4+4I,2I,4I\}.$$

**Theorem 2.1.1:** *Let* $(F,A)$ *and* $(H,A)$ *be two soft neutrosophic groups over* $N(G)$. *Then their intersection* $(F,A) \cap (H,A)$ *is again a soft neutrosophic group over* $N(G)$.

**Theorem 2.1.2:** Let $(F,A)$ and $(H,B)$ be two soft neutrosophic groups over $N(G)$. If $A \cap B = \phi$, then $(F,A) \cup (H,B)$ is a soft neutrosophic group over $N(G)$.

**Proposition 2.1.1:** *The extended intersection of two soft neutrosophic groups over* $N(G)$ *is soft neutrosophic group over* $N(G)$.

**Proposition 2.1.2:** *The restricted intersection of two soft neutrosophic groups over* $N(G)$ *is soft neutrosophic group over* $N(G)$.

**Proposition 2.1.3:** *The AND operation of two soft neutrosophic groups over* $N(G)$ *is soft neutrosophic group over* $N(G)$.

**Remark 2.1.1:** The extended union of two soft neutrosophic groups $(F,A)$ and $(K,B)$ over a neutrosophic group $N(G)$ is not a soft neutrosophic group over $N(G)$.

This is proved by the following example.

**Example 2.1.5:** Let $N(Z_2) = \{0, 1, I, 1+I\}$ be a neutrosophic group under addition modulo $2$. Let $A = \{a_1, a_2\}$ and $B = \{a_2, a_3\}$ be the set of parameters. Let $(F, A)$ and $(K, B)$ be two soft neutrosophic groups over $N(Z_2)$ under addition modulo $2$, where

$$F(a_1) = \{0, 1\}, F(a_2) = \{0, I\},$$

and

$$K(a_2) = \{0, 1\}, K(a_3) = \{0, 1+I\}.$$

Let

$$C = A \cap B = \{a_2\}.$$

Then clearly their extended union is not a soft neutrosophic group as

$$H(a_2) = F(a_2) \cup K(a_2) = \{0, 1, I\}$$

is not a neutrosophic subgroup of $N(Z_2)$.

**Reamrk 2.1.2:** *The restricted union of two soft neutrosophic groups $(F, A)$ and $(K, B)$ over $N(G)$ is not a soft neutrosophic group over $N(G)$.*

To prove the above remark, lets take a look to the following example.

**Example 2.1.6:** Let $N(Z_2) = \{0, 1, I, 1+I\}$ be a neutrosophic group under addition modulo $2$. Let $A = \{a_1, a_2\}$ and $B = \{a_2, a_3\}$ be the set of parameters. Let $(F, A)$ and $(K, B)$ be two soft neutrosophic groups over $N(Z_2)$ under addition modulo $2$, where

$$F(a_1) = \{0, 1\}, F(a_2) = \{0, I\},$$

*and*

$$K(a_2) = \{0,1\}, K(a_3) = \{0, 1+I\}.$$

Let

$$C = A \cap B = \{a_2\}.$$

Then clearly their extended union is not a soft neutrosophic group as

$$H(a_2) = F(a_2) \cup K(a_2) = \{0, 1, I\}$$

is not a neutrosophic subgroup of $N(Z_2)$.

Smilarly the $OR$ operation of two soft neutrosophic groups over $N(G)$ may not be a soft neutrosophic group.

**Definition 2.1.2:** A soft neutrosophic group $(F, A)$ over $N(G)$ which does not contain a proper soft group is called soft pseudo neutrosophic group.

Equivalently $(F, A)$ over $N(G)$ is a soft pseudo neutrosophic group if and only if each $F(a)$ is a pseudo neutrosophic group, for all $a \in A$.

The illustration is given by the following example.

**Example 2.1.7:** Let $N(Z_2) = \langle Z_2 \cup I \rangle = \{0, 1, I, 1+I\}$ be a neutrosophic group under addition modulo $2$. Let $A = \{a_1, a_2, a_3\}$ be the set of parameters. Then $(F, A)$ is a soft pseudo neutrosophic group over $N(G)$, where

$$F(a_1) = \{0, 1\},$$
$$F(a_2) = \{0, I\},$$
$$F(a_3) = \{0, 1+I\}.$$

**Theorem 2.1.3:** *Every soft pseudo neutrosophic group is a soft neutrosophic group.*

*The proof of this theorem is left as an exercise to the readers.*

The converse of the above theorem does not hold. The following example illustrate this fact.

**Example 2.1.8:** Let $N(Z_4)$ be a neutrosophic group and $(F, A)$ be a soft neutrosophic group over $N(Z_4)$, where

$$F(a_1) = \{0, 1, 2, 3\}, F(a_2) = \{0, I, 2I, 3I\},$$
$$F(a_3) = \{0, 2, 2I, 2+2I\}.$$

But $(F, A)$ is not a soft pseudo neutrosophic group as $(H, B)$ is clearly a proper soft subgroup of $(F, A)$, where

$$H(a_1) = \{0, 2\}, H(a_2) = \{0, 2\}.$$

**Theorem 2.1.4:** $(F,A)$ over $N(G)$ is a soft pseudo neutrosophic group if $N(G)$ is a pseudo neutrosophic group.

**Proof:** Suppose that $N(G)$ be a pseudo neutrosophic group. Then it does not contain a proper group and for all $x \in A$, the soft neutrosophic group $(F,A)$ over $N(G)$ is such that $F(x) \prec N(G)$. Since each $F(x)$ is a pseudo neutrosophic subgroup which does not contain a proper group and this makes $(F,A)$ is soft pseudo neutrosophic group.

**Proposition 2.1.4:** *The extended intersection of two soft pseudo neutrosophic groups $(F,A)$ and $(K,B)$ over $N(G)$ is a soft pseudo neutrosophic group over $N(G)$.*

**Proposition 2.1.5:** *The restricted intersection of two soft pseudo neutrosophic groups $(F,A)$ and $(K,B)$ over $N(G)$ is a soft pseudo neutrosophic group over $N(G)$.*

**Proposition 2.1.6:** *The AND operation of two soft pseudo neutrosophic groups over $N(G)$ is soft pseudo neutrosophic group over $N(G)$.*

**Remark 2.1.3:** *The extended union of two soft pseudo neutrosophic groups $(F,A)$ and $(K,B)$ over $N(G)$ is not a soft pseudo neutrosophic group over $N(G)$.*

The proof of this remark is shown in the following example.

**Example 2.1.9:** Let $N(Z_2) = \langle Z_2 \cup I \rangle = \{0, 1, I, 1+I\}$ be a neutrosophic group under addition modulo 2. Let $(F, A)$ and $(K, B)$ be two soft pseudo neutrosophic groups over $N(G)$, where

$$F(a_1) = \{0,1\}, F(a_2) = \{0, I\},$$
$$F(a_3) = \{0, 1+I\}.$$

and

$$K(a_1) = \{0, 1+I\}, K(a_2) = \{0,1\}.$$

Clearly their restricted union is not a soft pseudo neutrosophic group as union of two subgroups is not a subgroup.

**Remark 2.1.4:** *The restricted union of two soft pseudo neutrosophic groups $(F, A)$ and $(K, B)$ over $N(G)$ is not a soft pseudo neutrosophic group over $N(G)$.*

One can see it easily by checking an example.

**Note:** Similarly *the OR operation of two soft pseudo neutrosophic groups over $N(G)$ may not be a soft pseudo neutrosophic group.*

**Definition 2.1.3:** *Let $(F, A)$ and $(H, B)$ be two soft neutrosophic groups over $N(G)$. Then $(H, B)$ is a soft neutrosophic subgroup of $(F, A)$, denoted as $(H, B) < (F, A)$, if*

1. $B \subseteq A$
2. $H(a) \prec F(a)$, for all $a \in A$.

**Example 2.1.10:** Let $N(Z_4) = \langle Z_4 \cup I \rangle$ be a soft neutrosophic group under addition modulo $4$, that is

$$N(Z_4) = \begin{Bmatrix} 0,1,2,3,I,2I,3I,1+I,1+2I,1+3I, \\ 2+I,2+2I,2+3I,3+I,3+2I,3+3I \end{Bmatrix}.$$

Let $(F, A)$ be a soft neutrosophic group over $N(Z_4)$. Then

$$F(a_1) = \{0,1,2,3\}, F(a_2) = \{0, I, 2I, 3I\},$$
$$F(a_3) = \{0, 2, 2I, 2+2I\},$$
$$F(a_4) = \{0, I, 2I, 3I, 2, 2+2I, 2+I, 2+3I\}.$$

$(H, B)$ is a soft neutrosophic subgroup of $(F, A)$, where

$$H(a_1) = \{0, 2\}, H(a_2) = \{0, 2I\},$$
$$H(a_4) = \{0, I, 2I, 3I\}.$$

**Theorem 2.1.5:** *A soft group over $G$ is always a soft neutrosophic subgroup of a soft neutrosophic group over $N(G)$ if $A \subset B$.*

**Proof:** Let $(F, A)$ be a soft neutrosophic group over $N(G)$ and $(H, B)$ be a soft group over $G$. As $G \subset N(G)$ and for all $a \in B, H\ a\ \prec G \subset N\ G$. This implies $H\ a\ \prec F\ a$, for all $a \in A$ as $B \subset A$. Hence $(H, B) < (F, A)$.

**Example 2.1.11:** *Let* $(F, A)$ *be a soft neutrosophic group over* $N(Z_4)$.

*Then*
$$F(a_1) = \{0,1,2,3\}, F(a_2) = \{0, I, 2I, 3I\},$$
$$F(a_3) = \{0, 2, 2I, 2+2I\}.$$

*Let* $B = \{a_1, a_3\}$ *such that* $(H, B) \prec (F, A)$, *where*

$$H(a_1) = \{0,2\}, H(a_3) = \{0,2\}.$$

*Clearly* $B \subset A$ *and* $H(a) \prec F(a)$ *for all* $a \in B$.

**Theorem 2.1.6:** *A soft neutrosophic group over* $N(G)$ *always contains a soft group over G.*

**Proof:** The proof is left as an exercise for the readers.

**Definition 2.1.4:** *Let* $(F, A)$ *and* $(H, B)$ *be two soft pseudo neutrosophic groups over* $N(G)$. *Then* $(H, B)$ *is called soft pseudo neutrosophic subgroup of* $(F, A)$, *denoted as* $(H, B) < (F, A)$, *if*

1. $B \subseteq A$
2. $H(a) \prec F(a)$, for all $a \in A$.

We explain this situation by the following example.

**Example 2.1.12:** *Let* $(F,A)$ *be a soft pseudo neutrosophic group over* $N(Z_4)$, *where*

$$F(a_1) = \{0, I, 2I, 3I\}, F(a_2) = \{0, 2I\}.$$

Hence $(H, B) \prec (F, A)$ where

$$H(a_1) = \{0, 2I\}.$$

**Theorem 2.1.7:** Every soft neutrosophic group $(F, A)$ over $N(G)$ has soft neutrosophic subgroup as well as soft pseudo neutrosophic subgroup.

**Proof:** Straightforward.

**Definition 2.1.5:** Let $(F, A)$ be a soft neutrosophic group over $N(G)$. Then $(F, A)$ is called the identity soft neutrosophic group over $N(G)$ if $F(a) = \{e\}$, for all $a \in A$, where $e$ is the identity element of $N(G)$.

**Definition 2.1.6:** Let $(H, B)$ be a soft neutrosophic group over $N(G)$. Then $(F, A)$ is called an absolute-soft neutrosophic group over $N(G)$ if $F(a) = N(G)$, for all $a \in A$.

**Example 14:** *Let*

$$N(R) = \left\{ \begin{array}{l} a + bI \ : \ a,b \in R \text{ and} \\ I \text{ is indeterminacy} \end{array} \right\}$$

*is a neutrosophic real group where $R$ is set of real numbers and $I^2 = I$, therefore $I^n = I$, for $n$ a positive integer. Then $(F,A)$ is a an absolute-soft neutrosophic real group where*

$$F(a) = N(R), \text{ for all } a \in A.$$

**Theorem 2.1.8:** Every absolute-soft neutrosophic group over a neutrosophic group contain absolute soft group over a group.

**Theorem 2.1.9:** Every absolute soft group over $G$ is a soft neutrosophic subgroup of absolute-soft neutrosophic group over $N(G)$.

**Theorem 2.1.10:** Let $N(G)$ be a neutrosophic group. If order of $N(G)$ is prime number, then the soft neutrosophic group $(F,A)$ over $N(G)$ is either identity soft neutrosophic group or absolute-soft neutrosophic group.

**Proof** Straightforward.

**Definition 2.1.7:** Let $(F,A)$ be a soft neutrosophic group over $N(G)$. If for all $a \in A$, each $F(a)$ is Lagrange neutrosophic subgroup of $N(G)$, then $(F,A)$ is called soft Lagrange neutrosophic group over $N(G)$.

**Example 2.1.13:** *Let $N(Z_3/\{0\}) = \{1, 2, I, 2I\}$ be a neutrosophic group under multiplication modulo $3$. Now $\{1, 2\}$, $\{1, I\}$ are subgroups of $N(Z_3/\{0\})$ which divides order of $N(Z_3/\{0\})$. Then the $(F, A)$ is soft Lagrange neutrosophic group over $N(Z_3/\{0\})$, where*

$$F(a_1) = \{1, 2\}, F(a_2) = \{1, I\}.$$

**Theorem 2.1.11:** If $N(G)$ is Lagrange neutrosophic group, then $(F, A)$ over $N(G)$ is soft Lagrange neutrosophic group but the converse is not true in general.

The converse is left as an exercise for the readers.

**Theorem 2.1.12:** Every soft Lagrange neutrosophic group is a soft neutrosophic group.

**Proof** Straightforward.

**Remark 2** The converse of the above theorem does not hold.
We can shown it in the following example.

**Example 2.1.14:** *Let $N(G) = \{1, 2, 3, 4, I, 2I, 3I, 4I\}$ be a neutrosophic group under multiplication modulo $5$. Then $(F, A)$ be a soft neutrosophic group over $N(G)$, where*

$$F(a_1) = \{1, 4, I, 2I, 3I, 4I\}, F(a_2) = \{1, 2, 3, 4\},$$
$$F(a_3) = \{1, I, 2I, 3I, 4I\}.$$

*But clearly it is not soft Lagrange neutrosophic group as $F(a_1)$ which is a subgroup of $N(G)$ does not divide order of $N(G)$.*

**Theorem 2.1.13:** *If $N(G)$ is a neutrosophic group, then the soft Lagrange neutrosophic group is a soft neutrosophic group.*

**Proof**: Suppose that $N(G)$ be a neutrosophic group and $(F,A)$ be a soft Lagrange neutrosophic group over $N(G)$. Then by above theorem $(F,A)$ is also soft neutrosophic group.

**Example 2.1.15:** *Let $N(Z_4)$ be a neutrosophic group and $(F,A)$ is a soft Lagrange neutrosophic group over $N(Z_4)$ under addition modulo $4$, where*

$$F(a_1) = \{0,1,2,3\}, F(a_2) = \{0,I,2I,3I\},$$
$$F(a_3) = \{0,2,2I,2+2I\}.$$

*But $(F,A)$ has a proper soft group $(H,B)$, where*

$$H(a_1) = \{0,2\}, H(a_3) = \{0,2\}.$$

*Hence $(F,A)$ is soft neutrosophic group.*

**Remark 2.1.5:** Let $(F,A)$ and $(K,B)$ be two soft Lagrange neutrosophic groups over $N(G)$. Then

1. Their extended union $(F,A) \cup_E (K,B)$ over $N(G)$ is not soft Lagrange neutrosophic group over $N(G)$.
2. Their extended intersection $(F,A) \cap_E (K,B)$ over $N(G)$ is not soft Lagrange neutrosophic group over $N(G)$.
3. Their restricted union $(F,A) \cup_R (K,B)$ over $N(G)$ is not soft Lagrange neutrosophic group over $N(G)$.
4. Their restricted intersection $(F,A) \cap_R (K,B)$ over $N(G)$ is not soft Lagrange neutrosophic group over $N(G)$.

One can easily show 1,2,3 and 4 by the help of example.

**Remark 2.1.6:** Let $(F,A)$ and $(K,B)$ be two soft Lagrange neutrosophic groups over $N(G)$. Then

1. Their $AND$ operation $(F,A) \wedge (K,B)$ is not soft Lagrange neutrosophic group over $N(G)$.
2. Their $OR$ operation $(F,A) \vee (K,B)$ is not a soft Lagrange neutrosophic group over $N(G)$.

**Definition 2.1.8:** Let $(F,A)$ be a soft neutrosophic group over $N(G)$. Then $(F,A)$ is called soft weakly Lagrange neutrosophic group if atleast one $F(a)$ is a Lagrange neutrosophic subgroup of $N(G)$, for some $a \in A$.

**Example 2.1.16:** Let $N(G) = \{1, 2, 3, 4, I, 2I, 3I, 4I\}$ be a neutrosophic group under multiplication modulo $5$. Then $(F, A)$ is a soft weakly Lagrange neutrosophic group over $N(G)$, where

$$F(a_1) = \{1, 4, I, 2I, 3I, 4I\}, F(a_2) = \{1, 2, 3, 4\},$$
$$F(a_3) = \{1, I, 2I, 3I, 4I\}.$$

As $F(a_1)$ and $F(a_3)$ which are subgroups of $N(G)$ do not divide order of $N(G)$.

**Theorem 2.1.14:** Every soft weakly Lagrange neutrosophic group $(F, A)$ is soft neutrosophic group.

**Remark 2.1.7:** The converse of the above theorem does not hold in general.

This can be shown by the following example.

**Example 2.1.17:** Let $N(Z_4)$ be a neutrosophic group under addition modulo $4$ and $A = \{a_1, a_2\}$ be a set of parameters. Then $(F, A)$ is a soft neutrosophic group over $N(Z_4)$, where

$$F(a_1) = \{0, I, 2I, 3I\}, F(a_2) = \{0, 2I\}.$$

But not soft weakly Lagrange neutrosophic group over $N(Z_4)$

**Definition 2.1.9:** Let $(F,A)$ be a soft neutrosophic group over $N(G)$. Then $(F,A)$ is called soft Lagrange free neutrosophic group if $F(a)$ is not Lagrange neutrosophic subgroup of $N(G)$, for all $a \in A$.

**Example 2.1.18:** *Let* $N(G) = \{1,2,3,4,I,2I,3I,4I\}$ *be a neutrosophic group under multiplication modulo* $5$. *Then* $(F,A)$ *be a soft Lagrange free neutrosophic group over* $N(G)$, *where*

$$F(a_1) = \{1,4,I,2I,3I,4I\},$$
$$F(a_2) = \{1,I,2I,3I,4I\}.$$

*As* $F(a_1)$ *and* $F(a_2)$ *which are subgroups of* $N(G)$ *do not divide order of* $N(G)$.

**Theorem 2.1.15:** Every soft Lagrange free neutrosophic group $(F,A)$ over $N(G)$ is a soft neutrosophic group but the converse is not true.

**Definition 2.1.10:** Let $(F,A)$ be a soft neutrosophic group over $N(G)$. If for all $a \in A$, each $F(a)$ is a pseudo Lagrange neutrosophic subgroup of $N(G)$, then $(F,A)$ is called soft pseudo Lagrange neutrosophic group over $N(G)$.

**Example 2.1.19:** *Let* $N(Z_4)$ *be a neutrosophic group under addition modulo* $4$ *and* $A = \{a_1, a_2\}$ *be the set of parameters. Tthen* $(F,A)$ *is a soft pseudo Lagrange neutrosophic group over* $N(Z_4)$ *where*

$$F(a_1) = \{0, I, 2i, 3I, 4I\}, F(a_2) = \{0, 2I\}.$$

**Theorem 2.1.16:** Every soft pseudo Lagrange neutrosophic group is a soft neutrosophic group but the converse may not be true.

**Proof** Straightforward.

**Reamrk 2.1.8:** Let $(F,A)$ and $(K,B)$ be two soft pseudo Lagrange neutrosophic groups over $N(G)$. Then

1. Their extended union $(F,A) \cup_E (K,B)$ over $N(G)$ is not soft pseudo Lagrange neutrosophic group over $N(G)$.
2. Their extended intersection $(F,A) \cap_E (K,B)$ over $N(G)$ is not soft pseudo Lagrange neutrosophic group over $N(G)$.
3. Their restricted union $(F,A) \cup_R (K,B)$ over $N(G)$ is not soft pseudo Lagrange neutrosophic group over $N(G)$.
4. Their restricted intersection $(F,A) \cap_R (K,B)$ over $N(G)$ is not soft pseudo Lagrange neutrosophic group over $N(G)$.

One can easily establish these remarks.

**Remark 2.1.9:** Let $(F,A)$ and $(K,B)$ be two soft pseudo Lagrange neutrosophic groups over $N(G)$. Then

1. Their $AND$ operation $(F,A) \wedge (K,B)$ is not soft pseudo Lagrange neutrosophic group over $N(G)$.
2. Their $OR$ operation $(F,A) \vee (K,B)$ is not a soft pseudo Lagrange neutrosophic group over $N(G)$.

**Definition 2.1.11:** Let $(F, A)$ be a soft neutrosophic group over $N(G)$. Then $(F, A)$ is called soft weakly pseudo Lagrange neutrosophic group if atleast one $F(a)$ is a pseudo Lagrange neutrosophic subgroup of $N(G)$, for some $a \in A$.

**Example 2.1.20:** *Let $N(G) = \{1, 2, 3, 4, I, 2I, 3I, 4I\}$ be a neutrosophic group under multiplication modulo $5$ Then $(F, A)$ is a soft weakly pseudo Lagrange neutrosophic group over $N(G)$, where*

$$F(a_1) = \{1, I, 2I, 3I, 4I\}, F(a_2) = \{1, I\}.$$

*As $F(a_1)$ which is a subgroup of $N(G)$ does not divide order of $N(G)$.*

**Theorem 2.1.17:** Every soft weakly pseudo Lagrange neutrosophic group $(F, A)$ is soft neutrosophic group.

**Remark 2.1.10:** The converse of the above theorem is not true in general.

**Example 2.1.21:** Let $N(Z_4)$ be a neutrosophic group under addition modulo $4$ and $A = \{a_1, a_2\}$ be the set of parameters. Then $(F, A)$ is a soft neutrosophic group over $N(Z_4)$, where

$$F(a_1) = \{0, I, 2I, 3I\}, F(a_2) = \{0, 2I\}.$$

But it is not soft weakly pseudo Lagrange neutrosophic group.

**Remark 2.1.11:** Let $(F, A)$ and $(K, B)$ be two soft pseudo weakly Lagrange neutrosophic groups over $N(G)$. Then

1. Their extended union $(F, A) \cup_E (K, B)$ over $N(G)$ is not soft pseudo weakly Lagrange neutrosophic group over $N(G)$.
2. Their extended intersection $(F, A) \cap_E (K, B)$ over $N(G)$ is not soft pseudo weakly Lagrange neutrosophic group over $N(G)$.
3. Their restricted union $(F, A) \cup_R (K, B)$ over $N(G)$ is not soft pseudo weakly Lagrange neutrosophic group over $N(G)$.
4. Their restricted intersection $(F, A) \cap_R (K, B)$ over $N(G)$ is not soft pseudo weakly Lagrange neutrosophic group over $N(G)$.

**Proof:** The proof of 1,2,3 and 4 is easy.

**Remark 2.1.12:** Let $(F, A)$ and $(K, B)$ be two soft pseudo weakly Lagrange neutrosophic groups over $N(G)$. Then

1. Their $AND$ operation $(F, A) \wedge (K, B)$ is not soft pseudo weakly Lagrange neutrosophic group over $N(G)$.
2. Their $OR$ operation $(F, A) \vee (K, B)$ is not a soft pseudo weakly Lagrange neutrosophic group over $N(G)$.

**Definition 2.1.12:** Let $(F, A)$ be a soft neutrosophic group over $N(G)$. Then $(F, A)$ is called soft pseudo Lagrange free neutrosophic group if $F(a)$ is not pseudo Lagrange neutrosophic subgroup of $N(G)$, for all $a \in A$.

**Example 2.1.22:** Let $N(G) = \{1,2,3,,I,2I,3I,4I\}$ be a neutrosophic group under multiplication modulo $5$ Then $(F,A)$ is a soft pseudo Lagrange free neutrosophic group over $N(G)$, where

$$F(a_1) = \{1,I,2I,3I,4I\}, F(a_2) = \{1,I,2I,3I,4I\}.$$

As $F(a_1)$ and $F(a_2)$ which are subgroups of $N\ G$ do not divide order of $N(G)$.

**Theorem 2.1.18:** Every soft *pseudo Lagrange free neutrosophic group* $(F,A)$ over $N(G)$ is a soft neutrosophic group but the converse is not true.
 One can easily see the converse by taking number of examples.

**Remark 2.1.13:** Let $(F,A)$ and $(K,B)$ be two soft pseudo Lagrange free neutrosophic groups over $N(G)$. Then

1. Their extended union $(F,A) \cup_E (K,B)$ over $N(G)$ is not soft pseudo Lagrange free neutrosophic group over $N(G)$.
2. Their extended intersection $(F,A) \cap_E (K,B)$ over $N(G)$ is not soft pseudo Lagrange free neutrosophic group over $N(G)$.
3. Their restricted union $(F,A) \cup_R (K,B)$ over $N(G)$ is not soft pseudo Lagrange free neutrosophic group over $N(G)$.
4. Their restricted intersection $(F,A) \cap_R (K,B)$ over $N(G)$ is not soft pseudo Lagrange free neutrosophic group over $N(G)$.

**Proof:** The proof of 1,2,3 and 4 is easy.

**Remark 2.1.14:** Let $(F,A)$ and $(K,B)$ be two soft pseudo Lagrange free neutrosophic groups over $N(G)$. Then

1. Their $AND$ operation $(F,A) \wedge (K,B)$ is not soft pseudo Lagrange free neutrosophic group over $N(G)$.
2. Their $OR$ operation $(F,A) \vee (K,B)$ is not a soft pseudo Lagrange free neutrosophic group over $N(G)$.

**Definition 2.1.13:** A soft *neutrosophic group* $(F,A)$ over $N(G)$ is called soft *normal neutrosophic group* over $N(G)$ if $F(a)$ is a normal neutrosophic subgroup of $N(G)$, for all $a \in A$.

**Example 2.1.23:** *Let* $N(G) = \{e, a, b, c, I, aI, bI, cI\}$ *be a neutrosophic group under multiplication where*
$$a^2 = b^2 = c^2 = e, bc = cb = a, ac = ca = b, ab = ba = c.$$

*Then* $(F,A)$ *is a soft normal neutrosophic group over* $N(G)$ *where*

$$F(a_1) = \{e, a, I, aI\},$$
$$F(a_2) = \{e, b, I, bI\},$$
$$F(a_3) = \{e, c, I, cI\}.$$

**Theorem 2.1.19:** Every soft *normal neutrosophic group* $(F,A)$ over $N(G)$ is a soft *neutrosophic group* but the converse is not true.

**Remark 2.1.15:** Let $(F,A)$ and $(K,B)$ be two soft normal neutrosophic groups over $N(G)$. Then

1. Their extended union $(F,A) \cup_E (K,B)$ over $N(G)$ is not soft normal neutrosophic group over $N(G)$.
2. Their extended intersection $(F,A) \cap_E (K,B)$ over $N(G)$ is not soft normal neutrosophic group over $N(G)$.
3. Their restricted union $(F,A) \cup_R (K,B)$ over $N(G)$ is not soft normal neutrosophic group over $N(G)$.
4. Their restricted intersection $(F,A) \cap_R (K,B)$ over $N(G)$ is not soft normal neutrosophic group over $N(G)$.

**Proof:** The proof of 1,2,3 and 4 is easy.

**Remark 2.1.16:** Let $(F,A)$ and $(K,B)$ be two soft normal neutrosophic groups over $N(G)$. Then

1. Their $AND$ operation $(F,A) \wedge (K,B)$ is not soft normal neutrosophic group over $N(G)$.
2. Their $OR$ operation $(F,A) \vee (K,B)$ is not a soft normal neutrosophic group over $N(G)$.

**Definition 2.1.14:** Let $(F,A)$ be a soft neutrosophic group over $N(G)$. Then $(F,A)$ is called soft pseudo normal neutrosophic group if $F(a)$ is a pseudo normal neutrosophic subgroup of $N(G)$, for all $a \in A$.

**Example 2.1.24:** *Let $N(Z_2) = \{0, 1, I, 1 + I\}$ be a neutrosophic group under addition modulo 2 and let $A = \{a_1, a_2\}$ be the set of parameters. Then $(F, A)$ is soft pseudo normal neutrosophic group over $N(G)$ where*

$$F(a_1) = \{0, 1\}, F(a_2) = \{0, 1 + I\}.$$

**Theorem 2.1.20:** Every soft pseudo *normal neutrosophic group* $(F, A)$ over $N(G)$ is a soft *neutrosophic group* but the converse is not true.

**Remark 2.1.17:** Let $(F, A)$ and $(K, B)$ be two soft pseudo normal neutrosophic groups over $N(G)$. Then

1. Their extended union $(F, A) \cup_E (K, B)$ over $N(G)$ is not soft pseudo normal neutrosophic group over $N(G)$.
2. Their extended intersection $(F, A) \cap_E (K, B)$ over $N(G)$ is not soft pseudo normal neutrosophic group over $N(G)$.
3. Their restricted union $(F, A) \cup_R (K, B)$ over $N(G)$ is not soft pseudo normal neutrosophic group over $N(G)$.
4. Their restricted intersection $(F, A) \cap_R (K, B)$ over $N(G)$ is not soft pseudo normal neutrosophic group over $N(G)$.

**Proof:** The proof of 1,2,3 and 4 is easy.

**Remark 2.1.18:** Let $(F,A)$ and $(K,B)$ be two soft pseudo normal neutrosophic groups over $N(G)$. Then

1. Their $AND$ operation $(F,A) \wedge (K,B)$ is not soft pseudo normal neutrosophic group over $N(G)$.
2. Their $OR$ operation $(F,A) \vee (K,B)$ is not a soft pseudo normal neutrosophic group over $N(G)$.

**Definition 2.1.15:** Let $N(G)$ be a neutrosophic group. Then $(F,A)$ is called soft conjugate neutrosophic group over $N(G)$ if and only if $F(a)$ is conjugate neutrosophic subgroup of $N(G)$, for all $a \in A$.

**Example 2.1.25:** *Let*

$$N\ G\ = \begin{Bmatrix} 0,1,2,3,4,5,I,2I,3I,4I,5I, \\ 1+I, 2+I, 3+I, ..., 5+5I \end{Bmatrix}$$

*be a neutrosophic group under addition modulo* 6 *and let* $P = \{0,3,3I,3+3I\}$ *and* $K = \{0,2,4,2+2I,4+4I,2I,4I\}$ *are conjugate neutrosophic subgroups of* $N(G)$. *Then* $(F,A)$ *is soft conjugate neutrosophic group over* $N(G)$, *where*

$$F(a_1) = \{0,3,3I,3+3I\},$$
$$F(a_2) = \{0,2,4,2+2I,4+4I,2I,4I\}.$$

**Remark 2.1.19:** Let $(F,A)$ and $(K,B)$ be two soft conjugate neutrosophic groups over $N(G)$. Then

1. Their extended union $(F,A) \cup_E (K,B)$ over $N(G)$ is not soft conjugate neutrosophic group over $N(G)$.
2. Their extended intersection $(F,A) \cap_E (K,B)$ over $N(G)$ is not soft conjugate neutrosophic group over $N(G)$.
3. Their restricted union $(F,A) \cup_R (K,B)$ over $N(G)$ is not soft conjugate neutrosophic group over $N(G)$.
4. Their restricted intersection $(F,A) \cap_R (K,B)$ over $N(G)$ is not soft conjugate neutrosophic group over $N(G)$.

**Proof:** The proof of 1,2,3 and 4 is easy.

**Remark 2.1.20:** Let $(F,A)$ and $(K,B)$ be two soft conjugate neutrosophic groups over $N(G)$. Then

1. Their $AND$ operation $(F,A) \wedge (K,B)$ is not soft conjugate neutrosophic group over $N(G)$.
2. Their $OR$ operation $(F,A) \vee (K,B)$ is not a soft conjugate neutrosophic group over $N(G)$.

## Soft Neutrosophic Strong Group

In this section, we give the important notion soft neutrosophic strong group which is quite related to the strong part of neutrosophic group. We also introduce soft neutrosophic strong subgroup and give some of their basic characterization of this purely neutrosophic notion with many illustrative examples.

**Definition 2.1.16:** *Let $N(G)$ be a neutrosophic group and $(F,A)$ be soft set over $N(G)$. Then $(F,A)$ is called soft neutrosophic strong group over $N(G)$ if and only if $F(a)$ is a neutrosophic strong subgroup of $N(G)$, for all $a \in A$.*

This situation is explained with the help of following examples.

**Example 2.1.1:** Let

$$N(Z_4) = \begin{Bmatrix} 0,1,2,3,I,2I,3I,1+I,1+2I,1+3I, \\ 2+I,2+2I,2+3I,3+I,3+2I,3+3I \end{Bmatrix}$$

be a neutrosophic group under addition modulo $4$. and let $A = \{a_1, a_2, a_3, a_4\}$ be a set of parameters. Then $(F,A)$ is soft neutrosophic strong group over $N(Z_4)$, where

$$F(a_1) = \{0,3I\}, F(a_2) = \{0,I,2I,3I\}.$$

**Example 2.1.2:** Let
$$N(G) = \{e, a, b, c, I, aI, bI, cI\}$$
be a neutrosophic group under multiplication where
$$a^2 = b^2 = c^2 = e, bc = cb = a, ac = ca = b, ab = ba = c.$$
Then $(F, A)$ is a soft neutrosophic group over $N(G)$, where

$$F(a_1) = \{I, aI\},$$
$$F(a_2) = \{I, bI\},$$
$$F(a_3) = \{I, cI\}.$$

**Example 2.1.3:** Let $(F, A)$ be a soft neutrosophic groups over $N(Z_2)$ under addition modulo $2$, where

$$F(a_1) = \{0, 1+I\}, F(a_2) = \{0, I\}.$$

**Theorem 2.1.21:** Every soft neutrosophic strong group is trivially a soft neutrosophic group but the converse is not true.

The converse is obvious, so it is left for the readers as an exercise.

**Theorem 2.1.22:** If $N(G)$ is a neutrosophic strong group, then $(F, A)$ over $N(G)$ is also a soft neutrosophic strong group.

**Proof:** The proof is left as an exercise for the readers.

**Remark 2.1.21:** Let $(F,A)$ and $(K,B)$ be two soft neutrosophic strong groups over $N(G)$. Then

1. Their extended union $(F,A) \cup_E (K,B)$ over $N(G)$ is not soft neutrosophic strong group over $N(G)$.
2. Their extended intersection $(F,A) \cap_E (K,B)$ over $N(G)$ is not soft neutrosophic strong group over $N(G)$.
3. Their restricted union $(F,A) \cup_R (K,B)$ over $N(G)$ is not soft neutrosophic strong group over $N(G)$.
4. Their restricted intersection $(F,A) \cap_R (K,B)$ over $N(G)$ is not a soft neutrosophic strong group over $N(G)$.

**Proof:** The proof of 1,2,3 and 4 is easy.

**Remark 2.1.22:** Let $(F,A)$ and $(K,B)$ be two soft conjugate neutrosophic groups over $N(G)$. Then

1. Their $AND$ operation $(F,A) \wedge (K,B)$ is not a soft neutrosophic strong group over $N(G)$.
2. Their $OR$ operation $(F,A) \vee (K,B)$ is not a soft neutrosophic strong group over $N(G)$.

**Definition 2.1.17:** *Let $(F, A)$ and $(H, B)$ be two soft neutrosophic strong groups over $N(G)$. Then $(H, B)$ is called soft neutrosophic strong subgroup of $(F, A)$, denoted as $(H, B) < (F, A)$, if*

1. $B \subseteq A$
2. $H(a)$ is a neutrosophic strong subgroup of $F(a)$, for all $a \in A$.

We explain this situation by the following example.

**Example 2.1.10:** *Let $(F, A)$ be a soft neutrosophic strong group over $N(Z_4)$, where*

$$F(a_1) = \{0, I, 2I, 3I\}, F(a_2) = \{0, 2I\}.$$

Hence $(H, B) \prec (F, A)$ where

$$H(a_1) = \{0, 2I\}.$$

**Theorem 2.1.23:** Every soft neutrosophic strong subgroup of is trivially a soft neutrosophic subgroup but the converse is not true.

## 2.2 Soft Neutrosophic Bigroups and their Properties

Now we proceed onto define the notion of soft neutrosophic bigroups over neutrosophic bigroups. However soft neutrosophic bigroups are the parameterized family of the neutrosophic bigroups. We can also give soft Lagrange neutrosophic bigroup over a neutrosophic bigroup. Some of the important and interesting properties are also established with necessary examples to illustrate this theory.

**Definition 2.2.1:** Let $B_N(G) = \{B(G_1) \cup B(G_2), *_1, *_2\}$ be a neutrosophic bigroup and let $(F, A)$ be a soft set over $B_N(G)$. Then $(F, A)$ is said to be soft neutrosophic bigroup over $B_N(G)$ if and only if $F(a)$ is a subbigroup of $B_N(G)$ for all $a \in A$.

The following examples will help us in understanding this notion.

**Example 2.2.1:** Let $B_N(G) = \{B(G_1) \cup B(G_2), *_1, *_2\}$ be a neutrosophic bigroup, where
$$B(G_1) = \{0, 1, 2, 3, 4, I, 2I, 3I, 4I\}$$

is a neutrosophic group under multiplication modulo $5$.
$B(G_2) = \{g : g^{12} = 1\}$ is a cyclic group of order $12$.
Let $P(G) = \{P(G_1) \cup P(G_2), *_1, *_2\}$ be a neutrosophic subbigroup where
$P(G_1) = \{1, 4, I, 4I\}$ and $P(G_2) = \{1, g^2, g^4, g^6, g^8, g^{10}\}$.

Also $Q(G) = \{Q(G_1) \cup Q(G_2), *_1, *_2\}$ be another neutrosophic subbigroup where $Q(G_1) = \{1, I\}$ and $Q(G_2) = \{1, g^3, g^6, g^9\}$.

Then $(F, A)$ is a soft neutrosophic bigroup over $B_N(G)$, where

$$F(a_1) = \{1, 4, I, 4I, 1, g^2, g^4, g^6, g^8, g^{10}\}$$
$$F(a_2) = \{1, I, 1, g^3, g^6, g^9\}.$$

**Theorem 2.2.1:** Let $(F, A)$ and $(H, A)$ be two soft neutrosophic bigroup over $B_N(G)$. Then their intersection $(F, A) \cap (H, A)$ is again a soft neutrosophic bigroup over $B_N(G)$.

**Proof** Straight forward.

**Theorem 2.2.2:** Let $(F, A)$ and $(H, B)$ be two soft neutrosophic bigroups over $B_N(G)$ such that $A \cap B = \phi$, then their union is soft neutrosophic bigroup over $B_N(G)$.

**Proof** Straight forward.

**Remark 2.2.1:** The extended union of two soft neutrosophic bigroups $(F, A)$ and $(K, D)$ over $B_N(G)$ is not a soft neutrosophic bigroup over $B_N(G)$.

To prove it, see the following example.

**Example 2.2.2:** Let $B_N(G) = \{B(G_1) \cup B(G_2), *_1, *_2\}$, where
$B(G_1) = \{1, 2, 3, 4I, 2I, 3I, 4I\}$ and $B(G_2) = S_3$.
Let $P(G) = \{P(G_1) \cup P(G_2), *_1, *_2\}$ be a neutrosophic subbigroup where
$P(G_1) = \{1, 4, I, 4I\}$ and $P(G_2) = \{e, (12)\}$.
Also $Q(G) = \{Q(G_1) \cup Q(G_2), *_1, *_2\}$ be another neutrosophic subbigroup where $Q(G_1) = \{1, I\}$ and $Q(G_2) = \{e, (123), (132)\}$.
Then $(F, A)$ is a soft neutrosophic bigroup over $B_N(G)$, where

$$F(a_1) = \{1, 4, I, 4I, e, (12)\}$$
$$F(a_2) = \{1, I, e, (123), (132)\}.$$

Again let $R(G) = \{R(G_1) \cup R(G_2), *_1, *_2\}$ be another neutrosophic subbigroup
where $R(G_1) = \{1, 4, I, 4I\}$ and $R(G_2) = \{e, (13)\}$.
Also $T(G) = \{T(G_1) \cup T(G_2), *_1, *_2\}$ be a neutrosophic subbigroup where
$T(G_1) = \{1, I\}$ and $T(G_2) = \{e, (23)\}$.
Then $(K, D)$ is a soft neutrosophic bigroup over $B_N(G)$, where
$$K(a_2) = \{1, 4, I, 4I, e, (13)\},$$
$$K(a_3) = \{1, I, e, (23)\}.$$

The extended union $(F, A) \cup_\varepsilon (K, D) = (H, C)$ such that $C = A \cup D$ and for $a_2 \in C$, we have $H(a_2) = F(a_2) \cup K(a_2) = \{1, 4, I, 4I, e, (13)(123), (132)\}$ is not a subbigroup of $B_N(G)$.

**Proposition 2.2.1:** The extended intersection of two soft neutrosophic bigroups $(F, A)$ and $(K, D)$ over $B_N(G)$ is again a soft neutrosophic bigroup over $B_N(G)$.

**Remark 2.2.3:** The restricted union of two soft neutrosophic bigroups $(F,A)$ and $(K,D)$ over $B_N(G)$ is not a soft neutrosophic bigroup over $B_N(G)$.

**Proposition 2.2.2:** The restricted intersection of two soft neutrosophic bigroups $(F,A)$ and $(K,D)$ over $B_N(G)$ is a soft neutrosophic bigroup over $B_N(G)$.

**Proposition 2.2.3:** The $AND$ operation of two soft neutrosophic bigroups over $B_N(G)$ is again soft neutrosophic bigroup over $B_N(G)$.

**Remark 2.2.2:** The $OR$ operation of two soft neutrosophic bigroups over $B_N(G)$ may not be a soft nuetrosophic bigroup.

**Definition 2.2.2:** Let $(F,A)$ be a soft neutrosophic bigroup over $B_N(G)$. Then

1) $(F,A)$ is called identity soft neutrosophic bigroup if $F(a)=\{e_1,e_2\}$ for all $a \in A$, where $e_1$ and $e_2$ are the identities of $B(G_1)$ and $B(G_2)$ respectively.
2) $(F,A)$ is called an absolute-soft neutrosophic bigroup if $F(a)=B_N(G)$ for all $a \in A$.

**Theorem 2.2.3:** Let $B_N(G)$ be a neutrosophic bigroup of prime order $P$. Then $(F,A)$ over $B_N(G)$ is either identity soft neutrosophic bigroup or absolute soft neutrosophic bigroup.

**Definition 2.2.3:** Let $(F,A)$ and $(H,K)$ be two soft neutrosophic bigroups over $B_N(G)$. Then $(H,K)$ is soft neutrosophi subbigroup of $(F,A)$ written as $(H,K) \prec (F,A)$, if
1. $K \subseteq A$,
2. $H(a)$ is a neutrosophic subigroup of $F(a)$ for all $a \in A$.

**Example 2.2.4:** Let $B(G) = \{B(G_1) \cup B(G_2), *_1, *_2\}$ where

$$B(G_1) = \begin{cases} 0,1,2,3,4,I,2I,3I,4I,1+I,2+I,3+I,4+I, \\ 1+2I,2+2I,3+2I,4+2I,1+3I,2+3I, \\ 3+3I,4+3I,1+4I,2+4I,3+4I,4+4I \end{cases}$$

be a neutrosophic group under multiplication modulo 5 and $B(G_2) = \{g : g^{16} = 1\}$ a cyclic group of order 16.
Let $P(G) = \{P(G_1) \cup P(G_2), *_1, *_2\}$ be a neutrosophic subbigroup where $P(G_1) = \{0,1,2,3,4,I,2I,3I,4I\}$
and be another neutrosophic subbigroup where
$P(G_2) = \{g^2, g^4, g^6, g^8, g^{10}, g^{12}, g^{14}, 1\}$. Also $Q(G) = \{Q(G_1) \cup Q(G_2), *_1, *_2\}$,
$Q(G_1) = \{0,1,4,I,4I\}$ and $Q(G_2) = \{g^4, g^8, g^{12}, 1\}$.
Again let $R(G) = \{R(G_1) \cup R(G_2), *_1, *_2\}$ be a neutrosophic subbigroup where
$$R(G_1) = \{0,1,I\} \text{ and } R(G_2) = \{1, g^8\}.$$
Let $(F,A)$ be a soft neutrsophic bigroup over $B_N(G)$ where

$$F(a_1) = \{0,1,2,3,4,I,2I,3I,4I, g^2, g^4, g^6, g^8, g^{10}, g^{12}, g^{14}, 1\},$$
$$F(a_2) = \{0,1,4,I,4I, g^4, g^8, g^{12}, 1\},$$
$$F(a_3) = \{0,1,I, g^8, 1\}.$$

Let $(H, K)$ be another soft neutrosophic bigroup over $B_N(G)$, where

$$H(a_1) = \{0,1,2,3,4, g^4, g^8, g^{12}, 1\},$$
$$H(a_2) = \{0,1, I, g^8, 1\}.$$

Clearly $(H, K) \prec (F, A)$.

**Definition 2.2.4:** Let $B_N(G)$ be a neutrosophic bigroup. Then $(F, A)$ over $B_N(G)$ is called commutative soft neutrosophic bigroup if and only if $F(a)$ is a commutative subbigroup of $B_N(G)$ for all $a \in A$.

**Example 2.2.5:** Let $B(G) = \{B(G_1) \cup B(G_2), *_1, *_2\}$ be a neutrosophic bigroup where $B(G_1) = \{g : g^{10} = 1\}$ be a cyclic group of order 10 and $B(G_2) = \{1, 2, 3, 4, I, 2I, 3I, 4I\}$ be a neutrosophic group under mltiplication modulo 5.
Let $P(G) = \{P(G_1) \cup P(G_2), *_1, *_2\}$ be a commutative neutrosophic subbigroup where $P(G_1) = \{1, g^5\}$ and $P(G_2) = \{1, 4, I, 4I\}$. Also $Q(G) = \{Q(G_1) \cup Q(G_2), *_1, *_2\}$ be another commutative neutrosophic subbigroup where $Q(G_1) = \{1, g^2, g^4, g^6, g^8\}$ and $Q(G_2) = \{1, I\}$. Then $(F, A)$ is commutative soft neutrosophic bigroup over $B_N(G)$, where

$$F(a_1) = \{1, g^5, 1, 4, I, 4I\},$$
$$F(a_2) = \{1, g^2, g^4, g^6, g^8, 1, I\}.$$

**Theorem 2.2.4:** Every commutative *soft neutrosophic bigroup* $(F, A)$ over $B_N(G)$ is a soft *neutrosophic bigroup* but the converse is not true.

**Theorem 2.2.5:** If $B_N(G)$ is commutative neutrosophic bigroup. Then $(F,A)$ over $B_N(G)$ is commutative soft neutrosophic bigroup but the converse is not true.

**Theorem 2.2.6:** If $B_N(G)$ is cyclic neutrosophic bigroup. Then $(F,A)$ over $B_N(G)$ is commutative soft neutrosophic bigroup.

**Proposition 2.2.4:** Let $(F,A)$ and $(K,D)$ be two commutative soft neutrosophic bigroups over $B_N(G)$. Then

1. Their extended union $(F,A) \cup_\varepsilon (K,D)$ over $B_N(G)$ is not commutative soft neutrosophic bigroup over $B_N(G)$.
2. Their extended intersection $(F,A) \cap_\varepsilon (K,D)$ over $B_N(G)$ is commutative soft neutrosophic bigroup over $B_N(G)$.
3. Their restricted union $(F,A) \cup_R (K,D)$ over $B_N(G)$ is not commutative soft neutrosophic bigroup over $B_N(G)$.
4. Their restricted intersection $(F,A) \cap_R (K,D)$ over $B_N(G)$ is commutative soft neutrosophic bigroup over $B_N(G)$.

**Proposition 2.2.5:** Let $(F,A)$ and $(K,D)$ be two commutative soft neutrosophic bigroups over $B_N(G)$. Then

1. Their *AND* operation $(F,A) \wedge (K,D)$ is commutative soft neutrosophic bigroup over $B_N(G)$.
2. Their *OR* operation $(F,A) \vee (K,D)$ is not commutative soft neutrosophic bigroup over $B_N(G)$.

**Definition 2.2.5:** Let $B_N(G)$ be a neutrosophic bigroup. Then $(F, A)$ over $B_N(G)$ is called cyclic soft neutrosophic bigroup if and only if $F(a)$ is a cyclic subbigroup of $B_N(G)$ for all $a \in A$.

**Example 2.2.6:** Let $B(G) = \{B(G_1) \cup B(G_2), *_1, *_2\}$ be a neutrosophic bigroup where $B(G_1) = \{g : g^{10} = 1\}$ be a cyclic group of order $10$ and $B(G_2) = \{0, 1, 2, I, 2I, 1+I, 2+I, 1+2I, 2+2I\}$ be a neutrosophic group under multiplication modulo $3$. Le $P(G) = \{P(G_1) \cup P(G_2), *_1, *_2\}$ be a cyclic neutrosophic subbigroup where $P(G_1) = \{1, g^5\}$ and $\{1, 1+I\}$. Also $Q(G) = \{Q(G_1) \cup Q(G_2), *_1, *_2\}$ be another cyclic neutrosophic subbigroup where $Q(G_1) = \{1, g^2, g^4, g^6, g^8\}$ and $Q(G_2) = \{1, 2+2I\}$. Then $(F, A)$ is cyclic soft neutrosophic bigroup over $B_N(G)$, where

$$F(a_1) = \{1, g^5, 1, 1+I\},$$
$$F(a_2) = \{1, g^2, g^4, g^6, g^8, 1, 2+2I\}.$$

**Theorem 2.2.7:** If $B_N(G)$ is a cyclic neutrosophic soft bigroup, then $(F, A)$ over $B_N(G)$ is also cyclic soft neutrosophic bigroup.

**Theorem 2.2.8:** Every cyclic *soft neutrosophic bigroup* $(F, A)$ over $B_N(G)$ is a soft *neutrosophic bigroup* but the converse is not true.

**Proposition 2.2.6:** Let $(F,A)$ and $(K,D)$ be two cyclic soft neutrosophic bigroups over $B_N(G)$. Then

1. Their extended union $(F,A) \cup_\varepsilon (K,D)$ over $B_N(G)$ is not cyclic soft neutrosophic bigroup over $B_N(G)$.
2. Their extended intersection $(F,A) \cap_\varepsilon (K,D)$ over $B_N(G)$ is cyclic soft neutrosophic bigroup over $B_N(G)$.
3. Their restricted union $(F,A) \cup_R (K,D)$ over $B_N(G)$ is not cyclic soft neutrosophic bigroup over $B_N(G)$.
4. Their restricted intersection $(F,A) \cap_R (K,D)$ over $B_N(G)$ is cyclic soft neutrosophic bigroup over $B_N(G)$.

**Proposition 2.2.7:** Let $(F,A)$ and $(K,D)$ be two cyclic soft neutrosophic bigroups over $B_N(G)$. Then

1. Their *AND* operation $(F,A) \wedge (K,D)$ is cyclic soft neutrosophic bigroup over $B_N(G)$.
2. Their *OR* operation $(F,A) \vee (K,D)$ is not cyclic soft neutrosophic bigroup over $B_N(G)$.

**Definition 2.2.6:** Let $B_N(G)$ be a neutrosophic bigroup. Then $(F,A)$ over $B_N(G)$ is called normal soft neutrosophic bigroup if and only if $F(a)$ is normal subbigroup of $B_N(G)$ for all $a \in A$.

**Example 2.2.7:** Let $B(G) = \{B(G_1) \cup B(G_2), *_1, *_2\}$ be a neutrosophic bigroup, where

$$B(G_1) = \begin{Bmatrix} e, y, x, x^2, xy, x^2y, I, \\ Iy, Ix, Ix^2, Ixy, Ix^2y \end{Bmatrix}$$

is a neutrosophic group under multiplaction and $B(G_2) = \{g : g^6 = 1\}$ is a cyclic group of order 6.

Let $P(G) = \{P(G_1) \cup P(G_2), *_1, *_2\}$ be a normal neutrosophic subbigroup where $P(G_1) = \{e, y\}$ and $P(G_2) = \{1, g^2, g^4\}$.

Also $Q(G) = \{Q(G_1) \cup Q(G_2), *_1, *_2\}$ be another normal neutrosophic subbigroup where $Q(G_1) = \{e, x, x^2\}$ and $Q(G_2) = \{1, g^3\}$.

Then $(F, A)$ is a normal soft neutrosophic bigroup over $B_N(G)$ where

$$F(a_1) = \{e, y, 1, g^2, g^4\},$$
$$F(a_2) = \{e, x, x^2, 1, g^3\}.$$

**Theorem 2.2.9:** Every normal *soft neutrosophic bigroup* $(F, A)$ over $B_N(G)$ is a soft *neutrosophic bigroup* but the converse is not true.

**Theorem 2.2.10:** If $B_N(G)$ is a normal neutrosophic bigroup. Then $(F, A)$ over $B_N(G)$ is also normal soft neutrosophic bigroup.

**Theorem 2.2.11:** If $B_N(G)$ is a commutative neutrosophic bigroup. Then $(F, A)$ over $B_N(G)$ is normal soft neutrosophic bigroup.

**Theorem 2.2.11:** If $B_N(G)$ is a cyclic neutrosophic bigroup. Then $(F,A)$ over $B_N(G)$ is normal soft neutrosophic bigroup.

**Proposition 2.2.8:** Let $(F,A)$ and $(K,D)$ be two normal soft neutrosophic bigroups over $B_N(G)$. Then

1. Their extended union $(F,A) \cup_\varepsilon (K,D)$ over $B_N(G)$ is not normal soft neutrosophic bigroup over $B_N(G)$.
2. Their extended intersection $(F,A) \cap_\varepsilon (K,D)$ over $B_N(G)$ is normal soft neutrosophic bigroup over $B_N(G)$.
3. Their restricted union $(F,A) \cup_R (K,D)$ over $B_N(G)$ is not normal soft neutrosophic bigroup over $B_N(G)$.
4. Their restricted intersection $(F,A) \cap_R (K,D)$ over $B_N(G)$ is normal soft neutrosophic bigroup over $B_N(G)$.

**Proposition 2.2.9:** Let $(F,A)$ and $(K,D)$ be two normal soft neutrosophic bigroups over $B_N(G)$. Then

1. Their *AND* operation $(F,A) \wedge (K,D)$ is normal soft neutrosophic bigroup over $B_N(G)$.
2. Their *OR* operation $(F,A) \vee (K,D)$ is not normal soft neutrosophic bigroup over $B_N(G)$.

**Definition 2.2.7:** Let $(F, A)$ be a soft neutrosophic bigroup over $B_N(G)$. If for all $a \in A$, $F(a)$ is a Lagrange subbigroup of $B_N(G)$, then $(F, A)$ is called Lagrange soft neutosophic bigroup over $B_N(G)$.

**Example 2.2.8:** Let $B(G) = \{B(G_1) \cup B(G_2), *_1, *_2\}$ be a neutrosophic bigroup, where

$$B(G_1) = \begin{Bmatrix} e, y, x, x^2, xy, x^2y, I, \\ Iy, Ix, Ix^2, Ixy, Ix^2y \end{Bmatrix}$$

is a neutrosophic symmetric group of and $B(G_2) = \{0, 1, I, 1+I\}$ be a neutrosophic group under addition modulo $2$. Let $P(G) = \{P(G_1) \cup P(G_2), *_1, *_2\}$ be a neutrosophic subbigroup where $P(G_1) = \{e, y\}$ and $P(G_2) = \{0, 1\}$.

Also $Q(G) = \{Q(G_1) \cup Q(G_2), *_1, *_2\}$ be another neutrosophic subbigroup where $Q(G_1) = \{e, Iy\}$ and $Q(G_2) = \{0, 1+I\}$.

Then $(F, A)$ is Lagrange soft neutrosophic bigroup over $B_N(G)$, where

$$F(a_1) = \{e, y, 0, 1\},$$
$$F(a_2) = \{e, yI, 0, 1+I\}.$$

as a Lagrange soft neutrosophic bigroup.

**Theorem 2.2.12:** If $B_N(G)$ is a Lagrange neutrosophic bigroup, then $(F, A)$ over $B_N(G)$

**Theorem 2.2.13:** Every Lagrange soft *neutrosophic bigroup* $(F,A)$ over $B_N(G)$ is a soft *neutrosophic bigroup* but the converse is not true.

**Proposition 2.2.10:** Let $(F,A)$ and $(K,D)$ be two Lagrange soft neutrosophic bigroups over $B_N(G)$. Then

1. Their extended union $(F,A) \cup_\varepsilon (K,D)$ over $B_N(G)$ is not Lagrange soft neutrosophic bigroup over $B_N(G)$.
2. Their extended intersection $(F,A) \cap_\varepsilon (K,D)$ over $B_N(G)$ is not Lagrange soft neutrosophic bigroup over $B_N(G)$.
3. Their restricted union $(F,A) \cup_R (K,D)$ over $B_N(G)$ is not Lagrange soft neutrosophic bigroup over $B_N(G)$.
4. Their restricted intersection $(F,A) \cap_R (K,D)$ over $B_N(G)$ is not Lagrange soft neutrosophic bigroup over $B_N(G)$.

**Proposition 2.2.11:** Let $(F,A)$ and $(K,D)$ be two Lagrange soft neutrosophic bigroups over $B_N(G)$. Then

1. Their *AND* operation $(F,A) \wedge (K,D)$ is not Lagrange soft neutrosophic bigroup over $B_N(G)$.
2. Their *OR* operation $(F,A) \vee (K,D)$ is not Lagrange soft neutrosophic bigroup over $B_N(G)$.

**Definition 2.2.8:** Let $(F,A)$ be a soft neutrosophic bigroup over $B_N(G)$. Then $(F,A)$ is called weakly Lagrange soft neutosophic bigroup over $B_N(G)$ if atleast one $F(a)$ is a Lagrange subbigroup of $B_N(G)$, for some $a \in A$.

**Example 2.2.9:** Let $B(G) = \{B(G_1) \cup B(G_2), *_1, *_2\}$ be a neutrosophic bigroup, where

$$B(G_1) = \begin{Bmatrix} 0,1,2,3,4,I,2I,3I,4I,1+I,2+I,3+I,4+I, \\ 1+2I,2+2I,3+2I,4+2I,1+3I,2+3I, \\ 3+3I,4+3I,1+4I,2+4I,3+4I,4+4I \end{Bmatrix}$$

is a neutrosophic group under multiplication modulo 5 and $B(G_2) = \{g : g^{10} = 1\}$ is a cyclic group of order 10. Let $P(G) = \{P(G_1) \cup P(G_2), *_1, *_2\}$ be a neutrosophic subbigroup where $P(G_1) = \{0,1,4,I,4I\}$ and $P(G_2) = \{g^2, g^4, g^6, g^8, 1\}$. Also $Q(G) = \{Q(G_1) \cup Q(G_2), *_1, *_2\}$ be another neutrosophic subbigroup where $Q(G_1) = \{0,1,4,I,4I\}$ and $Q(G_2) = \{g^5, 1\}$. Then $(F,A)$ is a weakly Lagrange soft neutrosophic bigroup over $B_N(G)$, where

$$F(a_1) = \{0,1,4,I,4I,g^2,g^4,g^6,g^8,1\},$$
$$F(a_2) = \{0,1,4,I,4I,g^5,1\}.$$

**Theorem 2.2.14:** Every weakly Lagrange *soft neutrosophic bigroup* $(F,A)$ over $B_N(G)$ is a soft *neutrosophic bigroup* but the converse is not true.

**Proposition 2.2.12:** Let $(F,A)$ and $(K,D)$ be two weakly Lagrange soft neutrosophic bigroups over $B_N(G)$. Then

1. Their extended union $(F,A) \cup_\varepsilon (K,D)$ over $B_N(G)$ is not weakly Lagrange soft neutrosophic bigroup over $B_N(G)$.
2. Their extended intersection $(F,A) \cap_\varepsilon (K,D)$ over $B_N(G)$ is not weakly Lagrange soft neutrosophic bigroup over $B_N(G)$.
3. Their restricted union $(F,A) \cup_R (K,D)$ over $B_N(G)$ is not weakly Lagrange soft neutrosophic bigroup over $B_N(G)$.
4. Their restricted intersection $(F,A) \cap_R (K,D)$ over $B_N(G)$ is not weakly Lagrange soft neutrosophic bigroup over $B_N(G)$.

**Proposition 2.2.13:** Let $(F,A)$ and $(K,D)$ be two weakly Lagrange soft neutrosophic bigroups over $B_N(G)$. Then

1. Their $AND$ operation $(F,A) \wedge (K,D)$ is not weakly Lagrange soft neutrosophic bigroup over $B_N(G)$.
2. Their $OR$ operation $(F,A) \vee (K,D)$ is not weakly Lagrange soft neutrosophic bigroup over $B_N(G)$.

**Definition 2.2.9:** Let $(F,A)$ be a soft neutrosophic bigroup over $B_N(G)$. Then $(F,A)$ is called Lagrange free soft neutrosophic bigroup if each $F(a)$ is not Lagrange subbigroup of $B_N(G)$, for all $a \in A$.

**Example 2.2.10:** Let $B(G) = \{B(G_1) \cup B(G_2), *_1, *_2\}$ be a neutrosophic bigroup, where $B(G_1) = \{0, 1, I, 1+I\}$ is a neutrosophic group under addition modulo 2 of order 4 and $B(G_2) = \{g : g^{12} = 1\}$ is a cyclic group of order 12. Let $P(G) = \{P(G_1) \cup P(G_2), *_1, *_2\}$ be a neutrosophic subbigroup where $P(G_1) = \{0, I\}$ and $P(G_2) = \{g^4, g^8, 1\}$. Also $Q(G) = \{Q(G_1) \cup Q(G_2), *_1, *_2\}$ be another neutrosophic subbigroup where $Q(G_1) = \{0, 1+I\}$ and $Q(G_2) = \{1, g^3, g^6, g^9\}$. Then $(F, A)$ is Lagrange free soft neutrosophic bigroup over $B_N(G)$, where

$$F(a_1) = \{0, I, 1, g^4, g^8\},$$
$$F(a_2) = \{0, 1+I, 1, g^3, g^6, g^9\}.$$

**Theorem 2.2.15:** If $B_N(G)$ is Lagrange free neutrosophic bigroup, and then $(F, A)$ over $B_N(G)$ is Lagrange free soft neutrosophic bigroup.

**Proof:** This is obvious.

**Theorem 2.2.16:** Every Lagrange free *soft neutrosophic bigroup* $(F, A)$ over $B_N(G)$ is a soft *neutrosophic bigroup* but the converse is not true.

**Proof:** This is obvious.

**Proposition 2.2.14:** Let $(F,A)$ and $(K,D)$ be two Lagrange free soft neutrosophic bigroups over $B_N(G)$. Then

1. Their extended union $(F,A) \cup_\varepsilon (K,D)$ over $B_N(G)$ is not Lagrange free soft neutrosophic bigroup over $B_N(G)$.
2. Their extended intersection $(F,A) \cap_\varepsilon (K,D)$ over $B_N(G)$ is not Lagrange free soft neutrosophic bigroup over $B_N(G)$.
3. Their restricted union $(F,A) \cup_R (K,D)$ over $B_N(G)$ is not Lagrange free soft neutrosophic bigroup over $B_N(G)$.
4. Their restricted intersection $(F,A) \cap_R (K,D)$ over $B_N(G)$ is not Lagrange free soft neutrosophic bigroup over $B_N(G)$.

**Proposition 2.2.15:** Let $(F,A)$ and $(K,D)$ be two Lagrange free soft neutrosophic bigroups over $B_N(G)$. Then

1. Their AND operation $(F,A) \wedge (K,D)$ is not Lagrange free soft neutrosophic bigroup over $B_N(G)$.
2. Their OR operation $(F,A) \vee (K,D)$ is not Lagrange free soft neutrosophic bigroup over $B_N(G)$.

**Definition 2.2.10:** Let $B_N(G)$ be a neutrosophic bigroup. Then $(F,A)$ is called conjugate soft neutrosophic bigroup over $B_N(G)$ if and only if $F(a)$ is neutrosophic conjugate subbigroup of $B_N(G)$ for all $a \in A$.

**Example 2.2.11:** Let $B(G) = \{B(G_1) \cup B(G_2), *_1, *_2\}$ be a soft neutrosophic bigroup, where $B(G_1) = \{e, y, x, x^2, xy, x^2y\}$ is Klien $4$-group and
$B(G_2) = \begin{cases} 0,1,2,3,4,5,I,2I,3I,4I,5I, \\ 1+I, 2+I, 3+I, \ldots, 5+5I \end{cases}$ be a neutrosophic group under addition modulo $6$.

Let $P(G) = \{P(G_1) \cup P(G_2), *_1, *_2\}$ be a neutrosophic subbigroup of $B_N(G)$, where $P(G_1) = \{e, y\}$ and $P(G_2) = \{0, 3, 3I, 3+3I\}$.

Again let $Q(G) = \{Q(G_1) \cup Q(G_2), *_1, *_2\}$ be another neutrosophic subbigroup of $B_N(G)$, where $Q(G_1) = \{e, x, x^2\}$ and $Q(G_2) = \{0, 2, 4, 2+2I, 4+4I, 2I, 4I\}$. Then $(F, A)$ is conjugate soft neutrosophic bigroup over $B_N(G)$, where

$$F(a_1) = \{e, y, 0, 3, 3I, 3+3I\},$$
$$F(a_2) = \{e, x, x^2, 0, 2, 4, 2+2I, 4+4I, 2I, 4I\}.$$

**Theorem 2.2.17:** If $B_N(G)$ is conjugate neutrosophic bigroup, then $(F, A)$ over $B_N(G)$ is conjugate soft neutrosophic bigroup.

**Proof:** This is obvious.

**Theorem 2.2.18:** Every conjugate *soft neutrosophic bigroup* $(F, A)$ over $B_N(G)$ is a soft *neutrosophic bigroup* but the converse is not true.

**Proof:** This is left as an exercise for the readers.

**Proposition 2.2.16:** Let $(F,A)$ and $(K,D)$ be two conjugate soft neutrosophic bigroups over $B_N(G)$. Then

1. Their extended union $(F,A) \cup_\varepsilon (K,D)$ over $B_N(G)$ is not conjugate soft neutrosophic bigroup over $B_N(G)$.
2. Their extended intersection $(F,A) \cap_\varepsilon (K,D)$ over $B_N(G)$ is conjugate soft neutrosophic bigroup over $B_N(G)$.
3. Their restricted union $(F,A) \cup_R (K,D)$ over $B_N(G)$ is not conjugate soft neutrosophic bigroup over $B_N(G)$.
4. Their restricted intersection $(F,A) \cap_R (K,D)$ over $B_N(G)$ is conjgate soft neutrosophic bigroup over $B_N(G)$.

**Proposition 2.2.17:** Let $(F,A)$ and $(K,D)$ be two conjugate soft neutrosophic bigroups over $B_N(G)$. Then

1. Their $AND$ operation $(F,A) \wedge (K,D)$ is conjugate soft neutrosophic bigroup over $B_N(G)$.
2. Their $OR$ operation $(F,A) \vee (K,D)$ is not conjugate soft neutrosophic bigroup over $B_N(G)$.

## Soft Neutrosophic Strong Bigroup

Here we define soft neutrosophic strong bigroup over a neutrosophic strong bigroup which is of pure neutrosophic character. We have also made some fundamental characterization of this purely neutrosophic notion.

**Definition 2.2.11:** Let $(\langle G \cup I \rangle, *_1, *_2)$ be a strong neutrosophic bigroup. Then $(F, A)$ over $(\langle G \cup I \rangle, *_1, *_2)$ is called soft strong neutrosophic bigroup if and only if $F(a)$ is a strong neutrosophic subbigroup of $(\langle G \cup I \rangle, *_1, *_2)$ for all $a \in A$.

**Example 2.2.12:** Let $(\langle G \cup I \rangle, *_1, *_2)$ be a strong neutrosophic bigroup, where $\langle G \cup I \rangle = \langle G_1 \cup I \rangle \cup \langle G_2 \cup I \rangle$ with $\langle G_1 \cup I \rangle = \langle Z \cup I \rangle$, the neutrosophic group under addition and $\langle G_2 \cup I \rangle = \{0,1,2,3,4,I,2I,3I,4I\}$ a neutrosophic group under multiplication modulo 5. Let $H = H_1 \cup H_2$ be a strong neutrosophic subbigroup of $(\langle G \cup I \rangle, *_1, *_2)$, where $H_1 = \{\langle 2Z \cup I \rangle, +\}$ is a neutrosophic subgroup and $H_2 = \{0,1,4,I,4I\}$ is a neutrosophic subgroup. Again let $K = K_1 \cup K_2$ be another strong neutrosophic subbigroup of $(\langle G \cup I \rangle, *_1, *_2)$, where $K_1 = \{\langle 3Z \cup I \rangle, +\}$ is a neutrosophic subgroup and $K_2 = \{0,1,I,2I,3I,4I\}$ is a neutrosophic subgroup. Then clearly $(F, A)$ is a soft neutrosophic strong bigroup over $(\langle G \cup I \rangle, *_1, *_2)$, where

$$F(a_1) = \{0, \pm 2, \pm 4, ..., 1, 4, I, 4I\},$$
$$F(a_2) = \{0, \pm 3, \pm 6, ..., 1, I, 2I, 3I, 4I\}.$$

**Theorem 2.2.19:** Every soft neutrosophic strong bigroup $(F,A)$ is a soft neutrosophic bigroup but the converse is not true.

**Theorem 2.2.20:** If $(\langle G\cup I\rangle,*_1,*_2)$ is a strong neutrosophic bigroup, then $(F,A)$ over $(\langle G\cup I\rangle,*_1,*_2)$ is soft neutrosophic strong bigroup.

**Proposition 2.2.18:** Let $(F,A)$ and $(K,D)$ be two soft neutrosophic strong bigroups over $(\langle G\cup I\rangle,*_1,*_2)$. Then

1) Their extended union $(F,A)\cup_\varepsilon (K,D)$ over $(\langle G\cup I\rangle,*_1,*_2)$ is not soft neutrosophic strong bigroup over $(\langle G\cup I\rangle,*_1,*_2)$.
2) Their extended intersection $(F,A)\cap_\varepsilon (K,D)$ over $(\langle G\cup I\rangle,*_1,*_2)$ is soft neutrosophic strong bigroup over $(\langle G\cup I\rangle,*_1,*_2)$.
3) Their restricted union $(F,A)\cup_R (K,D)$ over $(\langle G\cup I\rangle,*_1,*_2)$ is not soft neutrosophic strong bigroup over $(\langle G\cup I\rangle,*_1,*_2)$.
4) Their restricted intersection $(F,A)\cap_R (K,D)$ over $(\langle G\cup I\rangle,*_1,*_2)$ is soft neutrosophic strong bigroup over $(\langle G\cup I\rangle,*_1,*_2)$.

**Proposition 2.2.19:** Let $(F,A)$ and $(K,D)$ be two soft neutrosophic strong bigroups over $(\langle G\cup I\rangle,*_1,*_2)$. Then

1) Their *AND* operation $(F,A)\wedge(K,D)$ is soft neutrosophic strong bigroup over $(\langle G\cup I\rangle,*_1,*_2)$.
2) Their *OR* operation $(F,A)\vee(K,D)$ is not soft neutrosophic strong bigroup over $(\langle G\cup I\rangle,*_1,*_2)$.

**Definition 2.2.12:** Let $(\langle G \cup I \rangle, *_1, *_2)$ be a neutrosophic strong bigroup. Then $(F,A)$ over $(\langle G \cup I \rangle, *_1, *_2)$ is called Lagrange soft neutrosophic strong bigroup if and only if $F(a)$ is Lagrange subbigroup of $(\langle G \cup I \rangle, *_1, *_2)$ for all $a \in A$.

**Example 2.2.13:** Let $(\langle G \cup I \rangle, *_1, *_2)$ be a strong neutrosophic bigroup of order $15$, where $\langle G \cup I \rangle = \langle G_1 \cup I \rangle \cup \langle G_2 \cup I \rangle$ with $\langle G_1 \cup I \rangle = \{0,1,2,1+I,I,2I,2+I,2+2I,1+2I\}$, the neutrosophic group under mltiplication modulo $3$ and $\langle G_2 \cup I \rangle = \langle A_3 \cup I \rangle = \{e, x, x^2, I, xI, x^2I\}$. Let $H = H_1 \cup H_2$ be a strong neutrosophic subbigroup of $(\langle G \cup I \rangle, *_1, *_2)$, where $H_1 = \{1, 2+2I\}$ is a neutrosophic subgroup and $H_2 = \{e, x, x^2\}$ is a neutrosophic subgroup. Again let $K = K_1 \cup K_2$ be another strong neutrosophic subbigroup of $(\langle G \cup I \rangle, *_1, *_2)$, where $K_1 = \{1, 1+I\}$ is a neutrosophic subgroup and $K_2 = \{I, xI, x^2I\}$ is a neutrosophic subgroup. Then clearly $(F,A)$ is Lagrange soft strong neutrosophic bigroup over $(\langle G \cup I \rangle, *_1, *_2)$, where

$$F(a_1) = \{1, 2+2I, e, x, x^2\},$$
$$F(a_2) = \{1, 1+I, I, xI, x^2I\}.$$

**Theorem 2.2.21:** Every Lagrange soft strong neutrosophic bigroup $(F,A)$ is a soft neutrosophic bigroup but the converse is not true.

**Theorem 2.2.22:** Every Lagrange soft strong neutrosophic bigroup $(F,A)$ is a soft strong neutrosophic bigroup but the converse is not true.

**Theorem 2.2.23:** If $(\langle G \cup I \rangle, *_1, *_2)$ is a Lagrange strong neutrosophic bigroup, then $(F, A)$ over $(\langle G \cup I \rangle, *_1, *_2)$ is a Lagrange soft strong neutrosophic soft bigroup.

**Proposition 2.2.20:** Let $(F, A)$ and $(K, D)$ be two Lagrange soft neutrosophic strong bigroups over $(\langle G \cup I \rangle, *_1, *_2)$. Then

1) Their extended union $(F, A) \cup_\varepsilon (K, D)$ over $(\langle G \cup I \rangle, *_1, *_2)$ is not Lagrange soft neutrosophic strong bigroup over $(\langle G \cup I \rangle, *_1, *_2)$.
2) Their extended intersection $(F, A) \cap_\varepsilon (K, D)$ over $(\langle G \cup I \rangle, *_1, *_2)$ is not Lagrange soft neutrosophic strong bigroup over $(\langle G \cup I \rangle, *_1, *_2)$.
3) Their restricted union $(F, A) \cup_R (K, D)$ over $(\langle G \cup I \rangle, *_1, *_2)$ is not Lagrange soft neutrosophic strong bigroup over $(\langle G \cup I \rangle, *_1, *_2)$.
4) Their restricted intersection $(F, A) \cap_R (K, D)$ over $(\langle G \cup I \rangle, *_1, *_2)$ is not Lagrange soft neutrosophic strong bigroup over $(\langle G \cup I \rangle, *_1, *_2)$.

**Proposition 2.2.21:** Let $(F, A)$ and $(K, D)$ be two Lagrange soft neutrosophic strong bigroups over $(\langle G \cup I \rangle, *_1, *_2)$. Then

1) Their *AND* operation $(F, A) \wedge (K, D)$ is not Lagrange soft neutrosophic strong bigroup over $(\langle G \cup I \rangle, *_1, *_2)$.
2) Their *OR* operation $(F, A) \vee (K, D)$ is not Lagrange soft neutrosophic strong bigroup over $(\langle G \cup I \rangle, *_1, *_2)$.

**Definition 2.2.13:** Let $(\langle G \cup I \rangle, *_1, *_2)$ be a neutrosophic strong bigroup. Then $(F, A)$ over $(\langle G \cup I \rangle, *_1, *_2)$ is called weakly Lagrange soft neutrosophic strong bigroup if atleast one $F(a)$ is a Lagrange subbigroup of $(\langle G \cup I \rangle, *_1, *_2)$ for some $a \in A$.

**Example 2.2.14:** Let $(\langle G \cup I \rangle, *_1, *_2)$ be a strong neutrosophic bigroup of order $15$, where $\langle G \cup I \rangle = \langle G_1 \cup I \rangle \cup \langle G_2 \cup I \rangle$ with $\langle G_1 \cup I \rangle = \{0, 1, 2, 1+I, I, 2I, 2+I, 2+2I, 1+2I\}$, the neutrosophic under mltiplication modulo $3$ and $\langle G_2 \cup I \rangle = \{e, x, x^2, I, xI, x^2I\}$. Let $H = H_1 \cup H_2$ be a strong neutrosophic subbigroup of $(\langle G \cup I \rangle, *_1, *_2)$, where $H_1 = \{1, 2, I, 2I\}$ is a neutrosophic subgroup and $H_2 = \{e, x, x^2\}$ is a neutrosophic subgroup. Again let $K = K_1 \cup K_2$ be another strong neutrosophic subbigroup of $(\langle G \cup I \rangle, *_1, *_2)$, where $K_1 = \{1, 1+I\}$ is a neutrosophic subgroup and $K_2 = \{e, I, xI, x^2I\}$ is a neutrosophic subgroup.
Then clearly $(F, A)$ is weakly Lagrange soft strong neutrosophic bigroup over $(\langle G \cup I \rangle, *_1, *_2)$, where

$$F(a_1) = \{1, 2, I, 2I, e, x, x^2\},$$
$$F(a_2) = \{1, 1+I, e, I, xI, x^2I\}.$$

**Theorem 2.2.24:** Every weakly Lagrange soft neutrosophic strong bigroup $(F, A)$ is a soft neutrosophic bigroup but the converse is not true.

**Theorem 2.2.25:** Every weakly Lagrange soft neutrosophic strong bigroup $(F, A)$ is a soft neutrosophic strong bigroup but the converse is not true.

**Proposition 2.2.22:** Let $(F,A)$ and $(K,D)$ be two weakly Lagrange soft neutrosophic strong bigroups over $(\langle G\cup I\rangle,*_1,*_2)$. Then

1) Their extended union $(F,A)\cup_\varepsilon (K,D)$ over $(\langle G\cup I\rangle,*_1,*_2)$ is not weakly Lagrange soft neutrosophic strong bigroup over $(\langle G\cup I\rangle,*_1,*_2)$. .
2) Their extended intersection $(F,A)\cap_\varepsilon (K,D)$ over $(\langle G\cup I\rangle,*_1,*_2)$ is not weakly Lagrange soft neutrosophic strong bigroup over $(\langle G\cup I\rangle,*_1,*_2)$.
3) Their restricted union $(F,A)\cup_R (K,D)$ over $(\langle G\cup I\rangle,*_1,*_2)$ is not weakly Lagrange soft neutrosophic strong bigroup over $(\langle G\cup I\rangle,*_1,*_2)$. .
4) Their restricted intersection $(F,A)\cap_R (K,D)$ over $(\langle G\cup I\rangle,*_1,*_2)$ is not weakly Lagrange soft neutrosophic strong bigroup over $(\langle G\cup I\rangle,*_1,*_2)$.

**Proposition 2.2.23:** Let $(F,A)$ and $(K,D)$ be two weakly Lagrange soft neutrosophic strong bigroups over $(\langle G\cup I\rangle,*_1,*_2)$. Then

1) Their *AND* operation $(F,A)\wedge(K,D)$ is not weakly Lagrange soft neutrosophic strong bigroup over $(\langle G\cup I\rangle,*_1,*_2)$. .
2) Their *OR* operation $(F,A)\vee(K,D)$ is not weakly Lagrange soft neutrosophic strong bigroup over $(\langle G\cup I\rangle,*_1,*_2)$.

**Definition 2.2.14:** Let $(\langle G\cup I\rangle,*_1,*_2)$ be a neutrosophic strong bigroup. Then $(F,A)$ over $(\langle G\cup I\rangle,*_1,*_2)$ is called Lagrange free soft neutrosophic strong bigroup if and only if $F(a)$ is not Lagrange subbigroup of $(\langle G\cup I\rangle,*_1,*_2)$ for all $a\in A$.

**Example 2.2.14:** Let $(\langle G \cup I \rangle, *_1, *_2)$ be a strong neutrosophic bigroup of order $15$, where $\langle G \cup I \rangle = \langle G_1 \cup I \rangle \cup \langle G_2 \cup I \rangle$ with $\langle G_1 \cup I \rangle = \{0,1,2,3,4,I,2I,3I,4I\}$, the neutrosophic under mltiplication modulo $5$ and $\langle G_2 \cup I \rangle = \{e, x, x^2, I, xI, x^2I\}$, a neutrosophic symmetric group. Let $H = H_1 \cup H_2$ be a strong neutrosophic subbigroup of $(\langle G \cup I \rangle, *_1, *_2)$, where $H_1 = \{1,4,I,4I\}$ is a neutrosophic subgroup and $H_2 = \{e, x, x^2\}$ is a neutrosophic subgroup. Again let $K = K_1 \cup K_2$ be another strong neutrosophic subbigroup of $(\langle G \cup I \rangle, *_1, *_2)$, where $K_1 = \{1, I, 2I, 3I, 4I\}$ is a neutrosophic subgroup and $K_2 = \{e, x, x^2\}$ is a neutrosophic subgroup. Then clearly $(F, A)$ is Lagrange free soft strong neutrosophic bigroup over $(\langle G \cup I \rangle, *_1, *_2)$, where

$$F(a_1) = \{1, 4, I, 4I, e, x, x^2\},$$
$$F(a_2) = \{1, I, 2I, 3I, 4I, e, x, x^2\}.$$

**Theorem 2.2.26:** Every Lagrange free soft neutrosophic strong bigroup $(F, A)$ is a soft neutrosophic bigroup but the converse is not true.

**Theorem 2.2.27:** Every Lagrange free soft neutrosophic strong bigroup $(F, A)$ is a soft neutrosophic strong bigroup but the converse is not true.

**Theorem 2.2.28:** If $(\langle G \cup I \rangle, *_1, *_2)$ is a Lagrange free neutrosophic strong bigroup, then $(F, A)$ over $(\langle G \cup I \rangle, *_1, *_2)$ is also Lagrange free soft neutrosophic strong bigroup.

**Proposition 2.2.24:** Let $(F,A)$ and $(K,D)$ be weakly Lagrange free soft neutrosophic strong bigroups over $(\langle G \cup I \rangle, *_1, *_2)$. Then

1) Their extended union $(F,A) \cup_\varepsilon (K,D)$ over $(\langle G \cup I \rangle, *_1, *_2)$ is not Lagrange free soft neutrosophic strong bigroup over $(\langle G \cup I \rangle, *_1, *_2)$.
2) Their extended intersection $(F,A) \cap_\varepsilon (K,D)$ over $(\langle G \cup I \rangle, *_1, *_2)$ is not Lagrange free soft neutrosophic strong bigroup over $(\langle G \cup I \rangle, *_1, *_2)$.
3) Their restricted union $(F,A) \cup_R (K,D)$ over $(\langle G \cup I \rangle, *_1, *_2)$ is not Lagrange free soft neutrosophic strong bigroup over $(\langle G \cup I \rangle, *_1, *_2)$.
4) Their restricted intersection $(F,A) \cap_R (K,D)$ over $(\langle G \cup I \rangle, *_1, *_2)$ is not Lagrange free soft neutrosophic strong bigroup over $(\langle G \cup I \rangle, *_1, *_2)$.

**Proposition 2.2.25:** Let $(F,A)$ and $(K,D)$ be two Lagrange free soft neutrosophic strong bigroups over $(\langle G \cup I \rangle, *_1, *_2)$. Then

1) Their *AND* operation $(F,A) \wedge (K,D)$ is not Lagrange free soft neutrosophic strong bigroup over $(\langle G \cup I \rangle, *_1, *_2)$.
2) Their *OR* operation $(F,A) \vee (K,D)$ is not Lagrange free soft neutrosophic strong bigroup over $(\langle G \cup I \rangle, *_1, *_2)$.

**Definition 2.2.15:** Let $(\langle G \cup I \rangle, *_1, *_2)$ be a neutrosophic strong bigroup. Then $(F,A)$ over $(\langle G \cup I \rangle, *_1, *_2)$ is called soft normal neutrosophic strong bigroup if and only if $F(a)$ is normal neutrosophic strong subbigroup of $(\langle G \cup I \rangle, *_1, *_2)$ for all $a \in A$.

**Example 2.2.15:** Let $(\langle G \cup I \rangle, *_1, *_2)$ be a neutrosophic strong bigroup of order $15$, where $\langle G \cup I \rangle = \langle G_1 \cup I \rangle \cup \langle G_2 \cup I \rangle$ with $\langle G_1 \cup I \rangle = \{0, 1, 2, 3, 4, I, 2I, 3I, 4I\}$, the neutrosophic under mltiplication modulo $5$ and $\langle G_2 \cup I \rangle = \{e, x, x^2, I, xI, x^2I\}$, a neutrosophic symmetric group.
Then clearly $(F, A)$ is soft normal neutrosophic strong bigroup over $(\langle G \cup I \rangle, *_1, *_2)$, where

$$F(x_1) = \{1, 4, I, 4I, e, x, x^2\},$$
$$F(x_2) = \{1, I, 2I, 3I, 4I, e, x, x^2\}.$$

**Theorem 2.229:** Every soft normal strong neutrosophic bigroup $(F, A)$ over $(\langle G \cup I \rangle, *_1, *_2)$ is a soft neutrosophic bigroup but the converse is not true.

**Proof:** This is obvious.

**Theorem 2.2.30:** Every soft normal strong neutrosophic bigroup $(F, A)$ over $(\langle G \cup I \rangle, *_1, *_2)$ is a soft strong neutrosophic bigroup but the converse is not true.

**Proof:** This is obvious.

**Proposition 2.2.26:** Let $(F,A)$ and $(K,D)$ be two soft normal strong neutrosophic bigroups over $(\langle G \cup I \rangle, *_1, *_2)$. Then

1) Their extended union $(F,A) \cup_\varepsilon (K,D)$ over $(\langle G \cup I \rangle, *_1, *_2)$ is not soft normal strong neutrosophic bigroup over $(\langle G \cup I \rangle, *_1, *_2)$.
2) Their extended intersection $(F,A) \cap_\varepsilon (K,D)$ over $(\langle G \cup I \rangle, *_1, *_2)$ is soft normal strong neutrosophic bigroup over $(\langle G \cup I \rangle, *_1, *_2)$.
3) Their restricted union $(F,A) \cup_R (K,D)$ over $(\langle G \cup I \rangle, *_1, *_2)$ is not soft normal strong neutrosophic bigroup over $(\langle G \cup I \rangle, *_1, *_2)$.
4) Their restricted intersection $(F,A) \cap_R (K,D)$ over $(\langle G \cup I \rangle, *_1, *_2)$ is soft normal strong neutrosophic bigroup over $(\langle G \cup I \rangle, *_1, *_2)$.

**Proposition 2.2.27** Let $(F,A)$ and $(K,D)$ be two soft normal strong neutrosophic bigroups over $(\langle G \cup I \rangle, *_1, *_2)$. Then

1) Their AND operation $(F,A) \wedge (K,D)$ is soft normal strong neutrosophic bigroup over $(\langle G \cup I \rangle, *_1, *_2)$.
2) Their OR operation $(F,A) \vee (K,D)$ is not soft normal strong neutrosophic bigroup over $(\langle G \cup I \rangle, *_1, *_2)$.

**Definition 2.2.16:** Let $(\langle G \cup I \rangle, *_1, *_2)$ be a neutrosophic strong bigroup. Then $(F,A)$ over $(\langle G \cup I \rangle, *_1, *_2)$ is called soft conjugate neutrosophic strong bigroup if and only if $F(a)$ is conjugate neutrosophic subbigroup of $(\langle G \cup I \rangle, *_1, *_2)$ for all $a \in A$.

**Example 2.2.16:** Let $(\langle G \cup I\rangle, *_1, *_2)$ be a strong neutrosophic bigroup, where $\langle G \cup I\rangle = \langle G_1 \cup I\rangle \cup \langle G_2 \cup I\rangle$ with
$\langle G_1 \cup I\rangle = \{0,1,2,1+I,I,2I,2+I,2+2I,1+2I\}$, the neutrosophic under mltiplication modulo 3 and $\langle G_2 \cup I\rangle = \{e, x, x^2, I, xI, x^2I\}$.
Then clearly $(F, A)$ is soft conjugate neutrosophic strong bigroup over $(\langle G \cup I\rangle, *_1, *_2)$, where

$$F(x_1) = \{1, 2, I, 2I, e, x, x^2\},$$
$$F(x_2) = \{1, 1+I, e, I, xI, x^2I\}.$$

**Theorem 2.2.31:** Every soft conjugate neutrosophic strong bigroup $(F, A)$ over $(\langle G \cup I\rangle, *_1, *_2)$ is a soft neutrosophic bigroup but the converse is not true.

**Proof:** This is obvious.

**Theorem 2.2.32:** Every soft conjugate neutrosophic strong bigroup $(F, A)$ over $(\langle G \cup I\rangle, *_1, *_2)$ is a soft neutrosophic strong bigroup but the converse is not true.

**Proof:** This is obvious.

**Proposition 2.2.28:** Let $(F,A)$ and $(K,D)$ be two soft conjugate neutrosophic strong bigroups over $(\langle G \cup I \rangle, *_1, *_2)$. Then

1) Their extended union $(F,A) \cup_\varepsilon (K,D)$ over $(\langle G \cup I \rangle, *_1, *_2)$ is not soft conjugate neutrosophic strong bigroup over $(\langle G \cup I \rangle, *_1, *_2)$.
2) Their extended intersection $(F,A) \cap_\varepsilon (K,D)$ over $(\langle G \cup I \rangle, *_1, *_2)$ is soft conjugate neutrosophic strong bigroup over $(\langle G \cup I \rangle, *_1, *_2)$.
3) Their restricted union $(F,A) \cup_R (K,D)$ over $(\langle G \cup I \rangle, *_1, *_2)$ is not soft conjugate neutrosophic strong bigroup over $(\langle G \cup I \rangle, *_1, *_2)$.
4) Their restricted intersection $(F,A) \cap_R (K,D)$ over $(\langle G \cup I \rangle, *_1, *_2)$ is soft conjugate neutrosophic strong bigroup over $(\langle G \cup I \rangle, *_1, *_2)$.

**Proposition 2.2.29:** Let $(F,A)$ and $(K,D)$ be two soft conjugate neutrosophic strong bigroups over $(\langle G \cup I \rangle, *_1, *_2)$. Then

1) Their *AND* operation $(F,A) \wedge (K,D)$ is soft conjugate neutrosophic strong bigroup over $(\langle G \cup I \rangle, *_1, *_2)$.
2) Their *OR* operation $(F,A) \vee (K,D)$ is not soft conjgate neutrosophic strong bigroup over $(\langle G \cup I \rangle, *_1, *_2)$.

## 2.3 Soft Neutrosophic N-Group

In this section, we extend soft sets to neutrosophic N-groups and introduce soft neutrosophic N-groups. This is the generalization of soft neutrosophic groups. Some of their impotant facts and figures are also presented here with illustrative examples. We also initiated the strong part of neutrosophy in this section. Now we proceed onto define soft neutrosophic N-groups as follows.

**Definition 2.3.1:** Let $(\langle G \cup I \rangle, *_1, \ldots, *_N)$ be a neutrosophic $N$-group and $(F, A)$ be a soft set over $(\langle G \cup I \rangle, *_1, \ldots, *_2)$. Then $(F, A)$ over $(\langle G \cup I \rangle, *_1, \ldots, *_2)$ is called soft neutrosophic $N$-group if and only if $F(a)$ is a sub $N$-group of $(\langle G \cup I \rangle, *_1, \ldots, *_2)$ for all $a \in A$.

For further understanding, we give the following examples.

**Example 2.3.1:** Let $(\langle G \cup I \rangle = \langle G_1 \cup I \rangle \cup \langle G_2 \cup I \rangle \cup \langle G_3 \cup I \rangle, *_1, *_2, *_3)$ be a neutrosophic 3-group, where $\langle G_1 \cup I \rangle = \langle Q \cup I \rangle$ a neutrosophic group under multiplication. $\langle G_2 \cup I \rangle = \{0,1,2,3,4,I,2I,3I,4I\}$ neutrosophic group under multiplication modulo 5 and $\langle G_3 \cup I \rangle = \{0,1,2,1+I,2+I,I,2I,1+2I,2+2I\}$ a neutrosophic group under multiplication modulo 3. Let
$P = \left\{ \left\{ \left\langle \frac{1}{2^n}, 2^n, \frac{1}{(2I)^n}, (2I)^n, I, 1 \right\rangle \right\}, (1,4,I,4I), (1,2,I,2I) \right\}$, $T = \{Q \setminus \{0\}, \{1,2,3,4\}, \{1,2\}\}$ and $X = \{Q \setminus \{0\}, \{1,2,I,2I\}, \{1,4,I,4I\}\}$ are sub 3-groups.

Then $(F, A)$ is clearly soft neutrosophic $3$-group over
$(\langle G \cup I \rangle = \langle G_1 \cup I \rangle \cup \langle G_2 \cup I \rangle \cup \langle G_3 \cup I \rangle, *_1, *_2, *_3)$, where

$$F(a_1) = \left\{ \left\{ \left\langle \frac{1}{2^n}, 2^n, \frac{1}{(2I)^n}, (2I)^n, I, 1 \right\rangle \right\}, (1, 4, I, 4I), (1, 2, I, 2I) \right\},$$

$$F(a_2) = \{Q \setminus \{0\}, \{1, 2, 3, 4\}, \{1, 2\}\},$$

$$F(a_3) = \{Q \setminus \{0\}, \{1, 2, I, 2I\}, \{1, 4, I, 4I\}\}.$$

**Example 2.3.2:** Let $(\langle G \cup I \rangle = \langle G_1 \cup I \rangle \cup G_2 \cup G_3, *_1, *_2, *_3)$ be neutrosophic $N$-group, where $\langle G_1 \cup I \rangle = \{\langle Z_6 \cup I \rangle\}$ is a group under addition modulo $6$, $G_2 = A_4$ and $G_3 = \langle g : g^{12} = 1 \rangle$, a cyclic group.
Take $P = (\langle P_1 \cup I \rangle \cup P_2 \cup P_3, *_1, *_2, *_3)$, a neutrosophic sub
$3$-group where $\langle T_1 \cup I \rangle = \{0, 3, 3I, 3 + 3I\}$, $P_2 = \left\{ \begin{pmatrix} 1234 \\ 1234 \end{pmatrix}, \begin{pmatrix} 1234 \\ 2143 \end{pmatrix}, \begin{pmatrix} 1234 \\ 4321 \end{pmatrix}, \begin{pmatrix} 1234 \\ 3412 \end{pmatrix} \right\}$,
$P_3 = \{1, g^6\}$. Since $P$ is a Lagrange neutrosophic sub $3$-group.
Let us Take $T = (\langle T_1 \cup I \rangle \cup T_2 \cup T_3, *_1, *_2, *_3)$, where $\langle T_1 \cup I \rangle = \{0, 3, 3I, 3 + 3I\}$, $T_2 = P_2$
and $T_3 = \{g^3, g^6, g^9, 1\}$ is another Lagrange sub $3$-group.
Let $(F, A)$ be a soft neutrosophic $N$-group over
$(\langle G \cup I \rangle = \langle G_1 \cup I \rangle \cup G_2 \cup G_3, *_1, *_2, *_3)$, where

$$F(a_1) = \left\{ 0, 3, 3I, 3 + 3I, 1, g^6, \begin{pmatrix} 1234 \\ 1234 \end{pmatrix}, \begin{pmatrix} 1234 \\ 2143 \end{pmatrix}, \begin{pmatrix} 1234 \\ 4321 \end{pmatrix}, \begin{pmatrix} 1234 \\ 3412 \end{pmatrix} \right\},$$

$$F(a_2) = \left\{ 0, 3, 3I, 3 + 3I, 1, g^3, g^6, g^9, \begin{pmatrix} 1234 \\ 1234 \end{pmatrix}, \begin{pmatrix} 1234 \\ 2143 \end{pmatrix}, \begin{pmatrix} 1234 \\ 4321 \end{pmatrix}, \begin{pmatrix} 1234 \\ 3412 \end{pmatrix} \right\}.$$

**Theorem 2.3.1:** Let $(F,A)$ and $(H,A)$ be two soft neutrosophic $N$-groups over $(\langle G \cup I \rangle, *_1,...,*_N)$. Then their intersection $(F,A) \cap (H,A)$ is again a soft neutrosophic $N$-group over $(\langle G \cup I \rangle, *_1,...,*_N)$.

**Proof** The proof is straight forward.

**Theorem 2.3.2:** Let $(F,A)$ and $(H,B)$ be two soft neutrosophic $N$-groups over $(\langle G \cup I \rangle, *_1,...,*_N)$ such that $A \cap B = \phi$, then their union is soft neutrosophic $N$-group over $(\langle G \cup I \rangle, *_1,...,*_N)$.

**Proof** The proof can be easily established.

**Proposition 2.3.1:** Let $(F,A)$ and $(K,D)$ be two soft neutrosophic $N$-groups over $(\langle G \cup I \rangle, *_1,...,*_N)$. Then

1) Their extended union $(F,A) \cup_\varepsilon (K,D)$ is not soft neutrosophic $N$-group over $(\langle G \cup I \rangle, *_1,...,*_N)$.
2) Their extended intersection $(F,A) \cap_\varepsilon (K,D)$ is soft neutrosophic $N$-group over $(\langle G \cup I \rangle, *_1,...,*_N)$.
3) Their restricted union $(F,A) \cup_R (K,D)$ is not soft neutrosophic $N$-group over $(\langle G \cup I \rangle, *_1,...,*_N)$.
4) Their restricted intersection $(F,A) \cap_R (K,D)$ is soft neutrosophic $N$-group over $(\langle G \cup I \rangle, *_1,...,*_N)$.

**Proposition 2.3.2:** Let $(F,A)$ and $(K,D)$ be two soft neutrosophic $N$-groups over $(\langle G \cup I \rangle, *_1,...,*_N)$. Then

1) Their *AND* operation $(F,A) \wedge (K,D)$ is soft neutrosophic $N$-group over $(\langle G \cup I \rangle, *_1,...,*_N)$.
2) Their *OR* operation $(F,A) \vee (K,D)$ is not soft neutrosophic $N$-group over $(\langle G \cup I \rangle, *_1,...,*_N)$.

**Definition 2.3.2:** Let $(F,A)$ be a soft neutrosophic $N$-group over $(\langle G \cup I \rangle, *_1,...,*_N)$. Then

1) $(F,A)$ is called identity soft neutrosophic $N$-group if $F(a) = \{e_1,...,e_N\}$ for all $a \in A$, where $e_1,...,e_N$ are the identities of $\langle G_1 \cup I \rangle,...,\langle G_N \cup I \rangle$ respectively.
2) $(F,A)$ is called Full soft neutrosophic $N$-group if $F(a) = (\langle G \cup I \rangle, *_1,...,*_N)$ for all $a \in A$.

**Definition 2.3.3:** Let $(F,A)$ and $(K,D)$ be two soft neutrosophic $N$-groups over $(\langle G \cup I \rangle, *_1,...,*_N)$. Then $(K,D)$ is soft neutrosophic sub $N$-group of $(F,A)$ written as $(K,D) \prec (F,A)$, if

1. $D \subseteq A$
2. $K(a) \prec F(a)$ for all $a \in A$.

**Example 2.3.3:** Let $(F, A)$ be as in Example 22. Let $(K, D)$ be another soft neutrosophic soft $N$-group over $(\langle G \cup I \rangle = \langle G_1 \cup I \rangle \cup \langle G_2 \cup I \rangle \cup \langle G_3 \cup I \rangle, *_1, *_2, *_3)$, where

$$K(a_1) = \left\{ \left\{ \left\langle \frac{1}{2^n}, 2^n \right\rangle \right\}, \{1, 4, I, 4I\}, \{1, 2, I, 2I\} \right\},$$

$$K(a_2) = \{Q \setminus \{0\}, \{1, 4\}, \{1, 2\}\}.$$

Clearly $(K, D) \prec (F, A)$.

**Note:** Thus a soft neutrosophic $N$-group can have two types of soft neutrosophic sub $N$-groups, which are following.

**Definition 2.3.4:** A soft neutrosophic sub $N$-group $(K, D)$ of a soft neutrosophic $N$-group $(F, A)$ is called soft strong neutrosophic sub $N$-group if

1. $D \subseteq A$,
2. $K(a)$ is neutrosophic sub $N$-group of $F(a)$ for all $a \in A$.

**Definition 2.3.5:** A soft neutrosophic sub $N$-group $(K, D)$ of a soft neutrosophic $N$-group $(F, A)$ is called soft sub $N$-group if

1. $D \subseteq A$,
2. $K(a)$ is only sub $N$-group of $F(a)$ for all $a \in A$.

**Definition 2.3.6:** Let $(\langle G \cup I \rangle, *_1, ..., *_N)$ be a neutrosophic $N$-group. Then $(F, A)$ over $(\langle G \cup I \rangle, *_1, ..., *_N)$ is called soft Lagrange neutrosophic $N$-group if and only if $F(a)$ is Lagrange sub $N$-group of $(\langle G \cup I \rangle, *_1, ..., *_N)$ for all $a \in A$.

**Example 2.3.4:** Let $(\langle G \cup I \rangle = \langle G_1 \cup I \rangle \cup G_2 \cup G_3, *_1, *_2, *_3)$ be neutrosophic $N$-group, where $\langle G_1 \cup I \rangle = \{\langle Z_6 \cup I \rangle\}$ is a group under addition modulo $6$, $G_2 = A_4$ and $G_3 = \langle g : g^{12} = 1 \rangle$, a cyclic group of order $12$, $o(\langle G \cup I \rangle) = 60$. Take $P = (\langle P_1 \cup I \rangle \cup P_2 \cup P_3, *_1, *_2, *_3)$, a neutrosophic sub 3-group where $\langle T_1 \cup I \rangle = \{0, 3, 3I, 3+3I\}$, $P_2 = \left\{ \begin{pmatrix} 1234 \\ 1234 \end{pmatrix}, \begin{pmatrix} 1234 \\ 2143 \end{pmatrix}, \begin{pmatrix} 1234 \\ 4321 \end{pmatrix}, \begin{pmatrix} 1234 \\ 3412 \end{pmatrix} \right\}$, $P_3 = \{1, g^6\}$.

Since $P$ is a Lagrange neutrosophic sub 3-group where order of $P = 10$. Let us Take $T = (\langle T_1 \cup I \rangle \cup T_2 \cup T_3, *_1, *_2, *_3)$, where $\langle T_1 \cup I \rangle = \{0, 3, 3I, 3+3I\}$, $T_2 = P_2$ and $T_3 = \{g^3, g^6, g^9, 1\}$ is another Lagrange sub 3-group where $o(T) = 12$.

Let $(F, A)$ is soft Lagrange neutrosophic $N$-group over $(\langle G \cup I \rangle = \langle G_1 \cup I \rangle \cup G_2 \cup G_3, *_1, *_2, *_3)$, where

$$F(a_1) = \left\{ 0, 3, 3I, 3+3I, 1, g^6, \begin{pmatrix} 1234 \\ 1234 \end{pmatrix}, \begin{pmatrix} 1234 \\ 2143 \end{pmatrix}, \begin{pmatrix} 1234 \\ 4321 \end{pmatrix}, \begin{pmatrix} 1234 \\ 3412 \end{pmatrix} \right\},$$

$$F(a_2) = \left\{ 0, 3, 3I, 3+3I, 1, g^3, g^6, g^9, \begin{pmatrix} 1234 \\ 1234 \end{pmatrix}, \begin{pmatrix} 1234 \\ 2143 \end{pmatrix}, \begin{pmatrix} 1234 \\ 4321 \end{pmatrix}, \begin{pmatrix} 1234 \\ 3412 \end{pmatrix} \right\}.$$

**Theorem 2.3.3:** Every soft Lagrange neutrosophic $N$-group $(F,A)$ over $(\langle G\cup I\rangle,*_1,...,*_N)$ is a soft neutrosophic $N$-group but the converse is not true.

**Proof:** This is obvious.

**Theorem 2.3.4:** If $(\langle G\cup I\rangle,*_1,...,*_N)$ is a Lagrange neutrosophic $N$-group, then $(F,A)$ over $(\langle G\cup I\rangle,*_1,...,*_N)$ is also soft Lagrange neutrosophic $N$-group.

**Proof:** The proof is left as an exercise for the interested readers.

**Proposition 2.3.3:** Let $(F,A)$ and $(K,D)$ be two soft Lagrange neutrosophic $N$-groups over $(\langle G\cup I\rangle,*_1,...,*_N)$. Then

1. Their extended union $(F,A)\cup_\varepsilon (K,D)$ is not soft Lagrange neutrosophic $N$-group over $(\langle G\cup I\rangle,*_1,...,*_N)$.
2. Their extended intersection $(F,A)\cap_\varepsilon (K,D)$ is not soft Lagrange neutrosophic $N$-group over $(\langle G\cup I\rangle,*_1,...,*_N)$.
3. Their restricted union $(F,A)\cup_R (K,D)$ is not soft Lagrange neutrosophic $N$-group over $(\langle G\cup I\rangle,*_1,...,*_N)$.
4. Their restricted intersection $(F,A)\cap_R (K,D)$ is not soft Lagrange neutrosophic $N$-group over $(\langle G\cup I\rangle,*_1,...,*_N)$.

**Proposition 2.3.4:** Let $(F, A)$ and $(K, D)$ be two soft Lagrange neutrosophic $N$-groups over $(\langle G \cup I \rangle, *_1, ..., *_N)$. Then

1. Their *AND* operation $(F, A) \wedge (K, D)$ is not soft Lagrange neutrosophic $N$-group over $(\langle G \cup I \rangle, *_1, ..., *_N)$.
2. Their *OR* operation $(F, A) \vee (K, D)$ is not soft Lagrange neutrosophic $N$-group over $(\langle G \cup I \rangle, *_1, ..., *_N)$.

**Definition 2.3.7:** Let $(\langle G \cup I \rangle, *_1, ..., *_N)$ be a neutrosophic $N$-group. Then $(F, A)$ over $(\langle G \cup I \rangle, *_1, ..., *_N)$ is called soft weakly Lagrange neutrosophic $N$-group if atleast one $F(a)$ is a Lagrange sub $N$-group of $(\langle G \cup I \rangle, *_1, ..., *_N)$ for some $a \in A$.

**Examp 2.3.5:** Let $(\langle G \cup I \rangle = \langle G_1 \cup I \rangle \cup G_2 \cup G_3, *_1, *_2, *_3)$ be neutrosophic $N$-group, where $\langle G_1 \cup I \rangle = \{\langle Z_6 \cup I \rangle\}$ is a group under addition modulo $6$, $G_2 = A_4$ and $G_3 = \langle g : g^{12} = 1 \rangle$, a cyclic group of order 12, $o(\langle G \cup I \rangle) = 60$. Take $P = (\langle P_1 \cup I \rangle \cup P_2 \cup P_3, *_1, *_2, *_3)$, a neutrosophic sub 3-group where $\langle T_1 \cup I \rangle = \{0, 3, 3I, 3+3I\}$, $P_2 = \left\{ \begin{pmatrix} 1234 \\ 1234 \end{pmatrix}, \begin{pmatrix} 1234 \\ 2143 \end{pmatrix}, \begin{pmatrix} 1234 \\ 4321 \end{pmatrix}, \begin{pmatrix} 1234 \\ 3412 \end{pmatrix} \right\}$, $P_3 = \{1, g^6\}$.

Since $P$ is a Lagrange neutrosophic sub 3-group where order of $P = 10$. Let us Take $T = (\langle T_1 \cup I \rangle \cup T_2 \cup T_3, *_1, *_2, *_3)$, where $\langle T_1 \cup I \rangle = \{0, 3, 3I, 3+3I\}, T_2 = P_2$ and $T_3 = \{g^4, g^8, 1\}$ is another Lagrange sub 3-group.

Then $(F,A)$ is a soft weakly Lagrange neutrosophic $N$-group over $(\langle G \cup I \rangle = \langle G_1 \cup I \rangle \cup G_2 \cup G_3, *_1, *_2, *_3)$, where

$$F(a_1) = \left\{ 0, 3, 3I, 3+3I, 1, g^6, \begin{pmatrix} 1234 \\ 1234 \end{pmatrix}, \begin{pmatrix} 1234 \\ 2143 \end{pmatrix}, \begin{pmatrix} 1234 \\ 4321 \end{pmatrix}, \begin{pmatrix} 1234 \\ 3412 \end{pmatrix} \right\},$$

$$F(a_2) = \left\{ 0, 3, 3I, 3+3I, 1, g^4, g^8, \begin{pmatrix} 1234 \\ 1234 \end{pmatrix}, \begin{pmatrix} 1234 \\ 2143 \end{pmatrix}, \begin{pmatrix} 1234 \\ 4321 \end{pmatrix}, \begin{pmatrix} 1234 \\ 3412 \end{pmatrix} \right\}.$$

**Theorem 2.3.5:** Every soft weakly Lagrange neutrosophic $N$-group $(F,A)$ over $(\langle G \cup I \rangle, *_1, ..., *_N)$ is a soft neutrosophic $N$-group but the converse is not tue.

**Proof:** This is obvious.

**Theorem 39** If $(\langle G \cup I \rangle, *_1, ..., *_N)$ is a weakly Lagrange neutrosophi $N$-group, then $(F,A)$ over $(\langle G \cup I \rangle, *_1, ..., *_N)$ is also soft weakly Lagrange neutrosophic $N$-group.

**Proof:** This is obvious.

**Proposition 2.3.5:** Let $(F,A)$ and $(K,D)$ be two soft weakly Lagrange neutrosophic $N$-groups over $(\langle G\cup I\rangle,*_1,...,*_N)$. Then

1. Their extended union $(F,A)\cup_\varepsilon (K,D)$ is not soft weakly Lagrange neutrosophic $N$-group over $(\langle G\cup I\rangle,*_1,...,*_N)$.
2. Their extended intersection $(F,A)\cap_\varepsilon (K,D)$ is not soft weakly Lagrange neutrosophic $N$-group over $(\langle G\cup I\rangle,*_1,...,*_N)$.
3. Their restricted union $(F,A)\cup_R (K,D)$ is not soft weakly Lagrange neutrosophic $N$-group over $(\langle G\cup I\rangle,*_1,...,*_N)$.
4. Their restricted intersection $(F,A)\cap_R (K,D)$ is not soft weakly Lagrange neutrosophic $N$-group over $(\langle G\cup I\rangle,*_1,...,*_N)$.

**Proposition 2.3.6:** Let $(F,A)$ and $(K,D)$ be two soft weakly Lagrange neutrosophic $N$-groups over $(\langle G\cup I\rangle,*_1,...,*_N)$. Then

1) Their $AND$ operation $(F,A)\wedge(K,D)$ is not soft weakly Lagrange neutrosophic $N$-group over $(\langle G\cup I\rangle,*_1,...,*_N)$.
2) Their $OR$ operation $(F,A)\vee(K,D)$ is not soft weakly Lagrange neutrosophic $N$-group over $(\langle G\cup I\rangle,*_1,...,*_N)$.

**Definition 2.3.8:** Let $(\langle G\cup I\rangle,*_1,...,*_N)$ be a neutrosophic $N$-group. Then $(F,A)$ over $(\langle G\cup I\rangle,*_1,...,*_N)$ is called soft Lagrange free neutro neutrosophic $N$-group if $F(a)$ is not Lagrange sub $N$-group of $(\langle G\cup I\rangle,*_1,...,*_N)$ for all $a\in A$.

**Example 2.3.6:** Let $(\langle G \cup I \rangle = \langle G_1 \cup I \rangle \cup G_2 \cup G_3, *_1, *_2, *_3)$ be neutrosophic 3-group, where $\langle G_1 \cup I \rangle = \{\langle Z_6 \cup I \rangle\}$ is a group under addition modulo 6, $G_2 = A_4$ and $G_3 = \langle g : g^{12} = 1 \rangle$, a cyclic group of order 12, $o(\langle G \cup I \rangle) = 60$. Take $P = (\langle P_1 \cup I \rangle \cup P_2 \cup P_3, *_1, *_2, *_3)$, a neutrosophic sub 3-group where

$$P_1 = \{0, 2, 4\}, \quad P_2 = \left\{ \begin{pmatrix} 1234 \\ 1234 \end{pmatrix}, \begin{pmatrix} 1234 \\ 2143 \end{pmatrix}, \begin{pmatrix} 1234 \\ 4321 \end{pmatrix}, \begin{pmatrix} 1234 \\ 3412 \end{pmatrix} \right\}, \quad P_3 = \{1, g^6\}.$$

Since $P$ is a Lagrange neutrosophic sub 3-group where order of $P = 10$. Let us Take $T = (\langle T_1 \cup I \rangle \cup T_2 \cup T_3, *_1, *_2, *_3)$, where $\langle T_1 \cup I \rangle = \{0, 3, 3I, 3+3I\}, T_2 = P_2$ and $T_3 = \{g^4, g^8, 1\}$ is another Lagrange sub 3-group.

Then $(F, A)$ is soft Lagrange free neutrosophic 3-group over $(\langle G \cup I \rangle = \langle G_1 \cup I \rangle \cup G_2 \cup G_3, *_1, *_2, *_3)$, where

$$F(a_1) = \left\{ 0, 2, 4, 1, g^6, \begin{pmatrix} 1234 \\ 1234 \end{pmatrix}, \begin{pmatrix} 1234 \\ 2143 \end{pmatrix}, \begin{pmatrix} 1234 \\ 4321 \end{pmatrix}, \begin{pmatrix} 1234 \\ 3412 \end{pmatrix} \right\},$$

$$F(a_2) = \left\{ 0, 3, 3I, 3+3I, 1, g^4, g^8, \begin{pmatrix} 1234 \\ 1234 \end{pmatrix}, \begin{pmatrix} 1234 \\ 2143 \end{pmatrix}, \begin{pmatrix} 1234 \\ 4321 \end{pmatrix}, \begin{pmatrix} 1234 \\ 3412 \end{pmatrix} \right\}$$

**Theorem 2.3.6:** Every soft Lagrange free neutrosophic $N$-group $(F, A)$ over $(\langle G \cup I \rangle, *_1, \ldots, *_N)$ is a soft neutrosophic $N$-group but the converse is not true.

**Theorem 2.3.7:** If $(\langle G \cup I \rangle, *_1, \ldots, *_N)$ is a Lagrange free neutrosophic $N$-group, then $(F, A)$ over $(\langle G \cup I \rangle, *_1, \ldots, *_N)$ is also soft Lagrange free neutrosophic $N$-group.

**Proposition 2.3.7:** Let $(F,A)$ and $(K,D)$ be two soft Lagrange free neutrosophic $N$-groups over $(\langle G \cup I \rangle, *_1, ..., *_N)$. Then

1) Their extended union $(F,A) \cup_\varepsilon (K,D)$ is not soft Lagrange free neutrosophic $N$-group over $(\langle G \cup I \rangle, *_1, ..., *_N)$.
2) Their extended intersection $(F,A) \cap_\varepsilon (K,D)$ is not soft Lagrange free neutrosophic $N$-group over $(\langle G \cup I \rangle, *_1, ..., *_N)$.
3) Their restricted union $(F,A) \cup_R (K,D)$ is not soft Lagrange free neutrosophic $N$-group over $(\langle G \cup I \rangle, *_1, ..., *_N)$.
4) Their restricted intersection $(F,A) \cap_R (K,D)$ is not soft Lagrange free neutrosophic $N$-group over $(\langle G \cup I \rangle, *_1, ..., *_N)$.

**Proposition 2.3.8:** Let $(F,A)$ and $(K,D)$ be two soft Lagrange free neutrosophic $N$-groups over $(\langle G \cup I \rangle, *_1, ..., *_N)$. Then

1) Their *AND* operation $(F,A) \wedge (K,D)$ is not soft Lagrange free neutrosophic $N$-group over $(\langle G \cup I \rangle, *_1, ..., *_N)$.
2) Their *OR* operation $(F,A) \vee (K,D)$ is not soft Lagrange free neutrosophic $N$-group over $(\langle G \cup I \rangle, *_1, ..., *_N)$.

**Definition 2.3.9:** Let $(\langle G \cup I \rangle, *_1, ..., *_N)$ be a neutrosophic $N$-group. Then $(F,A)$ over $(\langle G \cup I \rangle, *_1, ..., *_N)$ is called soft normal neutrosophic $N$-group if $F(a)$ is normal sub $N$-group of $(\langle G \cup I \rangle, *_1, ..., *_N)$ for all $a \in A$.

**Example 2.3.7:** Let $(\langle G_1 \cup I \rangle = \langle G_1 \cup I \rangle \cup G_2 \cup \langle G_3 \cup I \rangle, *_1, *_2, *_3)$ be a soft neutrosophic $N$-group, where $\langle G_1 \cup I \rangle = \{e, y, x, x^2, xy, x^2y, I, yI, xI, x^2I, xyI, x^2yI\}$ is a neutrosophic group under multiplaction, $G_2 = \{g : g^6 = 1\}$, a cyclic group of order 6 and $\langle G_3 \cup I \rangle = \langle Q_8 \cup I \rangle = \{\pm 1, \pm i, \pm j, \pm k, \pm I, \pm iI, \pm jI, \pm kI\}$ is a group under multiplication. Let $P = (\langle P_1 \cup I \rangle \cup P_2 \cup \langle P_3 \cup I \rangle, *_1, *_2, *_3)$, a normal sub 3-group where $P_1 = \{e, y, I, yI\}$, $P_2 = \{1, g^2, g^4\}$ and $P_3 = \{1, -1\}$.

Also $T = (\langle T_1 \cup I \rangle \cup T_2 \cup \langle T_3 \cup I \rangle, *_1, *_2, *_3)$ be another normal sub 3-group where $\langle T_1 \cup I \rangle = \{e, I, xI, x^2I\}, T_2 = \{1, g^3\}$ and $\langle T_3 \cup I \rangle = \{\pm 1, \pm i\}$.

Then $(F, A)$ is a soft normal neutrosophic $N$-group over $(\langle G_1 \cup I \rangle = \langle G_1 \cup I \rangle \cup G_2 \cup \langle G_3 \cup I \rangle, *_1, *_2, *_3)$, where

$$F(a_1) = \{e, y, I, yI, 1, g^2, g^4, \pm 1\},$$
$$F(a_2) = \{e, I, xI, x^2I, 1, g^3, \pm 1, \pm i\}.$$

**Theorem 2.3.8:** Every soft normal neutrosophic $N$-group $(F, A)$ over $(\langle G \cup I \rangle, *_1, \ldots, *_N)$ is a soft neutrosophic $N$-group but the converse is not true.

It is left as an exercise for the readers to prove the converse with the help of examples.

**Proposition 2.3.9:** Let $(F,A)$ and $(K,D)$ be two soft normal neutrosophic $N$-groups over $(\langle G \cup I \rangle, *_1, ..., *_N)$. Then

1. Their extended union $(F,A) \cup_\varepsilon (K,D)$ is not soft normal neutrosophic soft $N$-group over $(\langle G \cup I \rangle, *_1, ..., *_N)$.
2. Their extended intersection $(F,A) \cap_\varepsilon (K,D)$ is soft normal neutrosophic $N$-group over $(\langle G \cup I \rangle, *_1, ..., *_N)$.
3. Their restricted union $(F,A) \cup_R (K,D)$ is not soft normal neutrosophic $N$-group over $(\langle G \cup I \rangle, *_1, ..., *_N)$.
4. Their restricted intersection $(F,A) \cap_R (K,D)$ is soft normal neutrosophic $N$-group over $(\langle G \cup I \rangle, *_1, ..., *_N)$.

**Proposition 2.3.10:** Let $(F,A)$ and $(K,D)$ be two soft normal neutrosophic $N$-groups over $(\langle G \cup I \rangle, *_1, ..., *_N)$. Then

1. Their *AND* operation $(F,A) \wedge (K,D)$ is soft normal neutrosophic $N$-group over $(\langle G \cup I \rangle, *_1, ..., *_N)$.
2. Their *OR* operation $(F,A) \vee (K,D)$ is not soft normal neutrosophic $N$-group over $(\langle G \cup I \rangle, *_1, ..., *_N)$.

**Definition 2.3.10:** Let $(\langle G \cup I \rangle, *_1, ..., *_N)$ be a neutrosophic $N$-group. Then $(F,A)$ over $(\langle G \cup I \rangle, *_1, ..., *_N)$ is called soft conjugate neutrosophic $N$-group if $F(a)$ is a conjugate sub $N$-group of $(\langle G \cup I \rangle, *_1, ..., *_N)$ for all $a \in A$.

**Example 2.3.8:** Let $(\langle G_1 \cup I \rangle = \langle G_1 \cup I \rangle \cup G_2 \cup \langle G_3 \cup I \rangle, *_1, *_2, *_3)$ be a soft neutrosophic $N$-group, where $\langle G_1 \cup I \rangle = \{e, y, x, x^2, xy, x^2y, I, yI, xI, x^2I, xyI, x^2yI\}$ is a neutrosophic group under multiplaction, $G_2 = \{g : g^6 = 1\}$, a cyclic group of order 6 and $\langle G_3 \cup I \rangle = \langle Q_8 \cup I \rangle = \{\pm 1, \pm i, \pm j, \pm k, \pm I, \pm iI, \pm jI, \pm kI\}$ is a group under multiplication.

Then $(F, A)$ is a soft conjugate neutrosophic $N$-group over $(\langle G_1 \cup I \rangle = \langle G_1 \cup I \rangle \cup G_2 \cup \langle G_3 \cup I \rangle, *_1, *_2, *_3)$, where

$$F(a_1) = \{I, yI, 1, g^2, g^4, \pm 1\},$$
$$F(a_2) = \{I, xI, x^2I, 1, g^3, \pm 1, \pm i\}.$$

**Theorem 2.3.9:** Every soft conjugate neutrosophic $N$-group $(F, A)$ over $(\langle G \cup I \rangle, *_1, ..., *_N)$ is a soft neutrosophic $N$-group but the converse is not true.

**Proposition 2.3.11:** Let $(F, A)$ and $(K, D)$ be two soft conjugate neutrosophic $N$-groups over $(\langle G \cup I \rangle, *_1, ..., *_N)$. Then

1. Their extended union $(F, A) \cup_\varepsilon (K, D)$ is not soft conjugate neutrosophic $N$-group over $(\langle G \cup I \rangle, *_1, ..., *_N)$.
2. Their extended intersection $(F, A) \cap_\varepsilon (K, D)$ is soft conjugate neutrosophic $N$-group over $(\langle G \cup I \rangle, *_1, ..., *_N)$.
3. Their restricted union $(F, A) \cup_R (K, D)$ is not soft conjugate neutrosophic $N$-group over $(\langle G \cup I \rangle, *_1, ..., *_N)$.
4. Their restricted intersection $(F, A) \cap_R (K, D)$ is soft conjugate neutrosophic $N$-group over $(\langle G \cup I \rangle, *_1, ..., *_N)$.

**Proposition 2.3.12:** Let $(F,A)$ and $(K,D)$ be two soft conjugate neutrosophic $N$-groups over $(\langle G\cup I\rangle,*_1,...,*_N)$. Then

1. Their $AND$ operation $(F,A)\wedge(K,D)$ is soft conjugate neutrosophic $N$-group over $(\langle G\cup I\rangle,*_1,...,*_N)$.
2. Their $OR$ operation $(F,A)\vee(K,D)$ is not soft conjugate neutrosophic $N$-group over $(\langle G\cup I\rangle,*_1,...,*_N)$.

## Soft Neutrosophic Strong N-Group

The notions of soft strong neutrosophic N-groups over neutrosophic N-groups are introduced here. We give some basic definitions of soft neutrosophic strong N-groups and illustrated it with the help of exmaples and give some basic results.

**Definition 2.3.11:** Let $(\langle G\cup I\rangle,*_1,...,*_N)$ be a neutrosophic $N$-group. Then $(F,A)$ over $(\langle G\cup I\rangle,*_1,...,*_N)$ is called soft neutrosophic strong $N$-group if and only if $F(a)$ is a neutrosophic strong sub $N$-group for all $a\in A.$.

**Example 2.3.9:** Let $(\langle G \cup I \rangle = \langle G_1 \cup I \rangle \cup \langle G_2 \cup I \rangle \cup \langle G_3 \cup I \rangle, *_1, *_2, *_3)$ be a neutrosophic 3-group, where $\langle G_1 \cup I \rangle = \langle Z_2 \cup I \rangle = \{0, 1, I, 1+I\}$, a neutrosophic group under multiplication modulo 2. $\langle G_2 \cup I \rangle = \{O, 1, 2, 3, 4, I, 2I, 3I, 4I\}$, neutrosophic group under multiplication modulo 5 and $\langle G_3 \cup I \rangle = \{0, 1, 2, I, 2I\}$, a neutrosophic group under multiplication modulo 3.

Let $P = \left\{ \left\{ \left\langle \dfrac{1}{2^n}, 2^n, \dfrac{1}{(2I)^n}, (2I)^n, I, 1 \right\rangle \right\}, \{1, 4, I, 4I\}, \{1, 2, I, 2I\} \right\}$, and

$X = \{Q \setminus \{0\}, \{1, 2, I, 2I\}, \{1, I\}\}$ are neutrosophic sub 3-groups.
Then $(F, A)$ is clearly soft neutrosophic strong 3-group over $(\langle G \cup I \rangle = \langle G_1 \cup I \rangle \cup \langle G_2 \cup I \rangle \cup \langle G_3 \cup I \rangle, *_1, *_2, *_3)$, where

$$F(a_1) = \left\{ \left\{ \left\langle \dfrac{1}{2^n}, 2^n, \dfrac{1}{(2I)^n}, (2I)^n, I, 1 \right\rangle \right\}, \{1, 4, I, 4I\}, \{1, I\} \right\},$$

$$F(a_1) = \{Q \setminus \{0\}, \{1, 2, I, 2I\}, \{1, I\}\}$$

**Theorem 2.3.10:** Every soft strong neutrosophic soft $N$-group $(F, A)$ is a soft neutrosophic $N$-group but the converse is not true.

**Theorem 2.3.11:** $(F, A)$ over $(\langle G \cup I \rangle, *_1, ..., *_N)$ is soft neutrosophic strong $N$-group if $(\langle G \cup I \rangle, *_1, ..., *_N)$ is a neutrosophic strong $N$-group.

**Proposition 2.3.13:** Let $(F,A)$ and $(K,D)$ be two soft neutrosophic strong $N$-groups over $(\langle G \cup I \rangle, *_1,...,*_N)$. Then

1. Their extended union $(F,A) \cup_\varepsilon (K,D)$ is not soft neutrosophic strong $N$-group over $(\langle G \cup I \rangle, *_1,...,*_N)$.
2. Their extended intersection $(F,A) \cap_\varepsilon (K,D)$ is not soft neutrosophic strong $N$-group over $(\langle G \cup I \rangle, *_1,...,*_N)$.
3. Their restricted union $(F,A) \cup_R (K,D)$ is not soft neutrosophic strong $N$-group over $(\langle G \cup I \rangle, *_1,...,*_N)$.
4. Their restricted intersection $(F,A) \cap_R (K,D)$ is not soft neutrosophic strong $N$-group over $(\langle G \cup I \rangle, *_1,...,*_N)$.

**Proof:** These are left as an exercise for the interested readers.

**Proposition 2.3.14:** Let $(F,A)$ and $(K,D)$ be two soft neutrosophic strong $N$-groups over $(\langle G \cup I \rangle, *_1,...,*_N)$. Then

1. Their $AND$ operation $(F,A) \wedge (K,D)$ is not soft neutrosophic strong $N$-group over $(\langle G \cup I \rangle, *_1,...,*_N)$.
2. Their $OR$ operation $(F,A) \vee (K,D)$ is not soft neutrosophic strong $N$-group over $(\langle G \cup I \rangle, *_1,...,*_N)$.

**Proof:** These are left as an exercise for the interested readers.

**Definition 2.3.12:** Let $(F,A)$ and $(H,K)$ be two soft neutrosophic strong $N$-groups over $(\langle G \cup I \rangle, *_1, ..., *_N)$. Then $(H,K)$ is called soft neutrosophic strong sub $N$-group of $(F,A)$ written as $(H,K) \prec (F,A)$, if

1. $K \subseteq A$,
2. $H(a)$ is soft neutrosophic strong sub $N$-group of $F(a)$ for all $a \in A$..

**Theorem 2.3.12:** If $(\langle G \cup I \rangle, *_1, ..., *_N)$ is a neutrosophic strong $N$-group. Then every soft neutrosophic sub $N$-group of $(F,A)$ is soft neutosophic strong sub $N$-group.

**Definition 2.3.13:** Let $(\langle G \cup I \rangle, *_1, ..., *_N)$ be a neutrosophic strong $N$-group. Then $(F,A)$ over $(\langle G \cup I \rangle, *_1, ..., *_N)$ is called soft Lagrange neutrosophic strong $N$-group if $F(a)$ is a Lagrange neutrosophic sub $N$-group of $(\langle G \cup I \rangle, *_1, ..., *_N)$ for all $a \in A$.

**Theorem 2.3.13:** Every soft Lagrange neutrosophic strong $N$-group $(F,A)$ over $(\langle G \cup I \rangle, *_1, ..., *_N)$ is a soft neutrosophic soft $N$-group but the converse is not true.

**Theorem 2.3.14:** Every soft Lagrange neutrosophic strong $N$-group $(F,A)$ over $(\langle G \cup I \rangle, *_1, ..., *_N)$ is a soft neutrosophic strong $N$-group but the converse is not tue.

**Theorem 2.3.15:** If $(\langle G \cup I \rangle, *_1,...,*_N)$ is a Lagrange neutrosophic strong $N$-group, then $(F,A)$ over $(\langle G \cup I \rangle, *_1,...,*_N)$ is also soft Lagrange neutrosophic strong $N$-group.

**Proposition 2.3.15:** Let $(F,A)$ and $(K,D)$ be two soft Lagrange neutrosophic strong $N$-groups over $(\langle G \cup I \rangle, *_1,...,*_N)$. Then

1. Their extended union $(F,A) \cup_\varepsilon (K,D)$ is not soft Lagrange neutrosophic strong $N$-group over $(\langle G \cup I \rangle, *_1,...,*_N)$.
2. Their extended intersection $(F,A) \cap_\varepsilon (K,D)$ is not soft Lagrange neutrosophic strong $N$-group over $(\langle G \cup I \rangle, *_1,...,*_N)$.
3. Their restricted union $(F,A) \cup_R (K,D)$ is not soft Lagrange neutrosophic strong $N$-group over $(\langle G \cup I \rangle, *_1,...,*_N)$.
4. Their restricted intersection $(F,A) \cap_R (K,D)$ is not soft Lagrange neutrosophic strong $N$-group over $(\langle G \cup I \rangle, *_1,...,*_N)$.

**Proposition 2.3.16:** Let $(F,A)$ and $(K,D)$ be two soft Lagrange neutrosophic strong $N$-groups over $(\langle G \cup I \rangle, *_1,...,*_N)$. Then

1. Their *AND* operation $(F,A) \wedge (K,D)$ is not soft Lagrange neutrosophic strong $N$-group over $(\langle G \cup I \rangle, *_1,...,*_N)$.
2. Their *OR* operation $(F,A) \vee (K,D)$ is not soft Lagrange neutrosophic strong $N$-group over $(\langle G \cup I \rangle, *_1,...,*_N)$.

**Definition 2.3.14:** Let $(\langle G \cup I \rangle, *_1, ..., *_N)$ be a neutrosophic strong $N$-group. Then $(F, A)$ over $(\langle G \cup I \rangle, *_1, ..., *_N)$ is called soft weakly Lagrange neutrosophic strong $N$-group if atleast one $F(a)$ is a Lagrange neutrosophic sub $N$-group of $(\langle G \cup I \rangle, *_1, ..., *_N)$ for some $a \in A$.

**Theorem 2.3.16:** Every soft weakly Lagrange neutrosophic strong $N$-group $(F, A)$ over $(\langle G \cup I \rangle, *_1, ..., *_N)$ is a soft neutrosophic soft $N$-group but the converse is not true.

**Theorem 2.3.17:** Every soft weakly Lagrange neutrosophic strong $N$-group $(F, A)$ over $(\langle G \cup I \rangle, *_1, ..., *_N)$ is a soft neutrosophic strong $N$-group but the converse is not true.

**Proposition 2.3.17:** Let $(F, A)$ and $(K, D)$ be two soft weakly Lagrange neutrosophic strong $N$-groups over $(\langle G \cup I \rangle, *_1, ..., *_N)$. Then

1. Their extended union $(F, A) \cup_\varepsilon (K, D)$ is not soft weakly Lagrange neutrosophic strong $N$-group over $(\langle G \cup I \rangle, *_1, ..., *_N)$.
2. Their extended intersection $(F, A) \cap_\varepsilon (K, D)$ is not soft weakly Lagrange neutrosophic strong $N$-group over $(\langle G \cup I \rangle, *_1, ..., *_N)$.
3. Their restricted union $(F, A) \cup_R (K, D)$ is not soft weakly Lagrange neutrosophic strong $N$-group over $(\langle G \cup I \rangle, *_1, ..., *_N)$.
4. Their restricted intersection $(F, A) \cap_R (K, D)$ is not soft weakly Lagrange neutrosophic strong $N$-group over $(\langle G \cup I \rangle, *_1, ..., *_N)$.

**Proposition 2.3.18:** Let $(F,A)$ and $(K,D)$ be two soft weakly Lagrange neutrosophic strong $N$-groups over $(\langle G\cup I\rangle,*_1,...,*_N)$. Then
1. Their *AND* operation $(F,A)\wedge(K,D)$ is not soft weakly Lagrange neutrosophic strong $N$-group over $(\langle G\cup I\rangle,*_1,...,*_N)$.
2. Their *OR* operation $(F,A)\vee(K,D)$ is not soft weakly Lagrange neutrosophic strong $N$-group over $(\langle G\cup I\rangle,*_1,...,*_N)$.

**Definition 2.3.15:** Let $(\langle G\cup I\rangle,*_1,...,*_N)$ be a strong neutrosophic $N$-group. Then $(F,A)$ over $(\langle G\cup I\rangle,*_1,...,*_N)$ is called soft Lagrange free neutrosophic strong $N$-group if $F(a)$ is not a Lagrange neutrosophic sub $N$-group of $(\langle G\cup I\rangle,*_1,...,*_N)$ for all $a\in A$.

**Theorem 2.3.18:** Every soft Lagrange free neutrosophic strong $N$-group $(F,A)$ over $(\langle G\cup I\rangle,*_1,...,*_N)$ is a soft neutrosophic $N$-group but the converse is not true.

**Theorem 2.3.19:** Every soft Lagrange free neutrosophic strong $N$-group $(F,A)$ over $(\langle G\cup I\rangle,*_1,...,*_N)$ is a soft neutrosophic strong $N$-group but the converse is not true.

**Theorem 2.3.20:** If $(\langle G\cup I\rangle,*_1,...,*_N)$ is a Lagrange free neutrosophic strong $N$-group, then $(F,A)$ over $(\langle G\cup I\rangle,*_1,...,*_N)$ is also soft Lagrange free neutrosophic strong $N$-group.

**Proposition 2.3.19:** Let $(F, A)$ and $(K, D)$ be two soft Lagrange free neutrosophic strong $N$-groups over $(\langle G \cup I \rangle, *_1, ..., *_N)$. Then

1. Their extended union $(F, A) \cup_\varepsilon (K, D)$ is not soft Lagrange free neutrosophic strong $N$-group over $(\langle G \cup I \rangle, *_1, ..., *_N)$.
2. Their extended intersection $(F, A) \cap_\varepsilon (K, D)$ is not soft Lagrange free neutrosophic strong $N$-group over $(\langle G \cup I \rangle, *_1, ..., *_N)$.
3. Their restricted union $(F, A) \cup_R (K, D)$ is not soft Lagrange free neutrosophic strong $N$-group over $(\langle G \cup I \rangle, *_1, ..., *_N)$.
4. Their restricted intersection $(F, A) \cap_R (K, D)$ is not soft Lagrange free neutrosophic strong $N$-group over $(\langle G \cup I \rangle, *_1, ..., *_N)$.

**Proposition 2.3.20:** Let $(F, A)$ and $(K, D)$ be two soft Lagrange free neutrosophic strong $N$-groups over $(\langle G \cup I \rangle, *_1, ..., *_N)$. Then

1. Their $AND$ operation $(F, A) \wedge (K, D)$ is not soft Lagrange free neutrosophic strong $N$-group over $(\langle G \cup I \rangle, *_1, ..., *_N)$.
2. Their $OR$ operation $(F, A) \vee (K, D)$ is not soft Lagrange free neutrosophic strong $N$-group over $(\langle G \cup I \rangle, *_1, ..., *_N)$.

**Definition 2.3.16:** Let $N$ be a strong neutrosophic $N$-group. Then $(F, A)$ over $(\langle G \cup I \rangle, *_1, ..., *_N)$ is called sofyt normal neutrosophic strong $N$-group if $F(a)$ is normal neutrosophic sub $N$-group of $(\langle G \cup I \rangle, *_1, ..., *_N)$ for all $a \in A$.

**Theorem 2.3.21:** Every soft normal strong neutrosophic $N$-group $(F,A)$ over $(\langle G \cup I \rangle, *_1,...,*_N)$ is a soft neutrosophic $N$-group but the converse is not true.

**Proof:** The proof is left as an exercise for the interested readers.

**Theorem 2.3.22:** Every soft normal strong neutrosophic $N$-group $(F,A)$ over $(\langle G \cup I \rangle, *_1,...,*_N)$ is a soft strong neutrosophic $N$-group but the converse is not true.

One can easily see the converse by the help of example.

**Proposition 2.3.21:** Let $(F,A)$ and $(K,D)$ be two soft normal neutrosophic strong $N$-groups over $(\langle G \cup I \rangle, *_1,...,*_N)$. Then

1. Their extended union $(F,A) \cup_\varepsilon (K,D)$ is not soft normal neutrosophic strong $N$-group over $(\langle G \cup I \rangle, *_1,...,*_N)$.
2. Their extended intersection $(F,A) \cap_\varepsilon (K,D)$ is soft normal neutrosophic strong $N$-group over $(\langle G \cup I \rangle, *_1,...,*_N)$.
3. Their restricted union $(F,A) \cup_R (K,D)$ is not soft normal neutrosophic strong $N$-group over $(\langle G \cup I \rangle, *_1,...,*_N)$.
4. Their restricted intersection $(F,A) \cap_R (K,D)$ is soft normal neutrosophic strong $N$-group over $(\langle G \cup I \rangle, *_1,...,*_N)$.

**Proof:** These are straightforward.

**Proposition 2.3.22:** Let $(F,A)$ and $(K,D)$ be two soft normal neutrosophic strong $N$-groups over $(\langle G\cup I\rangle, *_1,...,*_N)$. Then

1. Their *AND* operation $(F,A)\wedge(K,D)$ is soft normal neutrosophic strong $N$-group over $(\langle G\cup I\rangle, *_1,...,*_N)$.
2. Their *OR* operation $(F,A)\vee(K,D)$ is not soft normal neutrosophic strong $N$-group over $(\langle G\cup I\rangle, *_1,...,*_N)$.

**Definition 2.3.17:** Let $(\langle G\cup I\rangle, *_1,...,*_N)$ be a strong neutrosophic $N$-group. Then $(F,A)$ over $(\langle G\cup I\rangle, *_1,...,*_N)$ is called soft conjugate neutrosophic strong $N$-group if $F(a)$ is a conjugate neutrosophic sub $N$-group of $(\langle G\cup I\rangle, *_1,...,*_N)$ for all $a\in A$.

**Theorem 2.3.23:** Every soft conjugate neutrosophic *strong* $N$-group $(F,A)$ over $(\langle G\cup I\rangle, *_1,...,*_N)$ is a soft neutrosophic $N$-group.

**Proof:** The proof is left as an exercise for the interested readers.

**Theorem 2.3.24:** Every soft conjugate neutrosophic strong $N$-group $(F,A)$ over $(\langle G\cup I\rangle, *_1,...,*_N)$ is a soft neutrosophic strong $N$-group but the converse is not true.

One can see the converse by the help of example.

**Proposition 2.3.23:** Let $(F, A)$ and $(K, D)$ be two soft conjugate neutrosophic strong $N$-groups over $(\langle G \cup I \rangle, *_1, ...., *_N)$. Then

1. Their extended union $(F, A) \cup_\varepsilon (K, D)$ is not soft conjugate neutrosophic strong $N$-group over $(\langle G \cup I \rangle, *_1, ...., *_N)$.
2. Their extended intersection $(F, A) \cap_\varepsilon (K, D)$ is soft conjugate neutrosophic strong $N$-group over $(\langle G \cup I \rangle, *_1, ...., *_N)$.
3. Their restricted union $(F, A) \cup_R (K, D)$ is not soft conjugate neutrosophic strong $N$-group over $(\langle G \cup I \rangle, *_1, ...., *_N)$.
4. Their restricted intersection $(F, A) \cap_R (K, D)$ is soft conjugate neutrosophic strong $N$-group over $(\langle G \cup I \rangle, *_1, ...., *_N)$.

**Proposition 2.3.24:** Let $(F, A)$ and $(K, D)$ be two soft conjugate neutrosophic strong $N$-groups over $(\langle G \cup I \rangle, *_1, ...., *_N)$. Then

1. Their *AND* operation $(F, A) \wedge (K, D)$ is soft conjugate neutrosophic strong $N$-group over $(\langle G \cup I \rangle, *_1, ...., *_N)$.
2. Their *OR* operation $(F, A) \vee (K, D)$ is not soft conjugate neutrosophic strong $N$-group over $(\langle G \cup I \rangle, *_1, ...., *_N)$.

# Chapter Three

**SOFT NEUTROSOPHIC SEMIGROUPS AND THEIR GENERALIZATION**

In this chapter the notion of soft neutrosophic semigroup with its generalization are given. This chapter has three sections. In first section, soft neutrosophic semigroup is introduced over a neutrosophic semigroup. Actually soft neutrosophic semigroups are parameterized collection of neutrosophic subsemigroups. We also investigate soft neutrosophic strong semigroup over a neutrosophic semigroup which is purely of neutrosophic behaviour. Now we proceed onto define soft neutrosophic semigroup over a neutrosophic semigroup as follows:

## 3.1 Soft Neutrosophic Semigroup

In this section, the authors define a soft neutrosophic semigroup over a neutrosophic semigroup and established some of the fundamental properties of it. Soft neutrosophic monoids are also introduce over neutrosophic monoids in the present section of this book. We also introduce soft neutrosophic strong semigroup over a neutrosophic semigroup.

**Definition 3.1.1:** Let $N(S)$ be a neutrosophic semigroup and $(F,A)$ be a soft set over $N(S)$. Then $(F,A)$ is called soft neutrosophic semigroup if and only if $F(a)$ is neutrosophic subsemigroup of $N(S)$, for all $a \in A$. Equivalently $(F,A)$ is a soft neutrosophic semigroup over $N(S)$ if $(F,A) \mathbin{\hat{\circ}} (F,A) \subseteq (F,A)$.

**Example 3.1.1:** Let $N(S) = \langle Z^+ \cup \{0\} \cup \{I\} \rangle$ be a neutrosophic semigroup under `+'. Let $A = \{a_1, a_2\}$ be a set fo parameters. Then clearly $(F, A)$ is a soft neutrosophic semigroup over $N(S)$, where

$$F(a_1) = \langle 2Z^+ \cup I \rangle, F(a_2) = \langle 3Z^+ \cup I \rangle.$$

**Theorem 3.1.1:** A soft neutrosophic semigroup over $N(S)$ always contain a soft semigroup over $S$.

**Proof:** The proof of this theorem is straightforward.

**Theorem 3.1.2:** Let $(F, A)$ and $(H, A)$ be two soft neutrosophic semigroups over $N(S)$. Then their intersection $(F, A) \cap (H, A)$ is again soft neutrosophic semigroup over $N(S)$.

**Proof:** The proof is staightforward.

**Theorem 3.1.3:** Let $(F, A)$ and $(H, B)$ be two soft neutrosophic semigroups over $N(S)$. If $A \cap B = \phi$, then $(F, A) \cup (H, B)$ is a soft neutrosophic semigroup over $N(S)$.

**Remark 3.1.1:** The extended union of two soft neutrosophic semigroups $(F, A)$ and $(K, B)$ over $N(S)$ is not a soft neutrosophic semigroup over $N(S)$.

We take the following example for the checking of above remark.

**Example 3.1.2:** Let $N(S) = \langle Z^+ \cup I \rangle$ be the neutrosophic semigroup under `+`. Let $A = \{a_1, a_2\}$ be a set of parameters and let $(F, A)$ be a soft neutrosophic semigroup over $N(S)$, where

$$F(a_1) = \langle 2Z^+ \cup I \rangle, F(a_2) = \langle 3Z^+ \cup I \rangle.$$

Let $(K, B)$ be another soft neutrosophic semigroup over $N(S)$, where

$$K(a_1) = \langle 5Z^+ \cup I \rangle, K(a_3) = \langle 4Z^+ \cup I \rangle.$$

Let $C = A \cap B = \{a_1\}$. The extended union $(F, A) \cup_E (K, B) = (H, C)$ where $H(a_2) = F(a_2) = \langle 3Z^+ \cup I \rangle$, $H(a_3) = K(a_3) = \langle 4Z^+ \cup I \rangle$ and for $a_1 \in C$, we have $H(a_1) = F(a_1) \cup K(a_1) = \langle 2Z^+ \cup I \rangle \cup \langle 5Z^+ \cup I \rangle$ which is not a neutrosophic subsemigroup as union of two neutrosophic subsemigroup is not neutrosophic subsemigroup.

**Proposition 3.1.1:** The extended intersection of two soft neutrosophic semigroups over $N(S)$ is a soft neutrosophic semigruop over $N(S)$.

**Remark 3.1.2:** The restricted union of two soft neutrosophic semigroups $(F, A)$ and $(K, B)$ over $N(S)$ is not a soft neutrosophic semigroup over $N(S)$.
One can easily check it in above Example (6).

**Proposition 3.1.2:** The restricted intersection of two soft neutrosophic semigroups over $N(S)$ is a soft neutrosophic semigroup over $N(S)$.

**Proposition 3.1.3:** The *AND* operation of two soft neutrosophic semigroups over $N(S)$ is a soft neutrosophic semigroup over $N(S)$.

**Remark 3.1.3:** The *OR* operation of two soft neutosophic semigroups over $N(S)$ may not be a soft nuetrosophic semigroup over $N(S)$.

**Definition 3.1.2:** Let $N(S)$ be a neutrosophic monoid and $(F,A)$ be a soft set over $N(S)$. Then $(F,A)$ is called soft neutrosophic monoid if and only if $F(a)$ is neutrosophic submonoid of $N(S)$, for all $a \in A$.

**Example 3.1.3:** Let $N(S) = \langle Z \cup I \rangle$ be a neutrosophic monoid and let $A = \{a_1, a_2\}$ be a set of parameters. Then $(F,A)$ is a soft neutrosophic monoid over $N(S)$, where

$$F(a_1) = \langle 2Z \cup I \rangle, F(a_2) = \langle 4Z \cup I \rangle.$$

**Theorem 3.1.4:** Every soft neutrosophic monoid over $N(S)$ is a soft neutrosophic semigroup over $N(S)$ but the converse is not true in general.

One can easily check the converse of the above theorem (4) with the help of Example.

**Proposition 3.1.4:** Let $(F,A)$ and $(K,B)$ be two soft neutrosophic monoids over $N(S)$. Then

1. Their extended intersection $(F,A) \cap_E (K,B)$ is a soft neutrosophic monoid over $N(S)$.
2. Their restricted intersection $(F,A) \cap_R (K,B)$ is a soft neutrosophic monoid over $N(S)$.
3. Their *AND* operation $(F,A) \wedge (K,B)$ is a soft neutrosophic monoid over $N(S)$.

**Remark 3.1.4:** Let $(F,A)$ and $(H,B)$ be two soft neutrosophic monoid over $N(S)$. Then

1. Their extended union $(F,A) \cup_E (K,B)$ is not a soft neutrosophic monoid over $N(S)$.
2. Their restricted union $(F,A) \cup_R (K,B)$ is not a soft neutrosophic monoid over $N(S)$.
3. Their *OR* operation $(F,A) \vee (K,B)$ is not a soft neutrosophic monoid over $N(S)$.

One can easily verify $(1), (2),$ and $(3)$ by the help of examples.

**Definition 3.1.3:** Let $(F,A)$ be a soft neutrosophic semigroup over $N(S)$. Then $(F,A)$ is called an absolute-soft neutrosophic semigroup over $N(S)$ if $F(a) = N(S)$, for all $a \in A$. We denote it by $\mathbb{F}_{N(S)}$.

**Theorem 3.1.5:** Every absolute-soft neutrosophic semigroup over $N(S)$ always contain absolute soft semigroup over $S$.

**Definition 3.1.4:** Let $(F,A)$ and $(H,B)$ be two soft neutrosophic semigroups over $N(S)$. Then $(H,B)$ is a soft neutrosophic subsemigroup of $(F,A)$, if

1. $B \subseteq A$, and
2. $H(a)$ is neutrosophic subsemigroup of $F(a)$, for all $a \in B$.

**Example 3.1.4:** Let $N(S) = \langle Z \cup I \rangle$ be a neutrosophic semigroup under `+'. $A = \{a_1, a_2, a_3\}$. Then $(F,A)$ is a soft neutrosophic semigroup over $N(S)$, where

$$F(a_1) = \langle 2Z \cup I \rangle, F(a_2) = \langle 3Z \cup I \rangle,$$
$$F(a_3) = \langle 5Z \cup I \rangle.$$

Let $B = \{a_1, a_2\} \subset A$. Then $(H,B)$ is soft neutrosophic subsemigroup of $(F,A)$ over $N(S)$, where

$$H(a_1) = \langle 4Z \cup I \rangle, H(a_2) = \langle 6Z \cup I \rangle.$$

**Theorem 3.1.6:** A soft neutrosophic semigroup over $N(S)$ have soft neutrosophic subsemigroups as well as soft subsemigroups over $N(S)$.

**Proof.** Obvious.

**Theorem 3.1.7:** Every soft semigroup over $S$ is always soft neutrosophic subsemigroup of soft neutrosophic semigroup over $N(S)$.

**Proof.** The proof is obvious.

**Proposition 3.1.5:** Let $(F, A)$ be a soft neutrosophic semigroup over $N(S)$ and $\{(H_i, B_i) : i \in J\}$ is a non empty family of soft neutrosophic subsemigroups of $(F, A)$. Then

1. $\bigcap_{i \in J}(H_i, B_i)$ is a soft neutrosophic subsemigroup of $(F, A)$.
2. $\bigwedge_{i \in J}(H_i, B_i)$ is a soft neutrosophic subsemigroup of $\bigwedge_{i \in J}(F, A)$.
3. $\bigcup_{i \in J}(H_i, B_i)$ is a soft neutrosophic subsemigroup of $(F, A)$ if $B_i \cap B_j = \phi$, for all $i \neq j$.

**Proof.** Straightforward.

**Definition 3.1.5:** A soft set $(F, A)$ over $N(S)$ is called soft neutrosophic left (right) ideal over $N(S)$ if $\mathbb{F}_{N(S)} \hat{\circ} (F, A) \subseteq (F, A)$, where $\mathbb{F}_{N(S)}$ is an absolute-soft neutrosophic semigroup over $N(S)$.
 A soft set over $N(S)$ is a soft neutrosophic ideal if it is both a soft neutrosophic left and a soft neutrosophic right ideal over $N(S)$.

**Example 3.1.5:** Let $N(S) = \langle Z \cup I \rangle$ be the neutrosophic semigroup under `+'. Let $A = \{a_1, a_2\}$ be a set of parameters. Then $(F, A)$ be a soft neutrosophic ideal over $N(S)$, where

$$F(a_1) = \langle 2Z \cup I \rangle, F(a_2) = \langle 4Z \cup I \rangle.$$

**Proposition 3.1.6:** A soft set $(F, A)$ over $N(S)$ is a soft neutrosophic ideal if and only if $F(a)$ is a neutrosophic ideal of $N(S)$, for all $a \in A$.

**Theorem 3.1.8:** Every soft neutrosophic ideal $(F,A)$ over $N(S)$ is a soft neutrosophic semigroup.

**Proposition 3.1.7:** Let $(F,A)$ and $(K,B)$ be two soft neutrosophic ideals over $N(S)$. Then

1. Their extended union $(F,A) \cup_E (K,B)$ over $N(S)$ is soft neutrosophic ideal over $N(S)$.
2. Their extended intersection $(F,A) \cap_E (K,B)$ over $N(S)$ is soft neutrosophic ideal over $N(S)$.
3. Their restricted union $(F,A) \cup_R (K,B)$ over $N(S)$ is soft neutrosophic ideal over $N(S)$.
4. Their restricted intersection $(F,A) \cap_R (K,B)$ over $N(S)$ is soft neutrosophic ideal over $N(S)$.

**Proof.** One can easily prove $(1),(2),(3)$ and $(4)$.

**Proposition 3.1.8:** Let $(F,A)$ and $(H,B)$ be two soft neutrosophic ideal over $N(S)$. Then

1. Their $AND$ operation $(F,A) \wedge (K,B)$ is a soft neutrosophic ideal over $N(S)$.
2. Their $OR$ operation $(F,A) \vee (K,B)$ is a soft neutrosophic ideal over $N(S)$.

**Theorem 3.1.9:** Let $(F,A)$ and $(K,B)$ be two soft neutrosophic semigroups (ideals) over $N(S)$ and $N(T)$ respectively. Then $(F,A) \times (K,B)$ is also a soft neutrosophic semigroup (ideal) over $N(S) \times N(T)$.

**Proof.** The proof is straight forward.

**Theorem 3.1.10:** Let $(F,A)$ be a soft neutrosophic semigroup over $N(S)$ and $\{(H_i,B_i) : i \in J\}$ is a non empty family of soft neutrosophic ideals of $(F,A)$. Then

1. $\bigcap_{i \in J}(H_i,B_i)$ is a soft neutrosophic ideal of $(F,A)$.
2. $\bigwedge_{i \in J}(H_i,B_i)$ is a soft neutrosophic ideal of $\bigwedge_{i \in J}(F,A)$.
3. $\bigcup_{i \in J}(H_i,B_i)$ is a soft neutrosophic ideal of $(F,A)$.
4. $\bigvee_{i \in J}(H_i,B_i)$ is a soft neutrosophic ideal of $\bigvee_{i \in J}(F,A)$.

**Proof.** We can easily prove $(1),(2),(3),$ and $(4)$.

**Definition 3.1.6:** A soft set $(F,A)$ over $N(S)$ is called soft neutrosophic principal ideal or soft neutrosophic cyclic ideal if and only if $F(a)$ is a principal or cyclic neutrosophic ideal of $N(S)$, for all $a \in A$.

**Proposition 3.1.9:** Let $(F,A)$ and $(K,B)$ be two soft neutrosophic principal ideals over $N(S)$. Then

1. Their extended intersection $(F,A) \cap_E (K,B)$ over $N(S)$ is soft neutrosophic principal ideal over $N(S)$.
2. Their restricted intersection $(F,A) \cap_R (K,B)$ over $N(S)$ is soft neutrosophic principal ideal over $N(S)$.
3. Their $AND$ operation $(F,A) \wedge (K,B)$ is soft neutrosophic principal ideal over $N(S)$.

**Remark 3.1.5:** Let $(F,A)$ and $(H,B)$ be two soft neutrosophic principal ideals over $N(S)$. Then

1. Their restricted union $(F,A) \cup_R (K,B)$ over $N(S)$ is not soft neutrosophic principal ideal over $N(S)$.
2. Their extended union $(F,A) \cup_E (K,B)$ over $N(S)$ is not soft neutrosophic principal ideal over $N(S)$.
3. Their $OR$ operation $(F,A) \vee (K,B)$ is not soft neutrosophic principal ideal over $N(S)$.

One can easily prove it by the help of examples.

## Soft Neutrosophic Strong Semigroup

The notion of soft neutrosophic strong semigroup over a neutrosophic semigroup is introduced here. We give the definition of soft neutrosophic strong semigroup and investigate some related properties with sufficient amount of illustrative examples.

**Definition 3.1.7:** Let $N(S)$ be a neutrosophic semigroup and let $(F,A)$ be a soft set over $N(S)$. Then $(F,A)$ is said to be soft neutrosophic strong semigroup over $N(S)$ if and only if $F(a)$ is neutrosophic strong subsemigroup of $N(S)$ for all $a \in A$.

**Example 3.1.6:** Let $N(S) = \{0,1,2,I,2I,\times,\}$ be a neutosophic semigroup. Let $A = \{a_1, a_2\}$ be a set of parameters. Then $(F, A)$ is clearly soft neutrosophic strong semigroup over $N(S)$, where

$$F(a_1) = \{0, I, 2I\}, F(a_2) = \{0, I\}$$

**Proposition 13.1.10:** Let $(F, A)$ and $(K, B)$ be two soft neutrosophic stron semigroups over $N(S)$. Then

1. Their extended intersection $(F, A) \cap_E (K, B)$ is soft neutrosophic strong semigroup over $N(S)$.
2. Their restricted intersection $(F, A) \cap_R (K, B)$ is soft neutrosophic strong semigroup over $N(S)$.
3. Their $AND$ operation $(F, A) \wedge (K, B)$ is soft neutrosophic strong semigroup over $N(S)$.

**Remark 3.1.6:** Let $(F, A)$ and $(K, B)$ be two soft neutrosophic strong semigroups over $N(S)$. Then

1. Their extended union $(F, A) \cup_E (K, B)$ is not soft neutrosophic strong semigroup over $N(S)$.
2. Their restricted union $(F, A) \cup_R (K, B)$ is not soft neutrosophic strong semigroup over $N(S)$.
3. Their $OR$ operation $(F, A) \vee (K, B)$ is not soft neutrosophic strong semigroup over $N(S)$.

One can easily verified (1), (2), and (3) with the help of examples.

**Definition 3.1.8:** Let $(F,A)$ and $(H,B)$ be two soft neutrosophic strong semigroups over $BN(S)$. Then $(H,B)$ is a soft neutrosophic strong subsemigroup of $(F,A)$, if

1. $B \subseteq A$, and
2. $H(a)$ is a neutrosophic strong subsemigroup of $F(a)$ for all $a \in A$.

**Example 3.1.7:** Let $N(S) = \{0,1,2,I,2I\}$ be a neutosophic strong semigroup under multiplication modulo $3$. Let $A = \{a_1, a_2, a_3\}$ be a set of parameters. Then $(F,A)$ is clearly soft neutrosophic strong semigroup over $N(S)$, where

$$F(a_1) = \{0, I, 2I\}, F(a_2) = \{0, I\},$$

$$F(a_3) = \{0, I, 2I\}.$$

Then $(H,B)$ is a soft neutrosophic subbisemigroup of $(F,A)$, where

$$H(a_1) = \{0, I\}, H(a_3) = \{0, I\}.$$

We now give some characterization of soft neutrosophic strong groups.

**Proposition 3.1.11:** Let $(F,A)$ be a soft neutrosophic strong semigroup over $BN(S)$ and $\{(H_i, B_i) : i \in J\}$ is a non-empty family of soft neutrosophic strong subsemigroups of $(F,A)$. Then

1. $\bigcap_{i \in J}(H_i, B_i)$ is a soft neutrosophic strong subsemigroup of $(F,A)$.
2. $\bigwedge_{i \in J}(H_i, B_i)$ is a soft neutrosophic strong subsemigroup of $\bigwedge_{i \in J}(F,A)$.
3. $\bigcup_{i \in J}(H_i, B_i)$ is a soft neutrosophic strong subsemigroup of $(F,A)$ if $B_i \cap B_j = \phi$, for all $i \neq j$.

**Proof:** Straightforward.

**Definition 3.1.9:** $(F,A)$ is called soft neutrosophic strong ideal over $N(S)$ if $F(a)$ is a neutrosophic strong ideal of $N(S)$, for all $a \in A$.

**Example 3.1.8:** Let $N(S) = \{0,1,2,I,2I\}$ be a neutosophic strong semigroup under multiplication modulo $3$. Let $A = \{a_1, a_2\}$ be a set of parameters. Then $(F,A)$ is clearly soft neutrosophic strong ideal over $N(S)$, where

$$F(a_1) = \{0, I, 2I\}, F(a_2) = \{0, I\},$$

**Theorem 3.1.11:** Every soft neutrosophic strong ideal $(F,A)$ over $N(S)$ is trivially a soft neutrosophic strong semigroup.

**Proof.** Straightforward.

**Theorem 3.1.12:** Every soft neutrosophic strong ideal $(F,A)$ over $N(S)$ is trivially a soft neutrosophic strong ideal.

**Proposition 3.1.12:** Let $(F,A)$ and $(K,B)$ be two soft neutrosophic strong ideals over $N(S)$. Then

1) Their extended intersection $(F,A) \cap_E (K,B)$ is soft neutrosophic strong ideal over $N(S)$.
2) Their restricted intersection $(F,A) \cap_R (K,B)$ is soft neutrosophic strong ideal over $N(S)$.
3) Their $AND$ operation $(F,A) \wedge (K,B)$ is soft neutrosophic strong ideal over $N(S)$.

**Remark 3.1.7:** Let $(F,A)$ and $(K,B)$ be two soft neutrosophic strong ideal over $N(S)$. Then

1) Their extended union $(F,A) \cup_E (K,B)$ is not soft neutrosophic strong ideal over $N(S)$.
2) Their restricted union $(F,A) \cup_R (K,B)$ is not soft neutrosophic strong ideal over $N(S)$.
3) Their $OR$ operation $(F,A) \vee (K,B)$ is not soft neutrosophic strong ideal over $N(S)$.

One can easily proved (1), (2), and (3) by the help of examples.

## 3.2 Soft Neutrosophic Bisemigroup

In this section, the definition of soft neutrosophic bisemigroup is given over a neutrosophic bisemigroup. Then soft neutrosophic biideals are defined over a neutrosophic bisemigroup. Some of their fundamental properties are also given in this section.

We now move on to define soft neutrosophic bisemigroups.

**Definition 3.2.1:** Let $(F,A)$ be a soft set over a neutrosophic bisemigroup $BN(S)$. Then $(F,A)$ is said to be soft neutrosophic bisemigroup over $BN(S)$ if and only if $F(a)$ is a neutrosophic subbisemigroup of $BN(S)$ for all $a \in A$.

**Example 3.2.1:** Let $BN(S) = \{\langle Z \cup I \rangle \cup \langle Z_3 \cup I \rangle\}$, where $\langle Z \cup I \rangle$ is a neutrosophic semigroup with respect to $+$ and $\langle Z_3 \cup I \rangle$ is a neutrosophic semigroup under multiplication modulo $3$. $BN(S)$. Let $A = \{a_1, a_2\}$ be a set of parameters. Then $(F,A)$ is soft neutrosophic bisemigroup over $BN(S)$, where

$$F(a_1) = \{\langle 2Z \cup I \rangle, 0, 1, I\}$$
$$F(a_2) = \{\langle 8Z \cup I \rangle, 0, 2, 2I\}.$$

**Proposition 3.2.1:** Let $(F,A)$ and $(K,B)$ be two soft neutrosophic bisemigroup over $BN(S)$. Then

1. Their extended intersection $(F,A) \cap_E (K,B)$ is soft neutrosophic bisemigroup over $BN(S)$.
2. Their restricted intersection $(F,A) \cap_R (K,B)$ is soft neutrosophic bisemigroup over $BN(S)$.
3. Their *AND* operation $(F,A) \wedge (K,B)$ is soft neutrosophic bisemigroup over $BN(S)$.

**Remark 3.2.1:** Let $(F,A)$ and $(K,B)$ be two soft neutrosophic bisemigroups over $BN(S)$. Then

1. Their extended union $(F,A) \cup_E (K,B)$ is not soft neutrosophic bisemigroup over $BN(S)$.
2. Their restricted union $(F,A) \cup_R (K,B)$ is not soft neutrosophic bisemigroup over $BN(S)$.
3. Their *OR* operation $(F,A) \vee (K,B)$ is not soft neutrosophic bisemigroup over $BN(S)$.

One can easily proved (1), (2), and (3) by the help of examples.

**Definition 3.2.2:** Let $(F,A)$ and $(H,B)$ be two soft neutrosophic bisemigroups over $BN(S)$. Then $(H,B)$ is a soft neutrosophic subbisemigroup of $(F,A)$, if

1. $B \subseteq A$, and
2. $H(a)$ is a neutrosophic subbisemigroup of $F(a)$ for all $a \in A$.

This situation can be explained in the following example.

**Example 3.2.2:** Let $BN(S) = \{\langle Z \cup I \rangle \cup \langle Z_3 \cup I \rangle\}$, where $\langle Z \cup I \rangle$ is a neutrosophic semigroup with respect to $+$ and $\langle Z_3 \cup I \rangle$ is a neutrosophic semigroup under multiplication modulo 3. $BN(S)$. Let $A = \{a_1, a_2\}$ be a set of parameters. Then $(F, A)$ is soft neutrosophic bisemigroup over $BN(S)$, where

$$F(a_1) = \{\langle 2Z \cup I \rangle, 0, I\}$$
$$F(a_2) = \{\langle 8Z \cup I \rangle, 0, 1, I\}$$

Then $(H, B)$ is a soft neutrosophic subbisemigroup of $(F, A)$, where

$$H(a_2) = \{\langle 4Z \cup I \rangle, 0, I\}.$$

**Proposition 3.2.2:** Let $(F, A)$ be a soft neutrosophic bisemigroup over $BN(S)$ and $\{(H_i, B_i) : i \in J\}$ is a non-empty family of soft neutrosophic subbisemigroups of $(F, A)$. Then

1. $\cap_{i \in J}(H_i, B_i)$ is a soft neutrosophic subbisemigroup of $(F, A)$.
2. $\wedge_{i \in J}(H_i, B_i)$ is a soft neutrosophic subbisemigroup of $\wedge_{i \in J}(F, A)$.
3. $\cup_{i \in J}(H_i, B_i)$ is a soft neutrosophic subbisemigroup of $(F, A)$ if $B_i \cap B_j = \phi$, for all $i \neq j$.

**Proof.** Straightforward.

**Definition 3.2.3:** $(F, A)$ is called soft neutrosophic biideal over $BN(S)$ if $F(a)$ is a neutrosophic biideal of $BN(S)$, for all $a \in A$.

**Example 3.2.3:** Let $BN(S) = \{\langle Z \cup I \rangle \cup \langle Z_3 \cup I \rangle\}$, where $\langle Z \cup I \rangle$ is a neutrosophic semigroup with respect to $+$ and $\langle Z_3 \cup I \rangle$ is a neutrosophic semigroup under multiplication modulo $3$. $BN(S)$. Let $A = \{a_1, a_2\}$ be a set of parameters. Then $(F, A)$ is soft neutrosophic biideal over $BN(S)$, where

$$F(a_1) = \{\langle 2Z \cup I \rangle, 0, I\}$$
$$F(a_2) = \{\langle 8Z \cup I \rangle, 0, 1, I\}$$

**Theorem 3.2.1:** Every soft neutrosophic biideal $(F, A)$ over $BN(S)$ is trivially a soft neutrosophic bisemigroup.

**Proof:** This is obvious.

**Proposition 3.2.3:** Let $(F, A)$ and $(K, B)$ be two soft neutrosophic biideals over $BN(S)$. Then

1. Their extended intersection $(F, A) \cap_E (K, B)$ is soft neutrosophic biideal over $BN(S)$.
2. Their restricted intersection $(F, A) \cap_R (K, B)$ is soft neutrosophic biideal over $BN(S)$.
3. Their $AND$ operation $(F, A) \wedge (K, B)$ is soft neutrosophic biideal over $BN(S)$.

**Remark 3.2.2:** Let $(F,A)$ and $(K,B)$ be two soft neutrosophic biideals over $BN(S)$. Then

1. Their extended union $(F,A) \cup_E (K,B)$ is not soft neutrosophic biideals over $BN(S)$.
2. Their restricted union $(F,A) \cup_R (K,B)$ is not soft neutrosophic biidleals over $BN(S)$.
3. Their $OR$ operation $(F,A) \vee (K,B)$ is not soft neutrosophic biideals over $BN(S)$.

One can easily proved $(1),(2),$ and $(3)$ by the help of examples

**Theorem 3.2.2:** Let $(F,A)$ be a soft neutrosophic biideal over $BN(S)$ and $\{(H_i, B_i) : i \in J\}$ is a non-empty family of soft neutrosophic biideals of $(F,A)$. Then

1. $\bigcap_{i \in J}(H_i, B_i)$ is a soft neutrosophic biideal of $(F,A)$.
2. $\wedge_{i \in J}(H_i, B_i)$ is a soft neutrosophic biideal of $\wedge_{i \in J}(F,A)$.

## Soft Neutrosophic Strong Bisemigroup

As usual, the theory of purely neutrosophics is alos exist in soft neutrosophic bisemigroup and so we define soft neutrosophic strong bisemigroup over a neutrosophic semigroup here and establish their related properties and characteristics.

**Definition 3.2.4:** Let $(F,A)$ be a soft set over a neutrosophic bisemigroup $BN(S)$. Then $(F,A)$ is said to be soft neutrosophic strong bisemigroup over $BN(S)$ if and only if $F(a)$ is a neutrosophic strong subbisemigroup of $BN(S)$ for all $a \in A$.

**Example 3.2.4:** Let $BN(S) = \{\langle Z \cup I \rangle \cup \langle Z_3 \cup I \rangle\}$, where $\langle Z \cup I \rangle$ is a neutrosophic semigroup with respect to $+$ and $\langle Z_3 \cup I \rangle$ is a neutrosophic semigroup under multiplication modulo $3$. $BN(S)$. Let $A = \{a_1, a_2\}$ be a set of parameters. Then $(F,A)$ is soft neutrosophic strong bisemigroup over $BN(S)$, where

$$F(a_1) = \{\langle 2Z \cup I \rangle, 0, I\}$$
$$F(a_2) = \{\langle 8Z \cup I \rangle, 0, 1, I\}$$

**Theorem 3.2.3:** Every soft neutrosophic strong bisemigroup is a soft neutrosophic bisemigroup.

**Proposition 3.2.4:** Let $(F,A)$ and $(K,B)$ be two soft neutrosophic strong bisemigroups over $BN(S)$. Then

1. Their extended intersection $(F,A) \cap_E (K,B)$ is soft neutrosophic strong bisemigroup over $BN(S)$.
2. Their restricted intersection $(F,A) \cap_R (K,B)$ is soft neutrosophic strong bisemigroup over $BN(S)$.
3. Their $AND$ operation $(F,A) \wedge (K,B)$ is soft neutrosophic strong bisemigroup over $BN(S)$.

**Remark 3.2.3:** Let $(F, A)$ and $(K, B)$ be two soft neutrosophic strong bisemigroups over $BN(S)$. Then

1. Their extended union $(F, A) \cup_E (K, B)$ is not soft neutrosophic strong bisemigroup over $BN(S)$.
2. Their restricted union $(F, A) \cup_R (K, B)$ is not soft neutrosophic strong bisemigroup over $BN(S)$.
3. Their $OR$ operation $(F, A) \vee (K, B)$ is not soft neutrosophic strong bisemigroup over $BN(S)$.

One can easily proved (1),(2), and (3) by the help of examples.

**Definition 3.2.5:** Let $(F, A)$ and $(H, B)$ be two soft neutrosophic strong bisemigroups over $BN(S)$. Then $(H, B)$ is a soft neutrosophic strong subbisemigroup of $(F, A)$, if

1. $B \subseteq A$, and
2. $H(a)$ is a neutrosophic strong subbisemigroup of $F(a)$ for all $a \in A$.

**Example 3.2.5:** Let $BN(S) = \{\langle Z \cup I \rangle \cup \langle Z_3 \cup I \rangle\}$, where $\langle Z \cup I \rangle$ is a neutrosophic semigroup with respect to $+$ and $\langle Z_3 \cup I \rangle$ is a neutrosophic semigroup under multiplication modulo $3$. $BN(S)$. Let $A = \{a_1, a_2\}$ be a set of parameters. Then $(F, A)$ is soft neutrosophic strong bisemigroup over $BN(S)$, where

$$F(a_1) = \{\langle 2Z \cup I \rangle, 0, I\}$$
$$F(a_2) = \{\langle 8Z \cup I \rangle, 0, 1, I\}$$

Then $(H, B)$ is a soft neutrosophic strong subbisemigroup of $(F, A)$, where
$$H(a_2) = \{\langle 4Z \cup I \rangle, 0, I\}.$$

**Proposition 3.2.5:** Let $(F, A)$ be a soft neutrosophic strong bisemigroup over $BN(S)$ and $\{(H_i, B_i) : i \in J\}$ is a non-empty family of soft neutrosophic strong subbisemigroups of $(F, A)$. Then

1. $\bigcap_{i \in J}(H_i, B_i)$ is a soft neutrosophic strong subbisemigroup of $(F, A)$.
2. $\bigwedge_{i \in J}(H_i, B_i)$ is a soft neutrosophic strong subbisemigroup of $\bigwedge_{i \in J}(F, A)$.
3. $\bigcup_{i \in J}(H_i, B_i)$ is a soft neutrosophic strong subbisemigroup of $(F, A)$ if $B_i \cap B_j = \phi$, for all $i \neq j$.

**Proof.** Straightforward.

**Definition 3.2.6:** $(F, A)$ is called soft neutrosophic strong biideal over $BN(S)$ if $F(a)$ is a neutrosophic strong biideal of $BN(S)$, for all $a \in A$.

**Example 3.2.6:** Let $BN(S) = \{\langle Z \cup I \rangle \cup \langle Z_3 \cup I \rangle\}$, where $\langle Z \cup I \rangle$ is a neutrosophic semigrouop with respect to $+$ and $\langle Z_3 \cup I \rangle$ is a neutrosophic semigroup under multiplication modulo $3$. $BN(S)$. Let $A = \{a_1, a_2\}$ be a set of parameters. Then $(F, A)$ is soft neutrosophic strong biideal over $BN(S)$, where

$$F(a_1) = \{\langle 2Z \cup I \rangle, 0, I\}$$
$$F(a_2) = \{\langle 8Z \cup I \rangle, 0, 1, I\}$$

**Theorem 3.2.4:** Every soft neutrosophic strong biideal $(F,A)$ over $BN(S)$ is a soft neutrosophic bisemigroup.

**Theorem 3.2.5:** Every soft neutrosophic strong biideal $(F,A)$ over $BN(S)$ is
a soft neutrosophic strong bisemigroup but the converse is not true.

**Proposition 3.2.6:** Let $(F,A)$ and $(K,B)$ be two soft neutrosophic strong biideals over $BN(S)$. Then

1. Their extended intersection $(F,A) \cap_E (K,B)$ is soft neutrosophic strong biideal over $BN(S)$.
2. Their restricted intersection $(F,A) \cap_R (K,B)$ is soft neutrosophic strong biideal over $BN(S)$.
3. Their $AND$ operation $(F,A) \wedge (K,B)$ is soft neutrosophic strong biideal over $BN(S)$.

**Remark 3.2.4:** Let $(F,A)$ and $(K,B)$ be two soft neutrosophic strong biideals over $BN(S)$. Then

1. Their extended union $(F,A) \cup_E (K,B)$ is not soft neutrosophic strong biideals over $BN(S)$.
2. Their restricted union $(F,A) \cup_R (K,B)$ is not soft neutrosophic strong biidleals over $BN(S)$.
3. Their $OR$ operation $(F,A) \vee (K,B)$ is not soft neutrosophic strong biideals over $BN(S)$.

One can easily proved $(1),(2),$ and $(3)$ by the help of examples

**Theorem 3.2.6:** Let $(F,A)$ be a soft neutrosophic strong biideal over $BN(S)$ and $\{(H_i, B_i) : i \in J\}$ is a non empty family of soft neutrosophic strong biideals of $(F,A)$. Then

1. $\bigcap_{i \in J}(H_i, B_i)$ is a soft neutrosophic strong biideal of $(F,A)$.
2. $\bigwedge_{i \in J}(H_i, B_i)$ is a soft neutrosophic strong biideal of $\bigwedge_{i \in J}(F,A)$.

**Proof:** The proof is left as an exercise for the readers.

### 3.3 Soft Neutrosophic N-semigroup

In this section, we extend soft sets to neutrosophic N-semigroups and introduce soft neutrosophic N-semigroups. This is the generalization of soft neutrosophic semigroups. Some of their impotant facts and figures are also presented here with illustrative examples. We also initiated the strong part of neutrosophy in this section. Now we proceed on to define soft neutrosophic N-semigroup as follows.

**Definition 3.3.1:** Let $\{S(N), *_1, ..., *_N\}$ be a neutrosophic $N$-semigroup and $(F,A)$ be a soft set over $\{S(N), *_1, ..., *_N\}$. Then $(F,A)$ is termed as soft neutrosophic $N$-semigroup if and only if $F(a)$ is a neutrosophic sub $N$-semigroup, for all $a \in A$.

We give further expaination in the following example.

**Example 3.3.1:** Let $S(N) = \{S_1 \cup S_2 \cup S_3 \cup S_4, *_1, *_2, *_3, *_4\}$ be a neutrosophic 4-semigroup where

$S_1 = \{Z_{12}, \text{semigroup under multiplication modulo } 12\}$,

$S_2 = \{0,1,2,3,I,2I,3I,4I$, semigroup under multiplication modulo $4\}$, a neutrosophic semigroup.

$S_3 = \left\{ \begin{pmatrix} a & b \\ c & d \end{pmatrix} : a,b,c,d \in \langle R \cup I \rangle \right\}$, neutrosophic semigroup under matrix multiplication and

$S_4 = \langle Z \cup I \rangle$, neutrosophic semigroup under multiplication.

Let $A = \{a_1, a_2\}$ be a set of parameters. Then $(F, A)$ is soft neutrosophic 4-semigroup over $S(4)$, where

$$F(a_1) = \{0,2,4,6,8,10\} \cup \{0,I,2I,3I\} \cup \left\{ \begin{pmatrix} a & b \\ c & d \end{pmatrix} : a,b,c,d \in \langle Q \cup I \rangle \right\} \cup \langle 5Z \cup I \rangle$$

$$F(a_1) = \{0,6,\} \cup \{0,1,I\} \cup \left\{ \begin{pmatrix} a & b \\ c & d \end{pmatrix} : a,b,c,d \in \langle Z \cup I \rangle \right\} \cup \langle 2Z \cup I \rangle$$

**Theorem 3.3.1:** Let $(F, A)$ and $(H, A)$ be two soft neutrosophic $N$-semigroup over $S(N)$. Then their intersection $(F, A) \cap (H, A)$ is again a soft neutrosophic $N$-semigroup over $S(N)$.

**Proof.** Straightforward.

**Theorem 3.3.2:** Let $(F, A)$ and $(H, B)$ be two soft neutrosophic $N$-semigroups over $S(N)$ such that $A \cap B = \phi$. Then their union is soft neutrosophic $N$-semigroup over $S(N)$.

**Proof.** Straightforward.

**Proposition 3.3.1:** Let $(F,A)$ and $(K,B)$ be two soft neutrosophic $N$-semigroup over $S(N)$. Then
1. Their extended intersection $(F,A) \cap_E (K,B)$ is soft neutrosophic $N$-semigroup over $S(N)$.
2. Their restricted intersection $(F,A) \cap_R (K,B)$ is soft neutrosophic $N$-semigroup over $S(N)$.
3. Their AND operation $(F,A) \wedge (K,B)$ is soft neutrosophic $N$-semigroup over $S(N)$.

**Proof:** These can be seen easily.

**Remark 3.3.1:** Let $(F,A)$ and $(K,B)$ be two soft neutrosophic $N$-semigroup over $S(N)$. Then
1. Their extended union $(F,A) \cup_E (K,B)$ is not soft neutrosophic $N$-semigroup over $S(N)$.
2. Their restricted union $(F,A) \cup_R (K,B)$ is not soft neutrosophic $N$-semigroup over $S(N)$.
3. Their OR operation $(F,A) \vee (K,B)$ is not soft neutrosophic $N$-semigroup over $S(N)$.

One can easily verified $(1), (2),$ and $(3)$.

**Definition 3.3.2:** Let $(F,A)$ be a soft neutrosophic $N$-semigroup over $S(N)$. Then $(F,A)$ is called absolute-soft neutrosophic $N$-semigroup over $S(N)$ if $F(a) = S(N)$, for all $a \in A$. We denote it by $\mathbb{F}_{S(N)}$.

**Definition 3.3.3:** Let $(F,A)$ and $(H,B)$ be two soft neutrosophic $N$-semigroup over $S(N)$. Then $(H,B)$ is a soft neutrosophic sub $N$-semigroup of $(F,A)$, if
1) $B \subset A$.
2) $H(a)$ is a neutrosophic sub $N$-semigroup of $F(a)$ for all $a \in A$.

**Example 3.3.2:** Let $S(N) = \{S_1 \cup S_2 \cup S_3 \cup S_4, *_1, *_2, *_3, *_4\}$ be a neutrosophic 4-semigroup where
$S_1 = \{Z_{12}, \text{semigroup under multiplication modulo } 12\}$,
$S_2 = \{0,1,2,3,I,2I,3I,4I, \text{semigroup under multiplication modulo } 4\}$, a neutrosophic semigroup.
$S_3 = \left\{ \begin{pmatrix} a & b \\ c & d \end{pmatrix} : a,b,c,d \in \langle R \cup I \rangle \right\}$, neutrosophic semigroup under matrix multiplication and
$S_4 = \langle Z \cup I \rangle$, neutrosophic semigroup under multiplication.
Let $A = \{a_1, a_2, a_3\}$ be a set of parameters. Then $(F,A)$ is soft neutrosophic 4-semigroup over $S(4)$, where

$F(a_1) = \{0,2,4,6,8,10\} \cup \{0,I,2I,3I\} \cup \left\{ \begin{pmatrix} a & b \\ c & d \end{pmatrix} : a,b,c,d \in \langle Q \cup I \rangle \right\} \cup \langle 5Z \cup I \rangle$,

$F(a_2) = \{0,6\} \cup \{0,1,I\} \cup \left\{ \begin{pmatrix} a & b \\ c & d \end{pmatrix} : a,b,c,d \in \langle Z \cup I \rangle \right\} \cup \langle 2Z \cup I \rangle$,

$F(a_3) = \{0,3,6,9\} \cup \{0,I,2I\} \cup \left\{ \begin{pmatrix} a & b \\ c & d \end{pmatrix} : a,b,c,d \in \langle 2Z \cup I \rangle \right\} \cup \langle 3Z \cup I \rangle$.

Clearly $(H,B)$ is a soft neutrosophic sub 4-semigroup of $(F,A)$, where

$$H(a_1) = \{0,4,8\} \cup \{0,I,2I\} \cup \left\{ \begin{pmatrix} a & b \\ c & d \end{pmatrix} : a,b,c,d \in \langle Z \cup I \rangle \right\} \cup \langle 10Z \cup I \rangle,$$

$$H(a_3) = \{0,6\} \cup \{0,I\} \cup \left\{ \begin{pmatrix} a & b \\ c & d \end{pmatrix} : a,b,c,d \in \langle 4Z \cup I \rangle \right\} \cup \langle 6Z \cup I \rangle.$$

**Proposition 3.3.2:** Let $(F,A)$ be a soft neutrosophic $N$-semigroup over $S(N)$ and $\{(H_i, B_i) : i \in J\}$ is a non empty family of soft neutrosophic $N$-semigroups of $(F,A)$. Then

1. $\bigcap_{i \in J}(H_i, B_i)$ is a soft neutrosophic $N$-semigroup of $(F,A)$.
2. $\bigwedge_{i \in J}(H_i, B_i)$ is a soft neutrosophic $N$-semigroup of $\bigwedge_{i \in J}(F,A)$.
3. $\bigcup_{i \in J}(H_i, B_i)$ is a soft neutrosophic $N$-semigroup of $(F,A)$ if $B_i \cap B_j = \phi$, for all $i \neq j$.

**Proof.** Straightforward.

**Definition 3.3.4:** $(F,A)$ is called soft neutrosophic $N$-ideal over $S(N)$ if $F(a)$ is a neutrosophic $N$-ideal of $S(N)$, for all $a \in A$.

**Theorem 3.3.3:** Every soft neutrosophic $N$-ideal $(F,A)$ over $S(N)$ is a soft neutrosophic -$N$ semigroup.

**Theorem 3.3.4:** Every soft neutrosophic $N$-ideal $(F,A)$ over $S(N)$ is a soft neutrosophic $N$-semigroup but the converse is not true.

**Proposition 3.3.3:** Let $(F,A)$ and $(K,B)$ be two soft neutrosophic $N$-ideals over $S(N)$. Then

1. Their extended intersection $(F,A) \cap_E (K,B)$ is soft neutrosophic $N$-ideal over $S(N)$.
2. Their restricted intersection $(F,A) \cap_R (K,B)$ is soft neutrosophic biideal over $S(N)$.
3. Their $AND$ operation $(F,A) \wedge (K,B)$ is soft neutrosophic $N$-ideal over $S(N)$.

**Remark 3.3.2:** Let $(F,A)$ and $(K,B)$ be two soft neutrosophic $N$-ideals over $S(N)$. Then

1. Their extended union $(F,A) \cup_E (K,B)$ is not soft neutrosophic $N$-ideal over $S(N)$.
2. Their restricted union $(F,A) \cup_R (K,B)$ is not soft neutrosophic $N$-ideal over $S(N)$.
3. Their $OR$ operation $(F,A) \vee (K,B)$ is not soft neutrosophic $N$-ideals over $S(N)$.

One can easily proved (1),(2), and (3) by the help of examples

**Theorem 3.3.5:** Let $(F,A)$ be a soft neutrosophic $N$-ideal over $S(N)$ and $\{(H_i, B_i) : i \in J\}$ is a non-empty family of soft neutrosophic $N$-ideals of $(F,A)$. Then

1. $\bigcap_{i \in J}(H_i, B_i)$ is a soft neutrosophic $N$-ideal of $(F,A)$.
2. $\bigwedge_{i \in J}(H_i, B_i)$ is a soft neutrosophic $N$-ideal of $\bigwedge_{i \in J}(F,A)$.

## Soft Neutrosophic Strong N-semigroup

The notions of soft neutrosophic strong N-semigroups over neutrosophic N-semiggroups are introduced here. We give some basic definitions of soft neutrosophic strong N-semigroups and illustrated it with the help of exmaples and give some basic results.

**Definition 3.3.5:** Let $\{S(N), *_1, ..., *_N\}$ be a neutrosophic $N$-semigroup and $(F, A)$ be a soft set over $\{S(N), *_1, ..., *_N\}$. Then $(F, A)$ is called soft neutrosophic strong $N$-semigroup if and only if $F(a)$ is a neutrosophic strong sub $N$-semigroup, for all $a \in A$.

**Example 3.3.3:** Let $S(N) = \{S_1 \cup S_2 \cup S_3 \cup S_4, *_1, *_2, *_3, *_4\}$ be a neutrosophic $4$-semigroup where
  $S_1 = \{Z_{12}, \text{semigroup under multiplication modulo } 12\}$,
  $S_2 = \{0, 1, 2, 3, I, 2I, 3I, 4I, \text{semigroup under multiplication modulo } 4\}$, a neutrosophic semigroup.
  $S_3 = \left\{ \begin{pmatrix} a & b \\ c & d \end{pmatrix} : a, b, c, d \in \langle R \cup I \rangle \right\}$, neutrosophic semigroup under matrix multiplication and
  $S_4 = \langle Z \cup I \rangle$, neutrosophic semigroup under multiplication.
Let $A = \{a_1, a_2\}$ be a set of parameters. Then $(F, A)$ is soft neutrosophic strong $4$-semigroup over $S(4)$, where

$$F(a_1) = \{0,3,3I\} \cup \{0,I,2I,3I\} \cup \left\{ \begin{pmatrix} a & b \\ c & d \end{pmatrix} : a,b,c,d \in \langle Q \cup I \rangle \right\} \cup \langle 5Z \cup I \rangle$$

$$F(a_1) = \{0,2I,4I\} \cup \{0,1,I\} \cup \left\{ \begin{pmatrix} a & b \\ c & d \end{pmatrix} : a,b,c,d \in \langle Z \cup I \rangle \right\} \cup \langle 2Z \cup I \rangle$$

**Theorem 3.3.6:** Every soft neutrosophic strong $N$-semigroup is trivially a soft neutrosophic
$N$-semigroup but the converse is not true.

**Proof:** It is left as an exercise for the readers.

**Proposition 3.3.4:** Let $(F, A)$ and $(K, B)$ be two soft neutrosophic strong $N$-semigroups over $S(N)$. Then

1. Their extended intersection $(F, A) \cap_E (K, B)$ is soft neutrosophic strong $N$-semigroup over $S(N)$.
2. Their restricted intersection $(F, A) \cap_R (K, B)$ is soft neutrosophic strong $N$-semigroup over $S(N)$.
3. Their $AND$ operation $(F, A) \wedge (K, B)$ is soft neutrosophic strong $N$-semigroup over $S(N)$.

**Proof:** These can be established easily.

**Remark 3.3.3:** Let $(F,A)$ and $(K,B)$ be two soft neutrosophic strong $N$-semigroup over $S(N)$. Then
1. Their extended union $(F,A) \cup_E (K,B)$ is not soft neutrosophic strong $N$-semigroup over $S(N)$.
2. Their restricted union $(F,A) \cup_R (K,B)$ is not soft neutrosophic strong $N$-semigroup over $S(N)$.
3. Their $OR$ operation $(F,A) \vee (K,B)$ is not soft neutrosophic strong $N$-semigroup over $S(N)$.

One can easily verified $(1), (2)$, and $(3)$.

**Definition 3.3.6:** Let $(F,A)$ and $(H,B)$ be two soft neutrosophic strong $N$-semigroup over $S(N)$. Then $(H,B)$ is a soft neutrosophic strong sub $N$-semigroup of $(F,A)$, if
3) $B \subset A$.
4) $H(a)$ is a neutrosophic strong sub $N$-semigroup of $F(a)$ for all $a \in A$.

**Proposition 3.3.5:** Let $(F,A)$ be a soft neutrosophic strong $N$-semigroup over $S(N)$ and $\{(H_i, B_i) : i \in J\}$ is a non-empty family of soft neutrosophic strong $N$-semigroups of $(F,A)$. Then

1. $\bigcap_{i \in J}(H_i, B_i)$ is a soft neutrosophic strong $N$-semigroup of $(F,A)$.
2. $\bigwedge_{i \in J}(H_i, B_i)$ is a soft neutrosophic strong $N$-semigroup of $\bigwedge_{i \in J}(F,A)$.
3. $\bigcup_{i \in J}(H_i, B_i)$ is a soft neutrosophic strong $N$-semigroup of $(F,A)$ if $B_i \cap B_j = \phi$, for all $i \neq j$.

**Proof.** Straightforward.

**Definition 3.3.7:** $(F,A)$ is called soft neutrosophic strong $N$-ideal over $S(N)$ if $F(a)$ is a neutrosophic strong $N$-ideal of $S(N)$, for all $a \in A$.

**Theorem 3.3.7:** Every soft neutrosophic strong $N$-ideal $(F,A)$ over $S(N)$ is a soft neutrosophic-$N$ semigroup.

**Theorem 3.3.8:** Every soft neutrosophic strong $N$-ideal $(F,A)$ over $S(N)$ is a soft neutrosophic strong $N$-semigroup but the converse is not true.

**Proposition 3.3.6:** Let $(F,A)$ and $(K,B)$ be two soft neutrosophic strong $N$-ideals over $S(N)$. Then

1. Their extended intersection $(F,A) \cap_E (K,B)$ is soft neutrosophic strong $N$-ideal over $S(N)$.
2. Their restricted intersection $(F,A) \cap_R (K,B)$ is soft neutrosophic strong $N$-ideal over $S(N)$.
3. Their AND operation $(F,A) \wedge (K,B)$ is soft neutrosophic strong $N$-ideal over $S(N)$.

**Remark 3.3.4:** Let $(F,A)$ and $(K,B)$ be two soft neutrosophic strong $N$-ideals over $S(N)$. Then

1. Their extended union $(F,A) \cup_E (K,B)$ is not soft neutrosophic strong $N$-ideal over $S(N)$.
2. Their restricted union $(F,A) \cup_R (K,B)$ is not soft neutrosophic strong $N$-ideal over $S(N)$.
3. Their OR operation $(F,A) \vee (K,B)$ is not soft neutrosophic strong $N$-ideals over $S(N)$.

One can easily proved (1),(2), and (3) by the help of examples.

**Theorem 3.3.9:** Let $(F,A)$ be a soft neutrosophic strong $N$-ideal over $S(N)$ and $\{(H_i,B_i):i\in J\}$ is a non-empty family of soft neutrosophic strong $N$-ideals of $(F,A)$. Then

1. $\underset{i\in J}{\cap}(H_i,B_i)$ is a soft neutrosophic strong $N$-ideal of $(F,A)$.

$\underset{i\in J}{\wedge}(H_i,B_i)$ is a soft neutrosophic strong $N$-ideal of $\underset{i\in J}{\wedge}(F,A)$.

# Chapter four

**SOFT NEUTROSOPHIC LOOPS AND THEIR GENERALIZATION**

In this chapter the notion of soft neutrosophic loop with its generalization are given. This chapter has also three sections. In first section, soft neutrosophic loop is introduced over a neutrosophic loop. Actually soft neutrosophic loops are parameterized family of neutrosophic subloops. We also investigate soft neutrosophic strong loop over a neutrosophic loop which is purely of neutrosophic character. Now we proceed on to define soft neutrosophic loop over a neutrosophic loop.

## 4.1 Soft Neutrosophic Loop

In this section, we defined a soft neutrosophic loop over a neutrosophic loop and established some of the fundamental properties of it. We also introduce soft neutrosophic strong loop over a neutrosophic semigroup.

**Definition 4.1.1:** Let $\langle L \cup I \rangle$ be a neutrosophic loop and $(F, A)$ be a soft set over $\langle L \cup I \rangle$. Then $(F, A)$ is called soft neutrosophic loop if and only if $F(a)$ is neutrosophic subloop of $\langle L \cup I \rangle$ for all $a \in A$.

**Example 4.1.1:** Let $\langle L \cup I \rangle = \langle L_7(4) \cup I \rangle$ be a neutrosophic loop where $L_7(4)$ is a loop. Then $(F,A)$ is a soft neutrosophic loop over $\langle L \cup I \rangle$, where

$$F(a_1) = \{\langle e, eI, 2, 2I \rangle\}, F(a_2) = \{\langle e, 3 \rangle\},$$
$$F(a_3) = \{\langle e, eI \rangle\}.$$

**Theorem 4.1.1:** Every soft neutrosophic loop over $\langle L \cup I \rangle$ contains a soft loop over $L$.

**Proof.** The proof is straightforward.

**Theorem 4.1.2:** Let $(F,A)$ and $(H,A)$ be two soft neutrosophic loops over $\langle L \cup I \rangle$. Then their intersection $(F,A) \cap (H,A)$ is again soft neutrosophic loop over $\langle L \cup I \rangle$.

**Proof.** The proof is staightforward.

**Theorem 4.1.3:** Let $(F,A)$ and $(H,C)$ be two soft neutrosophic loops over $\langle L \cup I \rangle$. If $A \cap C = \phi$, then $(F,A) \cup (H,C)$ is a soft neutrosophic loop over $\langle L \cup I \rangle$.

**Remark 4.1.1:** The extended union of two soft neutrosophic loops $(F,A)$ and $(K,C)$ over $\langle L \cup I \rangle$ is not a soft neutrosophic loop over $\langle L \cup I \rangle$.

With the help of example we can easily check the above remark.

**Proposition 4.1.1:** The extended intersection of two soft neutrosophic loopps over $\langle L \cup I \rangle$ is a soft neutrosophic loop over $\langle L \cup I \rangle$.

**Remark 4.1.2:** The restricted union of two soft neutrosophic loops $(F,A)$ and $(K,C)$ over $\langle L \cup I \rangle$ is not a soft neutrosophic loop over $\langle L \cup I \rangle$.

One can easily check it by the help of example.

**Proposition 4.1.2:** The restricted intersection of two soft neutrosophic loops over $\langle L \cup I \rangle$ is a soft neutrosophic loop over $\langle L \cup I \rangle$.

**Proposition 4.1.3:** The *AND* operation of two soft neutrosophic loops over $\langle L \cup I \rangle$ is a soft neutrosophic loop over $\langle L \cup I \rangle$.

**Remark 4.1.3:** The *OR* operation of two soft neutosophic loops over $\langle L \cup I \rangle$ may not be a soft nuetrosophic loop over $\langle L \cup I \rangle$.

One can easily check it by the help of example.

**Definition 4.1.2:** Let $\langle L_n(m) \cup I \rangle = \{e,1,2,...,n,eI,1I,2I,...,nI\}$ be a new class of neutrosophic loop and $(F,A)$ be a soft neutrosophic loop over $\langle L_n(m) \cup I \rangle$. Then $(F,A)$ is called soft new class neutrosophic loop if $F(a)$ is a neutrosophic subloop of $\langle L_n(m) \cup I \rangle$ for all $a \in A$.

**Example 4.1.2:** Let $\langle L_5(3) \cup I \rangle = \{e,1,2,3,4,5,eI,1I,2I,3I,4I,5I\}$ be a new class of neutrosophic loop. Let $A = \{a_1, a_2, a_3, a_4, a_5\}$ be a set of parameters. Then $(F,A)$ is soft new class neutrosophic loop over $\langle L_5(3) \cup I \rangle$, where

$$F(a_1) = \{e, eI, 1, 1I\}, F(a_2) = \{e, eI, 2, 2I\},$$
$$F(a_3) = \{e, eI, 3, 3I\}, F(a_3) = \{e, eI, 4, 4I\},$$
$$F(a_5) = \{e, eI, 5, 5I\}.$$

**Theorem 4.1.4:** Every soft new class neutrosophic loop over $\langle L_n(m) \cup I \rangle$ is a soft neutrosophic loop over $\langle L_n(m) \cup I \rangle$ but the converse is not true.

**Proposition 4.1.4:** Let $(F, A)$ and $(K, C)$ be two soft new class neutrosophic loops over $\langle L_n(m) \cup I \rangle$. Then

1. Their extended intersection $(F, A) \cap_E (K, C)$ is a soft new class neutrosophic loop over $\langle L_n(m) \cup I \rangle$.
2. Their restricted intersection $(F, A) \cap_R (K, C)$ is a soft new classes neutrosophic loop over $\langle L_n(m) \cup I \rangle$.
3. Their *AND* operation $(F, A) \wedge (K, C)$ is a soft new class neutrosophic loop over $\langle L_n(m) \cup I \rangle$.

**Remark 4.1.4:** Let $(F, A)$ and $(K, C)$ be two soft new class neutrosophic loops over $\langle L_n(m) \cup I \rangle$. Then

1. Their extended union $(F, A) \cup_E (K, C)$ is not a soft new class neutrosophic loop over $\langle L_n(m) \cup I \rangle$.
2. Their restricted union $(F, A) \cup_R (K, C)$ is not a soft new class neutrosophic loop over $\langle L_n(m) \cup I \rangle$.
3. Their *OR* operation $(F, A) \vee (K, C)$ is not a soft new class neutrosophic loop over $\langle L_n(m) \cup I \rangle$.

One can easily verify (1), (2), and (3) by the help of examples.

**Definition 4.1.3:** Let $(F,A)$ be a soft neutrosophic loop over $\langle L \cup I \rangle$. Then $(F,A)$ is called the identity soft neutrosophic loop over $\langle L \cup I \rangle$ if $F(a) = \{e\}$ for all $a \in A$, where $e$ is the identity element of $\langle L \cup I \rangle$.

**Definition 4.1.4:** Let $(F,A)$ be a soft neutrosophic loop over $\langle L \cup I \rangle$. Then $(F,A)$ is called an absolute soft neutrosophic loop over $\langle L \cup I \rangle$ if $F(a) = \langle L \cup I \rangle$ for all $a \in A$.

**Definition 4.1.5:** Let $(F,A)$ and $(H,C)$ be two soft neutrosophic loops over $\langle L \cup I \rangle$. Then $(H,C)$ is callsed soft neutrosophic subloop of $(F,A)$, if
1. $C \subseteq A$.
2. $H(a)$ is a neutrosophic subloop of $F(a)$ for all $a \in A$.

**Example 4.1.3:** Consider the neutrosophic loop

$$\langle L_{15}(2) \cup I \rangle = \{e, 1, 2, 3, 4, \ldots, 15, eI, 1I, 2I, \ldots, 14I, 15I\}$$

of order $32$. Let $A = \{a_1, a_2, a_3\}$ be a set of parameters. Then $(F,A)$ is a soft neutrosophic loop over $\langle L_{15}(2) \cup I \rangle$, where

$$F(a_1) = \{e, 2, 5, 8, 11, 14, eI, 2I, 5I, 8I, 11I, 14I\},$$
$$F(a_2) = \{e, 2, 5, 8, 11, 14\},$$
$$F(a_3) = \{e, 3, eI, 3I\}.$$

Thus $(H,C)$ is a soft neutrosophic subloop of $(F,A)$ over $\langle L_{15}(2) \cup I \rangle$, where

$$H(a_1) = \{e, eI, 2I, 5I, 8I, 11I, 14I\},$$
$$H(a_2) = \{e, 3\}.$$

**Theorem 4.1.5:** Every soft loop over $L$ is a soft neutrosophic subloop over $\langle L \cup I \rangle$.

**Proof:** It is left an exercise for the interested readers.

**Definition 4.1.6:** Let $\langle L \cup I \rangle$ be a neutrosophic loop and $(F,A)$ be a soft set over $\langle L \cup I \rangle$. Then $(F,A)$ is called soft normal neutrosophic loop if and only if $F(a)$ is normal neutrosophic subloop of $\langle L \cup I \rangle$ for all $a \in A$.

**Example 4.1.4:** Let $\langle L_5(3) \cup I \rangle = \{e, 1, 2, 3, 4, 5, eI, 1I, 2I, 3I, 4I, 5I\}$ be a neutrosophic loop. Let $A = \{a_1, a_2, a_3\}$ be a set of parameters. Then clearly $(F,A)$ is soft normal neutrosophic loop over $\langle L_5(3) \cup I \rangle$, where

$$F(a_1) = \{e, eI, 1, 1I\}, F(a_2) = \{e, eI, 2, 2I\},$$
$$F(a_3) = \{e, eI, 3, 3I\}.$$

**Theorem 4.1.6:** Every soft normal neutrosophic loop over $\langle L \cup I \rangle$ is a soft neutrosophic loop over $\langle L \cup I \rangle$ but the converse is not true.

One can easily check it by the help of example.

**Proposition 4.1.5:** Let $(F,A)$ and $(K,C)$ be two soft normal neutrosophic loops over $\langle L\cup I\rangle$. Then

1. Their extended intersection $(F,A)\cap_E (K,C)$ is a soft normal neutrosophic loop over $\langle L\cup I\rangle$.
2. Their restricted intersection $(F,A)\cap_R (K,C)$ is a soft normal neutrosophic loop over $\langle L\cup I\rangle$.
3. Their *AND* operation $(F,A)\wedge(K,C)$ is a soft normal neutrosophic loop over $\langle L\cup I\rangle$.

**Remark 4.1.5:** Let $(F,A)$ and $(K,C)$ be two soft normal neutrosophic loops over $\langle L\cup I\rangle$. Then

1. Their extended union $(F,A)\cup_E (K,C)$ is not a soft normal neutrosophic loop over $\langle L\cup I\rangle$.
2. Their restricted union $(F,A)\cup_R (K,C)$ is not a soft normal neutrosophic loop over $\langle L\cup I\rangle$.
3. Their *OR* operation $(F,A)\vee(K,C)$ is not a soft normal neutrosophic loop over $\langle L\cup I\rangle$.

One can easily verify $(1),(2),$ and $(3)$ by the help of examples.

**Definition 4.1.7:** Let $\langle L\cup I\rangle$ be a neutrosophic loop and $(F,A)$ be a soft neutrosophic loop over $\langle L\cup I\rangle$. Then $(F,A)$ is called soft Lagrange neutrosophic loop if $F(a)$ is a Lagrange neutrosophic subloop of $\langle L\cup I\rangle$ for all $a\in A$.

**Example 4.1.5:** In Example 4.1.1, $(F,A)$ is a soft Lagrange neutrosophic loop over $\langle L \cup I \rangle$.

**Theorem 4.1.7:** Every soft Lagrange neutrosophic loop over $\langle L \cup I \rangle$ is a soft neutrosophic loop over $\langle L \cup I \rangle$ but the converse is not true.

**Theorem 4.1.8:** If $\langle L \cup I \rangle$ is a Lagrange neutrosophic loop, then $(F,A)$ over $\langle L \cup I \rangle$ is a soft Lagrange neutrosophic loop but the converse is not true.

**Remark 4.1.6:** Let $(F,A)$ and $(K,C)$ be two soft Lagrange neutrosophic loops over $\langle L \cup I \rangle$. Then

1. Their extended intersection $(F,A) \cap_E (K,C)$ is not a soft Lagrange neutrosophic loop over $\langle L \cup I \rangle$.
2. Their restricted intersection $(F,A) \cap_R (K,C)$ is not a soft Lagrange neutrosophic loop over $\langle L \cup I \rangle$.
3. Their *AND* operation $(F,A) \wedge (K,C)$ is not a soft Lagrange neutrosophic loop over $\langle L \cup I \rangle$.
4. Their extended union $(F,A) \cup_E (K,C)$ is not a soft Lagrnage neutrosophic loop over $\langle L \cup I \rangle$.
5. Their restricted union $(F,A) \cup_R (K,C)$ is not a soft Lagrange neutrosophic loop over $\langle L \cup I \rangle$.
6. Their *OR* operation $(F,A) \vee (K,C)$ is not a soft Lagrange neutrosophic loop over $\langle L \cup I \rangle$.

One can easily verify (1),(2),(3),(4),(5) and (6) by the help of examples.

**Definition 4.1.8:** Let $\langle L \cup I \rangle$ be a neutrosophic loop and $(F,A)$ be a soft neutrosophic loop over $\langle L \cup I \rangle$. Then $(F,A)$ is called soft weak Lagrange neutrosophic loop if atleast one $F(a)$ is not a Lagrange neutrosophic subloop of $\langle L \cup I \rangle$ for some $a \in A$.

**Example 4.1.6:** Consider the neutrosophic loop

$$\langle L_{15}(2) \cup I \rangle = \{e, 1, 2, 3, 4, \ldots, 15, eI, 1I, 2I, \ldots, 14I, 15I\}$$

of order $32$. Let $A = \{a_1, a_2, a_3\}$ be a set of parameters. Then $(F,A)$ is a soft weakly Lagrange neutrosophic loop over $\langle L_{15}(2) \cup I \rangle$, where

$$F(a_1) = \{e, 2, 5, 8, 11, 14, eI, 2I, 5I, 8I, 11I, 14I\},$$
$$F(a_2) = \{e, 2, 5, 8, 11, 14\},$$
$$F(a_3) = \{e, 3, eI, 3I\}.$$

**Theorem 4.1.9:** Every soft weak Lagrange neutrosophic loop over $\langle L \cup I \rangle$ is a soft neutrosophic loop over $\langle L \cup I \rangle$ but the converse is not true.

One can easily check it by the help of example.

**Theorem 4.1.10:** If $\langle L \cup I \rangle$ is weak Lagrange neutrosophic loop, then $(F,A)$ over $\langle L \cup I \rangle$ is also soft weak Lagrange neutrosophic loop but the converse is not true.

One can easily check it by the help of example.

**Remark 4.1.7:** Let $(F,A)$ and $(K,C)$ be two soft weak Lagrange neutrosophic loops over $\langle L\cup I\rangle$. Then

1. Their extended intersection $(F,A)\cap_E (K,C)$ is not a soft weak Lagrange neutrosophic loop over $\langle L\cup I\rangle$.
2. Their restricted intersection $(F,A)\cap_R (K,C)$ is not a soft weak Lagrange neutrosophic loop over $\langle L\cup I\rangle$.
3. Their *AND* operation $(F,A)\wedge(K,C)$ is not a soft weak Lagrange neutrosophic loop over $\langle L\cup I\rangle$.
4. Their extended union $(F,A)\cup_E (K,C)$ is not a soft weak Lagrnage neutrosophic loop over $\langle L\cup I\rangle$.
5. Their restricted union $(F,A)\cup_R (K,C)$ is not a soft weak Lagrange neutrosophic loop over $\langle L\cup I\rangle$.
6. Their *OR* operation $(F,A)\vee(K,C)$ is not a soft weak Lagrange neutrosophic loop over $\langle L\cup I\rangle$.

One can easily verify $(1),(2),(3),(4),(5)$ and $(6)$ by the help of examples.

**Definition 4.1.9:** Let $\langle L\cup I\rangle$ be a neutrosophic loop and $(F,A)$ be a soft neutrosophic loop over $\langle L\cup I\rangle$. Then $(F,A)$ is called soft Lagrange free neutrosophic loop if $F(a)$ is not a lagrange neutrosophic subloop of $\langle L\cup I\rangle$ for all $a\in A$.

**Theorem 4.1.11:** Every soft Lagrange free neutrosophic loop over $\langle L\cup I\rangle$ is a soft neutrosophic loop over $\langle L\cup I\rangle$ but the converse is not true.

**Theorem 4.1.12:** If $\langle L \cup I \rangle$ is a Lagrange free neutrosophic loop, then $(F,A)$ over $\langle L \cup I \rangle$ is also a soft Lagrange free neutrosophic loop but the converse is not true.

**Remark 4.1.8:** Let $(F,A)$ and $(K,C)$ be two soft Lagrange free neutrosophic loops over $\langle L \cup I \rangle$. Then

1. Their extended intersection $(F,A) \cap_E (K,C)$ is not a soft Lagrange soft neutrosophic loop over $\langle L \cup I \rangle$.
2. Their restricted intersection $(F,A) \cap_R (K,C)$ is not a soft Lagrange soft neutrosophic loop over $\langle L \cup I \rangle$.
3. Their *AND* operation $(F,A) \wedge (K,C)$ is not a soft Lagrange free neutrosophic loop over $\langle L \cup I \rangle$.
4. Their extended union $(F,A) \cup_E (K,C)$ is not a soft Lagrnage soft neutrosophic loop over $\langle L \cup I \rangle$.
5. Their restricted union $(F,A) \cup_R (K,C)$ is not a soft Lagrange free neutrosophic loop over $\langle L \cup I \rangle$.
6. Their *OR* operation $(F,A) \vee (K,C)$ is not a soft Lagrange free neutrosophic loop over $\langle L \cup I \rangle$.

One can easily verify $(1),(2),(3),(4),(5)$ and $(6)$ by the help of examples.

## Soft Neutrosophic Strong Loop

The notion of soft neutrosophic strong loop over a neutrosophic loop is introduced here. We gave the definition of soft neutrosophic strong loop and investigated some related properties with sufficient amount of illustrative examples.

**Definition 4.1.10:** Let $\langle L \cup I \rangle$ be a neutrosophic loop and $(F, A)$ be a soft set over $\langle L \cup I \rangle$. Then $(F, A)$ is called soft neutrosophic strong loop if and only if $F(a)$ is a strong neutrosophic subloop of $\langle L \cup I \rangle$ for all $a \in A$.

**Example 4.1.7:** Consider the neutrosophic loop

$$\langle L_{15}(2) \cup I \rangle = \{e, 1, 2, 3, 4, \ldots, 15, eI, 1I, 2I, \ldots, 14I, 15I\}.$$

Let $A = \{a_1, a_2\}$ be a set of parameters. Then $(F, A)$ is a soft neutrosophic strong loop over $\langle L_{15}(2) \cup I \rangle$, where

$$F(a_1) = \{e, eI, 2I, 5I, 8I, 11I, 14I\},$$
$$F(a_2) = \{e, 3, eI, 3I\}.$$

**Proposition 4.1.6:** Let $(F, A)$ and $(K, C)$ be two soft neutrosophic strong loops over $\langle L \cup I \rangle$. Then

1. Their extended intersection $(F, A) \cap_E (K, C)$ is a soft neutrosophic strong loop over $\langle L \cup I \rangle$.
2. Their restricted intersection $(F, A) \cap_R (K, C)$ is a soft neutrosophic strong loop over $\langle L \cup I \rangle$.
3. Their AND operation $(F, A) \wedge (K, C)$ is a soft neutrosophic strong loop over $\langle L \cup I \rangle$.

**Proof:** These are left to the readers as exercises.

**Remark 4.1.9:** Let $(F,A)$ and $(K,C)$ be two soft neutrosophic strong loops over $\langle L \cup I \rangle$. Then

1. Their extended union $(F,A) \cup_E (K,C)$ is not a soft neutrosophic strong loop over $\langle L \cup I \rangle$.
2. Their restricted union $(F,A) \cup_R (K,C)$ is not a soft neutrosophic strong loop over $\langle L \cup I \rangle$.
3. Their $OR$ operation $(F,A) \vee (K,C)$ is not a soft neutrosophic strong loop over $\langle L \cup I \rangle$.

One can easily verify $(1), (2),$ and $(3)$ by the help of examples.

**Definition 4.1.11:** Let $(F,A)$ and $(H,C)$ be two soft neutrosophic strong loops over $\langle L \cup I \rangle$. Then $(H,C)$ is called soft neutrosophic strong subloop of $(F,A)$, if

1. $C \subseteq A$.
2. $H(a)$ is a neutrosophic strong subloop of $F(a)$ for all $a \in A$.

**Definition 4.1.12:** Let $\langle L \cup I \rangle$ be a neutrosophic strong loop and $(F,A)$ be a soft neutrosophic loop over $\langle L \cup I \rangle$. Then $(F,A)$ is called soft Lagrange neutrosophic strong loop if $F(a)$ is a Lagrange neutrosophic strong subloop of $\langle L \cup I \rangle$ for all $a \in A$.

**Theorem 4.1.13:** Every soft Lagrange neutrosophic strong loop over $\langle L \cup I \rangle$ is a soft neutrosophic loop over $\langle L \cup I \rangle$ but the converse is not true.

**Theorem 4.1.14:** If $\langle L \cup I \rangle$ is a Lagrange neutrosophic strong loop, then $(F,A)$ over $\langle L \cup I \rangle$ is a soft Lagrange neutrosophic loop but the converse is not true.

**Remark 4.1.10:** Let $(F,A)$ and $(K,C)$ be two soft Lagrange neutrosophic strong loops over $\langle L \cup I \rangle$. Then

1. Their extended intersection $(F,A) \cap_E (K,C)$ is not a soft Lagrange neutrosophic strong loop over $\langle L \cup I \rangle$.
2. Their restricted intersection $(F,A) \cap_R (K,C)$ is not a soft Lagrange strong neutrosophic loop over $\langle L \cup I \rangle$.
3. Their AND operation $(F,A) \wedge (K,C)$ is not a soft Lagrange neutrosophic strong loop over $\langle L \cup I \rangle$.
4. Their extended union $(F,A) \cup_E (K,C)$ is not a soft Lagrnage neutrosophic strong loop over $\langle L \cup I \rangle$.
5. Their restricted union $(F,A) \cup_R (K,C)$ is not a soft Lagrange neutrosophic strong loop over $\langle L \cup I \rangle$.
6. Their OR operation $(F,A) \vee (K,C)$ is not a soft Lagrange neutrosophic strong loop over $\langle L \cup I \rangle$.

One can easily verify $(1),(2),(3),(4),(5)$ and $(6)$ by the help of examples.

**Definition 4.1.13:** Let $\langle L \cup I \rangle$ be a neutrosophic strong loop and $(F,A)$ be a soft neutrosophic loop over $\langle L \cup I \rangle$. Then $(F,A)$ is called soft weak Lagrange neutrosophic strong loop if atleast one $F(a)$ is not a Lagrange neutrosophic strong subloop of $\langle L \cup I \rangle$ for some $a \in A$.

**Theorem 4.1.15:** Every soft weak Lagrange neutrosophic strong loop over $\langle L \cup I \rangle$ is a soft neutrosophic loop over $\langle L \cup I \rangle$ but the converse is not true.

**Theorem 4.1.16:** If $\langle L \cup I \rangle$ is weak Lagrange neutrosophic strong loop, then $(F,A)$ over $\langle L \cup I \rangle$ is also soft weak Lagrange neutrosophic strong loop but the converse is not true.

**Remark 4.1.11:** Let $(F,A)$ and $(K,C)$ be two soft weak Lagrange neutrosophic strong loops over $\langle L \cup I \rangle$. Then

1. Their extended intersection $(F,A) \cap_E (K,C)$ is not a soft weak Lagrange neutrosophic strong loop over $\langle L \cup I \rangle$.
2. Their restricted intersection $(F,A) \cap_R (K,C)$ is not a soft weak Lagrange neutrosophic strong loop over $\langle L \cup I \rangle$.
3. Their *AND* operation $(F,A) \wedge (K,C)$ is not a soft weak Lagrange neutrosophic strong loop over $\langle L \cup I \rangle$.
4. Their extended union $(F,A) \cup_E (K,C)$ is not a soft weak Lagrnage neutrosophic strong loop over $\langle L \cup I \rangle$.
5. Their restricted union $(F,A) \cup_R (K,C)$ is not a soft weak Lagrange neutrosophic strong loop over $\langle L \cup I \rangle$.
6. Their *OR* operation $(F,A) \vee (K,C)$ is not a soft weak Lagrange neutrosophic strong loop over $\langle L \cup I \rangle$.

One can easily verify $(1), (2), (3), (4), (5)$ and $(6)$ by the help of examples.

**Definition 4.1.14:** Let $\langle L \cup I \rangle$ be a neutrosophic strong loop and $(F,A)$ be a soft neutrosophic loop over $\langle L \cup I \rangle$. Then $(F,A)$ is called soft Lagrange free neutrosophic strong loop if $F(a)$ is not a lagrange neutrosophic strong subloop of $\langle L \cup I \rangle$ for all $a \in A$.

**Theorem 4.1.17:** Every soft Lagrange free neutrosophic strong loop over $\langle L \cup I \rangle$ is a soft neutrosophic loop over $\langle L \cup I \rangle$ but the converse is not true.

**Theorem 4.1.18:** If $\langle L \cup I \rangle$ is a Lagrange free neutrosophic strong loop, then $(F,A)$ over $\langle L \cup I \rangle$ is also a soft Lagrange free neutrosophic strong loop but the converse is not true.

**Remark 4.1.12:** Let $(F,A)$ and $(K,C)$ be two soft Lagrange free neutrosophic strong loops over $\langle L \cup I \rangle$. Then

1. Their extended intersection $(F,A) \cap_E (K,C)$ is not a soft Lagrange free neutrosophic strong loop over $\langle L \cup I \rangle$.
2. Their restricted intersection $(F,A) \cap_R (K,C)$ is not a soft Lagrange free neutrosophic strong loop over $\langle L \cup I \rangle$.
3. Their *AND* operation $(F,A) \wedge (K,C)$ is not a soft Lagrange free neutrosophic strong loop over $\langle L \cup I \rangle$.
4. Their extended union $(F,A) \cup_E (K,C)$ is not a soft Lagrnage free strong neutrosophic strong loop over $\langle L \cup I \rangle$.
5. Their restricted union $(F,A) \cup_R (K,C)$ is not a soft Lagrange free neutrosophic loop over $\langle L \cup I \rangle$.
6. Their *OR* operation $(F,A) \vee (K,C)$ is not a soft Lagrange free neutrosophic strong loop over $\langle L \cup I \rangle$.

One can easily verify $(1),(2),(3),(4),(5)$ and $(6)$ by the help of examples.

## 4.2 Soft Neutrosophic Biloop

In this section, the definition of soft neutrosophic biloop is given over a neutrosophic biloop. Some of their fundamental properties are also given in this section.

**Definition 4.2.1:** Let $(\langle B \cup I \rangle, *_1, *_2)$ be a neutrosophic biloop and $(F, A)$ be a soft set over $(\langle B \cup I \rangle, *_1, *_2)$. Then $(F, A)$ is called soft neutrosophic biloop if and only if $F(a)$ is a neutrosophic subbiloop of $(\langle B \cup I \rangle, *_1, *_2)$ for all $a \in A$.

**Example 4.2.1:** Let $(\langle B \cup I \rangle, *_1, *_2) = (\{e, 1, 2, 3, 4, 5, eI, 1I, 2I, 3I, 4I, 5I\} \cup \{g : g^6 = e\}$ be a neutrosophic biloop. Let $A = \{a_1, a_2\}$ be a set of parameters. Then $(F, A)$ is clearly soft neutrosophic biloop over $(\langle B \cup I \rangle, *_1, *_2)$, where

$$F(a_1) = \{e, 2, eI, 2I\} \cup \{g^2, g^4, e\},$$
$$F(a_2) = \{e, 3, eI, 3I\} \cup \{g^3, e\}.$$

**Theorem 4.2.1:** Let $(F, A)$ and $(H, A)$ be two soft neutrosophic biloops over $(\langle B \cup I \rangle, *_1, *_2)$. Then their intersection $(F, A) \cap (H, A)$ is again a soft neutrosophic biloop over $(\langle B \cup I \rangle, *_1, *_2)$.

**Proof.** Straightforward.

**Theorem 4.2.2:** Let $(F,A)$ and $(H,C)$ be two soft neutrosophic biloops over $(\langle B\cup I\rangle,*_1,*_2)$ such that $A\cap C=\phi$. Then their union is soft neutrosophic biloop over $(\langle B\cup I\rangle,*_1,*_2)$.

**Proof.** Straightforward.

**Proposition 4.2.1:** Let $(F,A)$ and $(K,C)$ be two soft neutrosophic biloops over $(\langle B\cup I\rangle,*_1,*_2)$. Then

1. Their extended intersection $(F,A)\cap_E (K,C)$ is a soft neutrosophic biloop over $(\langle B\cup I\rangle,*_1,*_2)$.
2. Their restricted intersection $(F,A)\cap_R (K,C)$ is a soft neutrosophic biloop over $(\langle B\cup I\rangle,*_1,*_2)$.
3. Their *AND* operation $(F,A)\wedge(K,C)$ is a soft neutrosophic biloop over $(\langle B\cup I\rangle,*_1,*_2)$.

**Remark 4.2.1:** Let $(F,A)$ and $(K,C)$ be two soft neutrosophic biloops over $(\langle B\cup I\rangle,*_1,*_2)$. Then

1. Their extended union $(F,A)\cup_E (K,C)$ is not a soft neutrosophic biloop over $(\langle B\cup I\rangle,*_1,*_2)$.
2. Their restricted union $(F,A)\cup_R (K,C)$ is not a soft neutrosophic biloop over $(\langle B\cup I\rangle,*_1,*_2)$.
3. Their *OR* operation $(F,A)\vee(K,C)$ is not a soft neutrosophic biloop over $(\langle B\cup I\rangle,*_1,*_2)$.

One can easily verify $(1),(2),$ and $(3)$ by the help of examples.

**Definition 4.2.2:** Let $B = (\langle L_n(m) \cup I \rangle \cup B_2, *_1, *_2)$ be a new class neutrosophic biloop and $(F, A)$ be a soft set over $B = (\langle L_n(m) \cup I \rangle \cup B_2, *_1, *_2)$. Then $B = (\langle L_n(m) \cup I \rangle \cup B_2, *_1, *_2)$ is called soft new class neutrosophic subbiloop if and only if $F(a)$ is a neutrosophic subbiloop of $B = (\langle L_n(m) \cup I \rangle \cup B_2, *_1, *_2)$ for all $a \in A$.

**Example 4.2.2:** Let $B = (B_1 \cup B_2, *_1, *_2)$ be a new class neutrosophic biloop, where $B_1 = \langle L_5(3) \cup I \rangle = \{e, 1, 2, 3, 4, 5, eI, 2I, 3I, 4I, 5I\}$ be a new class of neutrosophic loop and $B_2 = \{g : g^{12} = e\}$ is a group. $\{e, eI, 1, 1I\} \cup \{1, g^6\}$, $\{e, eI, 2, 2I\} \cup \{1, g^2, g^4, g^6, g^8, g^{10}\}$, $\{e, eI, 3, 3I\} \cup \{1, g^3, g^6, g^9\}$, $\{e, eI, 4, 4I\} \cup \{1, g^4, g^8\}$ are neutrosophic subloops of $B$. Let $A = \{a_1, a_2, a_3, a_4\}$ be a set of parameters. Then $(F, A)$ is soft new class neutrosophic biloop over $B$, where

$$F(a_1) = \{e, eI, 1, 1I\} \cup \{e, g^6\},$$
$$F(a_2) = \{e, eI, 2, 2I\} \cup \{e, g^2, g^4, g^6, g^8, g^{10}\},$$
$$F(a_3) = \{e, eI, 3, 3I\} \cup \{e, g^3, g^6, g^6\},$$
$$F(a_4) = \{e, eI, 4, 4I\} \cup \{e, g^4, g^8\}.$$

**Theorem 4.2.3:** Every soft new class neutrosophic biloop over $B = (\langle L_n(m) \cup I \rangle \cup B_2, *_1, *_2)$ is trivially a soft neutrosophic biloop over but the converse is not true.

One can easily check it by the help of example.

**Proposition 4.2.2:** Let $(F,A)$ and $(K,C)$ be two soft new class neutrosophic biloops over $B = (\langle L_n(m) \cup I \rangle \cup B_2, *_1, *_2)$. Then

1. Their extended intersection $(F,A) \cap_E (K,C)$ is a soft new class neutrosophic biloop over $B = (\langle L_n(m) \cup I \rangle \cup B_2, *_1, *_2)$.
2. Their restricted intersection $(F,A) \cap_R (K,C)$ is a soft new class neutrosophic biloop over $B = (\langle L_n(m) \cup I \rangle \cup B_2, *_1, *_2)$.
3. Their $AND$ operation $(F,A) \wedge (K,C)$ is a soft new class neutrosophic biloop over $B = (\langle L_n(m) \cup I \rangle \cup B_2, *_1, *_2)$.

**Remark 4.2.2:** Let $(F,A)$ and $(K,C)$ be two soft new class neutrosophic biloops over $B = (\langle L_n(m) \cup I \rangle \cup B_2, *_1, *_2)$. Then

1. Their extended union $(F,A) \cup_E (K,C)$ is not a soft new class neutrosophic biloop over $B = (\langle L_n(m) \cup I \rangle \cup B_2, *_1, *_2)$.
2. Their restricted union $(F,A) \cup_R (K,C)$ is not a soft new class neutrosophic biloop over $B = (\langle L_n(m) \cup I \rangle \cup B_2, *_1, *_2)$.
3. Their $OR$ operation $(F,A) \vee (K,C)$ is not a soft new class neutrosophic biloop over $B = (\langle L_n(m) \cup I \rangle \cup B_2, *_1, *_2)$.

One can easily verify $(1), (2),$ and $(3)$ by the help of examples.

**Definition 4.2.3:** Let $(F,A)$ be a soft neutrosophic biloop over $B = (\langle B_1 \cup I \rangle \cup B_2, *_1, *_2)$. Then $(F,A)$ is called the identity soft neutrosophic biloop over $B = (\langle B_1 \cup I \rangle \cup B_2, *_1, *_2)$ if $F(a) = \{e_1, e_2\}$ for all $a \in A$, where $e_1, e_2$ are the identities of $B = (\langle B_1 \cup I \rangle \cup B_2, *_1, *_2)$ respectively.

**Definition 4.2.4:** Let $(F,A)$ be a soft neutrosophic biloop over $B = (\langle B_1 \cup I \rangle \cup B_2, *_1, *_2)$. Then $(F,A)$ is called an absolute-soft neutrosophic biloop over $B = (\langle B_1 \cup I \rangle \cup B_2, *_1, *_2)$ if $F(a) = (\langle B_1 \cup I \rangle \cup B_2, *_1, *_2)$ for all $a \in A$.

**Definition 4.2.5:** Let $(F,A)$ and $(H,C)$ be two soft neutrosophic biloops over $B = (\langle B_1 \cup I \rangle \cup B_2, *_1, *_2)$. Then $(H,C)$ is called soft neutrosophic subbiloop of $(F,A)$, if

1. $C \subseteq A$.
2. $H(a)$ is a neutrosophic subbiloop of $F(a)$ for all $a \in A$.

**Example 4.2.3:** Let $B = (B_1 \cup B_2, *_1, *_2)$ be a neutrosophic biloop, where $B_1 = \langle L_5(3) \cup I \rangle = \{e, 1, 2, 3, 4, 5, eI, 2I, 3I, 4I, 5I\}$ be a new class of neutrosophic loop and $B_2 = \{g : g^{12} = e\}$ is a group. Let $A = \{a_1, a_2, a_3, a_4\}$ be a set of parameters. Then $(F,A)$ is soft neutrosophic biloop over $B$, where

$$F(a_1) = \{e, eI, 1, 1I\} \cup \{e, g^6\},$$
$$F(a_2) = \{e, eI, 2, 2I\} \cup \{e, g^2, g^4, g^6, g^8, g^{10}\},$$
$$F(a_3) = \{e, eI, 3, 3I\} \cup \{e, g^3, g^6, g^6\},$$
$$F(a_4) = \{e, eI, 4, 4I\} \cup \{e, g^4, g^8\}.$$

Then $(H,C)$ is soft neutrosophic subbiloop of $(F,A)$, where

$$H(a_1) = \{e, 2\} \cup \{e, g^2\},$$
$$H(a_2) = \{e, eI, 3, 3I\} \cup \{e, g^6\}.$$

**Definition 4.2.6:** Let $(\langle B \cup I \rangle, *_1, *_2)$ be a neutrosophic biloop and $(F,A)$ be a soft set over $(\langle B \cup I \rangle, *_1, *_2)$. Then $(F,A)$ is called soft Lagrange neutrosophic biloop if and only if $F(a)$ is Lagrange neutrosophic subbiloop of $(\langle B \cup I \rangle, *_1, *_2)$ for all $a \in A$.

We furthere explain this fact in the following example.

**Example 4.2.4:** Let $B = (B_1 \cup B_2, *_1, *_2)$ be a neutrosophic biloop of order $20$, where $B_1 = \langle L_5(3) \cup I \rangle$ and $B_2 = \{g : g^8 = e\}$. Then clearly $(F,A)$ is a soft Lagrange soft neutrosophic biloop over $(\langle B \cup I \rangle, *_1, *_2)$, where

$$F(a_1) = \{e, eI, 2, 2I\} \cup \{e\},$$
$$F(a_2) = \{e, eI, 3, 3I\} \cup \{e\}.$$

**Theorem 4.2.4:** Every soft Lagrange neutrosophic biloop over $B = (\langle B_1 \cup I \rangle \cup B_2, *_1, *_2)$ is a soft neutrosophic biloop but the converse is not true.

One can easily check it by the help of example.

**Remark 4.2.3:** Let $(F,A)$ and $(K,C)$ be two soft Lagrange neutrosophic biloops over $B = (\langle B_1 \cup I \rangle \cup B_2, *_1, *_2)$. Then

1. Their extended intersection $(F,A) \cap_E (K,C)$ is not a soft Lagrange neutrosophic biloop over $B = (\langle B_1 \cup I \rangle \cup B_2, *_1, *_2)$.
2. Their restricted intersection $(F,A) \cap_R (K,C)$ is not a soft Lagrange neutrosophic biloop over $B = (\langle B_1 \cup I \rangle \cup B_2, *_1, *_2)$.
3. Their AND operation $(F,A) \wedge (K,C)$ is not a soft Lagrange neutrosophic biloop over $B = (\langle B_1 \cup I \rangle \cup B_2, *_1, *_2)$.
4. Their extended union $(F,A) \cup_E (K,C)$ is not a soft Lagrnage neutrosophic biloop over $B = (\langle B_1 \cup I \rangle \cup B_2, *_1, *_2)$.
5. Their restricted union $(F,A) \cup_R (K,C)$ is not a soft Lagrange neutrosophic biloop over $B = (\langle B_1 \cup I \rangle \cup B_2, *_1, *_2)$.
6. Their OR operation $(F,A) \vee (K,C)$ is not a soft Lagrange neutrosophic biloop over $B = (\langle B_1 \cup I \rangle \cup B_2, *_1, *_2)$.

One can easily verify $(1),(2),(3),(4),(5)$ and $(6)$ by the help of examples.

**Definition 4.2.6:** Let $(\langle B \cup I \rangle, *_1, *_2)$ be a neutrosophic biloop and $(F,A)$ be a soft set over $(\langle B \cup I \rangle, *_1, *_2)$. Then $(F,A)$ is called soft weakly Lagrange neutrosophic biloop if atleast one $F(a)$ is not a Lagrange neutrosophic subbiloop of $(\langle B \cup I \rangle, *_1, *_2)$ for some $a \in A$.

**Example 4.2.5:** Let $B = (B_1 \cup B_2, *_1, *_2)$ be a neutrosophic biloop of order $20$, where $B_1 = \langle L_5(3) \cup I \rangle$ and $B_2 = \{g : g^8 = e\}$. Then clearly $(F,A)$ is a soft weakly Lagrange neutrosophic biloop over $(\langle B \cup I \rangle, *_1, *_2)$, where

$$F(a_1) = \{e, eI, 2, 2I\} \cup \{e\},$$
$$F(a_2) = \{e, eI, 3, 3I\} \cup \{e, g^4\}.$$

**Theorem 4.2.5:** Every soft weakly Lagrange neutrosophic biloop over $B = (\langle B_1 \cup I \rangle \cup B_2, *_1, *_2)$ is a soft neutrosophic biloop but the converse is not true.

**Theorem 4.2.6:** If $B = (\langle B_1 \cup I \rangle \cup B_2, *_1, *_2)$ is a weakly Lagrange neutrosophic biloop, then $(F, A)$ over $B$ is also soft weakly Lagrange neutrosophic biloop but the converse is not holds.

**Remark 4.2.4:** Let $(F, A)$ and $(K, C)$ be two soft weakly Lagrange neutrosophic biloops over $B = (\langle B_1 \cup I \rangle \cup B_2, *_1, *_2)$. Then

1. Their extended intersection $(F, A) \cap_E (K, C)$ is not a soft weakly Lagrange neutrosophic biloop over $B = (\langle B_1 \cup I \rangle \cup B_2, *_1, *_2)$.
2. Their restricted intersection $(F, A) \cap_R (K, C)$ is not a soft weakly Lagrange neutrosophic biloop over $B = (\langle B_1 \cup I \rangle \cup B_2, *_1, *_2)$.
3. Their AND operation $(F, A) \wedge (K, C)$ is not a soft weakly Lagrange neutrosophic biloop over $B = (\langle B_1 \cup I \rangle \cup B_2, *_1, *_2)$.
4. Their extended union $(F, A) \cup_E (K, C)$ is not a soft weakly Lagrnage neutrosophic biloop over $B = (\langle B_1 \cup I \rangle \cup B_2, *_1, *_2)$.
5. Their restricted union $(F, A) \cup_R (K, C)$ is not a soft weakly Lagrange neutrosophic biloop over $B = (\langle B_1 \cup I \rangle \cup B_2, *_1, *_2)$.
6. Their OR operation $(F, A) \vee (K, C)$ is not a soft weakly Lagrange neutrosophic biloop over $B = (\langle B_1 \cup I \rangle \cup B_2, *_1, *_2)$.

One can easily verify $(1), (2), (3), (4), (5)$ and $(6)$ by the help of examples.

**Definition 4.2.7:** Let $(\langle B \cup I \rangle, *_1, *_2)$ be a neutrosophic biloop and $(F, A)$ be a soft set over $(\langle B \cup I \rangle, *_1, *_2)$. Then $(F, A)$ is called soft Lagrange free neutrosophic biloop if and only if $F(a)$ is not a Lagrange neutrosophic subbiloop of $(\langle B \cup I \rangle, *_1, *_2)$ for all $a \in A$.

**Example 4.2.6:** Let $B = (B_1 \cup B_2, *_1, *_2)$ be a neutrosophic biloop of order $20$, where $B_1 = \langle L_5(3) \cup I \rangle$ and $B_2 = \{g : g^8 = e\}$. Then clearly $(F, A)$ is a soft Lagrange free neutrosophic biloop over $(\langle B \cup I \rangle, *_1, *_2)$, where

$$F(a_1) = \{e, eI, 2, 2I\} \cup \{e, g^2, g^4, g^6\},$$
$$F(a_2) = \{e, eI, 3, 3I\} \cup \{e, g^4\}.$$

**Theorem 4.2.7:** Every soft Lagrange free neutrosophic biloop over $B = (\langle B_1 \cup I \rangle \cup B_2, *_1, *_2)$ is a soft neutrosophic biloop but the converse is not true.

One can easily check it by the help of example.

**Theorem 4.2.8:** If $B = (\langle B_1 \cup I \rangle \cup B_2, *_1, *_2)$ is a Lagrange free neutrosophic biloop, then $(F, A)$ over $B$ is also soft Lagrange free neutrosophic biloop but the converse does not holds.

One can easily check it by the help of example.

**Remark 4.2.5:** Let $(F,A)$ and $(K,C)$ be two soft Lagrange free neutrosophic biloops over $B = (\langle B_1 \cup I\rangle \cup B_2, *_1, *_2)$. Then

1. Their extended intersection $(F,A) \cap_E (K,C)$ is not a soft Lagrange free neutrosophic biloop over $B = (\langle B_1 \cup I\rangle \cup B_2, *_1, *_2)$.
2. Their restricted intersection $(F,A) \cap_R (K,C)$ is not a soft Lagrange free neutrosophic biloop over $B = (\langle B_1 \cup I\rangle \cup B_2, *_1, *_2)$.
3. Their $AND$ operation $(F,A) \wedge (K,C)$ is not a soft Lagrange free neutrosophic biloop over $B = (\langle B_1 \cup I\rangle \cup B_2, *_1, *_2)$.
4. Their extended union $(F,A) \cup_E (K,C)$ is not a soft Lagrnage free neutrosophic biloop over $B = (\langle B_1 \cup I\rangle \cup B_2, *_1, *_2)$.
5. Their restricted union $(F,A) \cup_R (K,C)$ is not a soft Lagrange free neutrosophic biloop over $B = (\langle B_1 \cup I\rangle \cup B_2, *_1, *_2)$.
6. Their $OR$ operation $(F,A) \vee (K,C)$ is not a soft Lagrange free neutrosophic biloop over $B = (\langle B_1 \cup I\rangle \cup B_2, *_1, *_2)$.

One can easily verify $(1),(2),(3),(4),(5)$ and $(6)$ by the help of examples.

We now proceed to define the strong part of soft neutrosophic biloops over a neutosophic biloop.

**Soft Neutrosophic Strong Biloop**

As usual, the theory of purely neutrosophics is alos exist in soft neutrosophic biloops and so we defined soft neutrosophic strong biloop over a neutrosophic biloop here and establish their related properties and characteristics.

**Definition 4.2.8:** Let $B = (B_1 \cup B_2, *_1, *_2)$ be a neutrosophic biloop where $B_1$ is a neutrosopphic biloop and $B_2$ is a neutrosophic group and $(F,A)$ be soft set over $B$. Then $(F,A)$ over $B$ is called soft neutrosophic strong biloop if and only if $F(a)$ is a neutrosopchic strong subbiloop of $B$ for all $a \in A$.

**Example 4.2.7:** Let $B = (B_1 \cup B_2, *_1, *_2)$ where $B_1 = \langle L_5(2) \cup I \rangle$ is a neutrosophic loop and $B_2 = \{0,1,2,3,4,,1I,2I,3I,4I\}$ under multiplication modulo 5 is a neutrosophic group. Let $A = \{a_1, a_2\}$ be a set of parameters. Then $(F, A)$ is soft neutrosophic strong biloop over $B$, where

$$F(a_1) = \{e, 2, eI, 2I\} \cup \{1, I, 4I\},$$
$$F(a_2) = \{e, 3, eI, 3I\} \cup \{1, I, 4I\}.$$

**Theorem 4.2.9:** Every soft neutrosophic strong biloop over $B = (B_1 \cup B_2, *_1, *_2)$ is a soft neutrosophic biloop but the converse is not true.

One can easily check it by the help of example.

**Theorem 4.2.10:** If $B = (B_1 \cup B_2, *_1, *_2)$ is a neutrosophic strong biloop, then $(F, A)$ over $B$ is also soft neutrosophic strong biloop but the converse is not holds.

One can easily check it by the help of example.

**Proposition 4..2.3:** Let $(F,A)$ and $(K,C)$ be two soft neutrosophic strong biloops over $B = (B_1 \cup B_2, *_1, *_2)$. Then

1. Their extended intersection $(F,A) \cap_E (K,C)$ is a soft neutrosophic strong biloop over $B = (B_1 \cup B_2, *_1, *_2)$.
2. Their restricted intersection $(F,A) \cap_R (K,C)$ is a soft neutrosophic strong biloop over $B = (B_1 \cup B_2, *_1, *_2)$.
3. Their AND operation $(F,A) \wedge (K,C)$ is a soft neutrosophic strong biloop over $B = (B_1 \cup B_2, *_1, *_2)$.

**Remark 4.2.6:** Let $(F,A)$ and $(K,B)$ be two soft neutrosophic strong biloops over $B = (B_1 \cup B_2, *_1, *_2)$. Then

1. Their extended union $(F,A) \cup_E (K,C)$ is not a soft neutrosophic strong biloop over $B = (B_1 \cup B_2, *_1, *_2)$.
2. Their restricted union $(F,A) \cup_R (K,C)$ is not a soft neutrosophic strong biloop over $B = (B_1 \cup B_2, *_1, *_2)$.
3. Their OR operation $(F,A) \vee (K,C)$ is not a soft neutrosophic strong biloop over $B = (B_1 \cup B_2, *_1, *_2)$.

One can easily verify $(1), (2),$ and $(3)$ by the help of examples.

**Definition 4.2.9:** Let $(F,A)$ and $(H,C)$ be two soft neutrosophic strong biloops over $B = (B_1 \cup B_2, *_1, *_2)$. Then $(H,C)$ is called soft neutrosophic strong subbiloop of $(F,A)$, if

1. $C \subseteq A$.
2. $H(a)$ is a neutrosophic strong subbiloop of $F(a)$ for all $a \in A$.

**Definition 4.2.10:** Let $B = (B_1 \cup B_2, *_1, *_2)$ be a neutrosophic biloop and $(F, A)$ be a soft set over $B = (B_1 \cup B_2, *_1, *_2)$. Then $(F, A)$ is called soft Lagrange neutrosophic strong biloop if and only if $F(a)$ is a Lagrange neutrosophic strong subbiloop of $B = (B_1 \cup B_2, *_1, *_2)$ for all $a \in A$.

**Theorem 4.2.11:** Every soft Lagrange neutrosophic strong biloop over $B = (B_1 \cup B_2, *_1, *_2)$ is a soft neutrosophic biloop but the converse is not true.

One can easily check it by the help of example.

**Remark 4.2.7:** Let $(F, A)$ and $(K, C)$ be two soft Lagrange neutrosophic strong biloops over $B = (B_1 \cup B_2, *_1, *_2)$. Then

1. Their extended intersection $(F, A) \cap_E (K, C)$ is not a soft Lagrange neutrosophic strong biloop over $B = (B_1 \cup B_2, *_1, *_2)$.
2. Their restricted intersection $(F, A) \cap_R (K, C)$ is not a soft Lagrange neutrosophic strong biloop over $B = (B_1 \cup B_2, *_1, *_2)$.
3. Their $AND$ operation $(F, A) \wedge (K, C)$ is not a soft Lagrange neutrosophic strong biloop over $B = (B_1 \cup B_2, *_1, *_2)$.
4. Their extended union $(F, A) \cup_E (K, C)$ is not a soft Lagrnage neutrosophic strong biloop over $B = (B_1 \cup B_2, *_1, *_2)$.
5. Their restricted union $(F, A) \cup_R (K, C)$ is not a soft Lagrange neutrosophic strong biloop over $B = (B_1 \cup B_2, *_1, *_2)$.
6. Their $OR$ operation $(F, A) \vee (K, C)$ is not a soft Lagrange neutrosophic strong biloop over $B = (B_1 \cup B_2, *_1, *_2)$.

One can easily verify $(1), (2), (3), (4), (5)$ and $(6)$ by the help of examples.

**Definition 4.2.11:** Let $B = (B_1 \cup B_2, *_1, *_2)$ be a neutrosophic biloop and $(F, A)$ be a soft set over $B = (B_1 \cup B_2, *_1, *_2)$. Then $(F, A)$ is called soft weakly Lagrange neutrosophic strong biloop if atleast one $F(a)$ is not a Lagrange neutrosophic strong subbiloop of $B = (B_1 \cup B_2, *_1, *_2)$ for some $a \in A$.

**Theorem 4.2.12:** Every soft weakly Lagrange neutrosophic strong biloop over $B = (B_1 \cup B_2, *_1, *_2)$ is a soft neutrosophic biloop but the converse is not true.

**Theorem 4.2.13:** If $B = (B_1 \cup B_2, *_1, *_2)$ is a weakly Lagrange neutrosophic strong biloop, then $(F, A)$ over $B$ is also soft weakly Lagrange neutrosophic strong biloop but the converse does not holds.

**Remark 4.2.8:** Let $(F, A)$ and $(K, C)$ be two soft weakly Lagrange neutrosophic strong biloops over $B = (B_1 \cup B_2, *_1, *_2)$. Then

1. Their extended intersection $(F, A) \cap_E (K, C)$ is not a soft weakly Lagrange neutrosophic strong biloop over $B = (B_1 \cup B_2, *_1, *_2)$.
2. Their restricted intersection $(F, A) \cap_R (K, C)$ is not a soft weakly Lagrange neutrosophic strong biloop over $B = (B_1 \cup B_2, *_1, *_2)$.
3. Their *AND* operation $(F, A) \wedge (K, C)$ is not a soft weakly Lagrange neutrosophic strong biloop over $B = (B_1 \cup B_2, *_1, *_2)$.
4. Their extended union $(F, A) \cup_E (K, C)$ is not a soft weakly Lagrange neutrosophic strong biloop over $B = (B_1 \cup B_2, *_1, *_2)$.
5. Their restricted union $(F, A) \cup_R (K, C)$ is not a soft weakly Lagrange neutrosophic strong biloop over $B = (B_1 \cup B_2, *_1, *_2)$.
6. Their *OR* operation $(F, A) \vee (K, C)$ is not a soft weakly Lagrange neutrosophic strong biloop over $B = (B_1 \cup B_2, *_1, *_2)$.

One can easily verify $(1),(2),(3),(4),(5)$ and $(6)$ by the help of examples.

**Definition 4.2.12:** Let $B = (B_1 \cup B_2, *_1, *_2)$ be a neutrosophic biloop and $(F,A)$ be a soft set over $B = (B_1 \cup B_2, *_1, *_2)$. Then $(F,A)$ is called soft Lagrange free neutrosophic strong biloop if and only if $F(a)$ is not a Lagrange neutrosophic subbiloop of $B = (B_1 \cup B_2, *_1, *_2)$ for all $a \in A$.

**Theorem 4.2.13:** Every soft Lagrange free neutrosophic strong biloop over $B = (B_1 \cup B_2, *_1, *_2)$ is a soft neutrosophic biloop but the converse is not true.

**Theorem 4.2.14:** If $B = (B_1 \cup B_2, *_1, *_2)$ is a Lagrange free neutrosophic strong biloop, then $(F,A)$ over $B$ is also soft strong lagrange free neutrosophic strong biloop but the converse is not true.

**Remark 4.2.9:** Let $(F,A)$ and $(K,C)$ be two soft Lagrange free neutrosophic strong biloops over $B = (B_1 \cup B_2, *_1, *_2)$. Then

1. Their extended intersection $(F,A) \cap_E (K,C)$ is not a soft Lagrange free neutrosophic strong biloop over $B = (B_1 \cup B_2, *_1, *_2)$.
2. Their restricted intersection $(F,A) \cap_R (K,C)$ is not a soft Lagrange free neutrosophic strong biloop over $B = (B_1 \cup B_2, *_1, *_2)$.
3. Their $AND$ operation $(F,A) \wedge (K,C)$ is not a soft Lagrange free neutrosophic strong biloop over $B = (B_1 \cup B_2, *_1, *_2)$.
4. Their extended union $(F,A) \cup_E (K,C)$ is not a soft Lagrnage free neutrosophic strong biloop over $B = (B_1 \cup B_2, *_1, *_2)$.
5. Their restricted union $(F,A) \cup_R (K,C)$ is not a soft Lagrange free neutrosophic strong biloop over $B = (B_1 \cup B_2, *_1, *_2)$.
6. Their $OR$ operation $(F,A) \vee (K,C)$ is not a soft Lagrange free neutrosophic strong biloop over $B = (B_1 \cup B_2, *_1, *_2)$.

One can easily verify (1),(2),(3),(4),(5) and (6) by the help of examples.

## 4.3 Soft Neutrosophic N-loop

In this section, we extend soft sets to neutrosophic N-loops and introduce soft neutrosophic N-loops. This is the generalization of soft neutrosophic loops. Some of their impotant facts and figures are also presented here with illustrative examples. We also initiated the strong part of neutrosophy in this section. Now we proceed onto define soft neutrosophic N-loop as follows.

**Definition 4.3.1:** Let $S(B) = \{S(B_1) \cup S(B_2) \cup ... \cup S(B_N), *_1, *_2, ..., *_N\}$ be a neutrosophic $N$-loop and $(F, A)$ be a soft set over $S(B)$. Then $(F, A)$ is called soft neutrosophic $N$-loop if and only if $F(a)$ is a neutrosopchic sub $N$-loop of $S(B)$ for all $a \in A$.

**Example 4.3.1:** Let $S(B) = \{S(B_1) \cup S(B_2) \cup S(B_3), *_1, *_2, *_3\}$ be a neutrosophic $3$-loop, where $S(B_1) = \langle L_5(3) \cup I \rangle$, $S(B_2) = \{g : g^{12} = e\}$ and $S(B_3) = S_3$. Then $(F, A)$ is sof neutrosophic $N$-loop over $S(B)$, where
$$F(a_1) = \{e, eI, 2, 2I\} \cup \{e, g^6\} \cup \{e, (12)\},$$
$$F(a_2) = \{e, eI, 3, 3I\} \cup \{e, g^4, g^8\} \cup \{e, (13)\}.$$

**Theorem 4.3.1:** Let $(F, A)$ and $(H, A)$ be two soft neutrosophic $N$-loops over $S(B) = \{S(B_1) \cup S(B_2) \cup ... \cup S(B_N), *_1, *_2, ..., *_N\}$. Then their intersection $(F, A) \cap (H, A)$ is again a soft neutrosophic $N$-loop over $S(B)$.

**Theorem 4.3.2:** Let $(F,A)$ and $(H,C)$ be two soft neutrosophic $N$-loops over $S(B) = \{S(B_1) \cup S(B_2) \cup ... \cup S(B_N), *_1, *_2, ..., *_N\}$ such that $A \cap C = \phi$. Then their union is soft neutrosophic $N$-loop over $S(B)$.

**Proof.** Straightforward.

**Proposition 4.3.1:** Let $(F,A)$ and $(K,C)$ be two soft neutrosophic $N$-loops over $S(B) = \{S(B_1) \cup S(B_2) \cup ... \cup S(B_N), *_1, *_2, ..., *_N\}$. Then

1. Their extended intersection $(F,A) \cap_E (K,C)$ is a soft neutrosophic $N$-loop over $S(B)$.
2. Their restricted intersection $(F,A) \cap_R (K,C)$ is a soft neutrosophic $N$-loop over $S(B)$.
3. Their $AND$ operation $(F,A) \wedge (K,C)$ is a soft neutrosophic $N$-loop over $S(B)$.

**Remark 4.3.1:** Let $(F,A)$ and $(H,C)$ be two soft neutrosophic $N$-loops over $S(B) = \{S(B_1) \cup S(B_2) \cup ... \cup S(B_N), *_1, *_2, ..., *_N\}$. Then

1. Their extended union $(F,A) \cup_E (K,C)$ is not a soft neutrosophic $N$-loop over $S(B)$.
2. Their restricted union $(F,A) \cup_R (K,C)$ is not a soft neutrosophic $N$-loop over $S(B)$.
3. Their $OR$ operation $(F,A) \vee (K,C)$ is not a soft neutrosophic $N$-loop over $S(B)$.

One can easily verify $(1), (2),$ and $(3)$ by the help of examples.

**Definition 4.3.2:** Let $(F,A)$ be a soft neutrosophic $N$-loop over $S(B) = \{S(B_1) \cup S(B_2) \cup ... \cup S(B_N), *_1, *_2,...,*_N\}$. Then $(F,A)$ is called the identity soft neutrosophic $N$-loop over $S(B)$ if $F(a) = \{e_1, e_2,...,e_N\}$ for all $a \in A$, where $e_1, e_2,...,e_N$ are the identities element of $S(B_1), S(B_2),..., S(B_N)$ respectively.

**Definition 4.3.3:** Let $(F,A)$ be a soft neutrosophic $N$-loop over $S(B) = \{S(B_1) \cup S(B_2) \cup ... \cup S(B_N), *_1, *_2,...,*_N\}$. Then $(F,A)$ is called an absolute-soft neutrosophic $N$-loop over $S(B)$ if $F(a) = S(B)$ for all $a \in A$.

**Definition 4.3.4:** Let $(F,A)$ and $(H,C)$ be two soft neutrosophic $N$-loops over $S(B) = \{S(B_1) \cup S(B_2) \cup ... \cup S(B_N), *_1, *_2,...,*_N\}$. Then $(H,C)$ is called soft neutrosophic sub $N$-loop of $(F,A)$, if

1. $C \subseteq A$.
2. $H(a)$ is a neutrosophic sub $N$-loop of $F(a)$ for all $a \in A$.

**Definition 4.3.5:** Let $S(B) = \{S(B_1) \cup S(B_2) \cup ... \cup S(B_N), *_1, *_2,...,*_N\}$ be a neutrosophic $N$-loop and $(F,A)$ be a soft set over $S(B)$. Then $(F,A)$ is called soft Lagrange neutrosophic $N$-loop if and only if $F(a)$ is Lagrange neutrosophic sub $N$-loop of $S(B)$ for all $a \in A$.

**Theorem 4.3.3:** Every soft Lagrange neutrosophic $N$-loop over $S(B) = \{S(B_1) \cup S(B_2) \cup ... \cup S(B_N), *_1, *_2,...,*_N\}$ is a soft neutrosophic $N$-loop but the converse is not true.

One can easily check it by the help of example.

**Remark 4.3.2:** Let $(F,A)$ and $(K,C)$ be two soft Lagrange neutrosophic $N$-loops over $S(B) = \{S(B_1) \cup S(B_2) \cup ... \cup S(B_N), *_1, *_2, ..., *_N\}$. Then

1. Their extended intersection $(F,A) \cap_E (K,C)$ is not a soft Lagrange neutrosophic $N$-loop over $S(B)$.
2. Their restricted intersection $(F,A) \cap_R (K,C)$ is not a soft Lagrange neutrosophic $N$-loop over $S(B)$.
3. Their $AND$ operation $(F,A) \wedge (K,C)$ is not a soft Lagrange neutrosophic $N$-loop over $(B)$.
4. Their extended union $(F,A) \cup_E (K,C)$ is not a soft Lagrnage neutrosophic $N$-loop over $S(B)$.
5. Their restricted union $(F,A) \cup_R (K,C)$ is not a soft Lagrange neutrosophic $N$-loop over $S(B)$.
6. Their $OR$ operation $(F,A) \vee (K,C)$ is not a soft Lagrange neutrosophic $N$-loop over $S(B)$.

One can easily verify $(1),(2),(3),(4),(5)$ and $(6)$ by the help of examples.

**Definition 4.3.6:** Let $S(B) = \{S(B_1) \cup S(B_2) \cup ... \cup S(B_N), *_1, *_2, ..., *_N\}$ be a neutrosophic $N$-loop and $(F,A)$ be a soft set over $S(B)$. Then $(F,A)$ is called soft weakly Lagrange neutrosophic biloop if atleast one $F(a)$ is not a Lagrange neutrosophic sub $N$-loop of $S(B)$ for some $a \in A$.

**Theorem 4.3.4:** Every soft weakly Lagrange neutrosophic $N$-loop over $S(B) = \{S(B_1) \cup S(B_2) \cup ... \cup S(B_N), *_1, *_2, ..., *_N\}$ is a soft neutrosophic $N$-loop but the converse is not true.

**Theorem 4.3.5:** If $S(B) = \{S(B_1) \cup S(B_2) \cup ... \cup S(B_N), *_1, *_2, ..., *_N\}$ is a weakly Lagrange neutrosophic $N$-loop, then $(F, A)$ over $S(B)$ is also soft weakly Lagrange neutrosophic $N$-loop but the converse is not holds.

**Remark 4.3.3:** Let $(F, A)$ and $(K, C)$ be two soft weakly Lagrange neutrosophic $N$-loops over $S(B) = \{S(B_1) \cup S(B_2) \cup ... \cup S(B_N), *_1, *_2, ..., *_N\}$. Then

1. Their extended intersection $(F, A) \cap_E (K, C)$ is not a soft weakly Lagrange neutrosophic $N$-loop over $S(B)$.
2. Their restricted intersection $(F, A) \cap_R (K, C)$ is not a soft weakly Lagrange neutrosophic $N$-loop over $S(B)$.
3. Their *AND* operation $(F, A) \wedge (K, C)$ is not a soft weakly Lagrange neutrosophic $N$-loop over $S(B)$.
4. Their extended union $(F, A) \cup_E (K, C)$ is not a soft weakly Lagrnage neutrosophic $N$-loop over $S(B)$.
5. Their restricted union $(F, A) \cup_R (K, C)$ is not a soft weakly Lagrange neutrosophic $N$-loop over $S(B)$.
6. Their *OR* operation $(F, A) \vee (K, C)$ is not a soft weakly Lagrange neutrosophic $N$-loop over $S(B)$.

One can easily verify $(1), (2), (3), (4), (5)$ and $(6)$ by the help of examples.

**Definition 4.3.7:** Let $S(B) = \{S(B_1) \cup S(B_2) \cup ... \cup S(B_N), *_1, *_2, ..., *_N\}$ be a neutrosophic $N$-loop and $(F, A)$ be a soft set over $S(B)$. Then $(F, A)$ is called soft Lagrange free neutrosophic $N$-loop if and only if $F(a)$ is not a Lagrange neutrosophic sub $N$-loop of $S(B)$ for all $a \in A$.

**Theorem 4.3.6:** Every soft Lagrange free neutrosophic $N$-loop over $S(B) = \{S(B_1) \cup S(B_2) \cup ... \cup S(B_N), *_1, *_2, ..., *_N\}$ is a soft neutrosophic biloop but the converse is not true.

**Theorem 4.3.7:** If $S(B) = \{S(B_1) \cup S(B_2) \cup ... \cup S(B_N), *_1, *_2, ..., *_N\}$ is a Lagrange free neutrosophic $N$-loop, then $(F,A)$ over $S(B)$ is also soft lagrange free neutrosophic $N$-loop but the converse is not hold.

**Remark 4.3.4:** Let $(F,A)$ and $(K,C)$ be two soft Lagrange free neutrosophic $N$-loops over $S(B) = \{S(B_1) \cup S(B_2) \cup ... \cup S(B_N), *_1, *_2, ..., *_N\}$. Then

1. Their extended intersection $(F,A) \cap_E (K,C)$ is not a soft Lagrange free neutrosophic $N$-loop over $B = (\langle B_1 \cup I \rangle \cup B_2, *_1, *_2)$.
2. Their restricted intersection $(F,A) \cap_R (K,C)$ is not a soft Lagrange free neutrosophic $N$-loop over $S(B)$.
3. Their $AND$ operation $(F,A) \wedge (K,C)$ is not a soft Lagrange free neutrosophic $N$-loop over $S(B)$.
4. Their extended union $(F,A) \cup_E (K,C)$ is not a soft Lagrnage free neutrosophic $N$-loop over $S(B)$.
5. Their restricted union $(F,A) \cup_R (K,C)$ is not a soft Lagrange free neutrosophic $N$-loop over $S(B)$.
6. Their $OR$ operation $(F,A) \vee (K,C)$ is not a soft Lagrange free neutrosophic $N$-loop over $S(B)$.

One can easily verify $(1),(2),(3),(4),(5)$ and $(6)$ by the help of examples.

## Soft Neutrosophic Strong N-loop

The notions of soft neutrosophic strong N-loops over neutrosophic N-loops are introduced here. We give some basic definitions of soft neutrosophic strong N-loops and illustrated it with the help of exmaples and give some basic results.

**Definition 4.3.8:** Let $\langle L \cup I \rangle = \{L_1 \cup L_2 \cup ... \cup L_N, *_1, *_2, ..., *_N\}$ be a neutrosophic $N$-loop and $(F, A)$ be a soft set over $\langle L \cup I \rangle = \{L_1 \cup L_2 \cup ... \cup L_N, *_1, *_2, ..., *_N\}$. Then $(F, A)$ is called soft neutrosophic strong $N$-loop if and only if $F(a)$ is a neutrosopchic strong sub $N$-loop of $\langle L \cup I \rangle = \{L_1 \cup L_2 \cup ... \cup L_N, *_1, *_2, ..., *_N\}$ for all $a \in A$.

**Example 4.3.2:** Let $\langle L \cup I \rangle = \{L_1 \cup L_2 \cup L_3, *_1, *_2, *_3\}$ where $L_1 = \langle L_5(3) \cup I \rangle, L_2 = \langle L_7(3) \cup I \rangle$ and $L_3 = \{1, 2, 1I, 2I\}$. Then $(F, A)$ is a soft neutrosophic strong $N$-loop over $\langle L \cup I \rangle$, where

$$F(a_1) = \{e, 2, eI, 2I\} \cup \{e, 2, eI, 2I\} \cup \{1, I\},$$
$$F(a_2) = \{e, 3, eI, 3I\} \cup \{e, 3, eI, 3I\} \cup \{1, 2, 2I\}.$$

**Theorem 4.3.8:** All soft neutrosophic strong $N$-loops are soft neutrosophic $N$-loops but the converse is not true.

One can easily see the converse with the help of example.

**Proposition 4.3.2:** Let $(F,A)$ and $(K,C)$ be two soft neutrosophic strong $N$-loops over $\langle L \cup I \rangle = \{L_1 \cup L_2 \cup ... \cup L_N, *_1, *_2, ..., *_N\}$. Then

1. Their extended intersection $(F,A) \cap_E (K,C)$ is a soft neutrosophic strong $N$-loop over $\langle L \cup I \rangle$.
2. Their restricted intersection $(F,A) \cap_R (K,C)$ is a soft neutrosophic strong $N$-loop over $\langle L \cup I \rangle$.
3. Their $AND$ operation $(F,A) \wedge (K,C)$ is a soft neutrosophic strong $N$-loop over $\langle L \cup I \rangle$.

**Remark 4.3.5:** Let $(F,A)$ and $(K,C)$ be two soft neutrosophic strong $N$-loops over $\langle L \cup I \rangle = \{L_1 \cup L_2 \cup ... \cup L_N, *_1, *_2, ..., *_N\}$. Then

1. Their extended union $(F,A) \cup_E (K,C)$ is not a soft neutrosophic strong $N$-loop over $\langle L \cup I \rangle$.
2. Their restricted union $(F,A) \cup_R (K,C)$ is not a soft neutrosophic strong $N$-loop over $\langle L \cup I \rangle$.
3. Their $OR$ operation $(F,A) \vee (K,C)$ is not a soft neutrosophic strong $N$-loop over $\langle L \cup I \rangle$.

One can easily verify $(1), (2),$ and $(3)$ by the help of examples.

**Definition 4.39:** Let $(F,A)$ and $(H,C)$ be two soft neutrosophic strong $N$-loops over $\langle L \cup I \rangle = \{L_1 \cup L_2 \cup ... \cup L_N, *_1, *_2, ..., *_N\}$. Then $(H,C)$ is called soft neutrosophic strong sub $N$-loop of $(F,A)$, if

1. $C \subseteq A$.
2. $H(a)$ is a neutrosophic strong sub $N$-loop of $F(a)$ for all $a \in A$.

**Definition 4.3.10:** Let $\langle L \cup I \rangle = \{L_1 \cup L_2 \cup ... \cup L_N, *_1, *_2, ..., *_N\}$ be a neutrosophic strong $N$-loop and $(F, A)$ be a soft set over $\langle L \cup I \rangle$. Then $(F, A)$ is called soft Lagrange neutrosophic strong $N$-loop if and only if $F(a)$ is a Lagrange neutrosophic strong sub $N$-loop of $\langle L \cup I \rangle$ for all $a \in A$.

**Theorem 4.3.9:** Every soft Lagrange neutrosophic strong $N$-loop over $\langle L \cup I \rangle = \{L_1 \cup L_2 \cup ... \cup L_N, *_1, *_2, ..., *_N\}$ is a soft neutrosophic $N$-loop but the converse is not true.

**Remark 4.3.6:** Let $(F, A)$ and $(K, C)$ be two soft Lagrange neutrosophic strong $N$-loops over $\langle L \cup I \rangle = \{L_1 \cup L_2 \cup ... \cup L_N, *_1, *_2, ..., *_N\}$. Then

1. Their extended intersection $(F, A) \cap_E (K, C)$ is not a soft Lagrange neutrosophic strong $N$-loop over $\langle L \cup I \rangle$.
2. Their restricted intersection $(F, A) \cap_R (K, C)$ is not a soft Lagrange neutrosophic strong $N$-loop over $\langle L \cup I \rangle$.
3. Their $AND$ operation $(F, A) \wedge (K, C)$ is not a soft Lagrange neutrosophic strong $N$-loop over $\langle L \cup I \rangle$.
4. Their extended union $(F, A) \cup_E (K, C)$ is not a soft Lagrnage neutrosophic strong $N$-loop over $\langle L \cup I \rangle$.
5. Their restricted union $(F, A) \cup_R (K, C)$ is not a soft Lagrange neutrosophic strong $N$-loop over $\langle L \cup I \rangle$.
6. Their $OR$ operation $(F, A) \vee (K, C)$ is not a soft Lagrange neutrosophic strong $N$-loop over $\langle L \cup I \rangle$.

One can easily verify $(1), (2), (3), (4), (5)$ and $(6)$ by the help of examples.

**Definition 4.3.11:** Let $\langle L \cup I \rangle = \{L_1 \cup L_2 \cup ... \cup L_N, *_1, *_2, ..., *_N\}$ be a neutrosophic strong $N$-loop and $(F, A)$ be a soft set over $\langle L \cup I \rangle$. Then $(F, A)$ is called soft weakly Lagrange neutrosophic strong $N$-loop if atleast one $F(a)$ is not a Lagrange neutrosophic strong sub $N$-loop of $\langle L \cup I \rangle$ for some $a \in A$.

**Theorem 4.3.10:** Every soft weakly Lagrange neutrosophic strong $N$-loop over $\langle L \cup I \rangle = \{L_1 \cup L_2 \cup ... \cup L_N, *_1, *_2, ..., *_N\}$ is a soft neutrosophic $N$-loop but the converse is not true.

**Theorem 4.3.11:** If $\langle L \cup I \rangle = \{L_1 \cup L_2 \cup ... \cup L_N, *_1, *_2, ..., *_N\}$ is a weakly Lagrange neutrosophic strong $N$-loop, then $(F, A)$ over $\langle L \cup I \rangle$ is also a soft weakly Lagrange neutrosophic strong $N$-loop but the converse is not true.

**Remark 4.3.7:** Let $(F, A)$ and $(K, C)$ be two soft weakly Lagrange neutrosophic strong $N$-loops over $\langle L \cup I \rangle = \{L_1 \cup L_2 \cup ... \cup L_N, *_1, *_2, ..., *_N\}$. Then

1. Their extended intersection $(F, A) \cap_E (K, C)$ is not a soft weakly Lagrange neutrosophic strong $N$-loop over $\langle L \cup I \rangle$.
2. Their restricted intersection $(F, A) \cap_R (K, C)$ is not a soft weakly Lagrange neutrosophic strong $N$-loop over $\langle L \cup I \rangle$.
3. Their $AND$ operation $(F, A) \wedge (K, C)$ is not a soft weakly Lagrange neutrosophic strong $N$-loop over $\langle L \cup I \rangle$.
4. Their extended union $(F, A) \cup_E (K, C)$ is not a soft weakly Lagrnage neutrosophic strong $N$-loop over $\langle L \cup I \rangle$.
5. Their restricted union $(F, A) \cup_R (K, C)$ is not a soft weakly Lagrange neutrosophic strong $N$-loop over $\langle L \cup I \rangle$.
6. Their $OR$ operation $(F, A) \vee (K, C)$ is not a soft weakly Lagrange neutrosophic strong $N$-loop over $\langle L \cup I \rangle$.

**Definition 4.3.12:** Let $\langle L \cup I \rangle = \{L_1 \cup L_2 \cup ... \cup L_N, *_1, *_2, ..., *_N\}$ be a neutrosophic $N$-loop and $(F,A)$ be a soft set over $\langle L \cup I \rangle$. Then $(F,A)$ is called soft Lagrange free neutrosophic strong $N$-loop if and only if $F(a)$ is not a Lagrange neutrosophic strong sub $N$-loop of $\langle L \cup I \rangle$ for all $a \in A$.

**Theorem 4.3.12:** Every soft Lagrange free neutrosophic strong $N$-loop over $\langle L \cup I \rangle = \{L_1 \cup L_2 \cup ... \cup L_N, *_1, *_2, ..., *_N\}$ is a soft neutrosophic $N$-loop but the converse is not true.

**Theorem 4.3.12:** If $\langle L \cup I \rangle = \{L_1 \cup L_2 \cup ... \cup L_N, *_1, *_2, ..., *_N\}$ is a Lagrange free neutrosophic strong $N$-loop, then $(F,A)$ over $\langle L \cup I \rangle$ is also a soft Lagrange free neutrosophic strong $N$-loop but the converse is not true.

**Remark 4.3.8:** Let $(F,A)$ and $(K,C)$ be two soft Lagrange free neutrosophic strong $N$-loops over $\langle L \cup I \rangle = \{L_1 \cup L_2 \cup ... \cup L_N, *_1, *_2, ..., *_N\}$. Then

1. Their extended intersection $(F,A) \cap_E (K,C)$ is not a soft Lagrange free neutrosophic strong $N$-loop over $\langle L \cup I \rangle$.
2. Their restricted intersection $(F,A) \cap_R (K,C)$ is not a soft Lagrange free neutrosophic strong $N$-loop over $\langle L \cup I \rangle$.
3. Their *AND* operation $(F,A) \wedge (K,C)$ is not a soft Lagrange free neutrosophic strong $N$-loop over $\langle L \cup I \rangle$.
4. Their extended union $(F,A) \cup_E (K,C)$ is not a soft Lagrnage free neutrosophic strong $N$-loop over $\langle L \cup I \rangle$.
5. Their restricted union $(F,A) \cup_R (K,C)$ is not a soft Lagrange free neutrosophic strong $N$-loop over $\langle L \cup I \rangle$.
6. Their *OR* operation $(F,A) \vee (K,C)$ is not a soft Lagrange free neutrosophic strong $N$-loop over $\langle L \cup I \rangle$.

# Chapter Five

## SOFT NEUTROSOSPHIC LA-SEMIGROUP AND THEIR GENERALIZATION

This chapter has three sections. In first section, we introduce the important notions of soft neutrosophic LA-semigroups over neutrosophic LA-semi groups which is infact a collection of parameterized family of neutrosophic sub LA-semigroups and we also examin some of their characterization with sufficient amount of examples in this section. The second section deal with soft neutrosophic bi-LA-semigroups which are basically defined over neutrosophic bi-LA-semigroups. We also establish some basic and fundamental results of soft neutrosophic bi-LA-semigroups. The third section is about the generalization of soft neutrosophic LA-semigroups. We defined soft neutrosophic N-LA-semigroups over neutrosophic N-LA-semigroups in this section and some of their basic properties are also investigated.

### 5.1 Soft Neutrosophic LA-semigroups

The definition of soft neutrosophic LA-semigroup is introduced in this section and we also examine some of their properties. Throughout this section $N(S)$ will dnote a neutrosophic LA-semigroup unless stated otherwise.

**Definition 5.1.1:** Let $(F,A)$ be a soft set over $N(S)$. Then $(F,A)$ over $N(S)$ is called soft neutrosophic LA-semigroup if $(F,A) \odot (F,A) \subseteq (F,A)$.

**Proposition 5.1.1:** A soft set $(F,A)$ over $N(S)$ is a soft neutrosophic LA-semigroup if and only if $\phi \neq F(a)$ is a neutrosophic sub LA-semigroup of $N(S)$ for all $a \in A$.

**Proof:** Let $(F,A)$ be a soft neutrosophic LA-semigroup over $N(S)$. By definition $(F,A) \odot (F,A) \subseteq (F,A)$, and so $F(a)F(a) \subseteq F(a)$ for all $a \in A$. This means $\phi \neq F(a)$ is a neutrosophic sub LA-semigroup of $N(S)$.

Conversely suppose that $(F,A)$ is a soft set over $N(S)$ and for all $a \in A$, $F(a)$ is a neutrosophic sub LA-semigruop of $N(S)$ whenever $F(a) \neq \phi$. Since $F(a)$ is a neutrosophic sub LA-semigroup of $N(S)$, therefore $F(a)F(a) \subseteq F(a)$ for all $a \in A$ and consequently $(F,A) \odot (F,A) \subseteq (F,A)$ which completes the proof.

**Example 5.1.1:** Let $N(S) = \{1, 2, 3, 4, 1I, 2I, 3I, 4I\}$ be a neutrosophic LA-semigroup with the following table.

| * | 1 | 2 | 3 | 4 | 1I | 2I | 3I | 4I |
|---|---|---|---|---|----|----|----|-----|
| 1 | 1 | 4 | 2 | 3 | 1I | 4I | 2I | 3I |
| 2 | 3 | 2 | 4 | 1 | 3I | 2I | 4I | 1I |
| 3 | 4 | 1 | 3 | 2 | 4I | 1I | 3I | 2I |
| 4 | 2 | 3 | 1 | 4 | 2I | 3I | 1I | 4I |
| 1I | 1I | 4I | 2I | 3I | 1I | 4I | 2I | 3I |
| 2I | 3I | 2I | 4I | 1I | 3I | 2I | 4I | 1I |
| 3I | 4I | 1I | 3I | 2I | 4I | 1I | 3I | 2I |
| 4I | 2I | 3I | 1I | 4I | 2I | 3I | 1I | 4I |

Let $(F, A)$ be a soft set over $N(S)$. Then clearly $(F, A)$ is a soft neutrosophic LA-semigroup over $N(S)$, where

$$F(a_1) = \{1, 1I\}, \quad F(a_2) = \{2, 2I\},$$
$$F(a_3) = \{3, 3I\}, \quad F(a_4) = \{4, 4I\}.$$

**Theorem 5.1.1:** A soft LA-semigroup over an LA-semigroup $S$ is contained in a soft neutrosophic LA-semigroup over $N(S)$.

**Proof:** Snice an LA-semigroup $S$ is always contained in the corresponding neutrosophic LA-semigroup and henc it follows that soft LA-semigroup over $S$ is contained in soft neutrosophic LA-semigroup over $N(S)$.

**Proposition 5.1.2:** Let $(F, A)$ and $(H, B)$ be two soft neutronsophic LA-semigroup over $N(S)$. Then

1) Their extended intersection $(F, A) \cap_\varepsilon (H, B)$ is a soft neutrosophic LA-semigroup over $N(S)$.
2) Their restricted intersection $(F, A) \cap_R (H, B)$ is also soft neutrosophic LA-semigroup over $N(S)$.

**Proof:** These are straightforward.

**Remark 5.1.1:** Let $(F,A)$ and $(H,B)$ be two soft neutrosophic LA-semigroup over $N(S)$. Then

1) Their extended union $(F,A) \cup_\varepsilon (H,B)$ is not soft neutrosophic LA-semigroup over $N(S)$.
2) Their restricted union $(F,A) \cup_R (H,B)$ is not a soft neutrosophic LA-semigroup over $N(S)$.

Let us take the following example to prove the remark.

**Example 5.1.2:** Let $N(S) = \{1,2,3,4,1I,2I,3I,4I\}$ be a neutrosophic LA-semigroup with the following table.

| *  | 1  | 2  | 3  | 4  | 1I | 2I | 3I | 4I |
|----|----|----|----|----|----|----|----|----|
| 1  | 1  | 4  | 2  | 3  | 1I | 4I | 2I | 3I |
| 2  | 3  | 2  | 4  | 1  | 3I | 2I | 4I | 1I |
| 3  | 4  | 1  | 3  | 2  | 4I | 1I | 3I | 2I |
| 4  | 2  | 3  | 1  | 4  | 2I | 3I | 1I | 4I |
| 1I | 1I | 4I | 2I | 3I | 1I | 4I | 2I | 3I |
| 2I | 3I | 2I | 4I | 1I | 3I | 2I | 4I | 1I |
| 3I | 4I | 1I | 3I | 2I | 4I | 1I | 3I | 2I |
| 4I | 2I | 3I | 1I | 4I | 2I | 3I | 1I | 4I |

Let $(F,A)$ be a soft set over $N(S)$. Let $(F,A)$ and $(H,B)$ be two soft neutosophic LA-semigroups over $N(S)$, where

$$F(a_1) = \{1,1I\}, \quad F(a_2) = \{2,2I\},$$

$$F(a_3) = \{3,3I\}, \quad F(a_4) = \{4,4I\},$$

and

$$H(a_1) = \{3,3I\},$$

$$H(a_3) = \{1,1I\}.$$

Then clearly their extended union $(F,A) \cup_\varepsilon (H,B)$ and restricted union $(F,A) \cup_R (H,B)$ is not soft neutrosophic LA-semigroup because union of two neutrosophic sub LA-semigroup may not be again a neutrosophic sub LA-semigroup.

**Proposition 5.1.3:** Let $(F,A)$ and $(G,B)$ be two soft neutrosophic LA-semigroup over $N(S)$. Then $(F,A) \wedge (H,B)$ is also soft neutrosophic LA-semigroup if it is non-empty.

**Proof:** Straightforward.

**Proposition 5.1.4:** Let $(F,A)$ and $(G,B)$ be two soft neutrosophic LA-semigroup over the neutosophic LA-semigroup $N(S)$. If $A \cap B = \phi$ Then their extended union $(F,A) \cup_\varepsilon (G,B)$ is a soft neutrosophic LA-semigroup over $N(S)$.

**Proof:** Straightforward.

**Definition 5.1.2:** A soft neutrosophic LA-semigroup $(F,A)$ over $N(S)$ is said to be a soft neutosophic LA-semigroup with left identity $e$ if for all $a \in A$, the parameterized set $F(a)$ is a neutrosophic sub LA-semigroup with left identity $e$ where $e$ is the left identity of $N(S)$.

**Lemma 5.1.1:** Let $(F,A)$ be a soft neutrosophic LA-semigroup with left identity $e$ over $N(S)$. Then
$$(F,A) \odot (F,A) = (F,A).$$
**Proof:** By definition $(F,A) \odot (F,A) = (H,C)$, where $C = A \cap A = A$ and so $H(a) = F(a)F(a)$ for all $a \in C$. Since each $F(a)$ is a neutrosophic sub LA-semigroup with left identity $e$ which follows that $F(a)F(a) = F(a)$. Hence $H(a) = F(a)$ and thus $(H,C) = (F,A)$.

**Proposition 5.1.5:** Let $(F,A)$ and $(G,B)$ be two soft neutronsophic LA-semigroups over $N(S)$. Then thecartesian product of $(F,A)$ and $(G,B)$ is also soft neutrosophic LA-semigroup over $N(S)$.

**Proof:** Since $(F,A) * (G,B) = (H, A \times B)$ and defined as $H(a,b) = F(a)G(b)$ for $a \in A$ and $b \in B$. Let $x, y \in H(a,b)$ such that $x = a_1 b_1$ and $y = a_2 b_2$, where $a_1, a_2 \in F(a)$ and $b_1, b_2 \in G(b)$. Since
$$xy = (a_1 b_1)(a_2 b_2) \text{ and by using medial law .}$$
$$xy = (a_1 a_2)(b_1 b_2)$$
Hence $H(a,b)$ is neutrosophic sub LA-semigroup and consequently $(H, A \times B)$ is a soft neutrosophic LA-semigroup.

**Definition 5.1.3:** Let $(F,A)$ be a soft neutrosophic LA-semigroup over $N(S)$. Then $(F,A)$ is called an absolute soft neutrosophic LA-semigroup if $F(a) = N(S)$ for all $a \in A$. We denote it by $A_{N(S)}$.

**Definition 5.1.4:** Let $(F,A)$ and $(G,B)$ be two soft neutrosophic LA-semigroup over $N(S)$. Then $(G,B)$ is soft sub neutrosophic LA-semigroup of $(F,A)$, if

1) $B \subseteq A$, and
2) $H(b)$ is a neutrosophic sub LA-semigroup of $F(b)$, for all $b \in B$.

**Example 5.1.3:** Let $(F,A)$ be a soft neutrosophic LA-semigroup over $N(S)$ in Example $(1)$. Let $(G,B)$ be another soft neutrosophic LA-semigroup over $N(S)$, where
$$G(b_1) = \{1\}, G(b_2) = \{2I\}$$

Then clearly $(G,B)$ is a soft sub neutosophic LA-semigroup of $(F,A)$ over $N(S)$.

**Theorem 5.1.2:** Every soft LA-semigroup over $S$ is a soft sub neutrosophic LA-semigroup of a soft neutrosophic LA-semigroup over $N(S)$.

**Proof:** The proof is obvious.

**Definition 5.4.5:** A soft set $(F,A)$ over a neutrosophic LA-semigroup $N(S)$ is called a soft neutrosophic left (right) ideal over $N(S)$ if
$A_{N(S)} \odot (F,A) \subseteq (F,A), ((F,A) \odot A_{N(S)} \subseteq (F,A))$ where $A_{N(S)}$ is the absolute soft neutrosophic LA-semigroup over $N(S)$.

A soft set $(F,A)$ over $N(S)$ is a soft neutrosophic ideal if it is soft neutrosophic left ideal as well as soft neutrosophic right ideal over $N(S)$.

**Proposition 5.1.6:** Let $(F,A)$ be a soft set over $N(S)$. Then $(F,A)$ is a soft neutrosophic ideal over $N(S)$ if and only if $F(a) \neq \phi$ is a neutrosophic ideal of $N(S)$, for all $a \in A$.

**Proof:** Suppose that $(F,A)$ be a soft neutrosophic ideal over $N(S)$. Then by definition
$$A_{N(S)} \odot (F,A) \subseteq (F,A) \text{ and } (F,A) \odot A_{N(S)} \subseteq (F,A)$$
This implies that $N(S)F(a) \subseteq F(a)$ and $F(a)N(S) \subseteq F(a)$ for all $a \in A$.
Thus $F(a)$ is a neutrosophic ideal of $N(S)$ for all $a \in A$.

Conversely, suppose that $(F,A)$ is a soft set over $N(S)$ such that each $F(a)$ is a neutrosophic ideal of $N(S)$ for all $a \in A$. So
$N(S)F(a) \subseteq F(a)$ and $F(a)N(S) \subseteq F(a)$ for all $a \in A$.
Therefore $A_{N(S)} \odot (F,A) \subseteq (F,A)$ and $(F,A) \odot A_{N(S)} \subseteq (F,A)$. Hence $(F,A)$ is a soft neutrosophic ideal of $N(S)$.

**Example 5.1.4:** Let $N(S) = \{1, 2, 3, 1I, 2I, 3I\}$ be a neutrosophic LA-semigroup with the following table.

| *  | 1  | 2  | 3  | 1I | 2I | 3I |
|----|----|----|----|----|----|----|
| 1  | 3  | 3  | 3  | 3I | 3I | 3I |
| 2  | 3  | 3  | 3  | 3I | 3I | 3I |
| 3  | 1  | 3  | 3  | 1I | 3I | 3I |
| 1I | 3I | 3I | 3I | 3I | 3I | 3I |
| 2I | 3I | 3I | 3I | 3I | 3I | 3I |
| 3I | 1I | 3I | 3I | 1I | 3I | 3I |

Then clearly $(F, A)$ is a soft neutrosophic ideal over $N(S)$, where

$$F(a_1) = \{2, 3, 3I\}, \quad F(a_2) = \{1, 3, 1I, 3I\}.$$

**Proposition 5.1.7:** Let $(F,A)$ and $(G,B)$ be two soft neutrosophic ideals over $N(S)$. Then

1) Their restricted union $(F,A) \cup_R (G,B)$ is a soft neutrosophic ideal over $N(S)$.

2) Their restricted intersection $(F,A) \cap_R (G,B)$ is a soft neutrosophic ideal over $N(S)$.

3) Their extended union $(F,A) \cup_\varepsilon (G,B)$ is also a soft neutrosophic ideal over $N(S)$.

4) Their extended intersection $(F,A) \cap_\varepsilon (G,B)$ is a soft neutrosophic ideal over $N(S)$.

**Proposition 5.1.8:** Let $(F,A)$ and $(G,B)$ be two soft neutrosophic ideals over $N(S)$. Then
1. Their *OR* operation $(F,A) \vee (G,B)$ is a soft neutrosophic ideal over $N(S)$.

2. Their *AND* operation $(F,A) \wedge (G,B)$ is a soft neutrosophic ideal over $N(S)$.

**Proposition 5.1.9:** Let $(F,A)$ and $(G,B)$ be two soft neutrosophic ideals over $N(S)$, where $N(S)$ is a neutrosophic LA-semigroup with left identity $e$. Then $(F,A) * (G,B) = (H, A \times B)$ is also a soft neutrosophic ideal over $N(S)$.

**Proof:** As $(F,A)*(G,B) = (H, A\times B)$, where $H(a,b) = F(a)G(b)$. Let $x \in H(a,b)$, then $x = yz$ where $y \in F(a)$ and $z \in G(b)$. Let $s \in N(S)$. Then

$$sx = s(yz)$$
$$= (es)(yz)$$
$$= (ey)(sz), \text{ by medial law}$$
$$= y(sz) \in F(a)G(b) = H(a,b)$$

Similarly $xs \in F(a)G(b) = H(a,b)$. Thus $H(a,b)$ is a neutrosophic ideal of $N(S)$ and hence $(F,A)*(G,B)$ is a soft neutrosophic ideal over $N(S)$.

**Proposition 5.1.10:** Let $(F,A)$ and $(G,B)$ be two soft neutrosophic ideals over $N(S)$ and $N(T)$. Then the cartesian product $(F,A) \times (G,B)$ is a soft neutrosophic ideal over $N(S) \times N(T)$.

**Proof:** Straightforward.

**Definition 5.1.6:** Let $(F,A)$ and $(G,B)$ be a soft neutrosophic LA-semigroups over $N(S)$. Then $(G,B)$ is soft neutrosophic ideal of $(F,A)$, if

1) $B \subseteq A$, and
2) $H(b)$ is a neutrosophic ideal of $F(b)$, for all $b \in B$.

**Example 5.1.5:** Let $N(S) = \{1, 2, 3, I, 2I, 3I\}$ be a neutrosophic LA-semigroup with the following table.

| . | 1 | 2 | 3 | I | 2I | 3I |
|---|---|---|---|---|---|---|
| 1 | 3 | 3 | 2 | 3I | 3I | 2I |
| 2 | 2 | 2 | 2 | 2I | 2I | 2I |
| 3 | 2 | 2 | 2 | 2I | 2I | 2I |
| I | 3I | 3I | 2I | 3I | 3I | 2I |
| 2I | 2I | 2I | 2I | 2I | 2I | 2I |
| 3I | 2I | 2I | 2I | 2I | 2I | 2I |

Then $(F, A)$ is a soft LA-semigroup over $N(S)$, where

$$F(a_1) = \{2, 2I\}, F(a_2) = \{2, 3, 2I, 3I\},$$
$$F(a_3) = \{1, 2, 3\}.$$

Let $(G, B)$ be a soft subset of $(F, A)$ over $N(S)$, where

$$G(a_2) = \{2, 2I\}, G(a_3) = \{2, 3\}.$$

Then clearly $(G, B)$ is a soft neutrosophic ideal of $(F, A)$.

**Proposition 5.1.11:** If $(F',A')$ and $(G',B')$ are soft neutrosophic ideals of soft neutrosophic LA-semigroup $(F,A)$ and $(G,B)$ over neutrosophic LA-semigroups $N(S)$ and $N(T)$ respectively. Then $(F',A') \times (G',B')$ is a soft neutrosophic ideal of soft neutrosophic LA-semigroup $(F,A) \times (G,B)$ over $N(S) \times N(T)$.

**Proof:** As $(F',A')$ and $(G',B')$ are soft neutrosophic ideals of soft neutrosophic LA-semigroups of $(F,A)$ and $(G,B)$. Then clearly $A' \subseteq A$ and $B' \subseteq B$ and so $A' \times B' \subseteq A \times B$. Therefore $F'(a')$ is a neutrosophic ideal of $F(a')$ for all $a' \in A'$ and also $G'(b')$ is a neutrosophic ideal of $G(b')$ for all $b' \in B'$. Consequently $F'(a') \times G'(b')$ is a neutrosophic ideal of $F(a') \times G(b')$. Hence $(F',A') \times (G',B')$ is a soft neutrosophic ideal of soft neutrosophic LA-semigroup $(F,A) \times (G,B)$ over $N(S) \times N(T)$.

**Theorem 5.1.3:** Let $(F,A)$ be a soft neutrosophic LA-semigroup over $N(S)$ and $\{(H_j, B_j) : j \in J\}$ be a non-empty family of soft neutrosophic sub LA-semigroups of $(F,A)$. Then

1) $\bigcap_{R \atop j \in J} (H_j, B_j)$ is a soft neutrosophic sub LA-semigroup of $(F,A)$.

2) $\wedge_{R \atop j \in J} (H_j, B_j)$ is a soft neutrosophic sub LA-semigroup of $(F,A)$.

3) $\bigcup_{\varepsilon \atop j \in J} (H_j, B_j)$ is a soft neutrosophic sub LA-semigroup of $(F,A)$ if $B_j \cap B_k = \phi$ for all $j, k \in J$.

**Theorem 5.1.4:** Let $(F,A)$ be a soft neutrosophic LA-semigroup over $N(S)$ and $\{(H_j, B_j) : j \in J\}$ be a non-empty family of soft neutrosophic ideals of $(F,A)$. Then

1) $\bigcap_{R_{j \in J}} (H_j, B_j)$ is a soft neutrosophic ideal of $(F,A)$.

2) $\bigwedge_{j \in J} (H_j, B_j)$ is a soft neutrosophic ideal of $(F,A)$.

3) $\bigcup_{\varepsilon_{j \in J}} (H_j, B_j)$ is a soft neutrosophic ideal of $(F,A)$.

4) $\bigvee_{j \in J} (H_j, B_j)$ is a soft neutrosophic ideal of $(F,A)$.

**Proof:** Straightforward.

**Proposition 5.1.12:** Let $(F,A)$ be a soft neutrosophic LAsemigroup with left identity $e$ over $N(S)$ and $(G,B)$ be a soft neutrosophic right ideal of $(F,A)$. Then $(G,B)$ is also soft neutrosophic left ideal of $(F,A)$.

**Proof:** Straightforward.

**Lemma 5.1.2:** Let $(F,A)$ be a soft neutrosophic LA-semigroup with left identity $e$ over $N(S)$ and $(G,B)$ be a soft neutrosophic right ideal of $(F,A)$. Then $(G,B) \odot (G,B)$ is a soft neutrosophic ideal of $(F,A)$.

**Proof**: The proof is followed by above proposition.

**Soft Neutrosophic Strong LA-semigroup**

The notion of soft neutrosophic strong LA-semigroup over a neutrosophic LA-semigroup is introduced here. We give the definition of soft neutrosophic strong LA-semigroup and investigate some related properties with sufficient amount of illustrative examples.

**Definition 5.1.7:** Let $N(S)$ be a neutrosophic LA-smigroup and $(F,A)$ be a soft set over $N(S)$. Then $(F,A)$ is called soft neutrosophic strong LA-semigroup if and only if $F(a)$ is a neutrosophic strong su LA-semigroup of $N(S)$ for all $a \in A$.

**Example. 5.1.6:** Let $N(S) = \{1,2,3,4,1I,2I,3I,4I\}$ be a neutrosophic LA-semigroup with the following table.

| * | 1 | 2 | 3 | 4 | 1I | 2I | 3I | 4I |
|---|---|---|---|---|----|----|----|----|
| 1 | 1 | 4 | 2 | 3 | 1I | 4I | 2I | 3I |
| 2 | 3 | 2 | 4 | 1 | 3I | 2I | 4I | 1I |
| 3 | 4 | 1 | 3 | 2 | 4I | 1I | 3I | 2I |
| 4 | 2 | 3 | 1 | 4 | 2I | 3I | 1I | 4I |
| 1I | 1I | 4I | 2I | 3I | 1I | 4I | 2I | 3I |
| 2I | 3I | 2I | 4I | 1I | 3I | 2I | 4I | 1I |
| 3I | 4I | 1I | 3I | 2I | 4I | 1I | 3I | 2I |
| 4I | 2I | 3I | 1I | 4I | 2I | 3I | 1I | 4I |

Let $(F,A)$ be a soft set over $N(S)$. Then $(F,A)$ is soft neutrosophic strong LA-semigroup over $N(S)$ in the following manner,

$$F(a_1) = \{1I\}, \quad F(a_2) = \{2I\},$$

$$F(a_3) = \{3I\}, \quad F(a_4) = \{4I\}.$$

**Theorem 5.1.5:** All soft strong neutrosophic LA-semigroups or pure neutrosophic LA-semigroups are trivially soft neutrosophic LA-semigroups but the converse is not true in general.

**Proof:** The proof of this theorem is obvious.

**Definition 5.1.8:** Let $(G, B)$ be a soft sub-neutrosophic LA-smigroup of a soft neutrosophic LA-semigroup $(F, A)$ over $N(S)$. Then $(G, B)$ is said to be soft strong or pure sub-neutrosophic LA-semigroup of $(F, A)$ if each $G(b)$ is strong or pure neutrosophic sub LA-semigroup of $F(b)$, for all $b \in B$.

**Example 5.1.7:** Let $(F, A)$ be a soft neutrosophic LA-semigroup over $N(S)$ in example 5.1.1 and let $(G, B)$ be a soft sub-neutrosophic LA-semigroup of $(F, A)$ over $N(S)$, where
$$G(b_1) = \{1I\}, G(b_2) = \{2I\},$$
$$G(b_4) = \{4I\}.$$

Then clearly $(G, B)$ is a soft strong or pure sub-neutosophic LA-semigroup of $(F, A)$ over $N(S)$.

**Theorem 5.1.6:** A soft neutrosophic LA-semigroup $(F,A)$ over $N(S)$ can have soft sub LA-semigroups, soft sub-neutrosophic LA-semigroups and soft strong or pure sub-neutrosophic LA-semigroups.

**Proof:** It is obvious.

**Theorem 5.1.7:** If $(F,A)$ over $N(S)$ is a soft strong or pure neutrosophic LA-semigroup, then every soft sub-neutrosophic LA-semigroup of $(F,A)$ is a soft strong or pure sub-neutrosophic LA-semigroup.

**Proof:** The proof is straight forward.

**Definition 5.1.9:** A soft neutrosophic ideal $(F,A)$ over $N(S)$ is called soft strong or pure neutrosophic ideal over $N(S)$ if $F(a)$ is a strong or pure neutrosophic ideal of $N(S)$, for all $a \in A$.

**Example 5.1.8:** Let $N(S) = \{1,2,3,1I,2I,3I\}$ be as in example 5.1.6. Then $(F,A)$ be a soft strong or pure neutrosophic ideal over $N(S)$, where

$$F(a_1) = \{2I,3I\}, F(a_2) = \{1I,3I\}.$$

**Theorem 5.1.8:** All soft strong or pure neutrosophic ideals over $N(S)$ are trivially soft neutrosophic ideals but the converse is not true.

To see the converse, lets take a look to the following example.

**Example 5.1.9:** Let $(F,A)$ be a soft neutrosophic ideal over $N(S)$ in example 5.1.6. It is clear that $(F,A)$ is not a strong or pure neutrosophic ideal over $N(S)$.

**Proposition 5.1.13:** Let $(F,A)$ and $(G,B)$ be two soft strong or pure neutrosophic ideals over $N(S)$. Then

1) Their restricted union $(F,A) \cup_R (G,B)$ is a soft strong or pure neutrosophic ideal over $N(S)$.

2) Their restricted intersection $(F,A) \cap_R (G,B)$ is a soft strong or pure neutrosophic ideal over $N(S)$.

3) Their extended union $(F,A) \cup_\varepsilon (G,B)$ is also a soft strong or pure neutrosophic ideal over $N(S)$.

4) Their extended intersection $(F,A) \cap_\varepsilon (G,B)$ is a soft strong or pure neutrosophic ideal over $N(S)$.

**Proposition 5.1.14:** Let $(F,A)$ and $(G,B)$ be two soft strong or pure neutrosophic ideals over $N(S)$. Then

1) Their *OR* operation $(F,A) \vee (G,B)$ is a soft strong or pure neutrosophic ideal over $N(S)$.

2) Their *AND* operation $(F,A) \wedge (G,B)$ is a soft strong or pure neutrosophic ideal over $N(S)$.

**Proposition 5.1.15:** Let $(F,A)$ and $(G,B)$ be two soft strong or pure neutrosophic ideals over $N(S)$, where $N(S)$ is a neutrosophic LA-semigroup with left identity $e$. Then $(F,A)*(G,B)=(H,A\times B)$ is also a soft strong or pure neutrosophic ideal over $N(S)$.

**Proposition 5.1.16:** Let $(F,A)$ and $(G,B)$ be two soft strong or pure neutrosophic ideals over $N(S)$ and $N(T)$ respectively. Then the cartesian product $(F,A)\times(G,B)$ is a soft strong or pure neutrosophic ideal over $N(S)\times N(T)$.

**Definition 37** A soft neutrosophic ideal $(G,B)$ of a soft neutrosophic LA-semigroup $(F,A)$ is called soft strong or pure neutrosophic ideal if $G(b)$ is a strong or pure neutrosophic ideal of $F(b)$ for all $b \in B$.

**Example 5.1.10:** Let $(F,A)$ be a soft neutrosophic LA-semigroup over $N(S)$ in Example $(9)$ and $(G,B)$ be a soft neutrosophic ideal of $(F,A)$, where
$$G(b_1)=\{2I\}, G(b_2)=\{2I,3I\}.$$

Then clearly $(G,B)$ is a soft strong or pure neutrosophic ideal of $(F,A)$ over $N(S)$.

**Theorem 5.1.9:** Every soft strong or pure neutrosophic ideal of a soft neutrosophic LA-semigroup is trivially a soft neutrosophic ideal but the converse is not true.

## 5.2 Soft Neutrosophic Bi-LA-semigroup

In this section, the definition of soft neutrosophic bi-LA-semigroup is given over a neutrosophic bi-LA-semigroup. Some of their fundamental properties are also given in this section.

**Definition 5.2.1:** Let $BN(S)$ be a neutrosophic bi-LA-semigroup and $(F,A)$ be a soft set over $BN(S)$. Then $(F,A)$ is called soft neutrosophic bi-LA-semigroup if and only if $F(a)$ is a neutrosophic sub bi-LA-semigroup of $BN(S)$ for all $a \in A$.

**Example 5.2.1:** Let $BN(S) = \{\langle S_1 \cup I \rangle \cup \langle S_2 \cup I \rangle\}$ be a neutrosophic bi-LA-semigroup where $\langle S_1 \cup I \rangle = \{1,2,3,4,1I,2I,3I,4I\}$ is a neutrosophic LA-semigroup with the following table.

| *  | 1  | 2  | 3  | 4  | 1I | 2I | 3I | 4I |
|----|----|----|----|----|----|----|----|----|
| 1  | 1  | 4  | 2  | 3  | 1I | 4I | 2I | 3I |
| 2  | 3  | 2  | 4  | 1  | 3I | 2I | 4I | 1I |
| 3  | 4  | 1  | 3  | 2  | 4I | 1I | 3I | 2I |
| 4  | 2  | 3  | 1  | 4  | 2I | 3I | 1I | 4I |
| 1I | 1I | 4I | 2I | 3I | 1I | 4I | 2I | 3I |
| 2I | 3I | 2I | 4I | 1I | 3I | 2I | 4I | 1I |
| 3I | 4I | 1I | 3I | 2I | 4I | 1I | 3I | 2I |
| 4I | 2I | 3I | 1I | 4I | 2I | 3I | 1I | 4I |

$\langle S_2 \cup I \rangle = \{1,2,3,1I,2I,3I\}$ be another neutrosophic bi-LA-semigroup with the following table.

| * | 1 | 2 | 3 | 1I | 2I | 3I |
|---|---|---|---|----|----|----|
| 1 | 3 | 3 | 3 | 3I | 3I | 3I |
| 2 | 3 | 3 | 3 | 3I | 3I | 3I |
| 3 | 1 | 3 | 3 | 1I | 3I | 3I |
| 1I | 3I | 3I | 3I | 3I | 3I | 3I |
| 2I | 3I | 3I | 3I | 3I | 3I | 3I |
| 3I | 1I | 3I | 3I | 1I | 3I | 3I |

Let $A = \{a_1, a_2, a_3\}$ be a set of parameters. Then clearly $(F, A)$ is a soft neutrosophic bi-LA-semigroup over $BN(S)$, where

$$F(a_1) = \{1, 1I\} \cup \{2, 3, 3I\},$$

$$F(a_2) = \{2, 2I\} \cup \{1, 3, 1I, 3I\},$$

$$F(a_3) = \{4, 4I\} \cup \{1I, 3I\}.$$

**Proposition 5.2.1:** Let $(F,A)$ and $(K,D)$ be two soft neutrosophic bi-LA-semigroups over $BN(S)$. Then

1. Their extended intersection $(F,A) \cap_E (K,D)$ is soft neutrosophic bi-LA-semigroup over $BN(S)$.
2. Their restricted intersection $(F,A) \cap_R (K,D)$ is soft neutrosophic bi-LA-semigroup over $BN(S)$.
3. Their $AND$ operation $(F,A) \wedge (K,D)$ is soft neutrosophic bi-LA-semigroup over $BN(S)$.

**Remark 5.2.1:** Let $(F,A)$ and $(K,D)$ be two soft neutrosophic bi-LA-semigroups over $BN(S)$. Then

1. Their extended union $(F,A) \cup_E (K,D)$ is not soft neutrosophic bi-LA-semigroup over $BN(S)$.
2. Their restricted union $(F,A) \cup_R (K,D)$ is not soft neutrosophic bi-LA-semigroup over $BN(S)$.
3. Their $OR$ operation $(F,A) \vee (K,D)$ is not soft neutrosophic bi-LA-semigroup over $BN(S)$.

One can easily proved $(1),(2),$ and $(3)$ by the help of examples.

**Definition 5.2.2:** Let $(F,A)$ and $(K,D)$ be two soft neutrosophic bi-LA-semigroups over $BN(S)$. Then $(K,D)$ is called soft neutrosophic sub bi-LA-semigroup of $(F,A)$, if

1. $D \subseteq A$.
2. $K(a)$ is a neutrosophic sub bi-LA-semigroup of $F(a)$ for all $a \in A$.

**Example 5.2.2:** Let $(F,A)$ be a soft neutrosophic bi-LA-semigroup over $BN(S)$ in Example 5.1.1. Then clearly $(K,D)$ is a soft neutrosophic sub bi-LA-semigroup of $(F,A)$ over $BN(S)$, where

$$K(a_1) = \{1,1I\} \cup \{3,3I\},$$
$$K(a_2) = \{2,2I\} \cup \{1,1I\}.$$

**Theorem 5.2.1:** Let $(F,A)$ be a soft neutrosophic bi-LA-semigroup over $BN(S)$ and $\{(H_j, B_j) : j \in J\}$ be a non-empty family of soft neutrosophic sub bi-LA-semigroups of $(F,A)$. Then

1. $\bigcap_{R_{j \in J}} (H_j, B_j)$ is a soft neutrosophic sub bi-LA-semigroup of $(F,A)$.

2. $\bigwedge_{R_{j \in J}} (H_j, B_j)$ is a soft neutrosophic sub bi-LA-semigroup of $(F,A)$.

3. $\bigcup_{\varepsilon_{j \in J}} (H_j, B_j)$ is a soft neutrosophic sub bi-LA-semigroup of $(F,A)$ if $B_j \cap B_k = \phi$ for all $j, k \in J$.

**Definition 5.2.3:** Let $(F,A)$ be a soft set over a neutrosophic bi-LA-semigroup $BN(S)$. Then $(F,A)$ is called soft neutrosophic biideal over $BN(S)$ if and only if $F(a)$ is a neutrosophic biideal of $BN(S)$, for all $a \in A$.

**Example 5.2.3:** Let $BN(S) = \{\langle S_1 \cup I \rangle \cup \langle S_2 \cup I \rangle\}$ be a neutrosophic bi-LA-semigroup, where $\langle S_1 \cup I \rangle = \{1,2,3,1I,2I,3I\}$ be a neutrosophic bi-LA-semigroup with the following table.

| * | 1 | 2 | 3 | 1I | 2I | 3I |
|---|---|---|---|----|----|----|
| 1 | 3 | 3 | 3 | 3I | 3I | 3I |
| 2 | 3 | 3 | 3 | 3I | 3I | 3I |
| 3 | 1 | 3 | 3 | 1I | 3I | 3I |
| 1I | 3I | 3I | 3I | 3I | 3I | 3I |
| 2I | 3I | 3I | 3I | 3I | 3I | 3I |
| 3I | 1I | 3I | 3I | 1I | 3I | 3I |

And $\langle S_2 \cup I \rangle = \{1,2,3,I,2I,3I\}$ be another neutrosophic LA-semigroup with the following table.

| . | 1 | 2 | 3 | I | 2I | 3I |
|---|---|---|---|---|----|----|
| 1 | 3 | 3 | 2 | 3I | 3I | 2I |
| 2 | 2 | 2 | 2 | 2I | 2I | 2I |
| 3 | 2 | 2 | 2 | 2I | 2I | 2I |
| I | 3I | 3I | 2I | 3I | 3I | 2I |
| 2I | 2I | 2I | 2I | 2I | 2I | 2I |
| 3I | 2I | 2I | 2I | 2I | 2I | 2I |

Let $A = \{a_1, a_2\}$ be a set of parameters. Then $(F, A)$ is a soft neutrosophic biideal over $BN(S)$, where

$$F(a_1) = \{1, 1I, 3, 3I\} \cup \{2, 2I\},$$

$$F(a_2) = \{1, 3, 1I, 3I\} \cup \{2, 3, 2I, 3I\}.$$

**Proposition 5.2.2:** Every soft neutrosophic biideal over a neutrosophic bi-LA-semigroup is trivially a soft neutrosophic bi-LA-semigroup but the conver is not true in general.

One can easily see the converse by the help of example.

**Proposition 5.2.3:** Let $(F, A)$ and $(K, D)$ be two soft neutrosophic biideals over $BN(S)$. Then

1. Their restricted union $(F, A) \cup_R (K, D)$ is a soft neutrosophic biideal over $BN(S)$.

2. Their restricted intersection $(F, A) \cap_R (K, D)$ is a soft neutrosophic biideal over $BN(S)$.

3. Their extended union $(F, A) \cup_\varepsilon (K, D)$ is also a soft neutrosophic biideal over $BN(S)$.

4. Their extended intersection $(F, A) \cap_\varepsilon (K, D)$ is a soft neutrosophic biideal over $BN(S)$.

5. Their $OR$ operation $(F, A) \vee (K, D)$ is a soft neutrosophic biideal over $BN(S)$.

6. Their $AND$ operation $(F, A) \wedge (K, D)$ is a soft neutrosophic biideal over $BN(S)$.

**Definition 5.2.4:** Let $(F,A)$ and $(K,D)$ be two soft neutrosophic bi- LA-semigroups over $BN(S)$. Then $(K,D)$ is called soft neutrosophic biideal of $(F,A)$, if

1. $B \subseteq A$, and

2. $K(a)$ is a neutrosophic biideal of $F(a)$, for all $a \in A$.

**Example 5.2.4:** Let $(F,A)$ be a soft neutrosophic bi-LA-semigroup over $BN(S)$ in Example 5.2.2. Then $(K,D)$ is a soft neutrosophic biideal of $(F,A)$ over $BN(S)$, where

$$K(a_1) = \{1I, 3I\} \cup \{2, 2I\},$$

$$K(a_2) = \{1, 3, 1I, 3I\} \cup \{2I, 3I\}.$$

**Theorem 5.2.2:** A soft neutrosophic biideal of a soft neutrosophic bi-LA-semigroup over a neutrosophic bi-LA_semigroup is trivially a soft neutosophic sub bi-LA-semigroup but the converse is not true in general.

**Proposition 5.2.4:** If $(F',A')$ and $(G',B')$ are soft neutrosophic biideals of soft neutrosophic bi-LA-semigroups $(F,A)$ and $(G,B)$ over neutrosophic bi-LA-semigroups $N(S)$ and $N(T)$ respectively. Then $(F',A') \times (G',B')$ is a soft neutrosophic biideal of soft neutrosophic bi-LA-semigroup $(F,A) \times (G,B)$ over $N(S) \times N(T)$.

**Theorem 5.2.3:** Let $(F,A)$ be a soft neutrosophic bi-LA-semigroup over $BN(S)$ and $\{(H_j, B_j): j \in J\}$ be a non-empty family of soft neutrosophic biideals of $(F,A)$. Then

1. $\bigcap_{R_{j \in J}} (H_j, B_j)$ is a soft neutrosophicbi ideal of $(F,A)$.

2. $\bigwedge_{j \in J} (H_j, B_j)$ is a soft neutrosophic biideal of $(F,A)$.

3. $\bigcup_{\varepsilon_{j \in J}} (H_j, B_j)$ is a soft neutrosophic biideal of $(F,A)$.

4. $\bigvee_{j \in J} (H_j, B_j)$ is a soft neutrosophic biideal of $(F,A)$.

### Soft Neutrosophic Storng Bi-LA-semigroup

The notion of soft neutrosophic strong bi-LA-semigroup over a neutrosophic bi-LA-semigroup is introduced here. We gave the definition of soft neutrosophic strong bi-LA-semigroup and investigated some related properties with sufficient amount of illustrative examples.

**Definition 5.2.5:** Let $BN(S)$ be a neutrosophic bi-LA-semigroup and $(F,A)$ be a soft set over $BN(S)$. Then $(F,A)$ is called soft neutrosophic strong bi-LA-semigroup if and only if $F(a)$ is a neutrosophic strong sub bi-LA-semigroup for all $a \in A$.

**Example 5.2.5:** Let $BN(S)$ be a neutrosophic bi-LA-semigroup in Example 5.1.1. Let $A = \{a_1, a_2\}$ be a set of parameters. Then $(F, A)$ is a soft neutrosophic strong bi-LA-semigroup over $BN(S)$, where

$$F(a_1) = \{1I, 2I, 3I, 4I\} \cup \{2I, 3I\},$$
$$F(a_2) = \{1I, 2I, 3I, 4I\} \cup \{1I, 3I\}.$$

**Proposition 5.2.5:** Let $(F, A)$ and $(K, D)$ be two soft neutrosophic strong bi-LA-semigroups over $BN(S)$. Then

1. Their extended intersection $(F, A) \cap_E (K, D)$ is soft neutrosophic strong bi-LA-semigroup over $BN(S)$.
2. Their restricted intersection $(F, A) \cap_R (K, D)$ is soft neutrosophic strong bi-LA-semigroup over $BN(S)$.
3. Their AND operation $(F, A) \wedge (K, D)$ is soft neutrosophic strong bi-LA-semigroup over $BN(S)$.

**Remark 5.2.2:** Let $(F, A)$ and $(K, D)$ be two soft neutrosophic strong bi-LA-semigroups over $BN(S)$. Then

1. Their extended union $(F, A) \cup_E (K, D)$ is not soft neutrosophic strong bi-LA-semigroup over $BN(S)$.
2. Their restricted union $(F, A) \cup_R (K, D)$ is not soft neutrosophic strong bi-LA-semigroup over $BN(S)$.
3. Their OR operation $(F, A) \vee (K, D)$ is not soft neutrosophic strong bi-LA-semigroup over $BN(S)$.

One can easily proved (1), (2), and (3) by the help of examples.

**Definition 5.2.6:** Let $(F,A)$ and $(K,D)$ be two soft neutrosophic strong bi-LA-semigroups over $BN(S)$. Then $(K,D)$ is called soft neutrosophic strong sub bi-LA-semigroup of $(F,A)$, if

1. $B \subseteq A$.
2. $K(a)$ is a neutrosophic strong sub bi-LA-semigroup of $F(a)$ for all $a \in A$.

**Theorem 5.2.4:** Let $(F,A)$ be a soft neutrosophic strong bi-LA-semigroup over $BN(S)$ and $\{(H_j, B_j) : j \in J\}$ be a non-empty family of soft neutrosophic strong sub bi-LA-semigroups of $(F,A)$. Then

1. $\bigcap_{j \in J} {}_R (H_j, B_j)$ is a soft neutrosophic strong sub bi-LA-semigroup of $(F,A)$.

2. $\bigwedge_{j \in J} {}_R (H_j, B_j)$ is a soft neutrosophic strong sub bi-LA-semigroup of $(F,A)$.

3. $\bigcup_{j \in J} {}_E (H_j, B_j)$ is a soft neutrosophic strong sub bi-LA-semigroup of $(F,A)$ if $B_j \cap B_k = \phi$ for all $j, k \in J$.

**Definition 5.2.7:** Let $(F,A)$ be a soft set over a neutrosophic bi-LA-semigroup $BN(S)$. Then $(F,A)$ is called soft neutrosophic strong biideal over $BN(S)$ if and only if $F(a)$ is a neutrosophic strong biideal of $BN(S)$, for all $a \in A$.

**Example 5.2.6:** Let $BN(S)$ be a neutrosophic bi-LA-semigroup in Example 5.2.2. Let $A = \{a_1, a_2\}$ be a set of parameters. Then clearly $(F, A)$ is a soft neutrosophic strong biideal over $BN(S)$, where

$$F(a_1) = \{1I, 3I\} \cup \{1I, 2I, 3I\},$$

$$F(a_2) = \{1I, 3I\} \cup \{2I, 3I\}.$$

**Theorem 5.2.5:** Every soft neutrosophic strong biideal over $BN(S)$ is a soft neutrosophic biideal but the converse is not true.

We can easily see the converse by the help of example.

**Proposition 5.2.6:** Every soft neutrosophic strong biideal over a neutrosophic bi-LA-semigroup is trivially a soft neutrosophic strong bi-LA-semigroup but the converse is not true in general.

**Proposition 5.2.7:** Every soft neutrosophic strong biideal over a neutrosophic bi-LA-semigroup is trivially a soft neutrosophic bi-LA-semigroup but the converse is not true in general.

One can easily see the converse by the help of example.

**Proposition 5.2.8:** Let $(F,A)$ and $(K,D)$ be two soft neutrosophic strong biideals over $BN(S)$. Then

1. Their restricted union $(F,A) \cup_R (K,D)$ is a soft neutrosophic strong biideal over $BN(S)$.

2. Their restricted intersection $(F,A) \cap_R (K,D)$ is a soft neutrosophic strong biideal over $BN(S)$.

3. Their extended union $(F,A) \cup_\varepsilon (K,D)$ is also a soft neutrosophic strong biideal over $BN(S)$.

4. Their extended intersection $(F,A) \cap_\varepsilon (K,D)$ is a soft neutrosophic strong biideal over $BN(S)$.

5. Their OR operation $(F,A) \vee (K,D)$ is a soft neutrosophic biideal over $BN(S)$.

6. Their AND operation $(F,A) \wedge (K,D)$ is a soft neutrosophic biideal over $BN(S)$.

**Definition 5.2.8:** Let $(F,A)$ and $(K,D)$ be two soft neutrosophic strong bi-LA-semigroups over $BN(S)$. Then $(K,D)$ is called soft neutrosophic strong biideal of $(F,A)$, if

1. $D \subseteq A$, and

2. $K(a)$ is a neutrosophic strong biideal of $F(a)$, for all $a \in A$.

**Theorem 5.2.7:** A soft neutrosophic strong biideal of a soft neutrosophic strong bi-LA-semigroup over a neutrosophic bi-LA_semigroup is trivially a soft neutosophic strong sub bi-LA-semigroup but the converse is not true in general.

**Proposition 5.2.9:** If $(F',A')$ and $(G',B')$ are soft neutrosophic strong biideals of soft neutrosophic bi-LA-semigroups $(F,A)$ and $(G,B)$ over neutrosophic bi-LA-semigroups $N(S)$ and $N(T)$ respectively. Then $(F',A') \times (G',B')$ is a soft neutrosophic strong biideal of soft neutrosophic bi-LA-semigroup $(F,A) \times (G,B)$ over $N(S) \times N(T)$.

**Theorem 5.2.8:** Let $(F,A)$ be a soft neutrosophic strong bi-LA-semigroup over $BN(S)$ and $\{(H_j, B_j): j \in J\}$ be a non-empty family of soft neutrosophic strong biideals of $(F,A)$. Then

1. $\bigcap_{R_{j \in J}} (H_j, B_j)$ is a soft neutrosophic strong bi ideal of $(F,A)$.

2. $\bigwedge_{j \in J} (H_j, B_j)$ is a soft neutrosophic strong biideal of $(F,A)$.

3. $\bigcup_{\varepsilon_{j \in J}} (H_j, B_j)$ is a soft neutrosophic strong biideal of $(F,A)$.

4. $\bigvee_{j \in J} (H_j, B_j)$ is a soft neutrosophic strong biideal of $(F,A)$.

## 5.3 Soft Neutrosophic N-LA-semigroup

In this section, we extend soft sets to neutrosophic N-LA-semigroups and introduce soft neutrosophic N-LA-semigroups. This is the generalization of soft neutrosophic LA-semigroups. Some of their impotant facts and figures are also presented here with illustrative examples. We also initiated the strong part of neutrosophy in this section. Now we proceed onto define soft neutrosophic N-LA-semigroup as follows.

**Definition 5.3.1:** Let $\{S(N), *_1, *_2, ..., *_N\}$ be a neutrosophic N-LA-semigroup and $(F, A)$ be a soft set over $S(N)$. Then $(F, A)$ is called soft neutrosophic N-LA-semigroup if and only if $F(a)$ is a neutrosophic sub N-LA-semigroup of $S(N)$ for all $a \in A$.

**Example 5.3.1:** Let $S(N) = \{S_1 \cup S_2 \cup S_3, *_1, *_2, *_3\}$ be a neutrosophic 3-LA-semigroup where $S_1 = \{1, 2, 3, 4, 1I, 2I, 3I, 4I\}$ is a neutrosophic LA-semigroup with the following table.

| *  | 1  | 2  | 3  | 4  | 1I | 2I | 3I | 4I |
|----|----|----|----|----|----|----|----|----|
| 1  | 1  | 4  | 2  | 3  | 1I | 4I | 2I | 3I |
| 2  | 3  | 2  | 4  | 1  | 3I | 2I | 4I | 1I |
| 3  | 4  | 1  | 3  | 2  | 4I | 1I | 3I | 2I |
| 4  | 2  | 3  | 1  | 4  | 2I | 3I | 1I | 4I |
| 1I | 1I | 4I | 2I | 3I | 1I | 4I | 2I | 3I |
| 2I | 3I | 2I | 4I | 1I | 3I | 2I | 4I | 1I |
| 3I | 4I | 1I | 3I | 2I | 4I | 1I | 3I | 2I |
| 4I | 2I | 3I | 1I | 4I | 2I | 3I | 1I | 4I |

$S_2 = \{1,2,3,1I,2I,3I\}$ be another neutrosophic bi-LA-semigroup with the following table.

| * | 1 | 2 | 3 | 1I | 2I | 3I |
|---|---|---|---|----|----|----|
| 1 | 3 | 3 | 3 | 3I | 3I | 3I |
| 2 | 3 | 3 | 3 | 3I | 3I | 3I |
| 3 | 1 | 3 | 3 | 1I | 3I | 3I |
| 1I | 3I | 3I | 3I | 3I | 3I | 3I |
| 2I | 3I | 3I | 3I | 3I | 3I | 3I |
| 3I | 1I | 3I | 3I | 1I | 3I | 3I |

And $S_3 = \{1,2,3,I,2I,3I\}$ is another neutrosophic LA-semigroup with the following table.

| . | 1 | 2 | 3 | I | 2I | 3I |
|---|---|---|---|---|----|----|
| 1 | 3 | 3 | 2 | 3I | 3I | 2I |
| 2 | 2 | 2 | 2 | 2I | 2I | 2I |
| 3 | 2 | 2 | 2 | 2I | 2I | 2I |
| I | 3I | 3I | 2I | 3I | 3I | 2I |
| 2I | 2I | 2I | 2I | 2I | 2I | 2I |
| 3I | 2I | 2I | 2I | 2I | 2I | 2I |

Let $A = \{a_1, a_2, a_3\}$ be a set of parameters. Then clearly $(F, A)$ is a soft neutrosophic 3-LA-semigroup over $S(N)$, where

$$F(a_1) = \{1, 1I\} \cup \{2, 3, 3I\} \cup \{2, 2I\},$$

$$F(a_2) = \{2, 2I\} \cup \{1, 3, 1I, 3I\} \cup \{2, 3, 2I, 3I\},$$

$$F(a_3) = \{4, 4I\} \cup \{1I, 3I\} \cup \{2I, 3I\}.$$

**Proposition 5.3.1:** Let $(F, A)$ and $(K, D)$ be two soft neutrosophic N-LA-semigroups over $S(N)$. Then

1. Their extended intersection $(F, A) \cap_E (K, D)$ is soft neutrosophic N-LA-semigroup over $S(N)$.
2. Their restricted intersection $(F, A) \cap_R (K, D)$ is soft neutrosophic N-LA-semigroup over $S(N)$.
3. Their AND operation $(F, A) \wedge (K, D)$ is soft neutrosophic N-LA-semigroup over $S(N)$.

**Remark 5.3.1:** Let $(F, A)$ and $(K, D)$ be two soft neutrosophic N-LA-semigroups over $S(N)$. Then

1. Their extended union $(F, A) \cup_E (K, D)$ is not soft neutrosophic N-LA-semigroup over $S(N)$.
2. Their restricted union $(F, A) \cup_R (K, D)$ is not soft neutrosophic N-LA-semigroup over $S(N)$.
3. Their OR operation $(F, A) \vee (K, D)$ is not soft neutrosophic N-LA-semigroup over $S(N)$.

One can easily proved (1), (2), and (3) by the help of examples.

**Definition 5.3.2:** Let $(F, A)$ and $(K, D)$ be two soft neutrosophic N-LA-semigroups over $S(N)$. Then $(K, D)$ is called soft neutrosophic sub N-LA-semigroup of $(F, A)$, if

1. $D \subseteq A$.
2. $K(a)$ is a neutrosophic sub N-LA-semigroup of $F(a)$ for all $a \in A$.

**Theorem 5.3.1:** Let $(F, A)$ be a soft neutrosophic N-LA-semigroup over $S(N)$ and $\{(H_j, B_j) : j \in J\}$ be a non-empty family of soft neutrosophic sub N-LA-semigroups of $(F, A)$. Then

1. $\bigcap_R_{j \in J} (H_j, B_j)$ is a soft neutrosophic sub N-LA-semigroup of $(F, A)$.

2. $\wedge_R_{j \in J} (H_j, B_j)$ is a soft neutrosophic sub N-LA-semigroup of $(F, A)$.

3. $\bigcup_{\varepsilon}_{j \in J} (H_j, B_j)$ is a soft neutrosophic sub N-LA-semigroup of $(F, A)$ if $B_j \cap B_k = \phi$ for all $j, k \in J$.

**Definition 5.3.2:** Let $(F, A)$ be a soft set over a neutrosophic N-LA-semigroup $S(N)$. Then $(F, A)$ is called soft neutrosophic N-ideal over $S(N)$ if and only if $F(a)$ is a neutrosophic N-ideal of $S(N)$ for all $a \in A$.

**Example 5.3.2:** Consider Example 5.3.1. Let $A = \{a_1, a_2\}$ be a set of parameters. Then $(F, A)$ is a soft neutrosophic 3-ideal over $S(N)$, where

$$F(a_1) = \{1, 1I\} \cup \{3, 3I\} \cup \{2, 2I\},$$

$$F(a_2) = \{2, 2I\} \cup \{1I, 3I\} \cup \{2, 3, 3I\}.$$

**Proposition 5.3.2:** Every soft neutrosophic N-ideal over a neutrosophic N-LA-semigroup is trivially a soft neutrosophic N-LA-semigroup but the converse is not true in general.

One can easily see the converse by the help of example.

**Proposition 5.3.3:** Let $(F, A)$ and $(K, D)$ be two soft neutrosophic N-ideals over $S(N)$. Then

1. Their restricted union $(F, A) \cup_R (K, D)$ is a soft neutrosophic N-ideal over $S(N)$.
2. Their restricted intersection $(F, A) \cap_R (K, D)$ is a soft neutrosophic N-ideal over $S(N)$.
3. Their extended union $(F, A) \cup_\varepsilon (K, D)$ is also a soft neutrosophic N-ideal over $S(N)$.
4. Their extended intersection $(F, A) \cap_\varepsilon (K, D)$ is a soft neutrosophic N-ideal over $S(N)$.

**Proposition 5.3.4:** Let $(F,A)$ and $(K,D)$ be two soft neutrosophic N-ideals over $S(N)$. Then

1. Their *OR* operation $(F,A) \vee (K,D)$ is a soft neutrosophic N-ideal over $S(N)$.

2. Their *AND* operation $(F,A) \wedge (K,D)$ is a soft neutrosophic N-ideal over $S(N)$.

**Definition 5.3.3:** Let $(F,A)$ and $(K,D)$ be two soft neutrosophic N- LA-semigroups over $S(N)$. Then $(K,D)$ is called soft neutrosophic N-ideal of $(F,A)$, if

1. $B \subseteq A$, and

2. $K(a)$ is a neutrosophic N-ideal of $F(a)$ for all $a \in A$.

**Theorem 5.3.2:** A soft neutrosophic N-ideal of a soft neutrosophic N-LA-semigroup over a neutrosophic N-LA-semigroup is trivially a soft neutosophic sub N-LA-semigroup but the converse is not true in general.

**Proposition 5.3.5:** If $(F',A')$ and $(G',B')$ are soft neutrosophic N-ideals of soft neutrosophic N-LA-semigroups $(F,A)$ and $(G,B)$ over neutrosophic N-LA-semigroups $N(S)$ and $N(T)$ respectively. Then $(F',A') \times (G',B')$ is a soft neutrosophic N-ideal of soft neutrosophic N-LA-semigroup $(F,A) \times (G,B)$ over $N(S) \times N(T)$.

**Theorem 5.3.3:** Let $(F,A)$ be a soft neutrosophic N-LA-semigroup over $S(N)$ and $\{(H_j, B_j): j \in J\}$ be a non-empty family of soft neutrosophic N-ideals of $(F,A)$. Then

1. $\bigcap_{j \in J}{}_R (H_j, B_j)$ is a soft neutrosophic N-ideal of $(F,A)$.

2. $\bigwedge_{j \in J} (H_j, B_j)$ is a soft neutrosophic N-ideal of $(F,A)$.

3. $\bigcup_{j \in J}{}_\varepsilon (H_j, B_j)$ is a soft neutrosophic N-ideal of $(F,A)$.

4. $\bigvee_{j \in J} (H_j, B_j)$ is a soft neutrosophic N-ideal of $(F,A)$.

**Soft Neutrosophic Strong N-LA-semigroup**

The notions of soft neutrosophic strong N-LA-semigroups over neutrosophic N-LA-semigroups are introduced here. We give some basic definitions of soft neutrosophic strong N-LA-semigroups and illustrated it with the help of exmaples and give some basic results.

**Definition 5.3.4:** Let $\{S(N), *_1, *_2, ..., *_N\}$ be a neutrosophic N-LA-semigroup and $(F,A)$ be a soft set over $S(N)$. Then $(F,A)$ is called soft neutrosophic strong N-LA-semigroup if and only if $F(a)$ is a neutrosophic strong sub N-LA-semigroup of $S(N)$ for all $a \in A$.

**Example 5.3.3:** Let $S(N) = \{S_1 \cup S_2 \cup S_3, *_1, *_2, *_3\}$ be a neutrosophic 3-LA-semigroup in Example (***). Let $A = \{a_1, a_2, a_3\}$ be a set of parameters. Then clearly $(F, A)$ is a soft neutrosophic strong 3-LA-semigroup over $S(N)$, where

$$F(a_1) = \{1I\} \cup \{2I, 3I\} \cup \{2I\},$$

$$F(a_2) = \{2I\} \cup \{1I, 3I\} \cup \{2I, 3I\},$$

**Theorem 5.3.4:** If $S(N)$ is a neutrosophic strong N-LA-semigroup, then $(F, A)$ is also a soft neutrosophic strong N-LA-semigroup over $S(N)$.

**Proof:** The proof is trivial.

**Proposition 5.3.6:** Let $(F, A)$ and $(K, D)$ be two soft neutrosophic strong N-LA-semigroups over $S(N)$. Then

1. Their extended intersection $(F, A) \cap_E (K, D)$ is soft neutrosophic strong N-LA-semigroup over $S(N)$.
2. Their restricted intersection $(F, A) \cap_R (K, D)$ is soft neutrosophic strong N-LA-semigroup over $S(N)$.
3. Their AND operation $(F, A) \wedge (K, D)$ is soft neutrosophic strong N-LA-semigroup over $S(N)$.

**Remark 5.3.2:** Let $(F,A)$ and $(K,D)$ be two soft neutrosophic strong N-LA-semigroups over $S(N)$. Then

1. Their extended union $(F,A) \cup_E (K,D)$ is not soft neutrosophic strong N-LA-semigroup over $S(N)$.
2. Their restricted union $(F,A) \cup_R (K,D)$ is not soft neutrosophic strong N-LA-semigroup over $S(N)$.
3. Their OR operation $(F,A) \vee (K,D)$ is not soft neutrosophic strong N-LA-semigroup over $S(N)$.

One can easily proved (1), (2), and (3) by the help of examples.

**Definition 5.3.5:** Let $(F,A)$ and $(K,D)$ be two soft neutrosophic strong N-LA-semigroups over $S(N)$. Then $(K,D)$ is called soft neutrosophic strong sub N-LA-semigroup of $(F,A)$, if

1. $D \subseteq A$.
2. $K(a)$ is a neutrosophic strong sub N-LA-semigroup of $F(a)$ for all $a \in A$.

**Theorem 5.3.5:** Let $(F,A)$ be a soft neutrosophic strong N-LA-semigroup over $S(N)$ and $\{(H_j, B_j) : j \in J\}$ be a non-empty family of soft neutrosophic strong sub N-LA-semigroups of $(F,A)$. Then

1. $\bigcap_{j \in J}{}_R (H_j, B_j)$ is a soft neutrosophic strong sub N-LA-semigroup of $(F,A)$.

2. $\bigwedge_{j \in J}{}_R (H_j, B_j)$ is a soft neutrosophic strong sub N-LA-semigroup of $(F,A)$.

3. $\bigcup_{j \in J}{}_\varepsilon (H_j, B_j)$ is a soft neutrosophic strong sub N-LA-semigroup of $(F,A)$ if $B_j \cap B_k = \phi$ for all $j, k \in J$.

**Definition 5.3.5:** Let $(F,A)$ be a soft set over a neutrosophic N-LA-semigroup $S(N)$. Then $(F,A)$ is called soft neutrosophic strong N-ideal over $S(N)$ if and only if $F(a)$ is a neutrosophic strong N-ideal of $S(N)$ for all $a \in A$.

**Proposition 5.3.7:** Every soft neutrosophic strong N-ideal over a neutrosophic N-LA-semigroup is trivially a soft neutrosophic strong N-LA-semigroup but the converse is not true in general.

One can easily see the converse by the help of example.

**Proposition 5.2.8:** Let $(F,A)$ and $(K,D)$ be two soft neutrosophic strong N-ideals over $S(N)$. Then

1. Their restricted union $(F,A) \cup_R (K,D)$ is a soft neutrosophic strong N-ideal over $S(N)$.

2. Their restricted intersection $(F,A) \cap_R (K,D)$ is a soft neutrosophic N-ideal over $S(N)$.

3. Their extended union $(F,A) \cup_\varepsilon (K,D)$ is also a soft neutrosophic strong N-ideal over $S(N)$.

4. Their extended intersection $(F,A) \cap_\varepsilon (K,D)$ is a soft neutrosophic strong N-ideal over $S(N)$.

5. Their $OR$ operation $(F,A) \vee (K,D)$ is a soft neutrosophic strong N-ideal over $S(N)$.

6. Their $AND$ operation $(F,A) \wedge (K,D)$ is a soft neutrosophic strong N-ideal over $S(N)$.

**Definition 5.3.6:** Let $(F,A)$ and $(K,D)$ be two soft neutrosophic strong N- LA-semigroups over $S(N)$. Then $(K,D)$ is called soft neutrosophic strong N-ideal of $(F,A)$, if

1. $B \subseteq A$, and

2. $K(a)$ is a neutrosophic strong N-ideal of $F(a)$ for all $a \in A$.

**Theorem 5.3.6:** A soft neutrosophic strong N-ideal of a soft neutrosophic strong N-LA-semigroup over a neutrosophic N-LA-semigroup is trivially a soft neutosophic strong sub N-LA-semigroup but the converse is not true in general.

**Theorem 5.3.7:** A soft neutrosophic strong N-ideal of a soft neutrosophic strong N-LA-semigroup over a neutrosophic N-LA-semigroup is trivially a soft neutosophic strong N-ideal but the converse is not true in general.

**Proposition 5.3.9:** If $(F', A')$ and $(G', B')$ are soft neutrosophic strong N-ideals of soft neutrosophic strong N-LA-semigroups $(F, A)$ and $(G, B)$ over neutrosophic N-LA-semigroups $N(S)$ and $N(T)$ respectively. Then $(F', A') \times (G', B')$ is a soft neutrosophic strong N-ideal of soft neutrosophic strong N-LA-semigroup $(F, A) \times (G, B)$ over $N(S) \times N(T)$.

**Theorem 5.3.8:** Let $(F, A)$ be a soft neutrosophic strong N-LA-semigroup over $S(N)$ and $\{(H_j, B_j) : j \in J\}$ be a non-empty family of soft neutrosophic strong N-ideals of $(F, A)$. Then

1. $\bigcap_{R_{j \in J}} (H_j, B_j)$ is a soft neutrosophic strong N-ideal of $(F, A)$.

2. $\bigwedge_{j \in J} (H_j, B_j)$ is a soft neutrosophic strong N-ideal of $(F, A)$.

3. $\bigcup_{\varepsilon_{j \in J}} (H_j, B_j)$ is a soft neutrosophic strong N-ideal of $(F, A)$.

4. $\bigvee_{j \in J} (H_j, B_j)$ is a soft neutrosophic strong N-ideal of $(F, A)$.

# Chapter Six

# SUGGESTED PROBLEMS

In this chapter some problems about the soft neutrosophic structures and soft neutrosophic N-structures are given.

These problems are as follows:

1.  Can one define soft P-sylow neutrosophic group over a neutrosophic group?

2.  How one can define soft weak sylow neutrosphic group and soft sylow free neutrosophic group over a neutrosophic group?

3.  Define soft sylow neutrosophic subgroup of a soft sylow neutrosophic group over a neutrosophic group?

4.  What are soft sylow pseudo neutosophic group, weak sylow pseudo neutrosophic group and free sylow pseudo neutrosophic group over a neutrosophic group?

5. How one can define soft neutrosophic cosets of a soft neutrosophic group over a neutrosophic group?

6. What is the soft centre of a soft neutrosophic group over a neutrosophic group?

7. Define soft neutrosophic normalizer of a soft neutrosophic group over a neutrosophic group?

8. Can one define soft Cauchy neutrosophic group and soft semi Cauchy neutrosophic group over a neutrosophic group?

9. Define soft strong Cauchy neutrosophic group over a neutrosophic group?

10. Define all the properties and results in above problems in case of soft neutrosophic group over a neutrosophic bigroup?

11. Also define these notions for soft neutrosophic N-groups over neutrosophic N-groups?

12. Give some examples of soft neutrosophic semigroups over neutrosophic semigroups?

13. Write a soft neutrosophic semigroup over a neutrosophic semigroup which has soft neutrosophic subsemigroups, soft neutrosophic strong subsemigroups and soft subsemigroups?

14. Also do these calculations for a soft neutrosophic bisemigroup over a neutrosophic bisemigroup and for a soft neutrosophic N-semigroup over a neutrosophic N-semigroup respectively?

15. Can one define a soft neutrosophic cyclic semigroup over a neutrosophic semigroup?

16. What are soft neutrosophic idempotent semigroups over a neutrosophic semigroups? Give some of their examples?

17. What is a soft weakly neutrosophic idempotent semigroup over a neutrosophic semigroup?

18. Give examples of soft neutrosophic bisemigroups over neutrosophic bisemigroups?

19. Also give examples of soft neutrosophic N-semigroups over neutrosophic N-semigroups?

20. Can one define a soft neutosophic maximal and a soft neutrosophic minimal ideal over a neutrosophic semigroup?

21. How can one define a soft neutrosophic bimonoid over a neutrosophic bimonoid? Also give some examples of soft neutrosophic bimonoid over neutrosophic bimonoid?

22. Give some examples of soft Lagrange neutrosophic semigroups, soft weakly Lagrange neutrosophic semigroup and soft Lagrange free neutrosophic semigroups over neutrosophic semigroups?

23. Also repeat the above exercise 22, for soft Lagrange neutrosophic 5-semigroup, soft weakly Lagrange neutrosophic 5-semigroup and soft Lagrange free neutrosophic 5-semigroup over a neutrosophic 5-semigroup?

24. Give diffrerent examples of soft neutrosophic 3-semigroup, soft neutrosophic 4-semigroup and soft neutrosophic 5-semigroup over a neutrosophic 3-semigroup, a neutrosophic 4-semigroup and a neutrosophic 5-semigroup respectively?

25. When a soft neutrosophic semigroup will be a soft Lagrange free neutrosophic semigroup over a neutrosophic semigroup?

26. Can one define soft Cauchy neutrosophic semigroup over a neutrosophic semigroup? Also goive some examples of soft Cauchy neutrosophic semigroups?

27. Also define soft Cauchy neutrosophic bisemigroups and soft Cauchy neutrosophic N-semigroups over neutrosophic bisemigroups and neutrosophic N-semigroups respectively?

28. What is a soft Cauchy free neutrosophic semigroup over a neutrosophic semigroup? Also give some examples?

29. Define soft weakly Cauchy neutrosophic semigroup over a neutrosophic semigroup? Also give some examples?

30. Can ove define soft p-sylow neutrosophic semigroup over a neutrosophic semigroup?

31. What are soft weak sylow neutrosophic semigroups over neutrosophic semigroups? Also give some examples?

32. What are soft sylow free neutrosophic semigroups over neutrosophic semigroups? Also give some examples?

33. Give some examples of soft neutrosophic loops, soft neutrosophic biloops and soft neutrosophic N-loops over neutrosophic loops, neutrosophic biloops and neutrosophic N-loops respectively?

34. What are soft Cauchy neutrosophic loops, soft Cauchy neutrosophic biloops and soft Cauchy neutrosophic N-loops over neutrosophic loops, neutrosophic biloops and neutrosophic N-loops respectively?

35. What are soft p-sylow neutrosophic loops, soft p-sylow neutrosophic biloops and soft p-sylow neutrosophic N-loops over neutrosophic loops, neutrosophic biloops and neutrosophic N-loops respectively?

36. Can one dfine soft neutrosophic Moufang loop over a neutrosophic loop? Give some exmples?

37. What is a soft neutrosophic Bruck loop over a neutrosophic loop?

38. What is a soft neutrosophic Bol loop over a neutrosophic loop?

39. Give two examples of soft neutrosophic Moufang loop, three examples of soft neutrosophic Bruck loop and also 2 examples of soft neutrosophic Bol loop?

40. Define soft neutrosophic right LA-semigroup, soft neutrosophic right Bi-LA-semigroup and soft neutrosophic right N-LA-semigroup over a right neutrosophic LA-semigroup, a right neutrosophic Bi-LA-semigroup and a right neutrosophic N-LA-semigroup respectively?

41. Also give 2 examples in each case of above exercise 40?

42. Under what condition a soft neutrosophic LA-semiggroup will be a soft neutrosophic semigroup?

43. Give some examples of soft neutosophic LA-semigroups over neutrosophic LA-semigroups?

44. Can one define soft Lagrange neutrosophic LA-semigroup over a neutrosophic LA-semigroup?

45. What are soft weak Lagrange neutrosophic LA-semigroups over beutrosophic LA-semigroups? Give examples?

46. How one can define a soft Lagrange free neutrosophic LA-semigroup over a neutrosophic LA-semigroup?

47. Give some examples of soft Lagrnage neutrosophic Bi-LA-semigroups, soft weak Lagrange Bi-LA-semigroups and soft Lagrange free neutrosophic Bi-LA-semigroups over neutrosophic Bi-LA-semigroups?

48. Give some examples of soft Lagrnage neutrosophic N-LA-semigroups, soft weak Lagrange N-LA-semigroups and soft Lagrange free neutrosophic N-LA-semigroups over neutrosophic N-LA-semigroups?

49. Can one define soft Cauchy neutrosophic LA-semigroup over a neutrosophic LA-semigroup?

50. Can one define soft weak Cauchy neutrosophic LA-semigroup over a neutrosophic LA-semigroup?

51. Can one define soft Cauchy free neutrosophic LA-semigroup over a neutrosophic LA-semigroup?

52. Give examples soft Cauchy neutrosophic LA-semigroups, soft weak Cauchy neutrosophic LA-semigroups and soft Cauchy free neutrosophic La-semigroups over neutrosophic LA-semigroups?

# References

1. H. Aktas, N. Cagman, Soft sets and soft groups, Inf. Sci. 177 (2007) 2726-2735.

2. K. Atanassov, Intuitionistic fuzzy sets, Fuzzy Sets Syst. 64(2)(1986) 87-96.

3. Florentin Smarandache , Mumtaz Ali, Munazza Naz, Muhammad Shabir , Neutrosophic LA-semigroup, Neutrosophic Sets and Systems. (Accepted).

4. M. Shabir, M. Ali, M. Naz, F. Smarandache, Soft neutrosophic group, Neutrosophic Sets and Systems. 1(2013) 5-1.

5. M. Ali, F. Smarandache,M. Shabir, M. Naz, Soft neutrosophic Bigroup, and Soft Neutrosophic N-group, Neutrosophic Sets and Systems. 2 (2014) 55-81.

6. M. I. Ali, F. Feng, X. Liu, W. K. Min, M. Shabir, On some new operationsin soft set theory. Comp. Math. Appl., 57(2009), 1547-1553.

7. S. Broumi, F. Smarandache, Intuitionistic Neutrosophic Soft Set, J. Inf. & Comput. Sc. 8(2013) 130-140.

8. D. Chen, E.C.C. Tsang, D.S. Yeung, X. Wang, The parameterization reduction of soft sets and its applications,Comput. Math. Appl. 49(2005) 757-763.

9. F. Feng, M. I. Ali, M. Shabir, Soft relations applied to semigroups, Filomat 27(7)(2013) 1183-1196.

10. M.B. Gorzalzany, A method of inference in approximate reasoning based on interval-valued fuzzy sets, Fuzzy Sets Syst. 21(1987) 1-17.

11. W. B. V. Kandasamy, F. Smarandache, Basic Neutrosophic Algebraic Structures and their Applications to Fuzzy and Neutrosophic Models, Hexis (2004).

12. W. B. V. Kandasamy, F. Smarandache, N-Algebraic Structures and S-N-Algebraic Structures, Hexis Phoenix (2006).

13. W. B. V. Kandasamy, F. Smarandache, Some Neutrosophic Algebraic Structures and Neutrosophic N-Algebraic Structures, Hexis (2006).

14. P.K. Maji, R. Biswas and A. R. Roy, Soft set theory, Comput. Math. Appl. 45(2003) 555-562.

15. P. K. Maji, Neutrosophic Soft Sets, Ann. Fuzzy Math. Inf. 5(1)(2013) 2093-9310.

16. D. Molodtsov, Soft set theory first results, Comput. Math. Appl. 37(1999) 19-31.

17. Z. Pawlak, Rough sets, Int. J. Inf. Comp. Sci. 11(1982) 341-356.

18. F. Smarandache, A Unifying Field in Logics. Neutrosophy: Neutrosophic Probability, Set and Logic. Rehoboth: American Research Press (1999).

19. L.A. Zadeh, Fuzzy sets, Inf. Cont. 8(1965) 338-353.

20. Albert, A.A., Non-associative algebra I, II, Ann. Math. (2),**43**, 685-707, (1942).

21. Birkhoff, G. and Bartee, T.C. Modern Applied Algebra, Mc-Graw Hill, New York, (1970).

22. Bruck, R. H., A survey of binary systems, Springer-Verlag, (1958).

23. Bruck, R.H, Some theorems on Moufang loops, Math. Z., **73**, 59-78 (1960).

24. Castillo J., The Smarandache Semigroup, International Conference on Combinatorial Methods in Mathematics, II Meeting of the project 'Algebra, Geometria e Combinatoria', Faculdade de Ciencias da Universidade do Porto, Portugal, 9-11 July 1998.

25. Chang Quan, Zhang, Inner commutative rings, Sictiuan Dascue Xuebao (Special issue), **26**, 95-97 (1989).

26. Chein, Orin and Goodaire, Edgar G., Loops whose loop rings in characteristic 2 are alternative, Comm. Algebra, **18**, 659-668 (1990).

27. Chein, Orin, and Goodaire, Edgar G., Moufang loops with unique identity commutator (associator, square), J. Algebra, **130**, 369-384 (1990).

28. Chein, Orin, and Pflugfelder, H.O., The smallest Moufang loop, Arch. Math., **22**, 573-576 (1971).

29. Chein.O, Pflugfelder.H.O and Smith.J.D.H, (eds), Quasigroups and loops: Theory and applications, Sigma Series in Pure Maths, Vol. 8, Heldermann Verlag, (1990).

30. Chein, Orin, Kinyon, Michael. K., Rajah, Andrew and Vojlechovsky, Peter, Loops and the Lagrange property, (2002). http://lanl.arxiv.org/pdf/math.GR/0205141.

31. Fenyves.F, Extra loops II: On loops with identities of Bol Moufang types, Publ. Math. Debrecen, Vol.16, 187-192 (1969).

32. Goodaire, E.G., and Parmenter, M.M., Semisimplicity of alternative loop rings, Acta. Math. Hung, **50**. 241-247 (1987).

33. P. Holgate: Groupoids satisfying a simple invertive law, The Math. Student 61 (1992).

34. M. Kazim and M. Naseeruddin: On almost semigroups, Alig. Bull. Math.2 (1972).

35. Q. Mushtaq and M. S. Kamran: On left almost groups, Proc. Pak. Acad. Sci. 331 (1996), 53-55.

36. M. Shabir, S. Naz, Pure spectrum of an ag-groupoid with left identity and zero, World Applied Sciences Journal 17 (2012) 1759-1768.

37. Protic, P.V and N. Stevanovic, AG-test and some general properties of Abel-grassmann's groupoids,PU. M. A, 4,6 (1995), 371 – 383.

38. Madad Khan and N. Ahmad, Characterizations of left almost semigroups by their ideals, Journal of Advanced Research in Pure mathematics, 2 (2010), 61 – 73.

39. Q. Mushtaq and M. Khan, Ideals in left almost semigroups, Proceedings of 4[th] International Pure mathematics Conference, 2003, 65 – 77.

40. Mumtaz Ali, Muhammad Shabir, Munazza Naz, Florentin Smarandache, Neutrosophic LA-semigroup, Neutrosophic Sets and Systems. (Accepted).

41. Muhammad Aslam, Muhammad Shabir, Asif Mehmood, Some studies in soft LA-semigroup, Journal of Advance Research in Pure Mathematics, 3 (2011), 128 – 150.

# ABOUT THE AUTHORS

**Dr. Florentin Smarandache** is a Professor of Mathematics at the University of New Mexico in USA. He published over 75 books and 100 articles and notes in mathematics, physics, philosophy, psychology, literature, rebus. In mathematics his research is in number theory, non-Euclidean geometry, synthetic geometry, algebraic structures, statistics, neutrosophic logic and set (generalizations of fuzzy logic and set respectively), neutrosophic probability (generalization of classical and imprecise probability). Also, small contributions to nuclear and particle physics, information fusion, neutrosophy (a generalization of dialectics), law of sensations and stimuli, etc.

He can be contacted at smarand@unm.edu

**Mumtaz Ali** is a student of M.Phil in the Department of Mathematics, Quaid-e-Azam University Isalmabad. He recently started the research on soft set theory, neutrosophic set theory, algebraic structures including Smarandache algebraic structures, fuzzy theory, coding/ communication theory.

He can be contacted at mumtazali770@yahoo.com, bloomy_boy2006@yahoo.com

**Dr. Muhammad Shabir** is a Professor in the Department of Mathematics, Quaid-e-Azam University Islamabad. His interests are in fuzzy theory, soft set theory, algebraic structures. In the past decade he has guided 3 Ph.D. scholars in the different fields of non-associative Algebras and logics. Currently, five Ph.D. scholars are working under his guidance. He has written 75 research papers and he is also an author of a book. This is his 2$^{nd}$ book.

He can be contacted at **mshbirbhatti@yahoo.co.uk**